# Artificial Intelligence/Operations Research Workshop 2 Report Out

**Workshop**
August 16-17, 2022

**Workshop Organized by**

John Dickerson, University of Maryland
Bistra Dilkina, University of Southern California
Yu Ding, Texas A&M University
Swati Gupta, Georgia Institute of Technology
Pascal Van Hentenryck, Georgia Institute of Technology
Sven Koenig, University of Southern California
Ramayya Krishnan, Carnegie Mellon University
Radhika Kulkarni, SAS Institute, Inc. (retired)

**With Support From**

Catherine Gill, Computing Community Consortium (CCC)
Haley Griffin, Computing Community Consortium (CCC)
Maddy Hunter, Computing Community Consortium (CCC)
Ann Schwartz, Computing Community Consortium (CCC)

## Background and Introduction

Artificial intelligence (AI) has received significant attention in recent years, primarily due to breakthroughs in game playing, computer vision, and natural language processing that captured the imagination of the scientific community and the public at large. Many businesses, industries, and academic disciplines are now contemplating the application of AI to their own challenges. The federal government in the US and other countries have also invested significantly in advancing AI research and created funding initiatives and programs to promote greater collaboration across multiple communities. Some of the investment examples in the US include the establishment of the National AI Initiative Office, the launch of the National AI Research Resource Task Force, and more recently, the establishment of the National AI Advisory Committee.

In 2021 INFORMS and ACM SIGAI joined together with the Computing Community Consortium (CCC) to organize a series of three workshops. The objective for this workshop series is to explore ways to exploit the synergies of the AI and Operations Research (OR) communities to transform decision making. The aim of the workshops is to establish a joint strategic research vision for AI/OR that will maximize the societal impact of AI and OR in a world being transformed by technological change and a heightened desire to tackle important societal challenges such as growing racial and social inequity, climate change, and sustainable solutions to our food-water and energy needs. The vision for the workshops is to exploit and expand on the emerging synergies between these two communities with complementary strengths. However, there are barriers and difficulties in realizing this vision due to cultural differences between AI and OR communities. The workshop series aims to overcome these difficulties and to provide a stepping stone for a strong and sustained collaboration between the two fields.

The first workshop was held virtually in September 2021 with speakers and participants drawn from leading researchers in both the Operations Research and Computing research communities. The expectations were to promote greater inter-disciplinary collaborations between the two areas, inspire the agenda for research, and address critical questions for the future of AI. The agenda included interspersed sessions on Methods and Applications with moderated Q&A and breakout group discussions. Some of the key outcomes of the workshop included:

- Dispelling common misconceptions across the two communities and discovering common objectives, albeit a few disparate assumptions leading to different tools and methods
- Exploring multiple opportunities for collaboration at the Data, Methods, and Policy layers

- Discussing the need for a science of making good decisions with collaboration across multiple disciplines in addition to AI and OR
- Listing some potential future directions including access to data sets for a variety of application domains, competitions requiring joint teams of OR and AI participants, starting an AI/OR summer school for students in both disciplines, etc.

For details including the presentations and summaries of the group discussions, see the AI/OR Workshop I report.

---

## Overall Theme for the Second Workshop

One key outcome from the first workshop was the decision that we need a deeper dive into the topics of Fairness and Ethics in AI and also include a discussion of the role of Causality in applications of AI as well as considerations of human computer interaction in the implementation of automated solutions. As a result, the second workshop was planned around Trustworthy AI with four sessions on related topics within the scope of AI and OR:

- Fairness
- Explainable AI / Causality
- Robustness / Privacy
- Human Alignment and Human-Computer Interaction

While the first workshop focused on articulating a strategic vision, this workshop focused on what it takes to develop and deploy intelligent systems based on AI and OR technology that are trustworthy. In particular, the workshop was designed to:

- Study the state of the art in Trustworthy AI from a multi-disciplinary lens and for various application domains.
- Articulate grand challenges that need to be overcome to deploy trustworthy AI systems in the wild. Specifically, what tools and technologies have to be developed and evaluated for each of the foundational elements of trustworthy AI systems.
- Select a few topics for summer schools and research programs to foster collaborations in AI/OR for these topics.

The speakers and participants were selected from computer science and OR communities to foster a healthy exchange of ideas between the two groups. A key requirement for this workshop was to ensure that all speakers as well as participants attended it in person, to ensure a robust and healthy discussion on the topics presented during the event. It was held in Atlanta, GA on the Georgia Tech campus on August 16-17, 2022.

This report out should be considered an interim report with a final report to be published after the end of the third workshop. We expect the next workshop to be held in the first half of 2023. Note that the workshops are held on behalf of the community, by invitation only.

The workshop was kicked off with welcoming remarks by Radhika Kulkarni and introductions of all the participants, followed by a few remarks from Mr. Murat Omay from the Department of Transportation on funding opportunities for OR and AI research, from a problem-driven perspective. The details of Mr. Omay's remarks and all the sessions are described in the remainder of the report.

---

## Brief Comments from the Department of Transportation

Murat Omay - Figure B-1: Overview of ITS JPO Programs

After introductions, Murat Omay, from the Department of Transportation, gave a brief presentation on the DoT's Joint Program Office (JPO) for Intelligent Transportation Systems (ITS). Omay began by discussing the Data Access and Exchanges Portfolio, created and managed by the JPO ITS program. The portfolio aims to effectively generate, acquire, govern, manage and analyze ITS data across all modes to advance multimodal research and to enable a safe, equitable, multimodal and resilient transportation network. One of the key goals of this effort is to position the DoT in a leading role in data and AI strategies, data-based automation and AI research, innovation, and transformative data culture. Omay listed many of the current activities in this area, and explained how each furthered progress toward the overall goal of the JPO ITS program. Many of these activities were cross-departmental efforts, with ongoing dialogues between the DoT and the Office of Science and Technology Policy (OSTP), the Networking and Information Technology Research and Development program (NITRD) and the Open Knowledge Network (OKN) to name a few. Omay invited the participants to follow up with him directly to learn more about the Data Access and Exchanges Program.

Omay then turned our attention to the AI for ITS program, also run by the DoT. This program advocates for utilizing ethical AI and ML technologies to create a more efficient logistics system to transport goods and people. The program aims to coordinate technology and policy research to more quickly integrate AI and ML into the existing transportation system. In 2021, the AI for ITS program sought input from AI researchers on the "deployment ready" AI technologies the program had created, as well as on the best AI areas for the DoT to invest in and any current DoT technologies which could utilize AI to improve their performance.

Omay listed some of the key challenges to AI adoption and implementation, which are not unique to ITS. These include explainability, liability, model drift, privacy, security, ethics and equity and others; addressing these challenges is an ongoing exercise. He called out the fact that maintaining a human-in-the-loop approach is helpful in identifying and mitigating these challenges. He acknowledged the need for a multi-pronged approach and welcomed input from the community to help in addressing these issues.

Omay then took questions from the workshop participants and asked for any feedback pertaining to the DoT's efforts to incorporate AI in their technologies. He recommended Robert Sheehan as the key point of contact to learn more about the AI for ITS program. Robert Sheehan (Robert.Sheehan@dot.gov) is Acting Chief of Policy, Architecture, and Knowledge Transfer, U.S. DOT ITS Joint Program Office.

---

## Panel A: Fairness

Moderated by John Dickerson

**Slide Presentations**

Dimitris Bertsimas - Figure B-2: Improving on Fairness/Bias
Maria De-Arteaga - Figure B-3: Social Norms Bias: Residuals Harms of Fairness-Aware Algorithms
Nikhil Garg - Figure B-4: Auditing and Designing for Equity in Government Service Allocation
David Shmoys - Figure B-5: Fairness as the Objective in Congressional Districting

This panel broadly covered topics including fairness in allocation, learning, and decision making. Speakers broadly touched on the practical implementations of different definitions of fairness in, for example, healthcare, academic talent sourcing settings, and government resource allocation such as inspections of infrastructure. Speakers also discussed situations where traditional group fairness interventions do not work due to (by definition) averaging across an entire group. Fairness in redistricting to combat gerrymandering, a topic at the intersection of AI and OR research with economic and policy considerations as well, was also presented during this panel. A lively discussion followed considering, amongst other topics, the efficacy and appropriateness of using fairness as a *constraint* in various settings, and new topics in allocation and market design. Specifically, discussions of fairness in two-sided matching platforms such as rideshare (e.g., Uber and Lyft, where drivers on one side of the market are matched to riders on the other) arose naturally, although that specific use case had not yet been covered in any of the talks.

We now dive deeper into each of the four talks. Each talk came from folks more on the "OR side" of the research world, although all four speakers have published in AI and/or more broadly-defined "AI/ML-first" venues such as the ACM Conference on Fairness, Accountability, and Transparency (FAccT), and each speaker maintains strong collaborations with those on the "CS side."

The first talk, by Dimitris Bertsimas of MIT, primarily discussed an ongoing collaboration between MIT and Massachusetts General Hospital (MGH) analyzing a part of the healthcare pipeline, the discharge of trauma patients to post-acute care (PAC). Here, a group-level disparity was identified: 60% of trauma patients at MGH are Black, but only 12% of those receiving post-acute care are Black. Yet, it has been shown that PAC results in lower readmission rates and other benefits; this implies a potential disparate impact issue across racial categories. The talk covered in depth many of the intricacies of predicting risk and potential outcomes, and then presented a mixed-integer-programming-based (MIP-based) method for enforcing a form of demographic parity in PAC assignment. This approach also had interesting overlap with improving predictive performance of the assignment as well. The talk also connected the general approach used for this specific healthcare setting to, potentially, other settings such as talent sourcing in academic departments.

Perhaps most related to that first talk was Nikhil Garg of Cornell Tech's third talk in the panel, where he gave a general framework for auditing government processes and subsequently working with various government agencies to improve on those processes. Roughly, the pitch was to Audit agency decisions along the entire pipeline, from problem description to data collection to data maintenance to model building to decisioning. Issues of efficiency and equitability arise throughout the full pipeline, and do not exist in isolation; that is, upstream deployment decisions feed into the current module, and that module's outputs then impact downstream sections of the pipeline. As OR and AI practitioners, a particular focus here was on the challenge of modeling capacity-constrained decisions under various forms of uncertainty. The talk then went in-depth to an ongoing collaboration between Cornell Tech and New York City, looking to improve its 311-based method to crowdsource infrastructure incidents that require intervention from the city. Here, concerns of equity arise due to many forms of uncertainty (e.g., different areas of the city may have different reporting rates) and also allocative decisioning (e.g., areas close to city services may receive more effective treatment than those further away). A key takeaway from this talk was that focusing on fairness means comparing decisions for comparable incidents/groups – AI/ML tends to focus on analyzing individual incidents, whereas OR tends to focus on making global decisions across groups, so a responsible approach to fairness necessarily requires techniques from both fields' core competencies.

The chronologically second talk in the panel was given by Maria de-Arteaga of UT Austin, which provided a nuanced and in-depth discussion of some surprising side effects of blindly applying traditional fairness interventions. Specifically, the talk focused on identifying residual harms due to biased inputs when running fairness-aware algorithms, such as those that may impose a group fairness concern as an explicit constraint in an optimization problem. The example given was in the natural language processing (NLP) space, where group membership in a study was inferred by automatically "reading" written biographies of various academics. That inference process was shown to be biased on its own, with those who identified as a particular gender *and* who followed societal norms regarding writing were more accurately identified as their correct group than those who did not. Then, using a fairness-aware algorithm that enforces group fairness of some type – but across those biased group labels that have been inferred by some upstream process – may in fact disproportionately and systematically harm specific types of individuals within their true groups. A key takeaway, then, is that the risks of individual harm can certainly exist even under strict adherence to group fairness definitions and constraints.

Finally, David Schmoys of Cornell closed out the session with a discussion of his group's recent work on gerrymandering and fairness in redistricting. Gerrymandering, at a high level, is a method of "unnaturally" shaping political districts such that the resulting vote aggregated across those districts artificially over- or under-represents a particular party. Fair redistricting, then, seeks to combat this form of unfairness by way of choosing a partitioning of a geospatial region that holds to some standards. This can be seen, and is often modeled, as a set partitioning problem in its ideal form, with a variety of geographic/geospatial desiderata such as compactness and demographic or representational desiderata such as proportional representation. This concluding talk discussed methods for defining and also implementing forms of fairness in redistricting, and kicked off a vibrant post-panel discussion amongst participants.

---

## Networking Lunch

### Brief remarks from Pascal Van Hentenryck (AI4OPT) and Andrew Kahng (TILOS)

During lunch Prof. Pascal Van Hentenryck and Prof. Andrew Kahng gave the participants a brief overview of their respective AI Institutes, which have both received $20 Million dollar grants from NSF for a period of five years. Below are brief descriptions of these two

institutes, AI4OPT and TILOS, obtained from their websites and augmented with some comments from Pascal and Andrew.

**TILOS**

The TILOS mission is to make impossible optimizations possible, at scale and in practice. The institute's research will pioneer learning-enabled optimizations that transform chip design, robotics, networks, and other use domains that are vital to our nation's health, prosperity and welfare. TILOS is a partnership of faculty from the University of California, San Diego, Massachusetts Institute of Technology, National University, University of Pennsylvania, University of Texas at Austin, and Yale University. Many faculty members associated with TILOS are working on Fairness and Explainable AI, two of the themes of this workshop. TILOS is partially supported by the Intel Corporation.

**AI4OPT**

This NSF Artificial Intelligence (AI) Research Institute for Advances in Optimization aims at delivering a paradigm shift in automated decision-making at massive scales by fusing AI and Mathematical Optimization (MO), to achieve breakthroughs that neither field can achieve independently. The Institute is driven by societal challenges in energy, logistics and supply chains, resilience and sustainability, and circuit design and control. Moreover, to address the widening gap in job opportunities, the Institute delivers an innovative longitudinal education and workforce development program with an initial focus on historically black high schools and colleges in Georgia, as well as Hispanic-serving high-schools and colleges in California. The Institute is also developing internship programs with national laboratories and industrial partners, and is building a strong, welcoming, and inclusive community, highlighting social mobility opportunities and the societal impact of AI technologies. The focus is on fundamental and use-inspired research, with a goal of fusing machine learning and optimization by merging both the data-driven and model-based paradigms which are the hallmarks of the two areas of research.

Both of these institutes give junior and mid-level faculty the opportunity to lead by exploring the potential and challenges of use-inspired research in high-stakes domains. They also have strong collaborations with several industry partners listed on their websites.

---

## Panel B: Human Alignment/HCXAI/HCI

Moderated by Swati Gupta

**Slide Presentations**
Hamsa Bastani - Figure B-6: Decision-Aware Reinforcement Learning
Peter Frazier - Figure B-7: Preference Learning for Stakeholder Management
Kristian Lum - Figure B-8: De-biasing “Bias” Measurement
Mark Riedl - Figure B-9: Toward Human-Centered Explainable Artificial Intelligence

Panel B broadly consisted of decision making in various contexts which need to account for human factors, such as interpretability, ways in which models are explained, adaptability of the models, and changes from the status quo.

The panel started with Hamsa Bastani (Wharton) presenting her work with Sierra Leone National Medical Supplies Agency, where they focused on the distribution of essential medicines and managing inventory at different health facilities in the region, while navigating highly uncertain demands. Hamsa talked about aligning the loss function used to train the machine learning model with the decision loss associated with the downstream optimization problem. She interpreted the gradient of their loss function as a simple re-weighting of the training data, allowing it to flexibly and scalably be incorporated into complex modern data science pipelines, yet producing sizable efficiency gains. She shared results showing the decrease in unmet demand of essential medicines with the use of a decision-aware loss function versus a decision-blind loss function.

Next, Peter Frazier (Cornell and Uber) discussed understanding stakeholders and their constraints, in the context of COVID testing policies for Cornell University. Peter posed the question: Can AI help with tasks that OR practitioners need to do manually? For example, can AI help us understand stakeholder preferences? He advocated that Bayesian Optimization is a black-box derivative-free non-convex optimization method, which can be used for preference learning to get utility functions from stakeholders. The perspective of using AI to even understand the stakeholder preferences (which they may not be able to even quantify as functions or utilities themselves) was perceived as an interesting human-AI use case that has not received enough attention. The Cornell study showed that there is plenty of room for AI-enabled stakeholder engagement for OR applications. In particular, this approach helps one:

- Understand stakeholder goals, beliefs and incentives
- Understand how groups of stakeholders influence each other
- Predict how stakeholders will react to communication and
- Manage trust (in the OR analyst and her models).

The third talk in the session was given by Kristian Lum (Twitter), where she discussed the implications of checking for bias in terms of any algorithm's impact on different demographic groups. She postulated that while much of the work in algorithmic fairness over the last several years has focused on developing various definitions of model fairness (the absence of group-wise model performance disparities) and eliminating such "bias," much less work has gone into rigorously measuring it. She argued that many of the metrics used to measure group-wise model performance disparities are themselves statistically biased estimators of the underlying quantities they purport to represent. For example, the amount of "bias" measured increases as the number of groups increase, but this is just statistical noise. Lum proposed a "double-corrected" variance estimator to use instead. The key message of her talk was to question the statistical properties of various metrics for fairness used in practice and research, and that a lot of attention is needed to understand what we mean by fairness, inequity, and diversity. Meta-metrics cannot capture the entirety of the impact of ML systems. Small measured disparities should not be taken as a guarantee that the system is fair or free from adverse impacts.

Finally, we ended the session with a thoughtful talk by Mark Reidl (Georgia Tech), on explainable AI for consumer facing AI, and how different ways of communicating explanations themselves have a different impact on the users. For example, non-experts value contextual accuracy, awareness, strategic detail, intelligibility, relatability, and had unwarranted faith in numbers (they would trust the system more when numbers were used in the explanation). He claimed that our goal should not be to get people to trust the AI but to appropriately trust the AI via trust calibration. He discussed new research in explainable reinforcement learning and experiential explanations.

Overall the session touched upon various aspects of human-AI collaboration: how to interpret models for decision-makers, how to understand stakeholder preferences, how to even understand bias predictors, as well as the importance of designing good explanations to increase trust in AI.

---

## Breakouts for Panels A and B

*Breakout 1*
*Moderator: Jon Owen*
*Notetaker: Catherine Gill*

- **Human Interactions:** We discussed an interesting opportunity of expanding the conversation beyond AI and OR communities to include disciplines related to human decision making and psychology. Several motivating examples were considered, including: (a) the risk of introducing bias through the questions we ask and how we ask them when forming our understanding of a system (e.g., OR modeling) or executing market research that guides modeling and provides input data, (b) the evaluation and verification of results and outcomes, and (c) our effectiveness towards influencing actions through AI/OR analyses and the communication of results and implications – including human-in-the-loop decision making. It was mentioned that there are 3 places where humans come in: early, like choosing/inputting data; in the middle, where decisions get made; and the end, in evaluation and verification. At the end of the day, we want recommendations to be robust and to make sense to the decision maker.

- **Public Perceptions of AI:** We also discussed public misperceptions about AI due to hype. It was noted that many naïve users view AI as almost “magical”, and that it will always be better at solving problems than humans. This led to discussion around (a) the need for a better understanding of these techniques as tools, not solutions, and (b) greater transparency around embedded assumptions and their potential implications on interpretation of results, including the role of bias.
- **Data Quality:** We discussed several points related to data and notions of data quality. Often you don’t know the quality of data, especially when underlying data generating functions are poorly understood, or a specific use case for the data differs from the original intent (e.g., when data is generated/collected for one purpose but used for another). With these limitations, the need for data learning was discussed, as well as an observed gap between recording state actions without recording the associated probabilities.
- **AI “Intelligence” and Automation:** The concept of AI “intelligence” was briefly discussed as both an aspiration and a moving target based on our expectations for automation. Should we care about automating everything through AI/ML? Also, there are limits for data-centric approaches moving beyond descriptive and predictive; for example, prescriptive recommendations require going beyond the data to be useful – this is an opportunity for AI and OR to combine.

*Breakout 2*
*Moderator: Harrison Schramm*
*Notetaker: Haley Griffin*

- **Fairness and Bias:** We discussed fairness and the many ways in which bias can be introduced to a program. By attempting to limit the bias in a program a researcher may actually introduce more bias. Trying to limit unwanted confounding variables may result in detrimental unforeseen consequences, which can actually increase the unfairness of a program. Also, by reducing the bias of a program for the majority of users, you may disproportionately increase the unfairness for minority groups of users, who may be the very people who need fairness protections the most.

- **Explainable AI:** We talked about explainable AI, and the need for explainable OR as well. Some participants took issue with the current state of explainable AI, insisting that it is overly vague, and the outputs of AI programs give users no information as to how to interpret these outputs. These participants advised creating more programs to analyze and interpret the results of AI programs to better inform users how an AI program came to a certain conclusion, rather than just returning an output with no explanation.

*Breakout 3*
*Moderator: Theodora Chaspari*
*Notetaker: Ann Schwartz*

- **Fairness**: Efficiency and fairness are defined differently throughout all systems. Participants discussed whether any general fairness rules can be applied to all models. A clear and globalized definition of fairness would clarify the metrics used to measure it, and improve the fairness and user trust of a system.
- **Explainable and Trustworthy Technology**: Trust of a system is difficult to measure, as there are no clearly defined and widely accepted metrics to do so. We discussed how much information users need to understand a system, and how AI should be made explainable. AI should be able to explain its decision making similarly to how a doctor explains a condition to a patient. A doctor does not need to explain down to the cellular level why a patient may be suffering from a condition, but the doctor does need to explain what may have occurred for this condition to be present in the patient (lifestyle choices, hereditary conditions, etc.). Similarly, an AI program does not need to explain why a certain weight was assigned to each node, but rather why an accumulation of weighted nodes is sufficient to make a certain decision. While some researchers have historically argued that people don’t require explanations but just want outcomes from AI, sometimes it is required especially to build trust in a system. There is a need for an established formal process to determine the trustworthiness of a system.

*Breakout 4*
*Moderator: George Lan*
*Notetaker: Jai Moondra*

- **Causality methods across disciplines**:

- Statistics - focus is on estimator properties and optimality.
    - CS - focus is on scale and data.
    - OR - focus is on engineering models and experimental data.
- An important question is how experiments can be controlled: there may be policy and regulatory questions and network effects in industry that prevent random control experiments. For example, similar prices **must** be shown to everyone in cab-sharing. Another question is how to integrate real data with simulation data? In some situations, historical data can be used to design experiments to generate synthetic data. But certain settings like wind turbines pose difficulties to generate synthetic data.
- **Major challenges in XAI / Causality and possible ways in which a multi-disciplinary approach can help:** There are several examples in social sciences where causality helps. One challenge is that policy makers make decisions hoping to influence outcomes, but they sometimes don't accomplish this because they don't understand or use causality. How can they be helped with causal inference? Practical factors like operational costs, etc. matter. Causal inference in decision making can again be used here. For example, housing. Changing leadership and policy makers can focus on areas other than long-term outcomes (they can base their decisions more on politics). Educate them that politics can be done using causality and decision-making. How do we align policy makers' incentives with producing the best outcomes? Also we must bear the burden of explaining our mechanisms to them.
- **Practical Considerations for Causal Inference Methods**:
    - It is sometimes difficult to conduct experiments (ex: gene editing and other medical situations)
    - Regulatory approvals can delay research
    - Confidence needs to be high – repeatability and reproducibility are key
    - Sometimes correlation is the best that can be shown
    - Need a large and representative sample - sometimes there is the risk of some researchers falsifying data due to the small amount of data available
    - **Chains of causality get complicated quickly.** If A implies B which implies C which implies D, conditional effects of A on D are not so easy to understand.

- **Potential/emerging application areas: XAI and Causality**
    - Policy design
    - Medicine
    - Engineering
    - Cybersecurity
    - Social science - interventions

- Security and intelligence
- Sustainability - protecting parks and preventing poaching (in Africa for instance)
- Education

---

## Panel C: Robustness/Privacy

Moderated by Bistra Dilkina

**Slide Presentations**

Bo Li - Figure B-10: Trustworthy Machine Learning: Robustness, Privacy, Generalization, and Their Interconnections
Kush Varshney - Figure B-11: Problem-Driven Robustness, Privacy, and Fairness
John Abowd - Figure B-12: Some Lessons from the 2020 U.S. Census Disclosure Avoidance System

The first presentation was given by Bo Li on the topic of Trustworthy Machine Learning. Bo gave several examples of the dangers of Machine Learning, from hacking attempts, such as the Associated Press hack which crashed the stock market in 2013, to the public's concern with using ML programs, such as biometric recognition being used at airports. The goal to improve these Machine Learning programs, as stated by Li, is to close the trustworthiness gap, which can be accomplished in a number of ways.

Firstly, improved Robustness can increase public trust in an ML program simply by making it functional in more situations. Increased robustness not only increases trust in a program by the public, because the program can react correctly to more unexpected inputs, but it also increases the usefulness and accuracy of the program, since the number of situations the program cannot handle is reduced. Generalization, namely, a program's ability to adapt to new and variable data, can also improve trustworthiness. Finally, increased privacy improves trustworthiness, by reducing the risk of a user's data being accessed by a third party. Dr. Li advocated for a holistic approach to improving trustworthy ML, by tackling these three problems simultaneously and in concert.

The second presentation was by Kush Varshney on the topic of Problem-Driven Robustness, Privacy, and Fairness. Varshney started with describing a few problems in the Healthcare industry, starting with the Patient Protection and Affordable Care Act which changed the landscape of the health insurance market in the United States. Insurance companies had to decide which new markets to enter. Markets are defined by geography, age group and other factors. Some considerations were:

- The desire to enroll low-cost (healthy) people to enroll in their plans
- Preventing from accepting or denying enrollment on an individual basis
- Whether or not to offer plans in well-defined markets
- How to use data-driven decision making for determining whether or not to offer plans in new markets

Some of the challenges in this problem included the fact that insurance companies needed cost and demographic data on people who will enroll in new markets whereas they only have this for those enrolled in existing markets. Given that the target domain is unknown, one needs distributionally robust methods. Varshney described a few approaches to solve this problem, including a game theoretic formulation for invariant risk minimization.

Next, Dr. Varshney discussed the topic of privacy concerns in the context of the Health Insurance Portability and Accountability Act which includes the mandate of privacy protection even for health insurance companies' internal planning purposes. He cited that k-Anonymity is a common mathematical interpretation of the privacy condition and one of the approaches is to use k-member clustering (grouping the records so that the smallest group has at least k elements). He briefly described Distribution-preserving k-Anonymity for transfer learning which is an alternative to standard clustering to allow the resulting data to follow the distribution of the original data.

Varshney also discussed the critical need for algorithmic fairness in the health care insurance industry where often health care cost is used as a very poor proxy for health care needs because it could lead to racial discrimination.

Dr. Varshney ended his presentation with some thoughts on Operations Research + Artificial Intelligence.

- He called for a shift in the way AI problems are done which typically use a problem-driven approach. He recommended the incorporation of model-based approaches and called for cross-fertilization with risk management, probability theory, robust optimization, audio signal processing and game theory.
- He commented that there is a lot of similarity between the notion of Explainability and Robustness – they address the same basic problem with different approaches.
- He posed several interesting questions related to OR and AI: Why do we have many discussions on trustworthy ML but not on trustworthy OR? Is it because trustworthiness is not important to OR practitioners or do they do this under a different name (with robustness, sensitivity analysis, etc.)? Is it due to different audiences that consume these results with the OR users being experts within organizations while ML is deployed to non-technical experts? Or is it because there is more media exposure (negative or otherwise) on AI applications and not as much on OR ones?

The final presentation in this session was by John Abowd. Dr. Abowd spoke about the 2020 Census, during which a total of more than 150 billion statistics from 15GB of total data were captured, and the precautions that were taken to secure the data of those who were interviewed. These precautions were necessary to adhere to the increasing privacy protection regulations in the US as across many other parts of the world. After the 2010 Census collected and processed its data, the tabulations were released to the public. Shortly afterwards, it was discovered that the confidential microdata from that census could be accurately reconstructed from the publicly released tabulations. Geographic identifiers were associated with every microdata point, meaning that those who responded to the census could be very accurately identified across the United States, especially those in rural areas.

To prevent this inadvertent disclosure of confidential data from reoccurring during the 2020 Census, the Census team took a number of precautions, such as adhering to a formal privacy protection framework written for the purpose. In particular, a TopDown Disclosure Avoidance System was developed with strict requirements with regard to formal privacy protections. Details are available in the presentation slide deck included with this report.

Some of the key takeaways from John's presentation are:

- Going from suppression to differential privacy is much easier than going from publishing all the microdata to differential privacy.
- 2020 Census data clients had accuracy expectations that modern privacy protection can't support (the 2010 Census basically released all the microdata, although not intentionally).
- It is safe to forecast that AI applications, particularly in industry, are going to face the same conundrum. For instance, advertising executives are not going to like the privacy-protected models (Conventional AI applications are inherently disclosive.)

---

## Wrap-up of Day 1

Ramayya Krishnan

Throughout Day 1, several big themes were discussed pertaining to the topic of Trustworthy AI, including humans in the loop, preference elicitation, robustness, explainability, privacy, etc. Fairness issues were presented in the context of multiple settings such as fairness in healthcare and multi-stakeholder fairness in rideshare scenarios like Uber/Lyft. Several ideas were presented that were worthy of pursuing collaboratively between the OR and the CS communities. Krishnan called upon the participants to write down sketches of these ideas for future collaboration between the two

groups – so that we could discuss them briefly in the final session on Day 2. The intention was to provide a platform to bring researchers together for joint projects.

The participants had robust discussions regarding the boundary between OR and AI: is it conceivable that the differences between the two areas will vanish in the near future or will they persist due to cultural differences? Some of these differences were already discussed in the first workshop and will need to be overcome for fruitful collaborations to emerge. For example, some barriers stem from the culture of conference papers for CS versus journal articles for OR. Is it possible to borrow ideas from each other's preferred mode to enable quicker dissemination of ideas between the two groups? Are there opportunities to co-locate meetings to bring together the AI and OR communities?

## Day 2

John Dickerson kicked off Day 2 with a summary of the presentations on Day 1 and asked Prof. Sven Koenig to share some of his recommendations with regard to opportunities for collaboration.

Sven talked about competitions in computer science which have sparked a great deal of interest among researchers, including graduate students exploring new techniques in the field. Some examples of competitions in CS are:

- Robocup logistics competition conducted by the RoboCup Foundation
- NeurIPS 2020 Flatland Competition
- Several competitions advertised by ICAPS 2021 including
    - Learning to run a power network with trust
    - Automatic reinforcement learning for dynamic job shop scheduling problem
    - The dynamic pickup and delivery problem

Several teams participate in such competitions and often the winning teams employ optimization techniques. Sven posed the question: How can OR create challenge problems through their groups? How can both communities collaborate as teams competing in such competitions?
He recommended that we create challenge problems that would spark interest in such collaborations.

Lavanya Marla proposed a sketch of how we could devise challenge problems that would bring together experts in both fields. Some of her suggestions include:

- Classical OR resource allocation problems which have interesting aspects for both groups:
    - Start with classical OR problem

- Resources have preferences/rules that they state - AI models learn rules and preferences
    - OR model makes decisions with added side-constraints
    - Iterate

- Some examples with rich data arise in hospital staffing, airline crew scheduling, semiconductor manufacturing, transportation, etc. The challenges are to get the right data and provide an appropriate infrastructure for solving such problems.

See the concluding section of this report for some challenge problems that were discussed at the end of the workshop.

---

## Panel D: XAI/Causality

Moderated by Yu Ding

**Slide Presentations**

Zachary Lipton - Figure B-13: Adapting Predictors under Causally Structured Distribution Shift

Ruoxuan Xiong - Figure B-14: Design and Analysis of Panel Data Experiments
Yu Ding - Figure B-15: Causal Inference in Engineering Applications

The theme of this session was on Explainable AI (XAI) and Causality, although the three talks in the session focused primarily on causality and less on XAI. Admittedly, insights gained from understanding causality helps provide explanation to AI methods. The three talks in this session come from three different angles---one talk was given by a computer science (CS) researcher, one by an operations research (OR) researcher, and one on an application of causal inference by an engineering researcher. Altogether the session provides a balanced view on causality research combining both AI/OR perspectives.

The CS talk by Zachary Lipton raises the issue of adapting predictors under causally structured distribution shifts. The speaker discussed the anatomy of a structured shift problem in the context of domain/environments, structure, visibility, manipulation rules, objective, and statistical capabilities, presented examples of structured shift, such as covariate shift, label shift, or missing data shift (source and target data missing at different rates), and stressed the challenges in handling high-dimensional, arbitrarily non-linear data.

The OR talk by Ruoxuan Xiong discussed the design and analysis of panel data experiments in which conventional A/B testing suffers from network interference or contamination effect. Such problems commonly exist in the experiments run by certain ride sharing companies for testing whether a new feature would improve driver participation rate. A switchback experiment may help but arbitrary switches between control and treatment run into impracticality constraints; for instance, it is not practical for drivers to see different versions of their app every day. The speaker contemplated on a solution using a linear mixed effect model with integer switching variables that are solved optimally through integer programming.

The application talk by Yu Ding presented a success story in which causal inference methods were used for estimating the effects of certain technical upgrades on wind turbines. Both issues of effect estimation and deciding conditional causal relation were discussed. The classical covariate matching method was tailored for estimating turbine upgrade effect in good accuracy, with the caveat of additional needs for bias reduction. The determination of the conditional causal relation is complicated by the presence of autocorrelation in the time-series data, another type of distribution shift different from but complementing the distribution shifts discussed in the CS talk by Zach Lipton. Causal inference methods are shown to make a great positive impact on an important renewable energy application.

---

## Breakouts for Panel D

*Breakout 1*
*Moderator: Xiao Fang*
*Notetaker: Catherine Gill*

- We discussed how AI and OR can contribute to causal inference from observational data and randomized experiments. One significant challenge of causal inference is the interference effect, which refers to the effect of people affecting each other (e.g., people in the treatment group communicating and affecting those in the control group). This challenge can be addressed by panel data experiments, for which OR methods can help decide optimal experiment parameters (e.g., treatment time) and reinforcement learning methods can be employed to improve treatment efficacy.
- We also discussed high-stakes applications of causality and how AI/OR can help. These include quick decisions such as driving decisions made by autonomous vehicles and longer-term decisions such as college admission decisions. We further discussed the implementation of a policy and how to gather data to understand the causal mechanism underneath the policy. For example, a tax policy is implemented at some sites for testing and survey data is then collected from its affected citizens to understand the causal mechanism underneath the policy. However, we have difficulties in getting truthful survey data because humans are incentivized to lie in surveys. Hence, there might be research opportunities for AI/OR scholars: how to design surveys that solicit truthful data for causal inference. In some other situations, we have to infer features on the basis of surrogate information; for instance we may look at variables which indicate poverty instead of actual poverty statistics. Are there methods to address this issue?

*Breakout 2*
*Moderator: Abhishek Chakrabortty*
*Notetaker: Haley Griffin*

- **Machine Learning and Causality Synergies**: Machine learning methods can be very helpful in guiding decision making, but to make truly informed decisions, we have to understand the causality involved.  On the other hand, in cases where you have many confounding factors, some of the ML approaches are better able to deal with those factors; so one can take ML approaches and wrap them in causal methods. AI/OR literature should approach the problem of explainable AI similarly to how a doctor treats a patient; identify the problem (the part of a program which is unexplained), prescribe a treatment (an explanation of the causality), and monitor the recovery (keep an eye on the improved program).

- **Empirical Validation Limited by Data**: There are no guarantees in machine learning outcomes. Researchers can validate them empirically, but it is limited by the data set. It is wrong for researchers to use models outside of the domain they were intended for and extrapolate data that may not be correct. It can work out, but oftentimes it doesn't serve the same purpose. When working with end users and funding organizations, researchers should always check to make sure they can rationalize what they are doing. The researcher might have to point out insights they didn't consider, and by fleshing them out they can figure out if their vision is realistic. The greater the use of observational data the better.

- **Need to be Careful before Attributing Causality**: When you are choosing an independent variable for a causal method you are making a choice, and you have to disregard a lot of variables. Researchers should not claim causation until they have made adjustments throughout their research that prove there is a causal relationship rather than just an association. Eliminating confounding variables in attempting to make a fair world may instead lead to a biased one; adjusting to the world requires careful consideration of causality. This can be key when facing complicated optimization problems. The solutions must be data-driven.

*Breakout 3*
*Moderator: Amanda Coston*
*Notetaker: Ann Schwartz*

- **Causality, XAI, and trust:** We discussed the often complicated relationships between causality, explainable AI, and users' trust in AI. We discussed how explanations can engender trust even when they are not causal, and we also debated how causality may improve the fidelity of explanations. We identified a major risk with XAI: explanations may engender misplaced trust in algorithms, and to guard against this, we discussed the importance of listening to and addressing users' concerns with AI.

- **Grand Challenge:** We identified a possible grand challenge that poses the question: What modalities are needed to empower us to determine whether (or not) to trust a data-driven model? How do we create tools and methods to use for this purpose?

*Breakout 4*
*Moderator: Berk Ustun*
*Notetaker: Cyrus Hettle*

This breakout group primarily discussed various aspects of designing good challenge problems and potential application areas for them, as well as suitable topics and participants for the next workshop.

- **Designing Good Challenge Problems:** We discussed how to design challenge problems that could be used to showcase the value of AI in OR and vice-versa. Participants agreed that designing the kinds of challenge problems would require "more than just identifying a real-world problem." One key issue was to strike a middle ground between the kinds of challenge problems across fields. In OR, challenge problems stem from clients, and may yield insights that are "too specific." In AI, challenge problems are too "stylized," which means they produce insights that are "too abstract." A second issue we discussed was developing measurable and well-motivated evaluation criteria (i.e., metrics that can be used to evaluate solutions along with explanations for why these metrics are useful for a given application).

- **Potential Applications for Challenge Problems:** Potential for good challenge problems include: transportation, lending, and hiring. In particular, transportation problems tend to foster interdisciplinary work, for instance, the large interdisciplinary teams working on these problems within companies. COVID data, such as the challenges faced when rolling out vaccines, is multi-dimensional and could be a valuable source of problems. In contrast, while the COMPAS data set is widely used in fairness work, it can be politically fraught and challenging to separate from baseline issues with the prison system, and we would prefer something more neutral. COMPAS is a landmark dataset to study algorithmic fairness. This dataset was used to predict recidivism (whether a criminal will reoffend or not) in the US. We discussed an example of work with an Alabama criminal justice organization which was presented as an example to some CEOs with the framing of "although this topic is political, the general problem occurs in lending, hiring, etc.”

- **Evaluation Methods for Challenge Problems:** In addition, we discussed evaluation methods for challenge problems, which can be a critical step in design (“90% of challenge formulation work is formulating the metric”). Laying out evaluation criteria very clearly (a strength of OR), including quantifying dimensions and specifying correctness criteria, is beneficial. Even giving logical explanations for the criteria could be as valuable as running a challenge. Doing this for 10-12 different examples could help people relate to various prototype scenarios. We discussed some other frameworks for multi-objective problems, including allowing solvers to choose a subset of the objectives or specifying the objectives, but concealing their relative weights. A comparison was made to kidney transplants, where people propose multiple objectives and debate their relative merits.

- **Fairness Problems.** We discussed two broad categories of fairness problems, of which we saw two very different examples in the earlier sessions. Peter Frazier’s work on fairness in ridesharing was a multi-stakeholder problem involving riders, drivers, the community, and regulators. In contrast, Dmitris Bertsimas’s scarce resource allocation problem on racial differences in post-acuity care compared two groups with similar objectives, with an interesting tension for how to assign costs for people who don’t receive care. The timescale of the impact of decisions can also have an effect (immediate vs. several years) or can stretch out over a period of time (changes in retail price, privacy and fairness issues in the Census). It would be useful to study the commonalities and differences between problems in different settings, as the Privacy Forum has done for similar kinds of harms in different applications.

- **Real-World Impact:** Our session discussed how solving "real-world" problems could have a "real-world" impact. In effect, it can be easy to write papers but hard to have an impact outside of the research ecosystem. We highlighted the need to engage with a broader community of stakeholders – e.g., domain experts, practitioners, and policymakers. For example, sometimes there are disconnects between lawyers and policymakers, both of whom can have important input into constraints and utility for algorithms. This broader engagement is important as it can help us identify relevant problems and solve them correctly.

- **Some Ideas for the Next Workshop:** We highlighted the need to actively recruit individuals from broader communities for future events (e.g., we could ask experts from the federal government to present problems at the next workshop). Bringing in communities and policy beneficiaries in addition to agencies should also be done, though it adds complications. We discussed some examples of crossovers and what we can take from them (Gale-Shapley, Sheldon Jacobson's work on TSA Pre-Check, FCC actions) and the idea of putting something in place at the regulatory level to provide a mechanism for evaluating the cost of solving problems, not just the utility. We also discussed fostering effective collaboration with these stakeholders outside of AI and OR. Specifically, we need to work effectively when eliciting their preferences and constraints to develop technical solutions, and describing the benefits/limitations of these solutions (i.e., promote the adoption and responsible use).

---

## Closing Session and Challenge Problems

In our concluding discussions, we brainstormed solutions with the intention of uniting AI and OR researchers and pushing the needle towards a more collaborative practice of both subjects. These challenge problems are discussed in further detail below:

We discussed the disparities in fairness between users of ridesharing services, specifically in the city of New York. Transportation Network Companies (TNC's) such as Uber, Lyft, and taxi services have been observed to provide less reliable service in historically disadvantaged areas, because these areas usually have lower rider demand leading to longer wait times between trips for drivers. To rectify this inequity, financial incentives need to be put in place to encourage more drivers to travel to these areas. Who pays for these incentives, however, is the main point of contention, whether it be the rideshare riders or the TNC's themselves. To calculate who optimally ought to pay at any given time, and how to best allocate drivers to riders in this multi-agent setting and balance multiple objectives, we propose a competition be created with teams vying to develop the most efficient, equitable, and trustworthy algorithm combining AI and OR practices.

We also recommend creating a summer program, convening Artificial Intelligence and Operation Research experts to educate Ph.D. students on multidisciplinary training in AI and OR. This program would take place annually, similar to Machine Learning summer workshops which have taken place in the past. The summer program could focus on a number of topics on which research has been done in both AI and OR (such as decision making under uncertainty, local search, etc.) and invite both an AI and an OR speaker for each topic. This way, the Ph.D. students understand the techniques that have been developed in both AI and OR as well as their commonalities and differences. The Ph.D. students could work in interdisciplinary teams to solve a challenge problem that requires the integration of AI and OR techniques, to allow these students to overcome some of the challenges listed above.

We also put forth the idea of a joint AI and OR conference on decision making for intelligent robots. OR often uses its optimization techniques to support human decision makers (in business settings), while AI often focuses on autonomous decision making by agents. Robots, for example, must plan their motions and tasks, both individually and as a team, which are independently complex optimization problems but also need to be coordinated. To create robots which are as efficient, accurate, and coordinated as possible, a convergence of AI and OR expertise will be required.

Finally, we suggest another challenge problem which may demand a joint AI/OR approach. The setting is scheduling of nursing and physician staff in healthcare facilities. Healthcare delivery settings have seen significant fluctuations in demand patterns from pre-COVID times, to during COVID and post-COVID, resulting in increased variability of patterns for emergency care, out-of-hospital care, and clinic care. The classical OR version of this problem is an open-loop problem, that is, resources (staff) are assigned to rosters to meet (deterministic or stochastic) demand. However, in reality the process is often iterative, that is, staff often have changing priority over days or weeks of the roster, along with varying demand, which may require that the schedule be re-calculated dynamically to make it human-friendly. To better predict healthcare demands while creating optimal and fair schedules for healthcare staff is a challenge that could be addressed by AI and OR approaches. To view the Challenge Problems document from the workshop, please use the link provided.

By pursuing each, or any of the proposed solutions above, we believe we can work towards closing the existing gaps between the fields of Artificial Intelligence and Operations Research. Increased collaboration between researchers in both of these fields over time will lead to an erosion of the disperate lexicons both groups use when referring to the same practices, and will lend new expertise to each practice, from which both fields can benefit.

# Appendix A: Pre-Workshop Materials

## Figure A-1: Workshop Participants

| First Name | Last Name | Institution |
|---|---|---|
| John | Abowd | U.S. Census Bureau |
| Hamsa | Bastani | University of Pennsylvania |
| Dimitris | Bertsimas | Massachusetts Institute of Technology |
| Tracy | Camp | Computing Research Association |
| Abhishek | Chakrabortty | Texas A&M University |
| Theodora | Chaspari | Texas A&M University |
| Amanda | Coston | Carnegie Mellon University |
| Tapas | Das | University of South Florida |
| Maria | De-Arteaga | University of Texas Austin |
| John | Dickerson | University of Maryland |
| Bistra | Dilkina | University of Southern California |
| Yu | Ding | Texas A&M University |
| Xiao | Fang | University of Delaware |
| Peter | Frazier | Cornell University |
| Cyrus | Hettle | Georgia Institute of Technology |
| Nikhil | Garg | Cornell Tech |
| Cat | Gill | Computing Community Consortium |
| Haley | Griffin | Computing Community Consortium |
| Swati | Gupta | Georgia Institute of Technology |
| Andrew | Kahng | University of California San Diego |
| Subbarao | Kambhampati | Arizona State University |
| Sven | Koenig | University of Southern California |
| Ramayya | Krishnan | Carnegie Mellon University |
| Radhika | Kulkarni | SAS Institute, Inc. (retired) |
| George | Lan | Georgia Institute of Technology |
| Bo | Li | University of Illinois at Urbana-Champaign |
| Jing | Li | Georgia Institute of Technology |
| Zachary | Lipton | Carnegie Mellon University |
| Daniel | Lopresti | Lehigh University |
| Kristian | Lum | Twitter |
| Lavanya | Marla | University of Illinois at Urbana-Champaign |
| Jai | Moondra | Georgia Institute of Technology |
| Murat | Omay | U.S. Department of Transportation |
| Jon | Owen | General Motors |
| Mark | Riedl | Georgia Institute of Technology |
| Harrison | Schramm | Group W |
| Ann | Schwartz | Computing Research Association |
| Thiago | Serra | Bucknell University |
| David | Shmoys | Cornell University |
| Alice | Smith | Auburn University |
| Berk | Ustun | University of California San Diego |
| Pascal | Van Hentenryck | Georgia Institute of Technology |
| Kush | Varshney | IBM Research |
| Phebe | Vayanos | University of Southern California |
| Cathy | Xia | Ohio State University |
| Ruoxuan | Xiong | Emory University |
| Jerry | Zhu | University of Madison-Wisconsin |

## Figure A-2: Workshop Agenda

**August 16, 2022 (Tuesday)**

| | |
|---|---|
| **07:30 AM** | **NETWORKING BREAKFAST \| Conference B** |
| **08:30 AM** | **Welcome and Introductions \| Conference A** |
| **09:10 AM** | **TBD: Brief comments from funding agencies about opportunities for AI funding \| Conference A** |
| **09:30 AM** | **Panel A: Fairness \| Conference A**<br><br>**Dmitris Bertsimas, Massachusetts Institute of Technology**<br><br>**Maria De-Arteaga, University of Texas at Austin**<br><br>**Nikhil Garg, Cornell Tech**<br><br>**David Shmoys, Cornell University** |
| **10:45 AM** | **BREAK \| Outside Conference A** |
| **11:00 AM** | **Breakout A \| Conference A, C, D, and E** |
| **11:45 AM** | **Report Back A \| Conference A** |
| **12:00 PM** | **NETWORKING LUNCH \| Conference B**<br><br>**Brief remarks from Pascal Van Hentenryck (AI4OPT) and Andrew Kahng (TILOS)** |
| 01:00 PM | Panel B: Human Alignment/HCXAI/HCI \| Conference A<br><br>Hamsa Bastani, University of Pennsylvania<br><br>Peter Frazier, Cornell University and Uber<br><br>Kristian Lum, Twitter<br><br>Mark Riedl, Georgia Institute of Technology |
| 02:15 PM | Breakout B \| Conference A, C, D, and E |
| 03:00 PM | Report Back B \| Conference A |
| 03:15 PM | BREAK \| Outside Conference A |
| 03:30 PM | Panel C: Robustness/Privacy \| Conference A<br><br>Bo Li, University of Illinois at Urbana-Champaign<br><br>Kush Varshney, IBM<br><br>John Abowd, US Census Bureau |
| 04:45 PM | Breakout C \| Conference A, C, D, and E |
| 05:30 PM | Report Back C \| Conference A |
| 07:00 PM | NETWORKING DINNER \| Lure, 1106 Crescent Ave NE, Atlanta, GA 30309 |

**August 17, 2022 (Wednesday)**

| | |
|---|---|
| **07:30 AM** | **NETWORKING BREAKFAST Day 2 \| Conference B** |
| **08:30 AM** | **Recap Day 1 \| Conference A** |
| **09:00 AM** | **Panel D: XAI/Causality \| Conference A**<br><br>**Yu Ding, Texas A&M**<br><br>**Zachary Lipton, Carnegie Mellon University**<br><br>**Ruoxuan Xiong, Emory University** |
| **10:15 AM** | **BREAK \| Outside Conference A** |
| **10:30 AM** | **Breakout D \| Conference A, C, D, and E** |
| **11:15 AM** | **Report Back D \| Conference A** |
| **11:30 AM** | **Bringing it all Together \| Conference A** |
| **12:30 PM** | **NETWORKING LUNCH Day 2 \| Conference B** |
| **01:15 PM** | **Report Writing \| Conference A** |
| **02:15 PM** | **End of Workshop** |

# Appendix B: Workshop Slides

## Figure B-1: Murat Omay - Overview of ITS JPO Programs

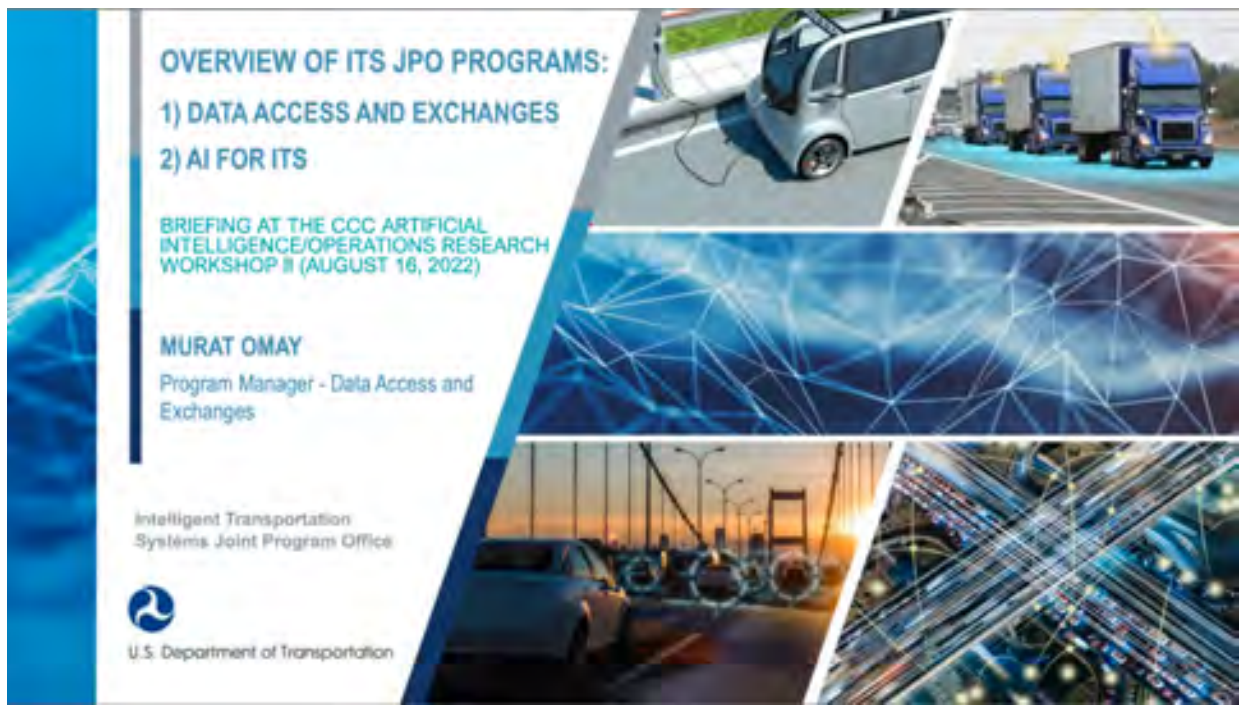

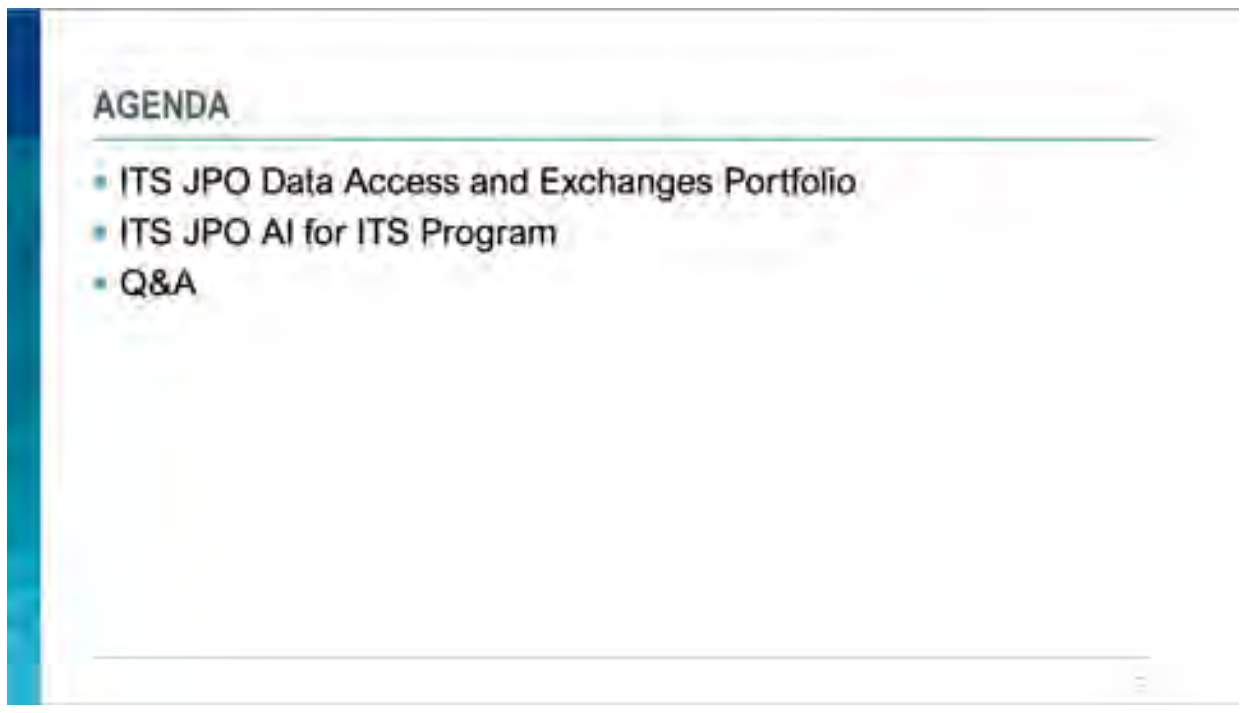

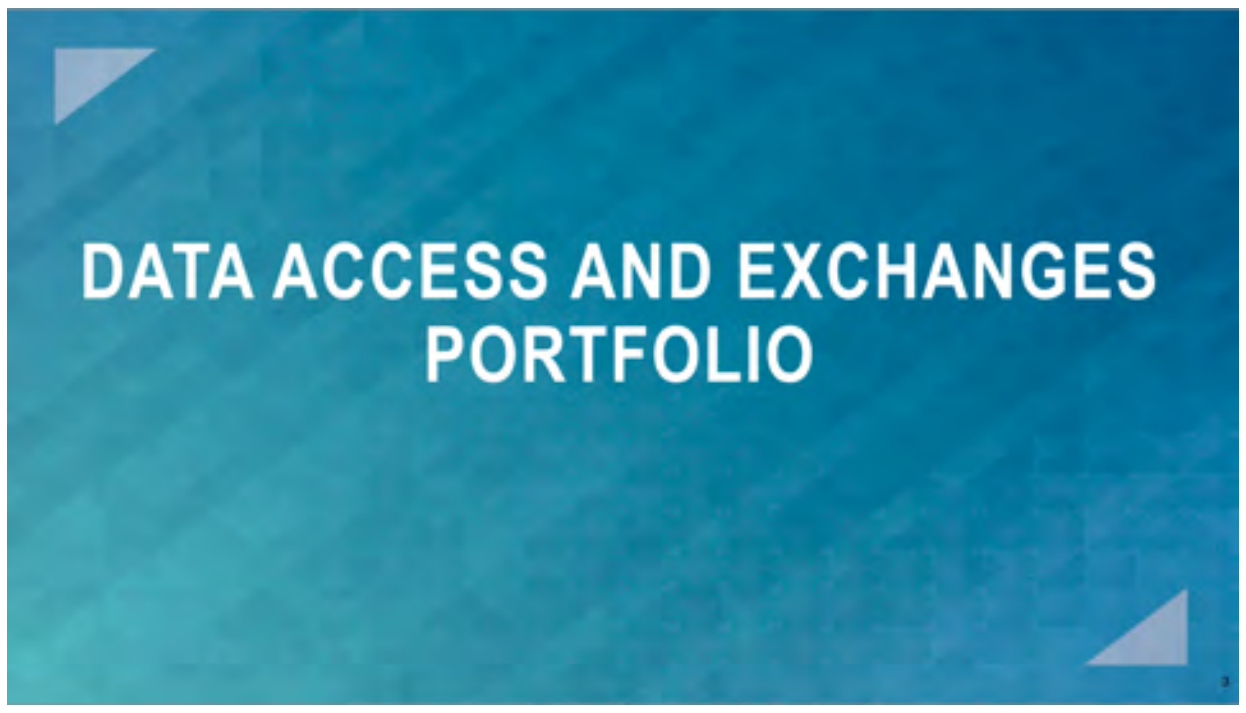

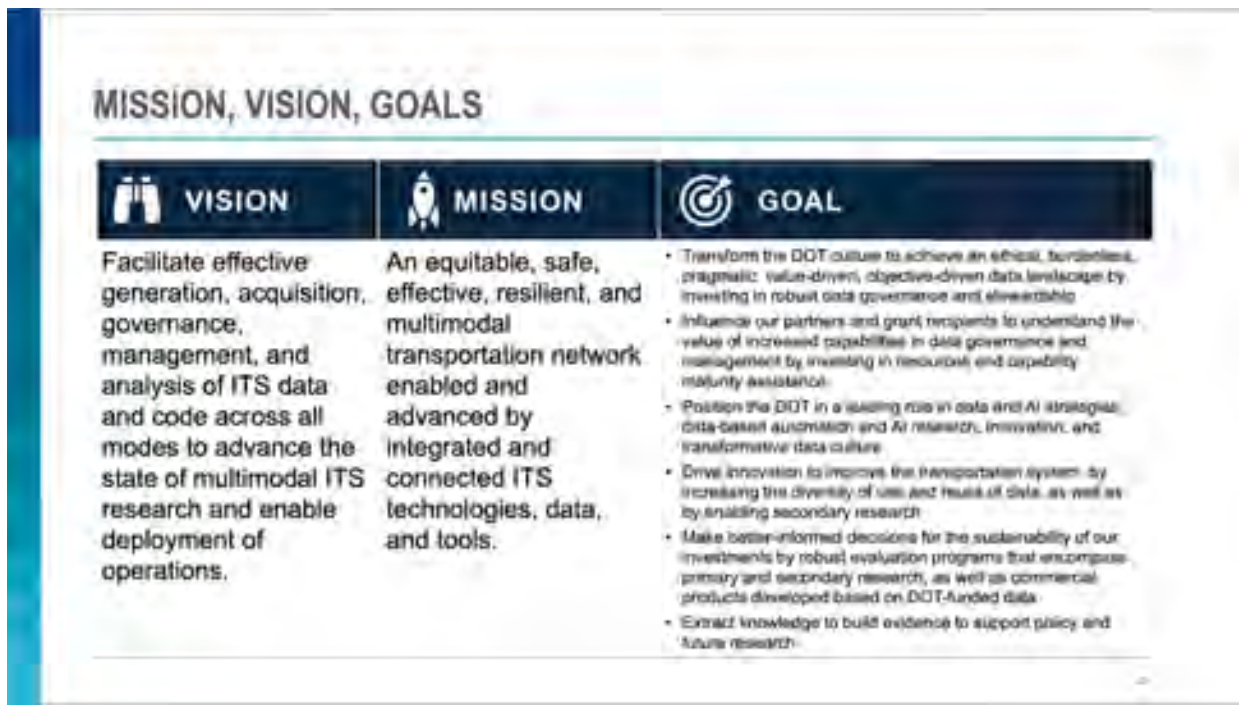

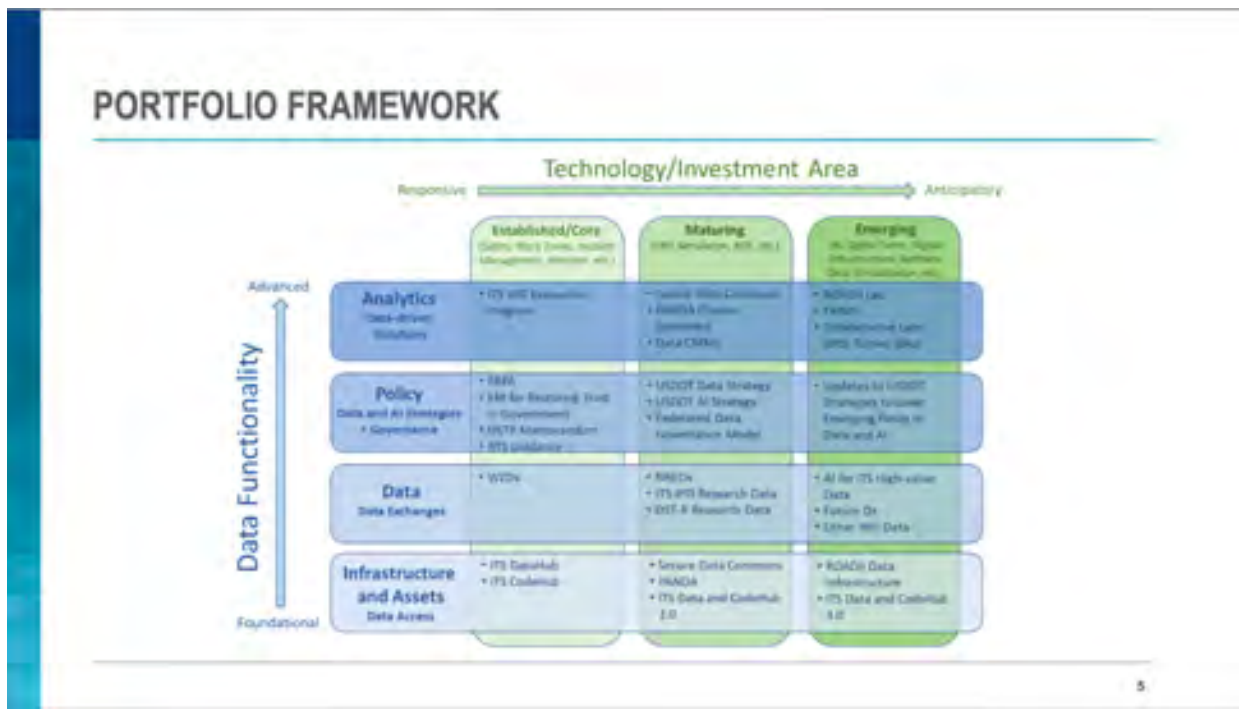

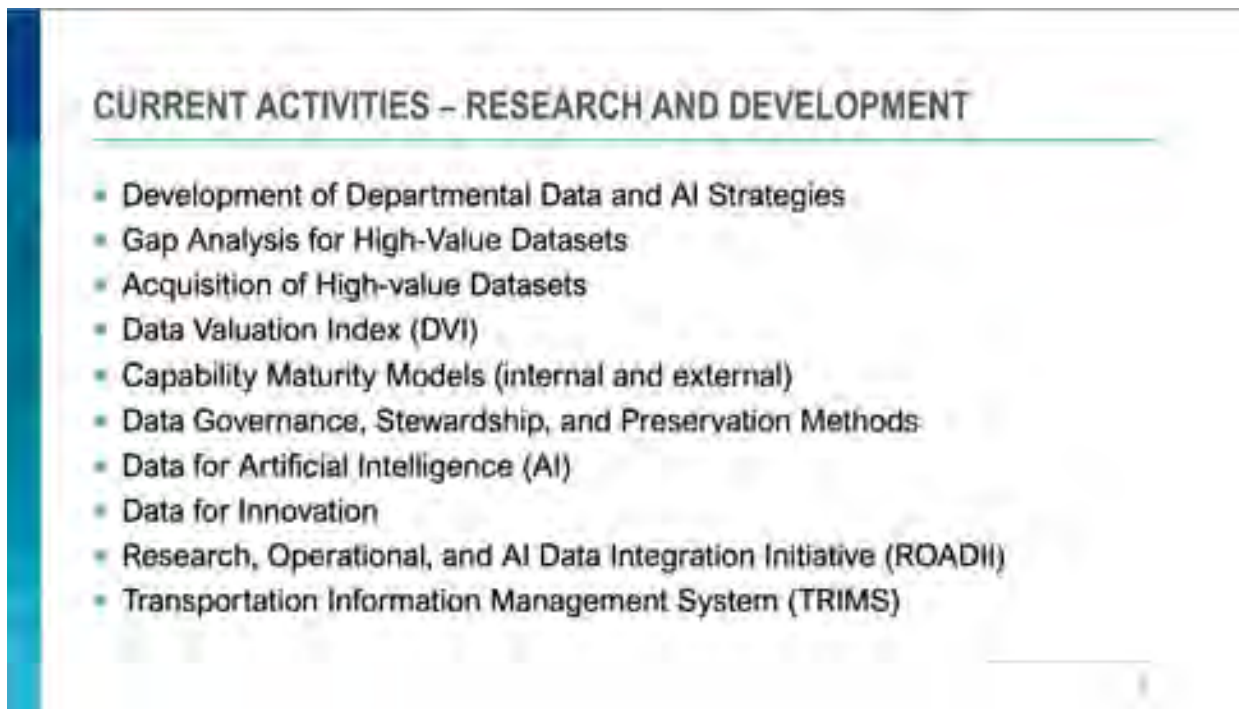

## CURRENT DIALOGUES

- Data Strategy Working Group led by USDOT OST-R/BTS
  - Goal: Using the NOFO Language as a steppingstone, develop a robust data culture within the DOT, its partners, and its grantees
  - Objective: Capture data governance and management objectives in a common language to be included as requirements in pre-award (NOFO) and post-award (contract/grant) stages, and operationalize the implementation of associated strategies
- CDOC - Data Ethics and Equity Working Group led by USDOT CDO
- Departmental Data Governance WG led by USDOT OCIO
- Data Science Task Force led by USDOT FHWA
- Artificial Intelligence Intermodal Working Group led by USDOT OST
- WH Office of Science and Technology Policy ML-AI Subgroup led by NITRD
- Open Knowledge Network (OKN) led by NSF
- Data Dialogue Series starting at the ITS World Congress in September 2022

## RESOURCES & CONTACT INFORMATION

- **Factsheets**
  - ITS Data Access and Exchanges Program: https://www.its.dot.gov/factsheets/pdf/PR_ITSDataAccessExchanges_Factsheet.pdf
  - Operational Data Environment (ODE): https://www.its.dot.gov/factsheets/pdf/ITSJPO_ODE.pdf
  - ITS Research Data and DataHub: https://www.its.dot.gov/factsheets/pdf/ITS_ResearchData.pdf
- **Data Access and Exchanges Portfolio Links**
  - ITS DataHub: https://www.its.dot.gov/data/index.htm
  - ITS CodeHub: https://www.its.dot.gov/code/index.htm
  - Secure Data Commons (SDC): https://www.transportation.gov/data/secure
  - OSS4ITS: https://usdot-oss4its.atlassian.net/wiki/spaces/OSSFITS/overview
  - V2X Hub: https://github.com/usdot-fhwa-OPS/V2X-Hub
  - CARMA: https://highways.dot.gov/research/operations/CARMA
  - Work Zone Data Exchange (WZDx): https://www.transportation.gov/av/data/wzdx

***To learn more about the Data Access and Exchanges Program, contact:***

**Murat Omay**
Program Manager
U.S. DOT ITS Joint Program Office
Murat.Omay@dot.gov

# AI FOR ITS PROGRAM

## PROGRAM VISION, MISSION, AND GOALS

| VISION | MISSION | GOALS |
|---|---|---|
| The AI for ITS Program's vision is to advance next generation transportation systems and services by leveraging trustworthy, ethical AI and Machine Learning (ML) for safer, more efficient, and accessible movement of people and goods. | The AI for ITS Program identifies, develops, implements, evaluates, and coordinates technology and policy research to advance the contextualization and integration of AI (including ML) into all aspects of the transportation system. | • Accelerate deployment and evaluation of AI for ITS applications<br>• Spur innovation of potentially transformative AI for ITS applications<br>• Maximize value, making the best use of USDOT resources<br>• Facilitate nationwide adoption of trustworthy, ethical AI-driven ITS |

## AI FOR ITS PROGRAM ORGANIZING PRINCIPLES

- Leverage AI to address critical **multimodal ITS challenges and needs**.
- Promote **accountability** by establishing processes to manage, operate, and oversee implementation.
- Prioritize **security** and **privacy** of sensitive data.
- Promote **quality, reliability, and representativeness of data** sources and processing.
- Identify precise, consistent, and reproducible **performance measures** that are consistent with program objectives and measure performance.
- Foster reliability and relevance over time through **continuous performance monitoring** and assessment of sustained and expanded use.
- Share **open data**, while protecting privacy, and facilitate **open-source development**, while preserving intellectual property, to increase richness of available data and code and promote innovation.
- Share **lessons learned and best practices** to facilitate reproducibility of applications and accelerate adoption of AI-driven next-generation ITS nationwide.

# UNDERSTANDING THE AI FOR ITS MARKET

## SOURCES SOUGHT NOTICE (2021)

| MOTIVATION | PURPOSE | KEY QUESTIONS |
|---|---|---|
| • Build upon AI for ITS stakeholder outreach and research activities<br>• Gather input from public, private, and academic sectors on AI-enabled solutions for ITS<br>• Identify opportunities for DOT investments to accelerate AI for ITS innovations to deployment | • Solicit feedback on:<br>– "Deployment-ready" applications that leverage AI for ITS needs<br>– DOT role and investment areas to facilitate next generation ITS leveraging AI<br>– Existing capabilities for developing and deploying AI for ITS | • Identification of deployment-ready AI-enabled ITS applications<br>• Experience with developing AI-enabled ITS applications<br>• Awareness of proven AI-enabled applications from other domains for adoption/integration<br>• Top three roles for DOT investments in AI |

## INSIGHTS FROM SOURCES SOUGHT RESPONSES (#1 OF 3)

- **Application Maturity:** Most of the applications identified are mature; degree of maturity varies. Understanding the scalability of applications across locations and network types would require additional research.
- **Type of AI:** Supervised ML techniques, including deep learning and computer vision. These techniques require a significant amount of labeled data for training.
- **Benefits Measurement:** Majority of the respondents did not specifically measure quantitative benefits, but notes perceived improvements in process efficiency, accuracy, and transportation system safety and mobility.
- **Cost Structures:** All costs were emphasized as highly dependent on data and computing requirements as well as geographic scale (e.g., the number of equipped intersections).

## INSIGHTS FROM SOURCES SOUGHT RESPONSES (#2 OF 3)

- **Data:**
  - Public domain data (specifically, imagery data) may suffer from a lack of quality control.
  - Individual devices/sensors are increasingly becoming primary data sources.
  - Real-time data are becoming increasingly available, bringing newer challenges with processing.
- **Cybersecurity:** Respondents did not conduct security analysis of their AI applications. A few followed their organization's best practices (e.g., data anonymization, encryption, minimizing access).
- **Trustworthy and Ethical AI:**
  - Consistently produce outputs that are reasonable, auditable, and explainable.
  - Operate within a data governance structure that is ethical, test constantly for bias, and adhere to federal, state, and local requirements, policies, standards, regulations, and laws.
  - Monitor ethical standards as these do not remain fixed and transform in response to evolving situations (what was acceptable 10 years ago, may no longer be considered ethical).

## INSIGHTS FROM SOURCES SOUGHT RESPONSES (#3 OF 3)

- **Top Challenges:**
  - Unseen data and model drift over time -- requiring ongoing maintenance of AI applications with a human-in-the-loop, especially for decisions of greater consequence
  - Other key challenges: lack of standards; lack of labeled and clean data; lack of compute power; lack of workforce and expertise in AI; skeptical attitude towards change and new technologies; potential proliferation of unethical AI systems; lack of funding
- **Top 3 Government Roles:**
  - Resolve AI-related policy issues (e.g., data governance and data sharing policies)
  - Develop standards to ensure easy access and sharing of data for execution
  - Conduct prototype testing/demonstrations of AI-enabled ITS applications

## CHALLENGES TO AI ADOPTION AND IMPLEMENTATION

- *AI for ITS Challenges and Potential Solutions, Insights, and Lessons Learned Report* (in development):
  - Based on SSN responses, review of literature, and other market research, and ongoing coordination with deployers and others.
- Key findings across the 12 identified challenges:
  - These challenges for AI adoption and successful implementation are not unique to ITS.
  - There may be tradeoffs between addressing different challenges.
  - Addressing these challenges is an ongoing exercise.
  - Maintaining a human-in-the-loop is helpful in identifying and mitigating these challenges.

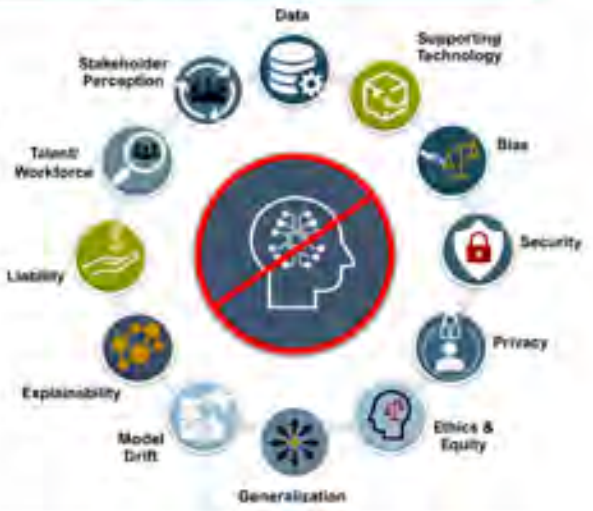

# POTENTIAL PROGRAM NEXT STEPS

## POTENTIAL MULTI-PRONGED APPROACH

**Goal 1: Accelerate deployment and evaluation** of AI for ITS where it can make an immediate impact on safety, mobility, equity
- BY focusing on near term, more mature applications for scalable rollout

**Goal 2: Spur innovation** regarding potentially transformative AI for ITS applications
- BY sponsoring a multi-stage innovation challenge, advancing from promising concepts to an operational prototype to a deployable capability

**Goal 3: Maximize value,** making the best use of ITS JPO and departmental resources
- BY coordinating complementary efforts across the ITS JPO and other USDOT programs as well as external activities

**Goal 4: Facilitate nationwide adoption** of trustworthy, ethical AI-driven ITS
- BY demonstrating value, disseminating insights and lessons learned, and facilitating peer exchanges and partnerships to support technology development and knowledge transfer

## RESOURCES & CONTACT INFORMATION

- **Factsheets**
  - AI for ITS Program Overview: https://www.its.dot.gov/research_areas/emerging_tech/pdf/ITSJPO_AIforITS_Program.pdf
  - Potential Application of AI in Transportation: https://www.its.dot.gov/factsheets/pdf/AIforITS_Program_Factsheet_RealWorld_AI_PotentialApps_in_Transportation_D4.pdf
  - AI Scenarios in Transportation for Possible Deployment: https://www.its.dot.gov/factsheets/pdf/AIforITS_Program_Factsheet_RealWorld_AI_Scenarios_in_Transportation_D4.pdf
- **AI for ITS Program Publications**
  - Summary of Potential Application of AI in Transportation https://rosap.ntl.bts.gov/view/dot/50651
  - AI Scenarios in Transportation for Possible Deployment https://rosap.ntl.bts.gov/view/dot/50752
  - Plan for AI for ITS Program https://rosap.ntl.bts.gov/view/dot/53932

***To learn more about the AI for ITS Program, contact:***

**Robert (Bob) Sheehan, P.E., PTOE**
Acting Chief of Policy, Architecture, and Knowledge Transfer
U.S. DOT ITS Joint Program Office
Robert.Sheehan@dot.gov

20

Figure B-2: Dimitris Bertsimas - Improving on Fairness/Bias

# Improving on Fairness/Bias

**Hari Bandi and Dimitris Bertsimas**

**MIT**

**Based on ``The price of Diversity"**

## Discharge to post acute care

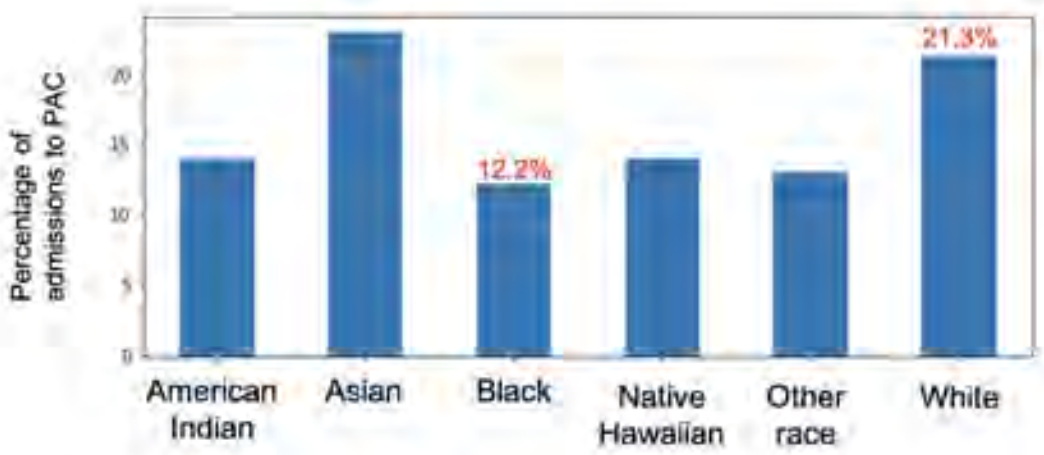

## The Problem: Systemic Bias

- Systemic *bias* with respect to *gender, race and ethnicity*, often *unconscious*, but prevalent in datasets involving *choices* made by people.
- Some examples include datasets related to human choices in *college admissions, hiring, lending*, or *parole* decisions that discriminate against *African-Americans* or *women*.

The Analytics Edge

## Summary

- We propose a novel optimization approach to train classification models on large datasets to *alleviate bias* and *enhance diversity* without significantly compromising on meritocracy.
- Key takeaway: The price of diversity is *low* and sometimes *negative*, that is we can modify our selection processes in a way that *enhances diversity without affecting meritocracy* significantly, and sometimes improving it.

The Analytics Edge

## Background: Massachusetts General Hospital

- Discharge planning is the development of an individualized discharge plan for a patient prior to leaving hospital for home or to a post acute care (PAC).
- Early prediction of PAC needs prior to discharge leads to
  - reducing hospital length of stay,
  - unplanned readmissions, and
  - improves patient outcomes.

## The problem

- The task is to determine discharge disposition for trauma patients within 48-hours after admission.
- Patients are either sent to a post acute care rehab center or home directly after discharge.
- A successful admission into a PAC depends on
  - Patients' needs,
  - Rehab center agreeing to admit the patient,
  - Patient agreeing to get admitted into a rehab center.

## Dataset

- The American College of Surgeons Trauma Quality Improvement Program (ACS-TQIP) database.
- Dataset is sourced from hospitals around the country.
- Features include:
  - patient demographics (age, gender),
  - comorbidities,
  - Emergency Department (ED) vital signs, and
  - injury characteristics (e.g., severity, mechanism).

## ML model

- Determine discharge disposition for trauma patients within 48-hours after admission.
- Patients are either sent to a post acute care rehab center or home directly after discharge.
- Build a Logistic Regression model to predict disposition with AUC =0.79

## Notation

- Each of the patients is assigned an *outcome $Y = \{-1, +1\}$* representing either *Entering PAC (+1)* or *not (−1)*.
  - $W$: set of white patients.
  - $B$: set of black patients.
  - $n_w$: total number of white patients.
  - $n_b$: total number of black patients.
  - $p_w$: total number of white patients who enter PAC.
  - $p_b$: total number of black patients who enter PAC.

## α-biased dataset

- We call a dataset α-biased if the *difference between the rates of positive observations* among a pair of subgroups $W$ and $B$ based on a protected variable (in this case, race) is at least α.

**Definition 1 (α-biased dataset)** *A dataset $\mathcal{X} = \{(x_i, y_i) | y_i \in \{-1, 1\}\}$ is said to be α-biased with respect to a pair of subgroups $\mathcal{W}, \mathcal{B} \subseteq \mathcal{X}$ if*

$$\left| \frac{\sum_{i \in \mathcal{W}} \mathbb{1}(y_i = +1)}{n_w} - \frac{\sum_{i \in \mathcal{B}} \mathbb{1}(y_i = +1)}{n_b} \right| \geq \alpha.$$

## Demographic parity

- Demographic parity imposes the condition that a classifier $H$ should *predict a positive outcome* for individuals across groups with *almost equal frequency*.

**Definition 2** (Demographic parity) *A classifier $\mathcal{H} : X \to \{-1, 1\}$ achieves demographic parity with bias $\epsilon$ with respect to groups $\mathcal{W}, \mathcal{B} \subseteq X$ if and only if*

$$\left| \frac{\sum_{i \in \mathcal{W}} \mathbb{1}(\mathcal{H}(x_i) = +1)}{n_w} - \frac{\sum_{i \in \mathcal{B}} \mathbb{1}(\mathcal{H}(x_i) = +1)}{n_b} \right| \leq \epsilon.$$

## Example of Demographic parity

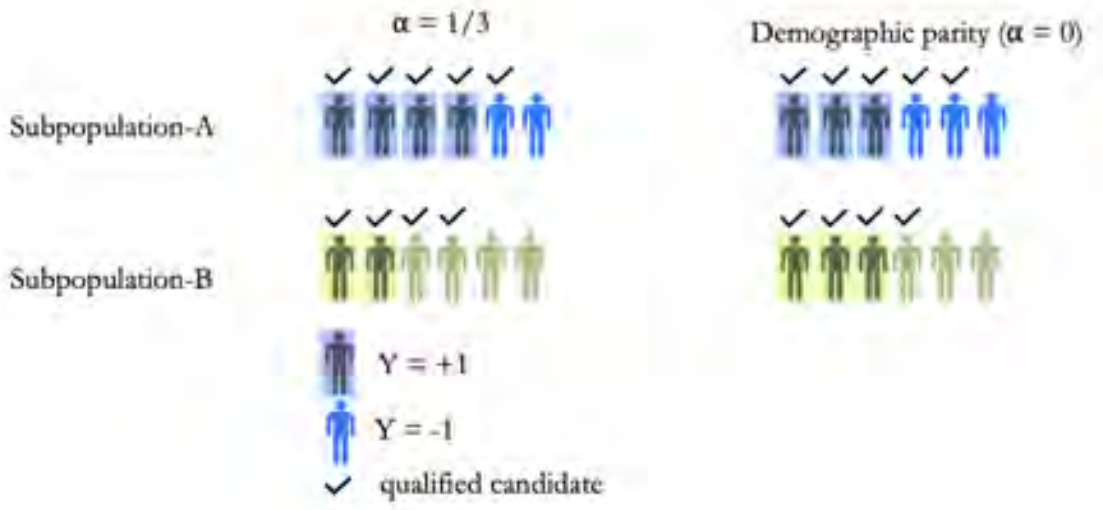

## Proposed solution

- *Flip outcome labels* (Y) while training your model to achieve demographic parity.
- We propose a Mixed-integer Optimization (MIO) problem that achieves this by introducing *binary variables* $z_i \in \{0, 1\}, i \in [n]$ to decide which outcome labels to flip.

## Proposed solution

- If we decide to *flip the outcome label* of the $i^{th}$ observation: $y_i \in \{-1, 1\}$, the resulting outcome label would be $\tilde{y}_i = y_i(1 - 2z_i)$.
- We define a set of $n$ binary variables ($z$) that flip at most $\tau_w$ proportion of labels in $W$ and $\tau_b$ proportion of labels in $B$ given by,

$$\mathcal{Z}_{\tau_w,\tau_b} = \left\{ \mathbf{z} \in \{0,1\}^n : \frac{\sum_{i \in W} z_i}{n_w} = \tau_w, \ \frac{\sum_{i \in B} z_i}{n_b} = \tau_b \right\}.$$

## Proposed solution

- The parameters $\tau_w$ and $\tau_b$ are estimated from the data so that the resulting classifier ensures *ε-demographic parity*.

$$\tau_w \le \frac{n_b \cdot p_w}{n_w(n_w + n_b)} - \frac{p_b}{n_w + n_b} + \frac{n_b \cdot \epsilon}{n_w(n_w + n_b)},$$

$$\tau_b \le \frac{p_w}{n_w + n_b} - \frac{n_w \cdot p_b}{n_b(n_w + n_b)} - \frac{n_w \cdot \epsilon}{n_b(n_w + n_b)}.$$

## Logistic Regression

- The dependent variable (Y) is a Bernoulli random variable
  - Y = +1 – "Entering PAC"
  - Y = -1 – "Not entering PAC"
- We seek to predict the probability of a success outcome of the dependent variable Y as a function of independent variables $x_1, x_2 .. x_k$
- We predict the *likelihood* that Y = +1 as follows:
  - $\Pr(Y = +1) = \frac{e^{\beta_0 + \beta_1 x_1 + \beta_2 x_2 + \cdots + \beta_k x_k}}{1 + e^{\beta_0 + \beta_1 x_1 + \beta_2 x_2 + \cdots + \beta_k x_k}}$

*This is guaranteed to be between 0 and 1*

## Logistic regression

Logistic regression model

$$\min_{\beta_0, \beta} \sum_{i=1}^{n} \log\left(1 + e^{-y_i(\beta^\top x_i + \beta_0)}\right)$$

Parameters of a Logistic regression model

MIO model

$$\min_{\mathbf{z} \in \mathcal{Z}_{\tau_w,\tau_b}} \min_{\beta_0, \beta} \sum_{i=1}^{n} \log\left(1 + e^{-y_i(1-2z_i)(\beta^\top x_i + \beta_0)}\right).$$

Binary variables

Updated outcome label

## MIO model (linearizing product terms)

$$\min_{\mathbf{z}} \min_{\beta, \gamma} \quad f(\beta, \gamma) := \sum_{i=1}^{n} \log\left(1 + e^{-y_i(\beta^\top x_i + \beta_0) + 2y_i(\gamma_i^\top x_i + \gamma_{i,0})}\right)$$

\# label flips

$$\text{s.t.} \quad \sum_{i \in W} z_i = \tau_w \cdot n_w,$$

$$\sum_{i \in B} z_i = \tau_b \cdot n_b,$$

Big-M constraints

$$-z_i M_j \le \gamma_{i,j} \le z_i M_j, \ i \in [n], j \in [p],$$

$$-(1 - z_i) M_j \le \gamma_{i,j} - \beta_j \le (1 - z_i) M_j, \ i \in [n], j \in [p],$$

Implied constraints (using binary variables)

$$\sum_{i \in W} \gamma_{i,j} = \tau_w \cdot n_w \cdot \beta_j, \ j \in [p],$$

$$\sum_{i \in B} \gamma_{i,j} = \tau_b \cdot n_b \cdot \beta_j, \ j \in [p],$$

$$z_i \in \{0, 1\}, i \in [n].$$

## Additional constraints

- *Maximize likelihood*
- *Demographic parity*
- *Severity of injuries unchanged*
- *Age and gender distribution unchanged*

## Predictive performance

AUC on original outcome labels AUC on modified outcome labels

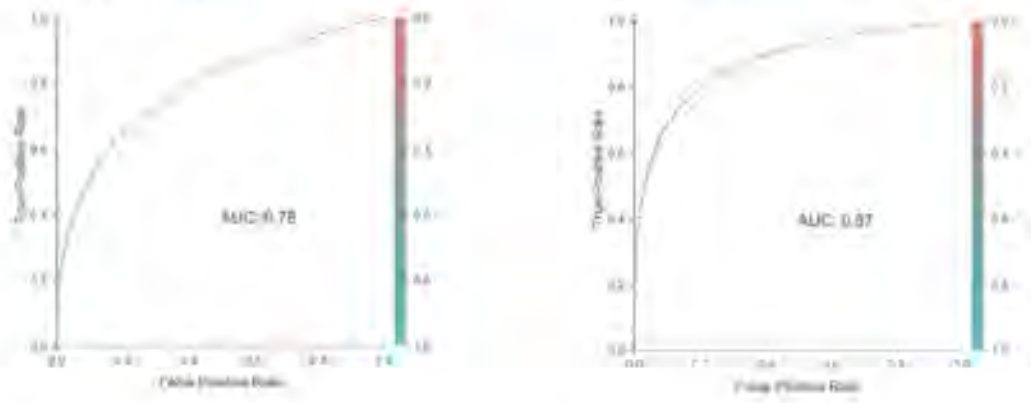

- Alleviating bias improves out-of-sample AUC of OCTs by 8-10%.

## Implementation tool

- Train Optimal Classification Trees (OCTs) to provide insights on which attributes of individuals lead to flipping of their labels.
- Construct a dataset based on output of the MIO model. Each defendant is labeled as one of the following:
  - negative (patient discharged to home),
  - high (patient discharged to PAC), or
  - no change (outcome label unchanged)

## Implementation tool

Left part of the OCT after splitting on Head severity ≤ 1.0

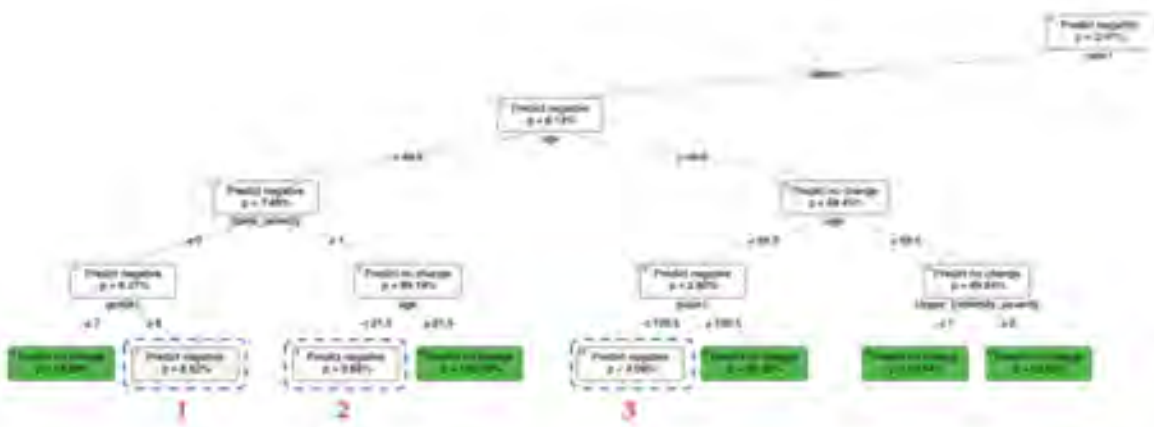

# Implementation tool

Right part of the OCT after splitting on Head severity ≥ 2.0.

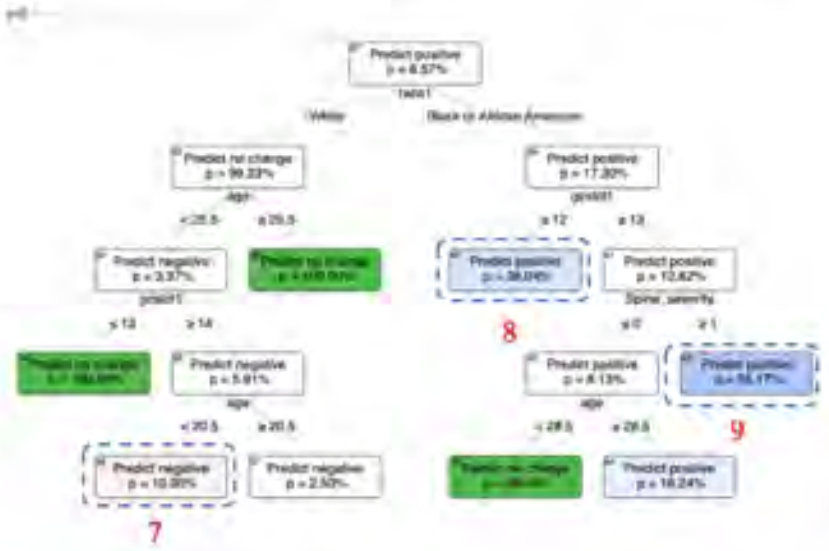

# Other applications

- Admissions
- Parole
- Bar exam

# Key takeaways

- Demonstrate how alleviating bias can improve selection processes in practice.
- Develop a highly interpretable implementation tool to make changes to the current selection processes to improve diversity.
- Alleviating bias improves predictive performance.

Figure B-3: Maria De-Arteaga - Social Norm Bias: Residual Harms of Fairness-Aware Algorithms

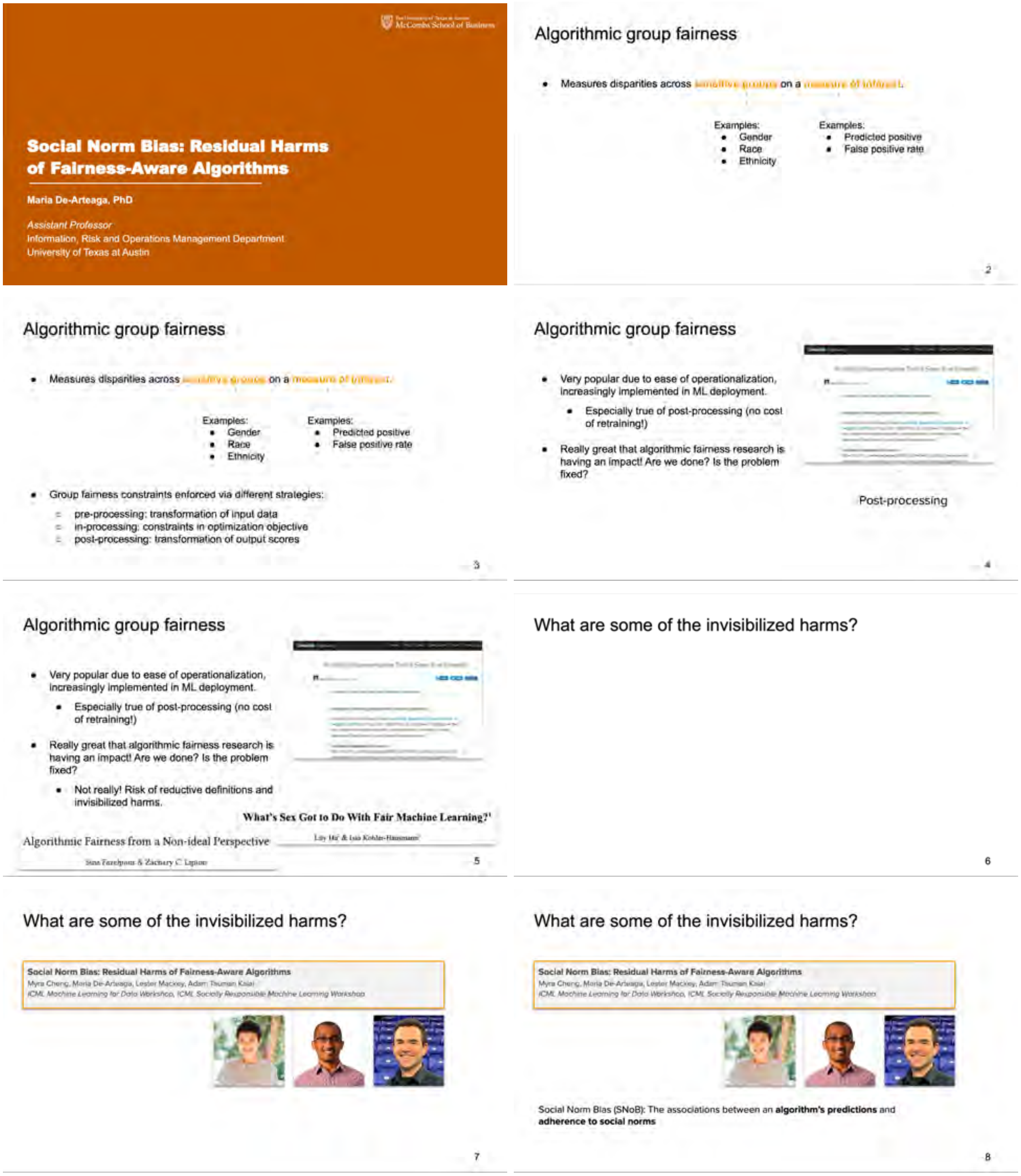

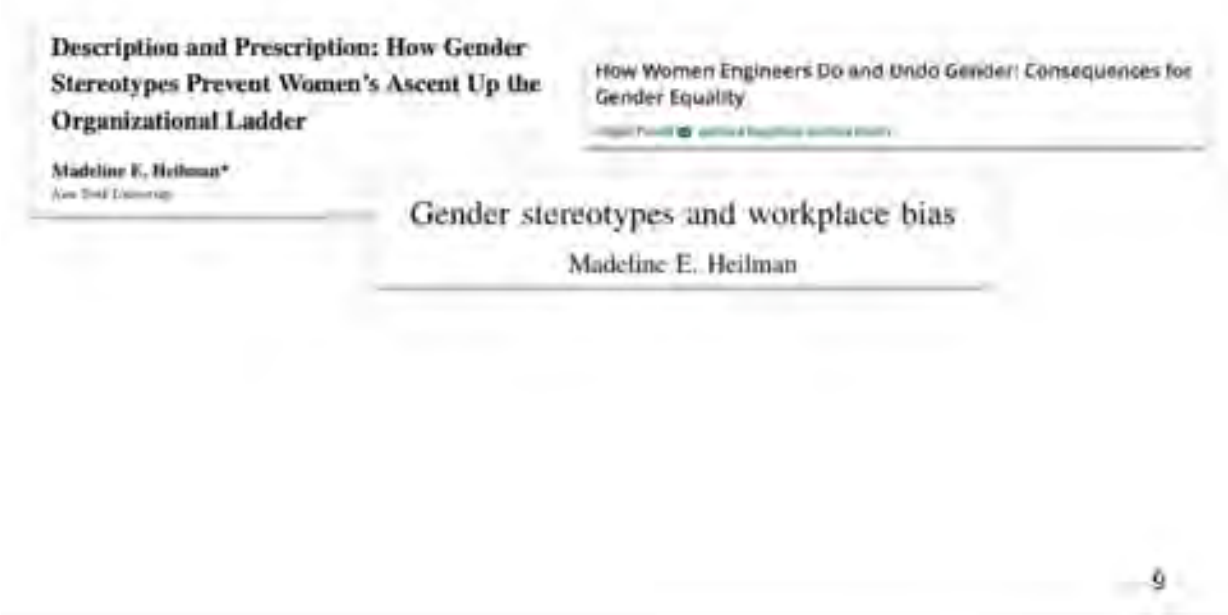
Description and Prescription: How Gender Stereotypes Prevent Women's Ascent Up the Organizational Ladder
Madeline E. Heilman*
How Women Engineers Do and Undo Gender: Consequences for Gender Equality
Gender stereotypes and workplace bias
Madeline E. Heilman
9

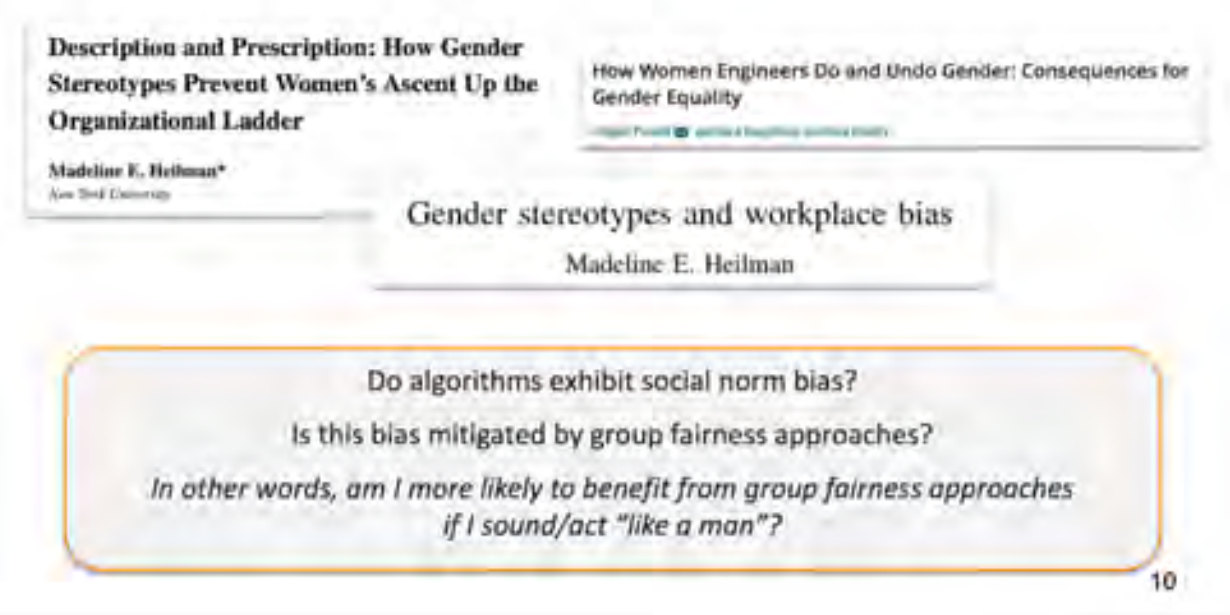
Description and Prescription: How Gender Stereotypes Prevent Women's Ascent Up the Organizational Ladder
Madeline E. Heilman*
How Women Engineers Do and Undo Gender: Consequences for Gender Equality
Gender stereotypes and workplace bias
Madeline E. Heilman
Do algorithms exhibit social norm bias?
Is this bias mitigated by group fairness approaches?
In other words, am I more likely to benefit from group fairness approaches if I sound/act "like a man"?
10

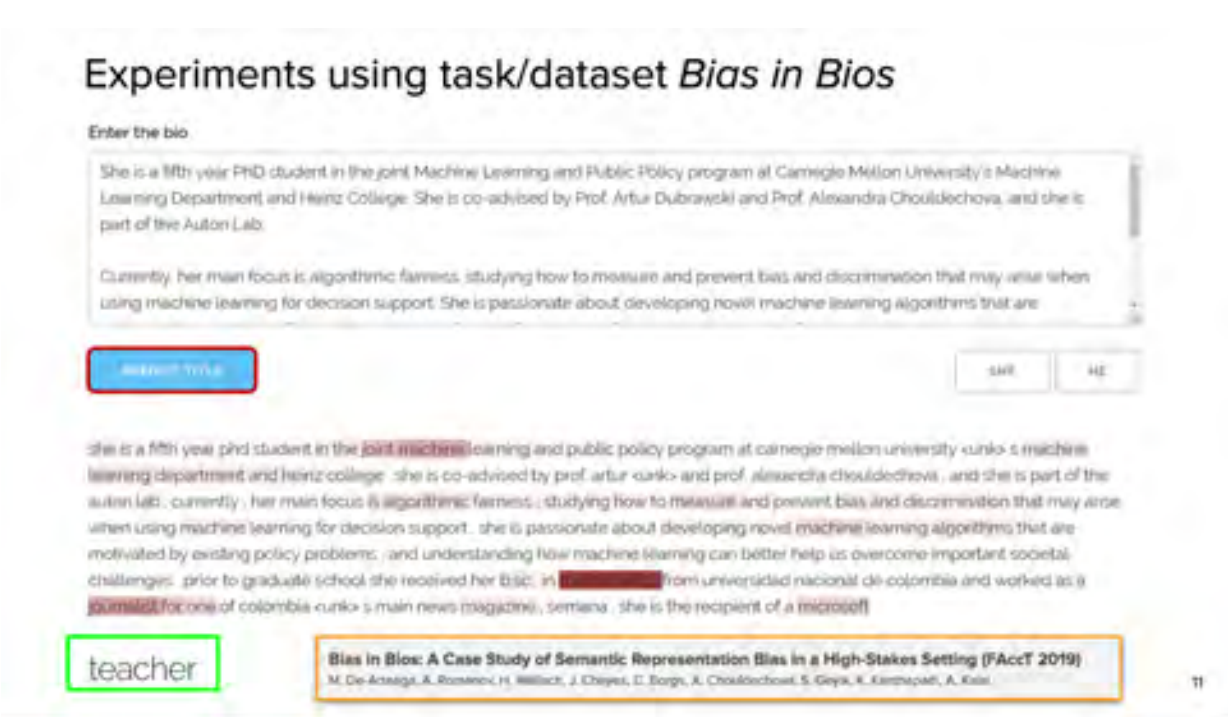
Experiments using task/dataset Bias in Bios
Enter the bio
teacher
Bias in Bios: A Case Study of Semantic Representation Bias in a High-Stakes Setting (FAccT 2019)
11

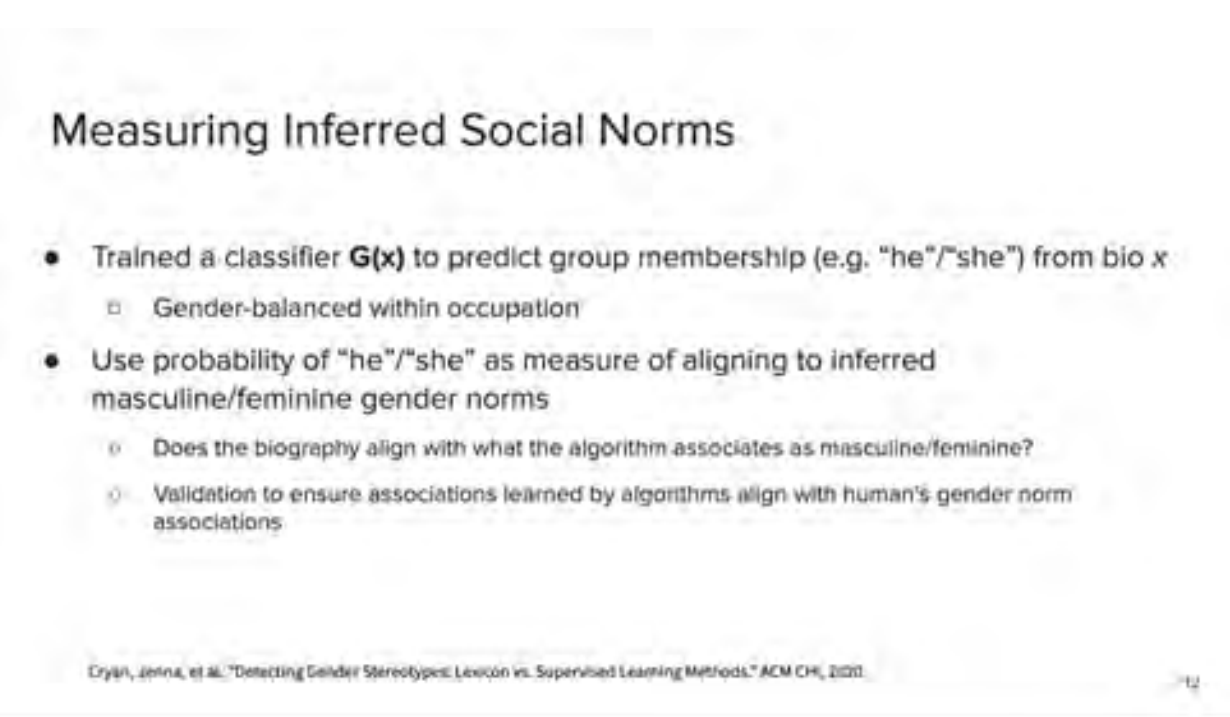
Measuring Inferred Social Norms
Trained a classifier G(x) to predict group membership (e.g. "he"/"she") from bio x
Gender-balanced within occupation
Use probability of "he"/"she" as measure of aligning to inferred masculine/feminine gender norms
Does the biography align with what the algorithm associates as masculine/feminine?
Validation to ensure associations learned by algorithms align with human's gender norm associations
Cryan, Jenna, et al. "Detecting Gender Stereotypes: Lexicon vs. Supervised Learning Methods." ACM CHI, 2020.
12

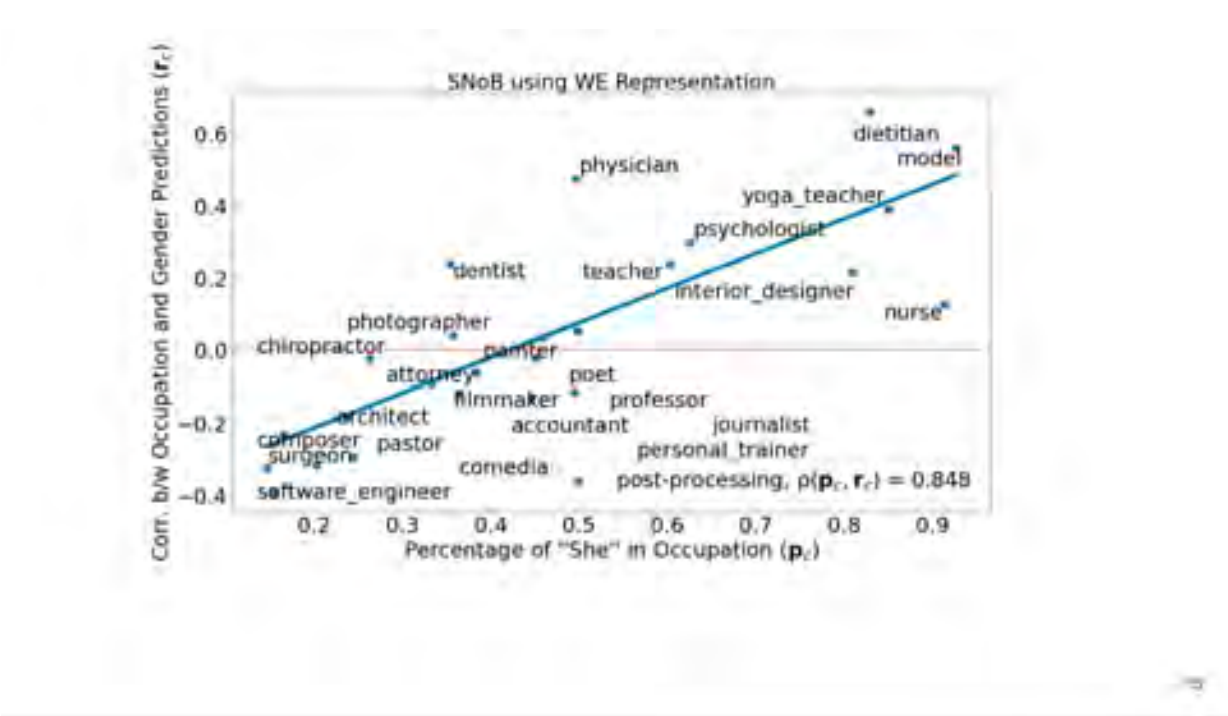

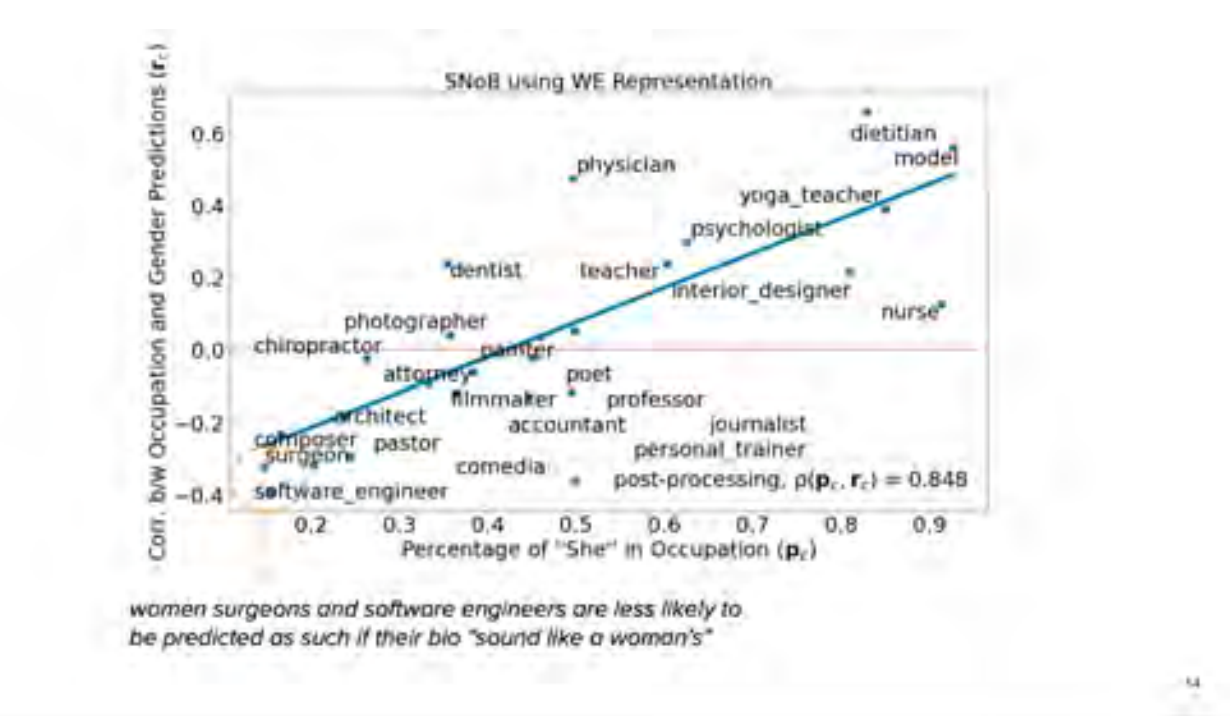
SNoB using WE Representation
Corr. b/w Occupation and Gender Predictions (r_c)
Percentage of "She" in Occupation (p_c)
dietitian
model
physician
yoga_teacher
psychologist
dentist
teacher
interior_designer
nurse
photographer
chiropractor
painter
attorney
poet
filmmaker
professor
architect
accountant
journalist
composer
surgeon
pastor
personal_trainer
comedian
software_engineer
post-processing, ρ(p_c, r_c) = 0.848
women surgeons and software engineers are less likely to be predicted as such if their bio "sound like a woman's"
14

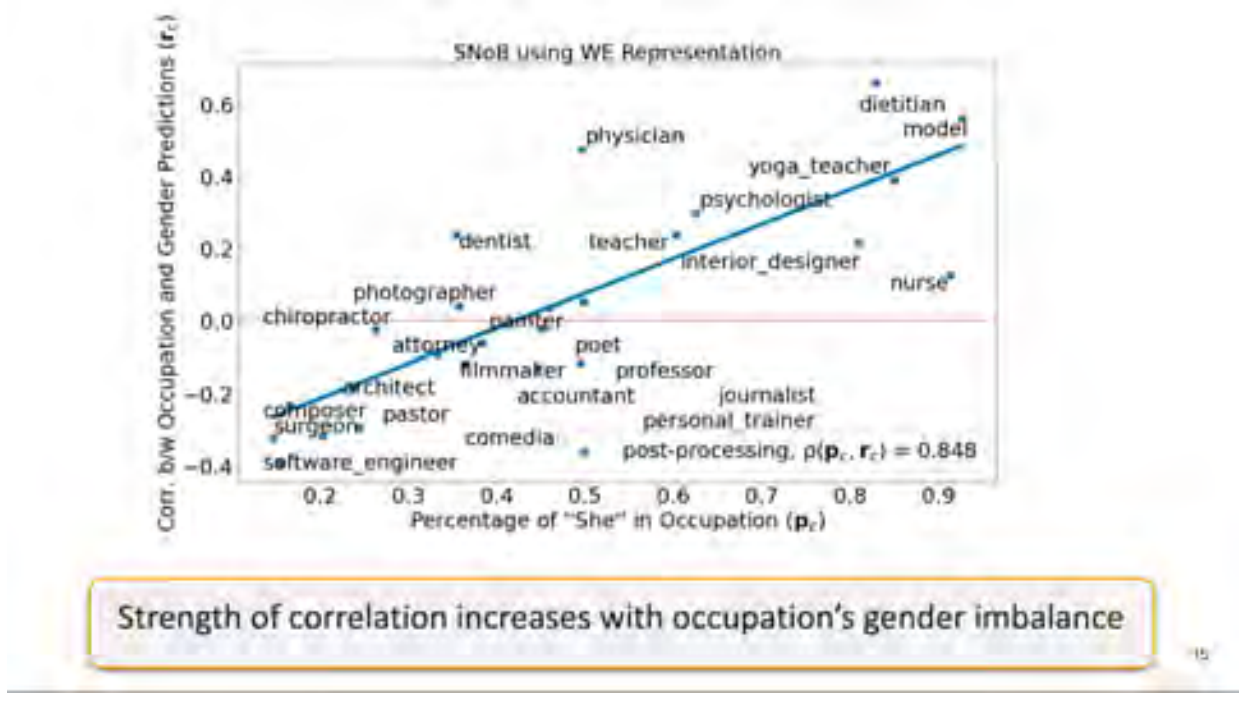
SNoB using WE Representation
Corr. b/w Occupation and Gender Predictions ($\mathbf{r}_c$)
Percentage of "She" in Occupation ($\mathbf{p}_c$)
dietitian
model
physician
yoga_teacher
psychologist
dentist
teacher
interior_designer
nurse
photographer
chiropractor
painter
attorney
poet
architect
filmmaker
professor
accountant
journalist
composer
surgeon
pastor
personal_trainer
comedian
software_engineer
post-processing, ρ($\mathbf{p}_c$, $\mathbf{r}_c$) = 0.848
Strength of correlation increases with occupation's gender imbalance

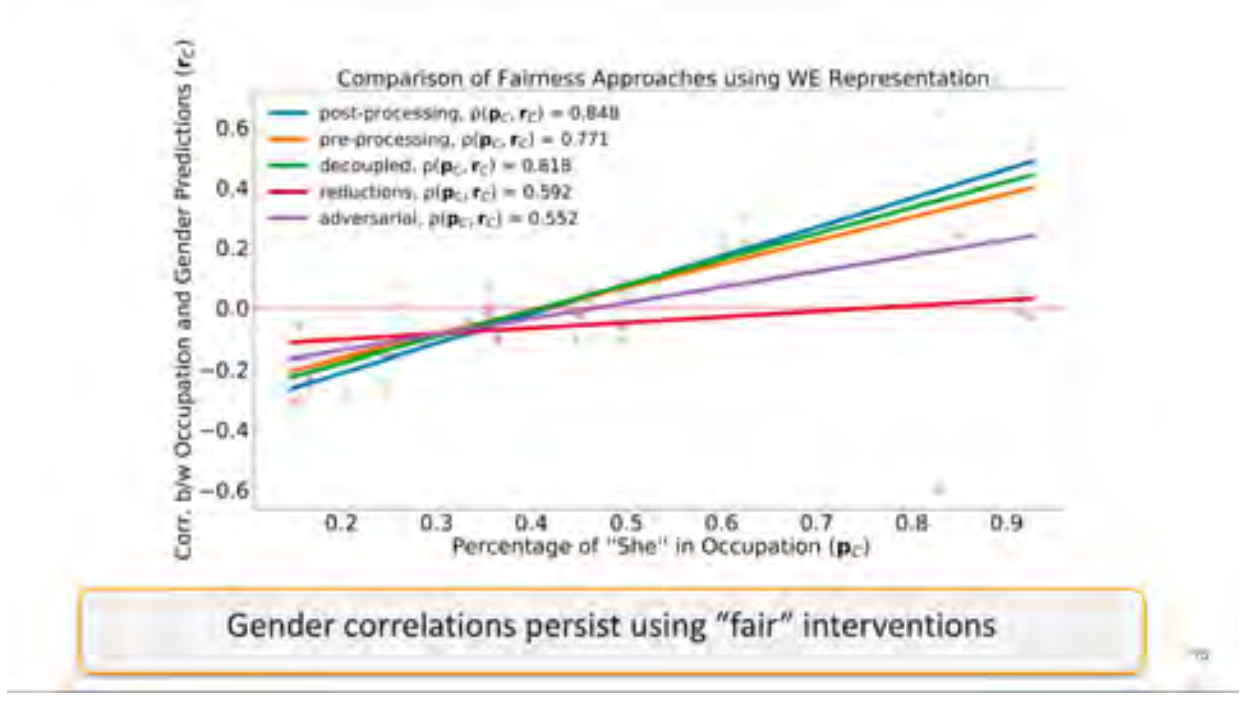
Comparison of Fairness Approaches using WE Representation
post-processing, ρ($\mathbf{p}_c$, $\mathbf{r}_c$) = 0.848
pre-processing, ρ($\mathbf{p}_c$, $\mathbf{r}_c$) = 0.771
decoupled, ρ($\mathbf{p}_c$, $\mathbf{r}_c$) = 0.818
reductions, ρ($\mathbf{p}_c$, $\mathbf{r}_c$) = 0.592
adversarial, ρ($\mathbf{p}_c$, $\mathbf{r}_c$) = 0.552
Corr. b/w Occupation and Gender Predictions ($\mathbf{r}_c$)
Percentage of "She" in Occupation ($\mathbf{p}_c$)
Gender correlations persist using "fair" interventions

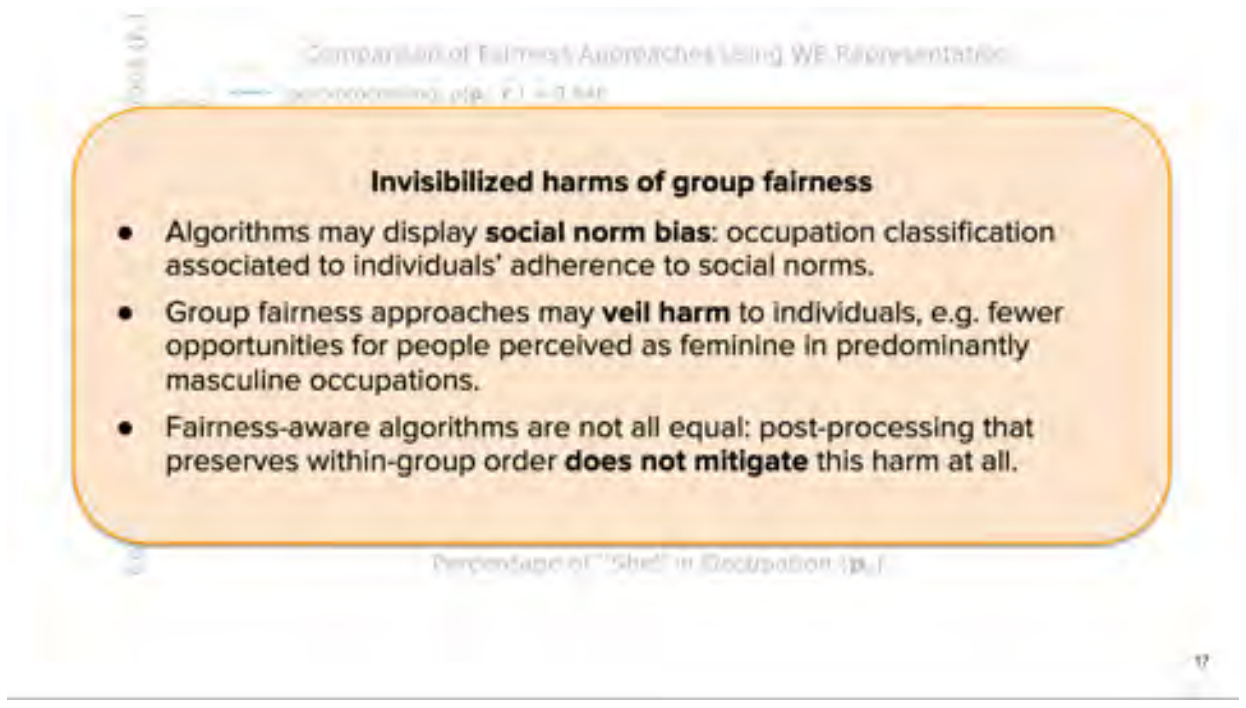
Invisibilized harms of group fairness
Algorithms may display **social norm bias**: occupation classification associated to individuals' adherence to social norms.
Group fairness approaches may **veil harm** to individuals, e.g. fewer opportunities for people perceived as feminine in predominantly masculine occupations.
Fairness-aware algorithms are not all equal: post-processing that preserves within-group order **does not mitigate** this harm at all.
17

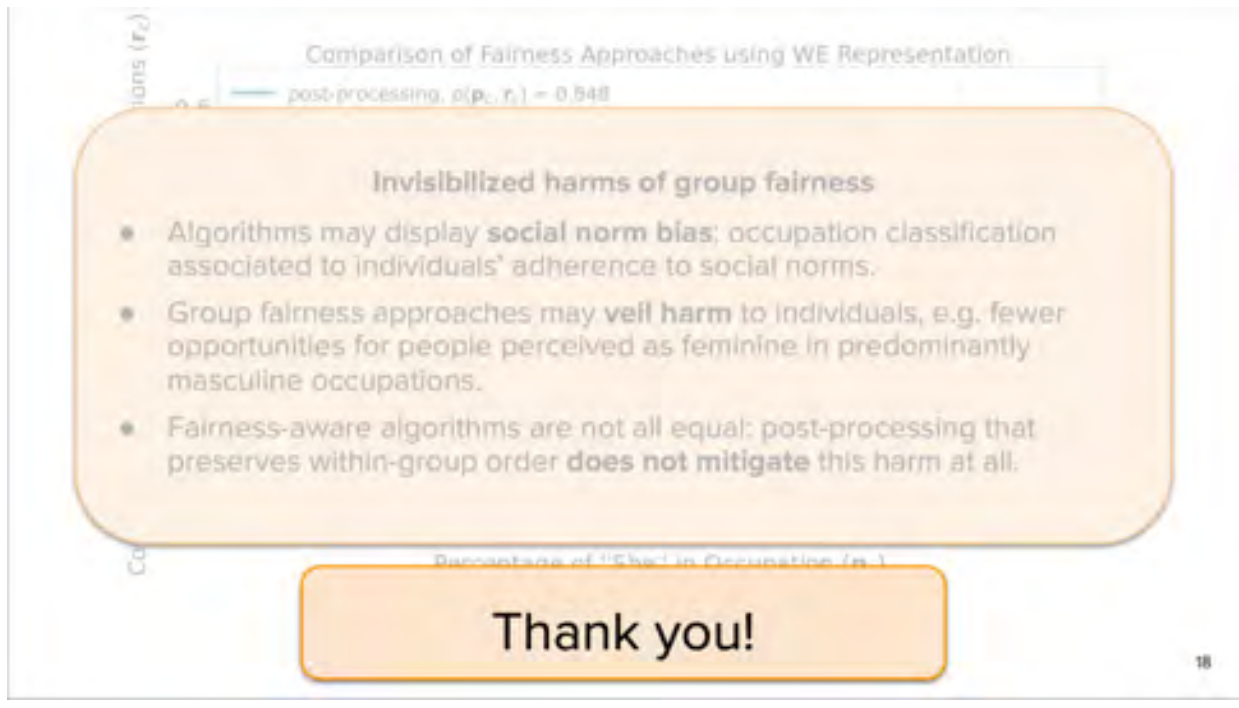
Thank you!
18

Figure B-4: Nikhil Garg - Auditing and Designing for Equity in Government Service Allocation

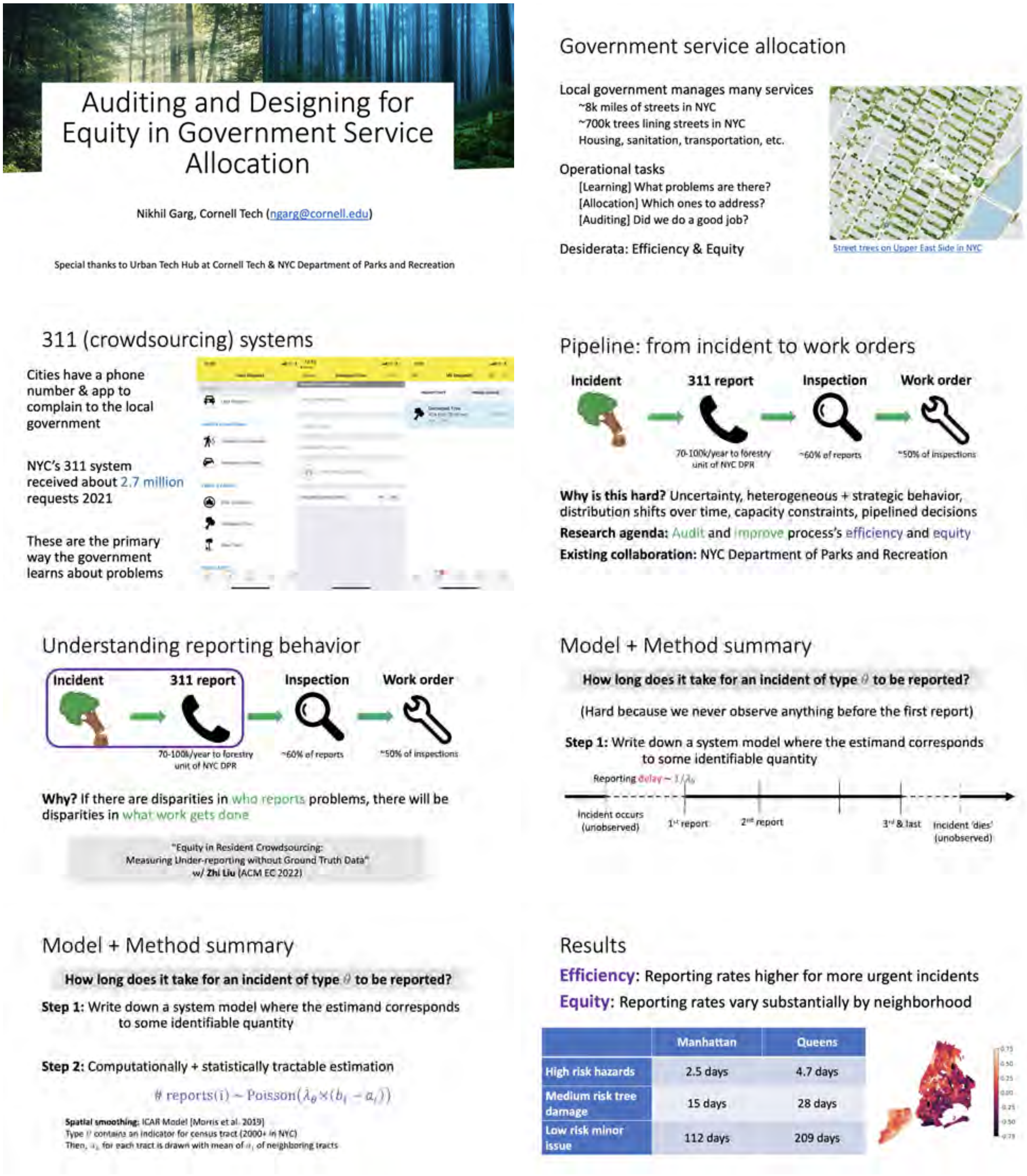

## Auditing agency decisions in entire pipeline

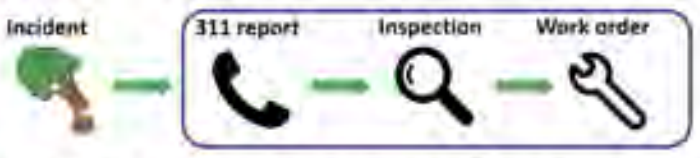

**Questions:** Is the agency inspecting the right reports? Are decisions efficient & equitable?

**Challenge:** Modeling capacity-constrained decisions under uncertainty

**Method**
(1) Use ML techniques to estimate incident risk given report characteristics
(2) Compare "optimal" set allocation decisions with empirical ones

"End-to-end Auditing of Decision Pipelines"
w/ **Benjamin Laufer** and **Emma Pierson**

## Improving agency decisions

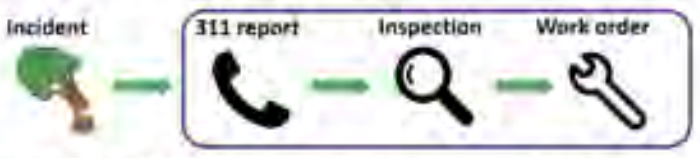

**Question:** Can we "optimally" re-prioritize inspections and work orders?

**Challenge:** Want "simple" policies that don't require maintaining an ML model

**Method**
(1) Use ML techniques to [robustly] estimate incident risk given report characteristics
(2) Come up with "service level agreements" for how quickly to address reports

"Making Inspection Decisions: Designing service level agreements"
w/ **Zhi Liu**

## Discussion: ML + Operations

**Machine learning** | **Operations**

Fairness requires both:
We want to compare decisions for comparable incidents/people/groups

Analyze individual incidents
(characterizing uncertainty, representing data)

Make global decisions
(comparing incidents, allocation under capacity constraints, modeling incentives)

## Another example: Recommendation systems

Old school ML view: Predict match between single item and user pair

But there are many global properties of recommender systems
- How users/items affect each other [competition effects]
- How users affect what items are produced [supply-side equilibria]
- How can we recommend *sets* of items [diverse recommendations]

Joint work with: Christian Borgs, Wenshuo Guo, Meena Jagadeesan, Michael I. Jordan, Karl Krauth, Lydia Liu, Laura Mitchell, Jacob Steinhardt, Gourab K Patro, Lorenzo Porcaro, Qiuyue Zhang, Meike Zehlike

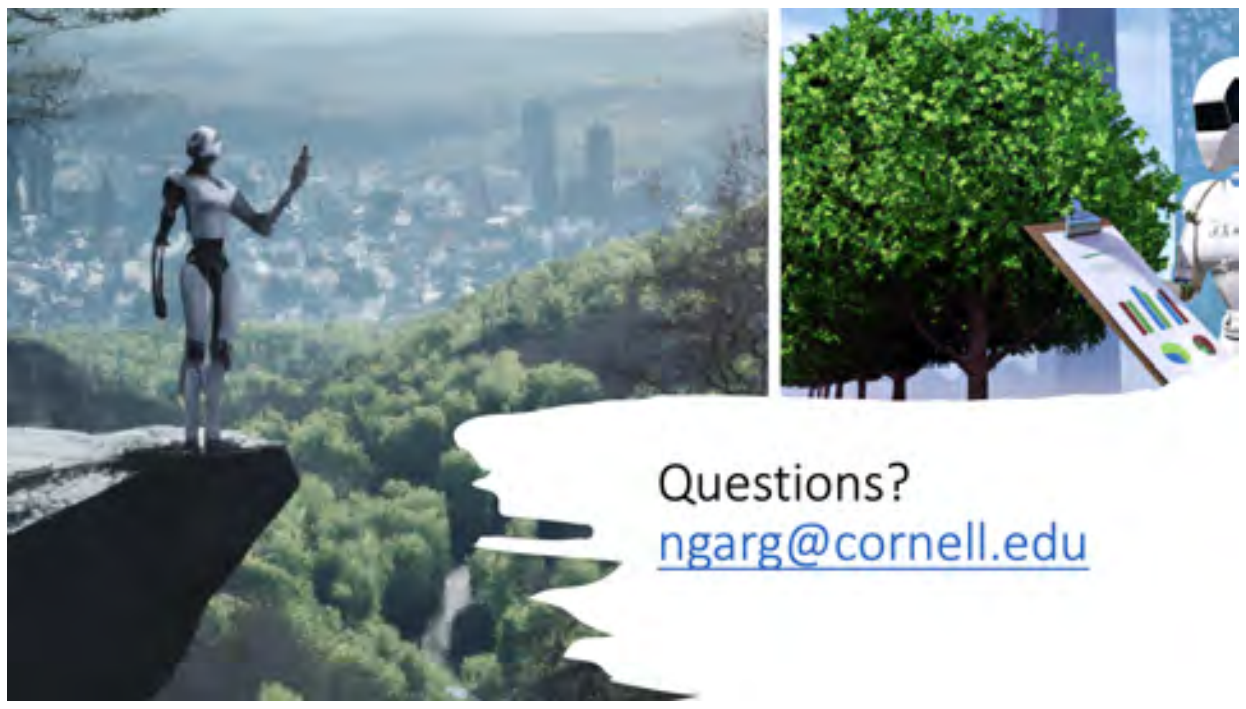

Figure B-5: David Shmoys - Fairness as the Objective in Congressional Districting

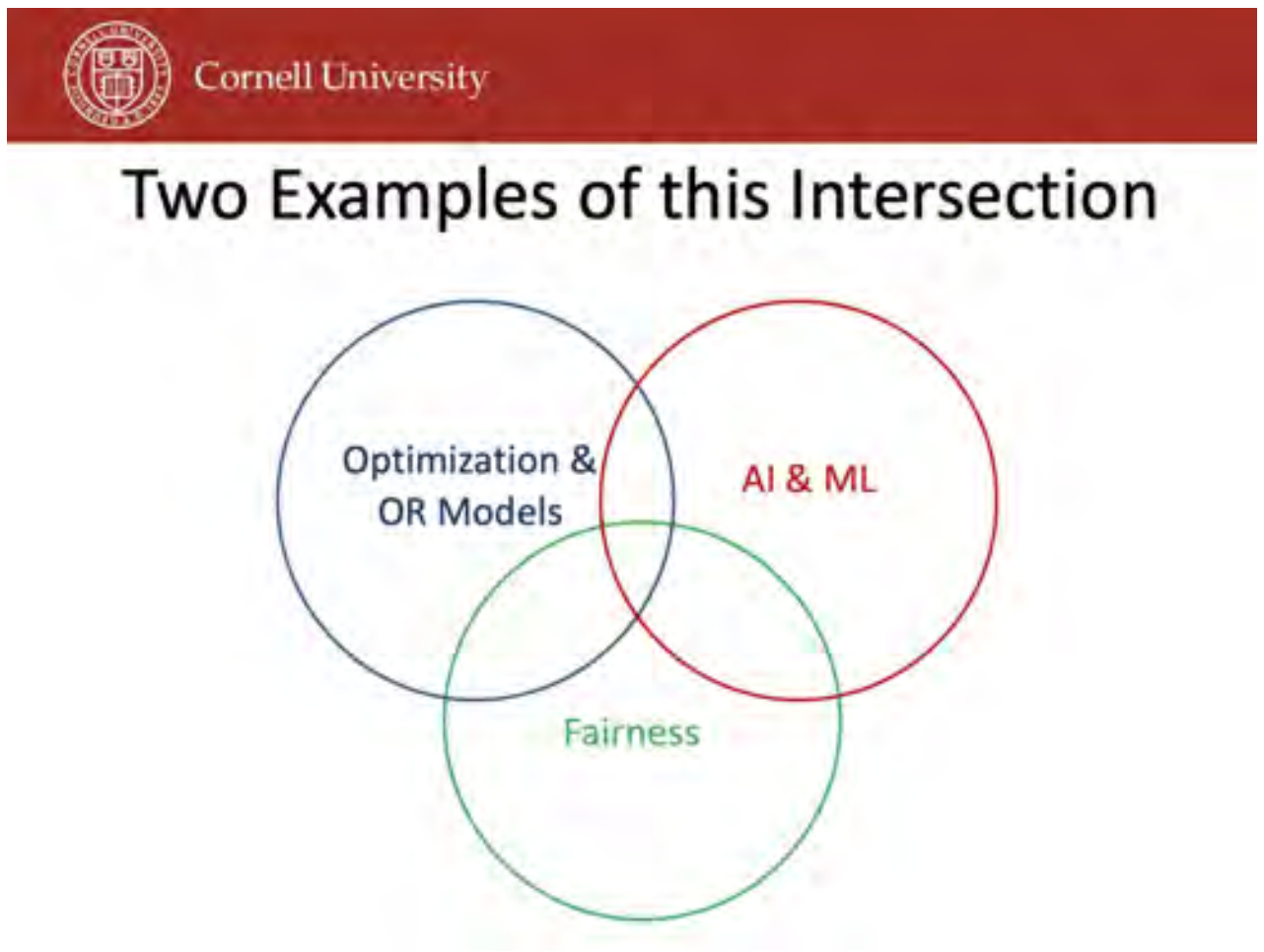

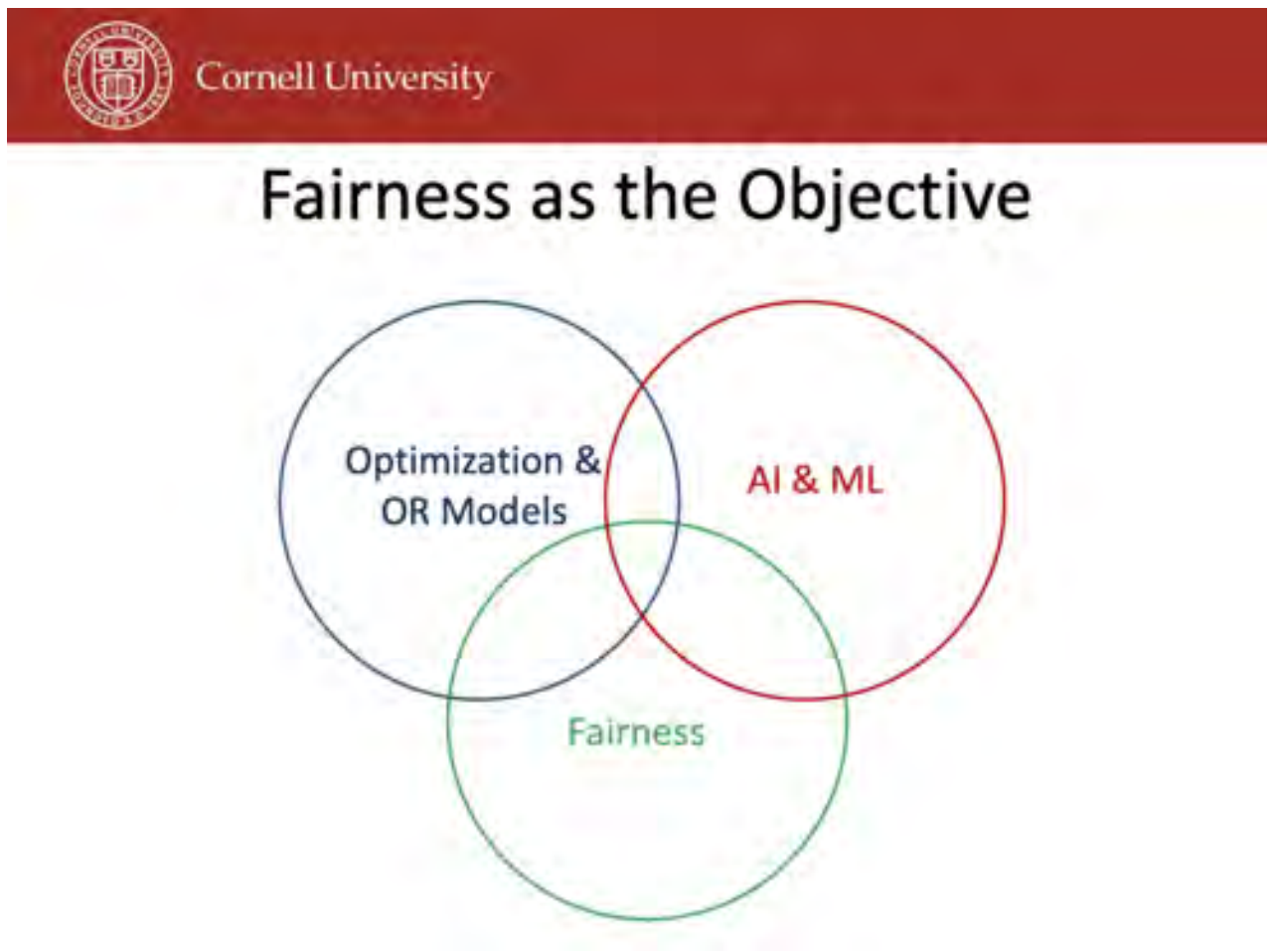

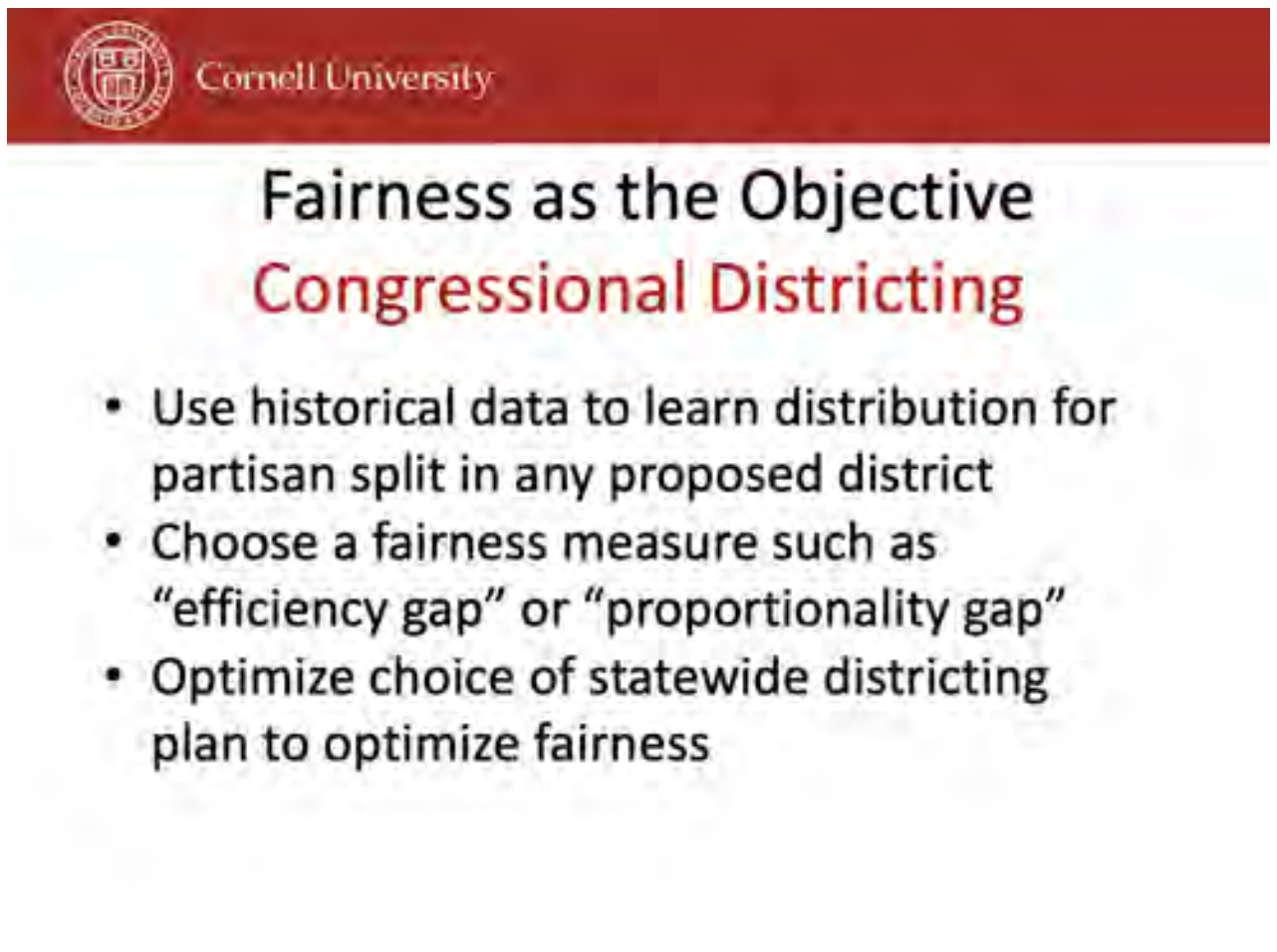

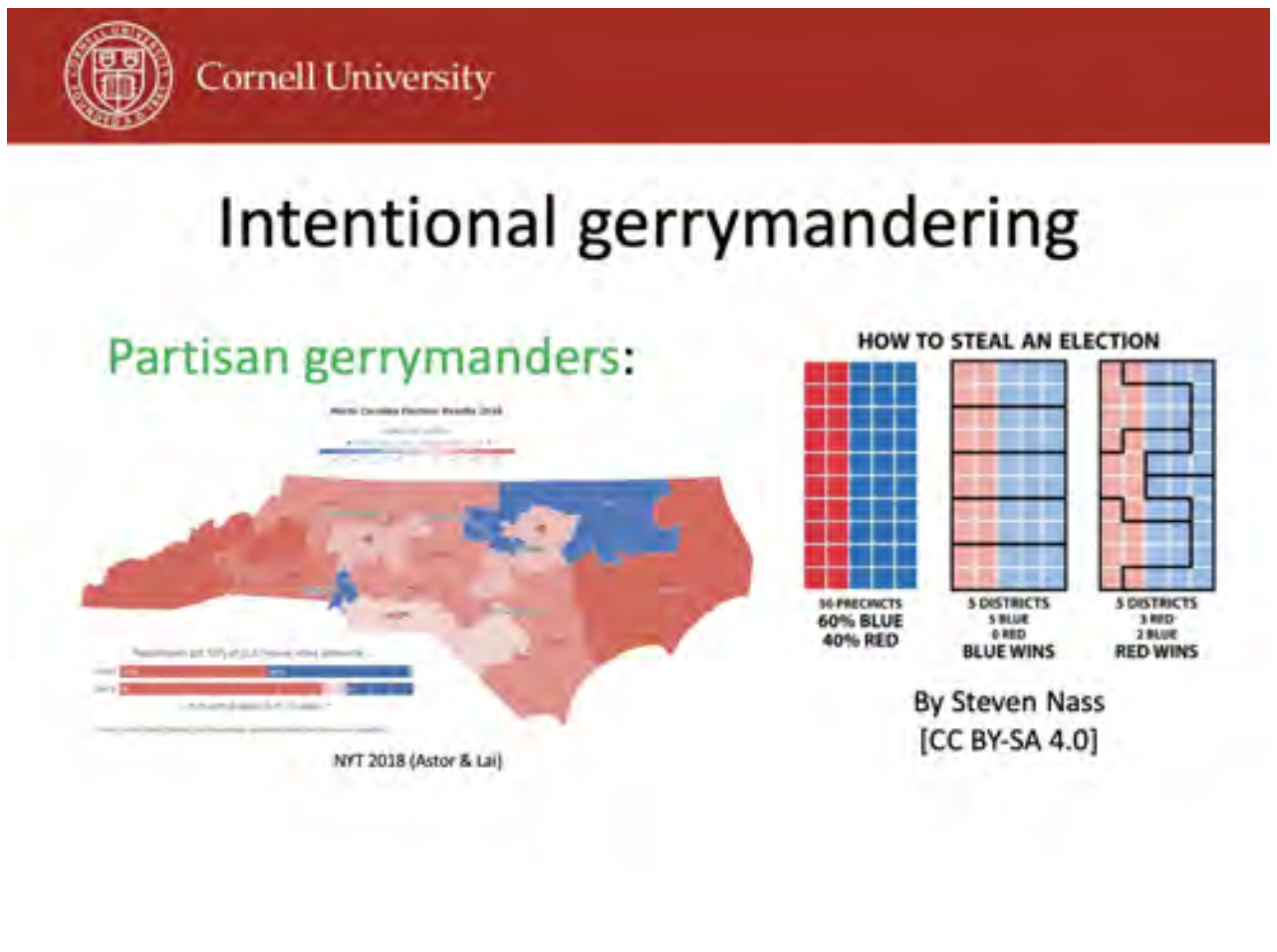

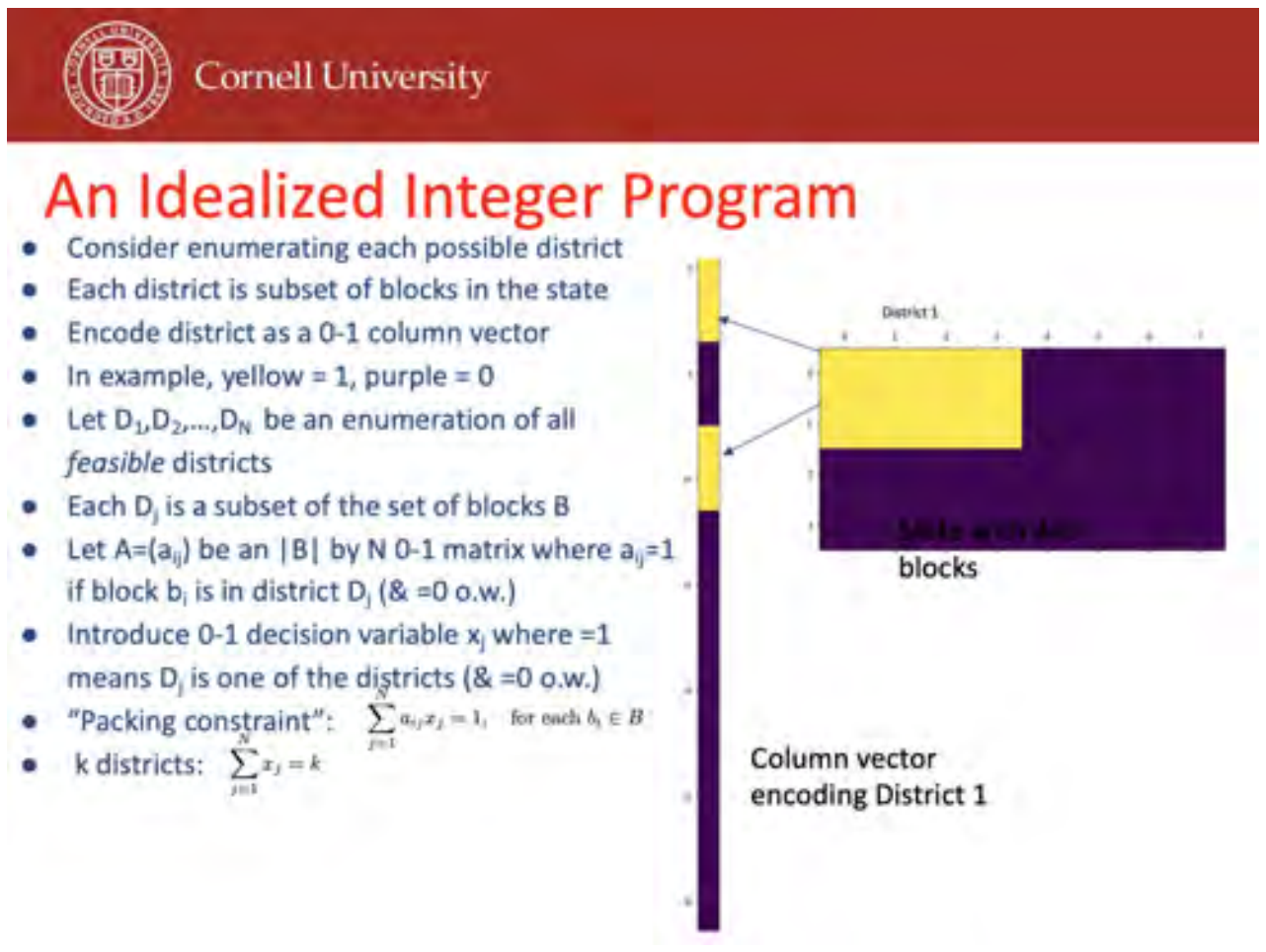

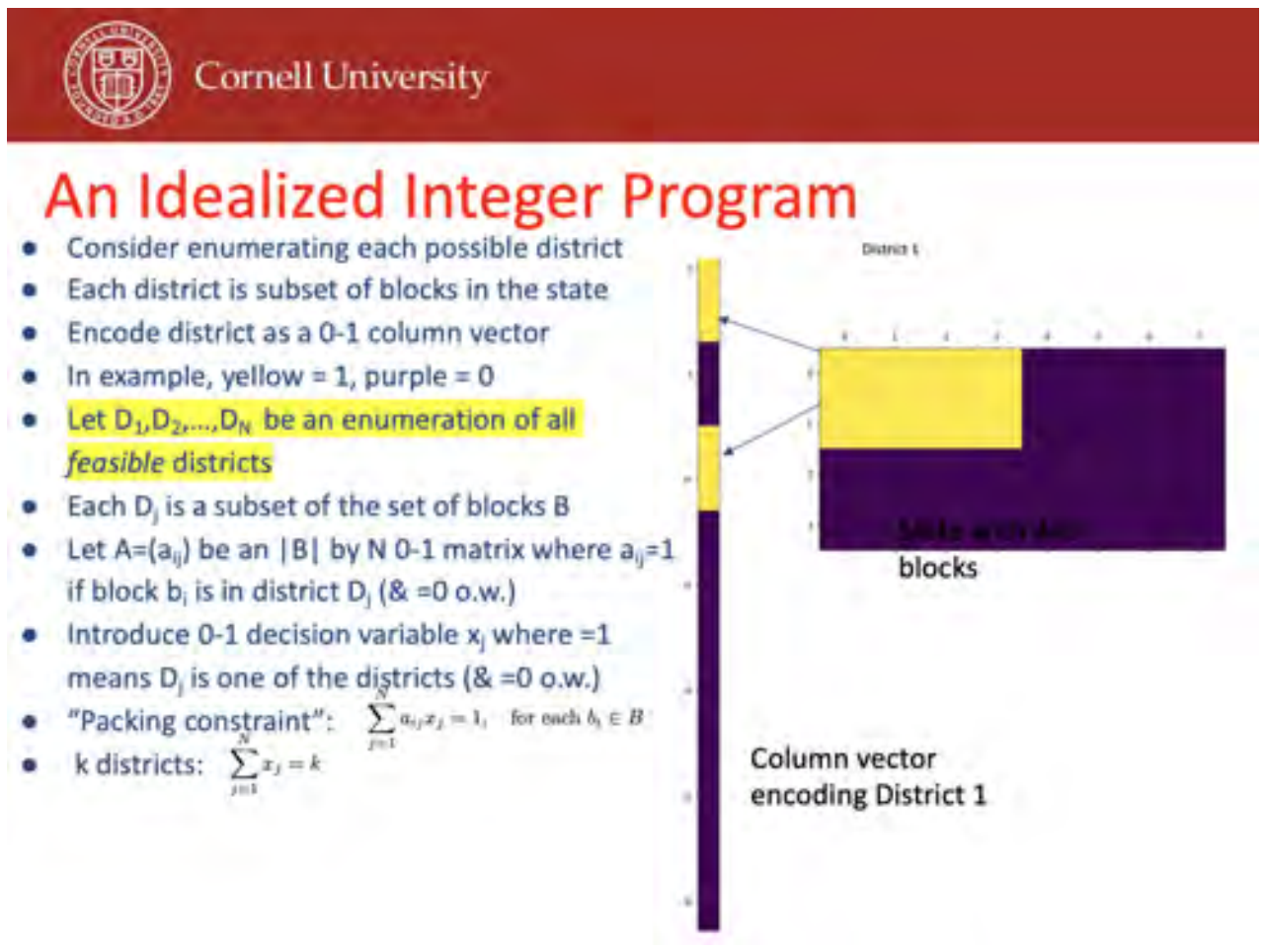

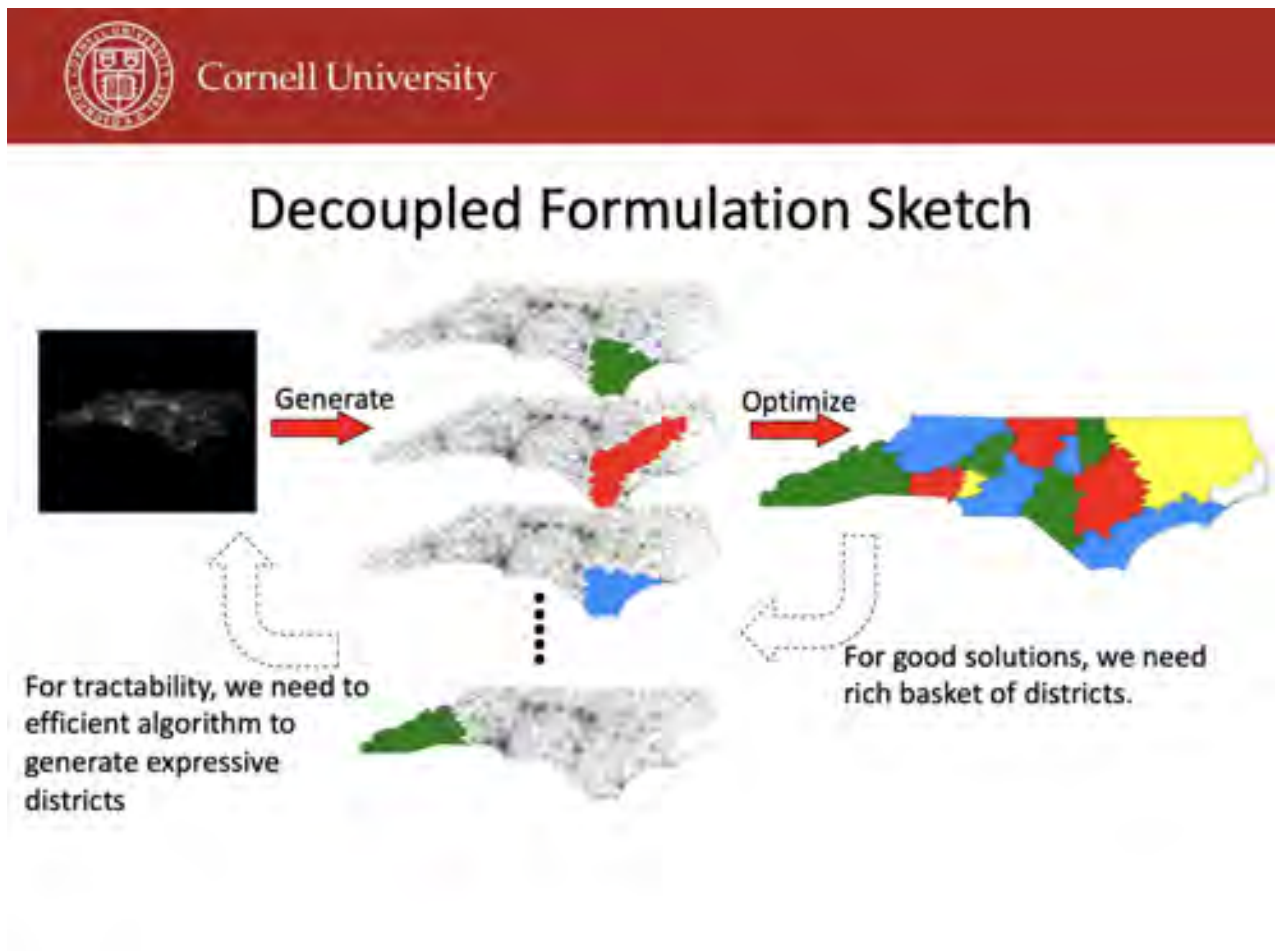
Cornell University
Decoupled Formulation Sketch
Generate
Optimize
For tractability, we need to efficient algorithm to generate expressive districts
For good solutions, we need rich basket of districts.

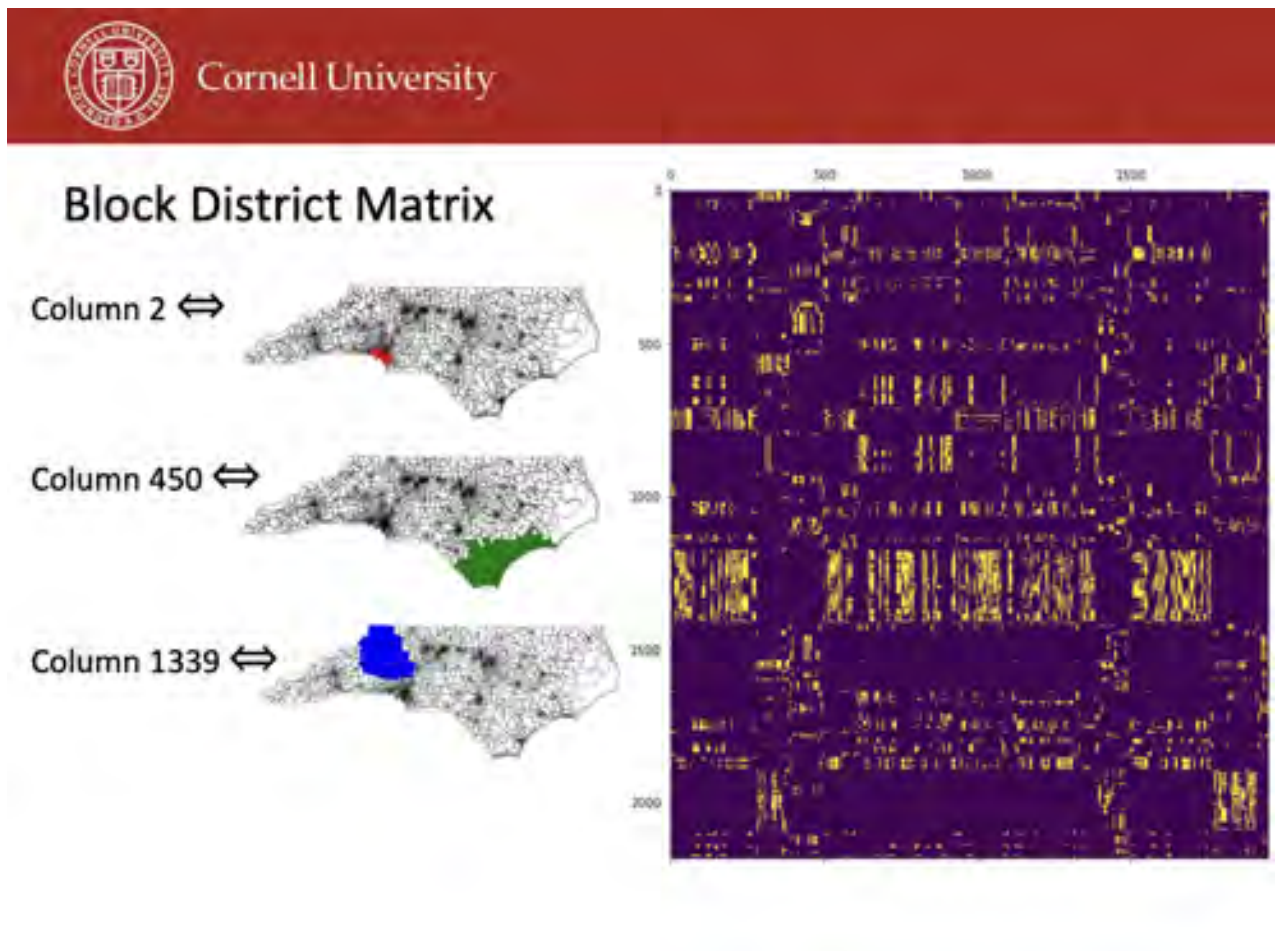
Cornell University
Block District Matrix
Column 2 ⇔
Column 450 ⇔
Column 1339 ⇔

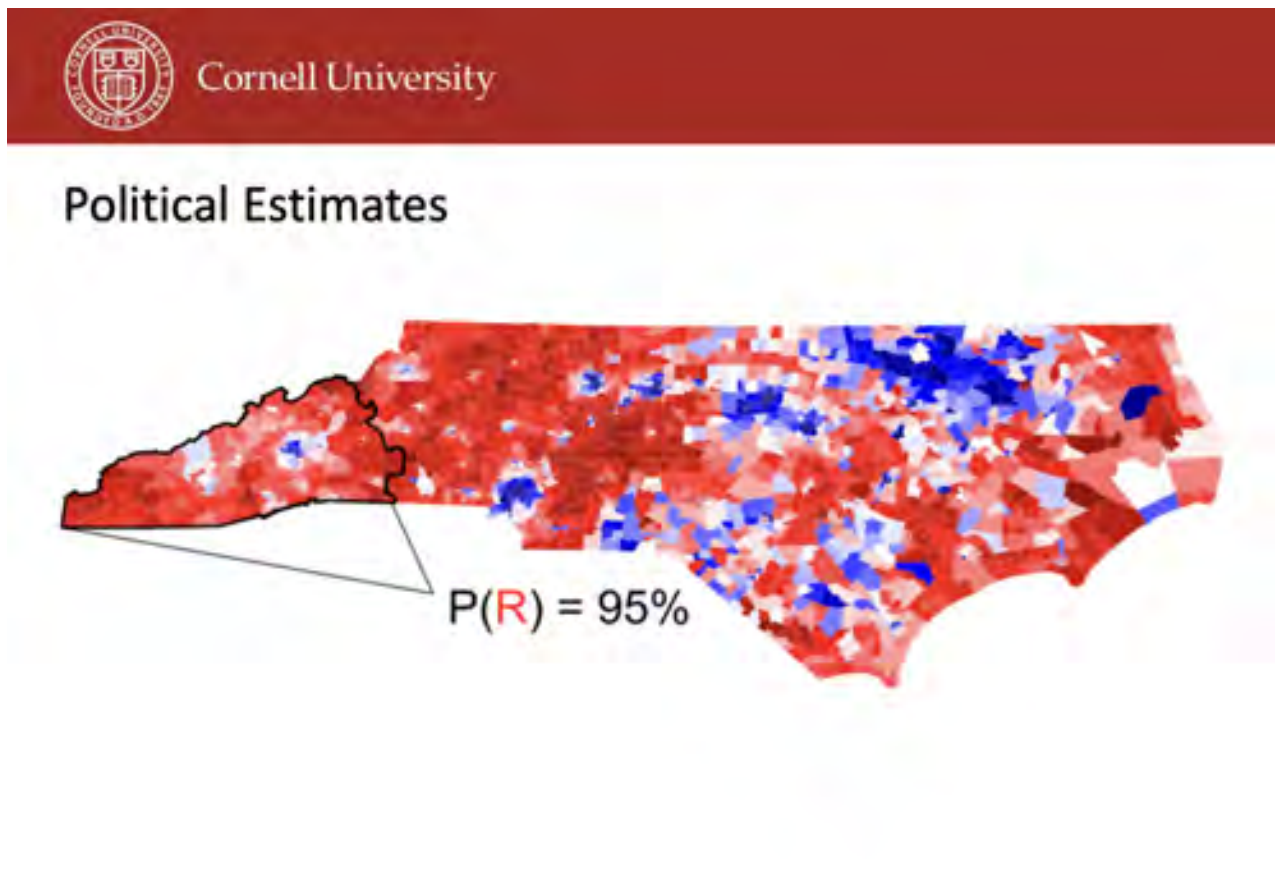
Cornell University
Political Estimates
P(R) = 95%

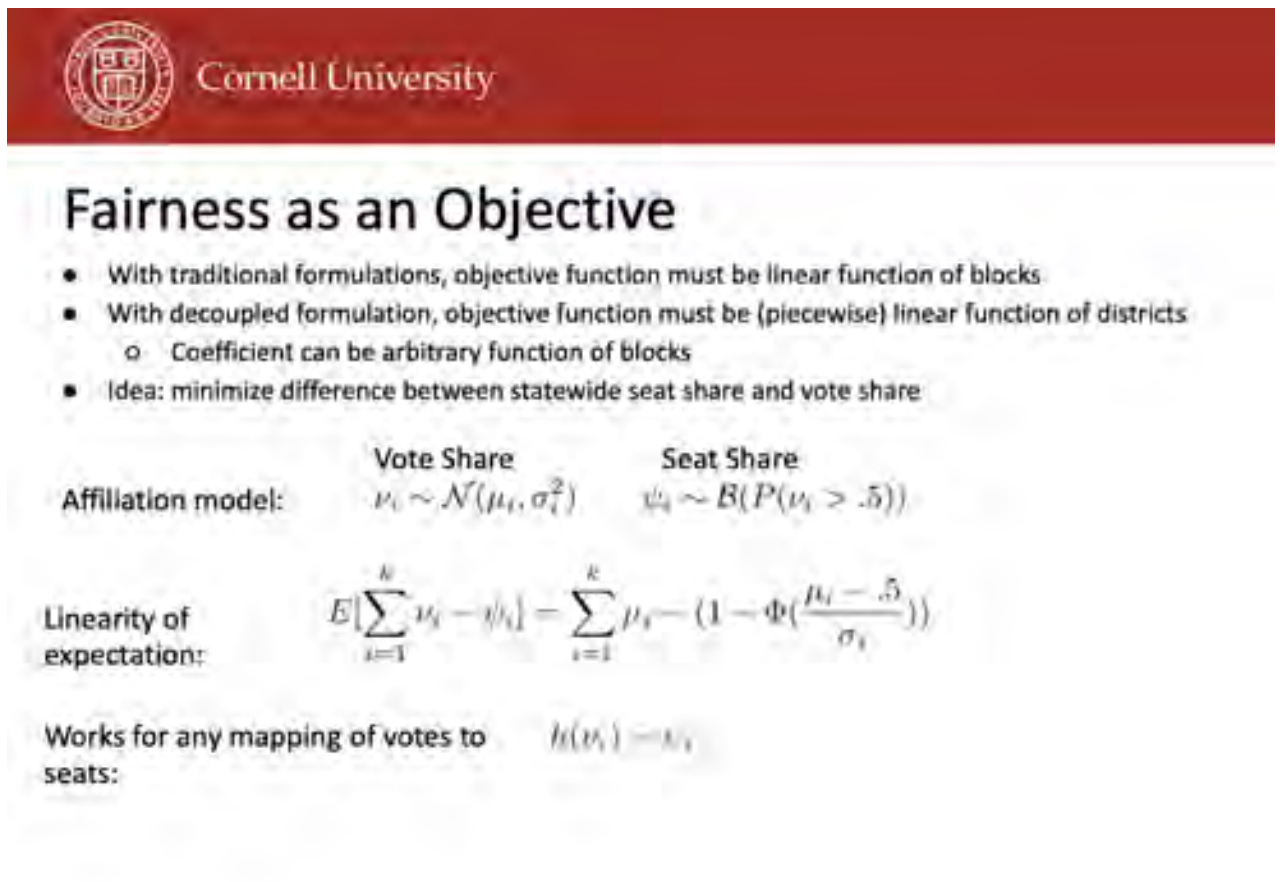
Cornell University
Fairness as an Objective
With traditional formulations, objective function must be linear function of blocks
With decoupled formulation, objective function must be (piecewise) linear function of districts
Coefficient can be arbitrary function of blocks
Idea: minimize difference between statewide seat share and vote share
Vote Share
Seat Share
Affiliation model:
Linearity of expectation:
Works for any mapping of votes to seats:

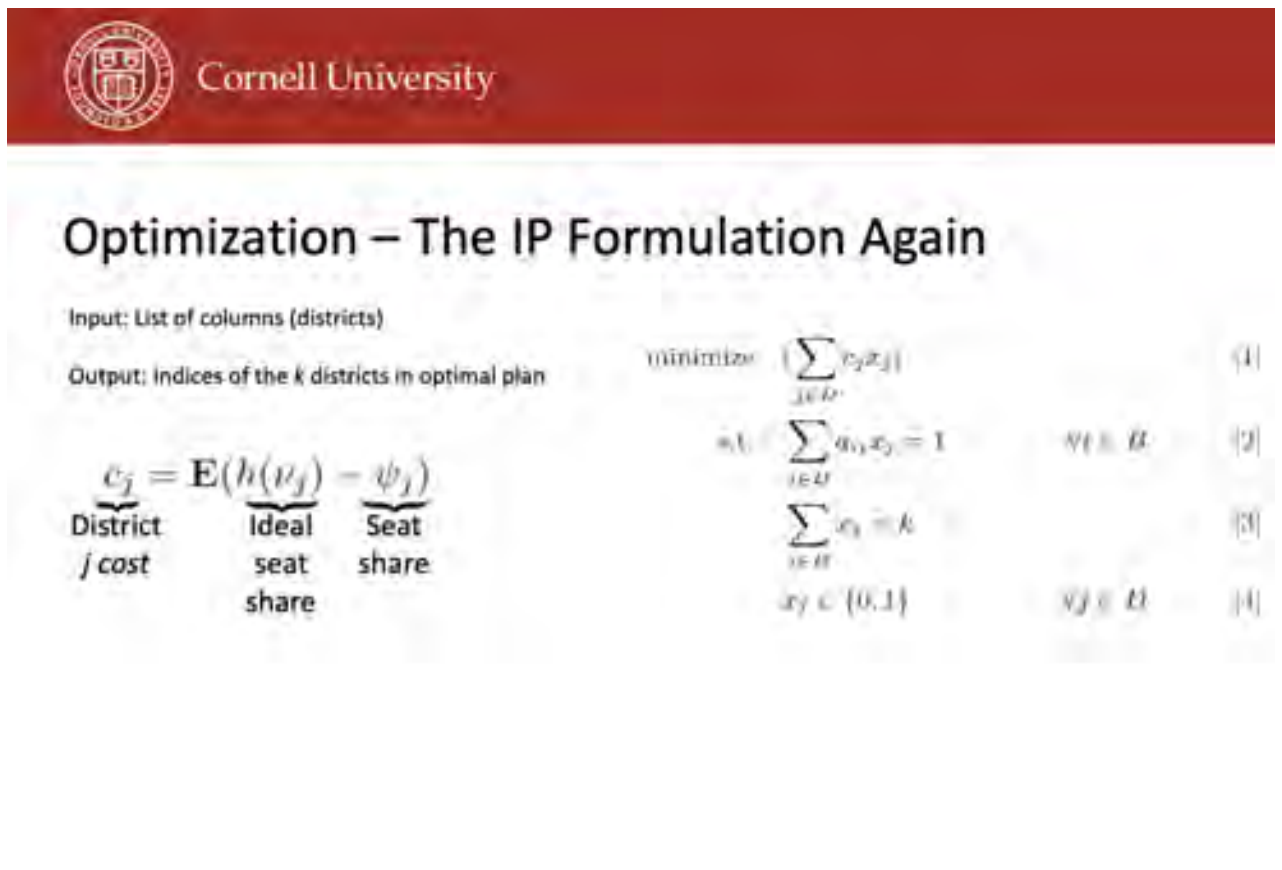
Cornell University
Optimization – The IP Formulation Again
Input: List of columns (districts)
Output: indices of the k districts in optimal plan
District
j cost
Ideal
seat
share
Seat
share
minimize

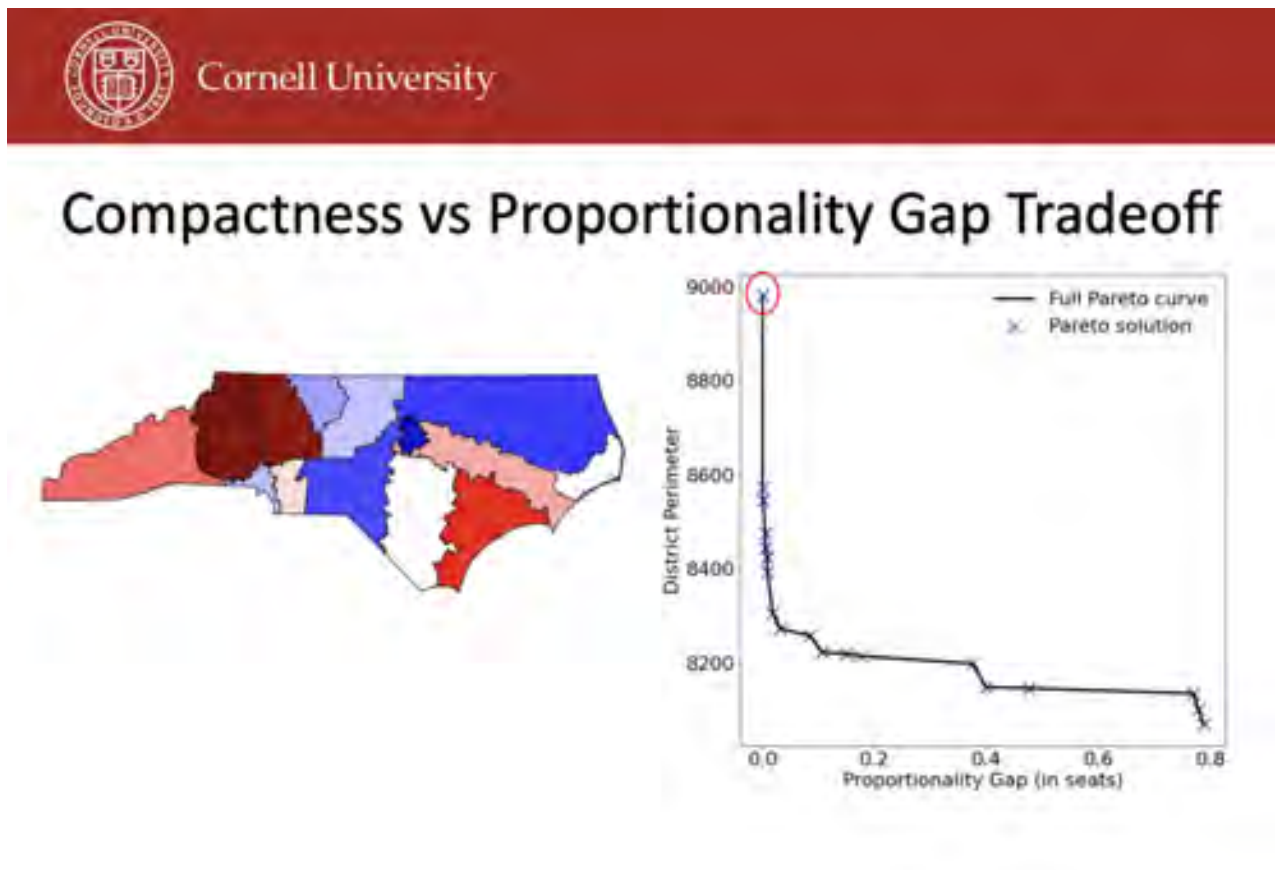

Cornell University

## Compactness vs Proportionality Gap Tradeoff

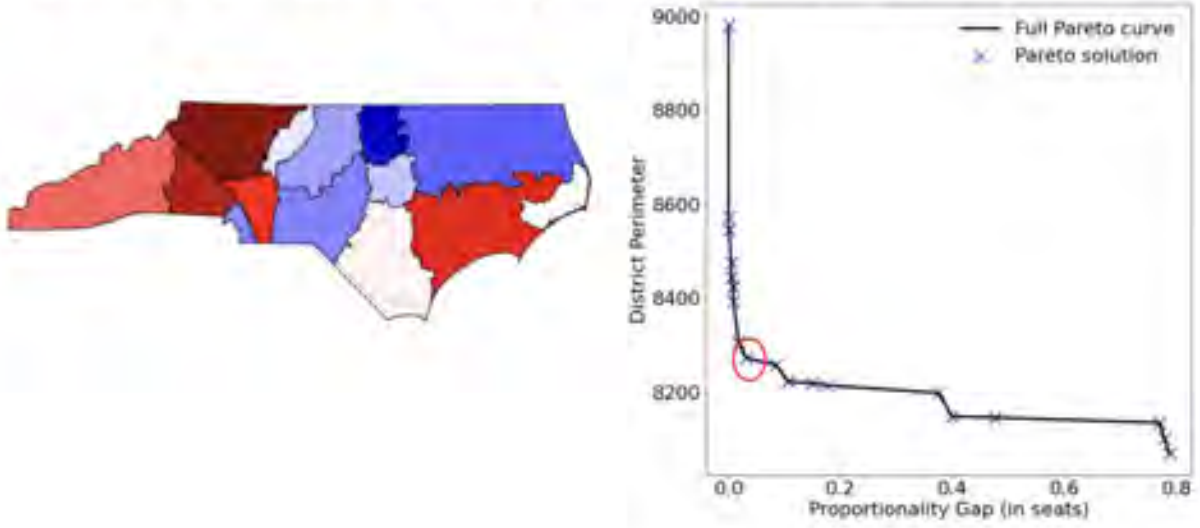

Cornell University

## Seat Share Possible Outcomes

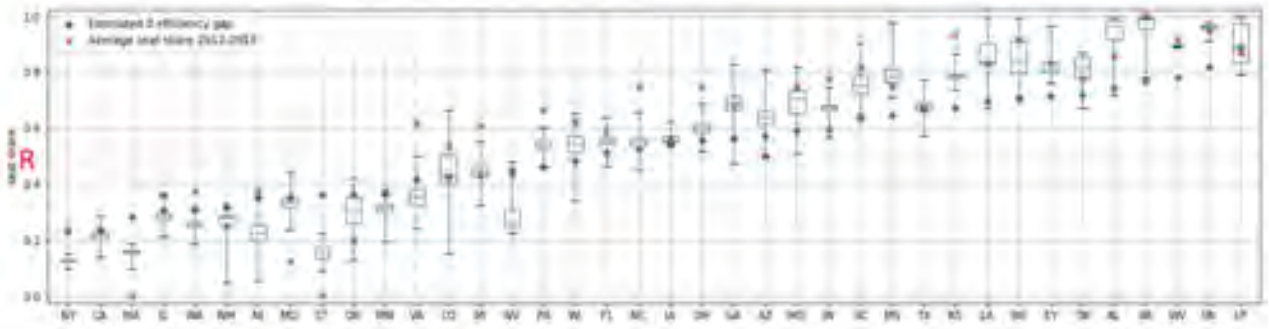

- 43% - 62% Expected Democratic Seat Share
- 32-106 Expected seat swaps
- Only 15 states admit proportional plan

Cornell University

This bill requires (1) that ranked choice voting . . . be used for all elections for Members of the House of Representatives, (2) that states entitled to six or more Representatives establish districts such that three to five Representatives are elected from each district, and (3) that states entitled to fewer than six Representatives elect all Representatives on an at-large basis

—Fair Representation Act, H.R. 4000, 2019

We show that 2- or 3-member districts with STV are enough to both *inhibit partisan gerrymanders* and *eliminate natural gerrymanders,* without sacrificing "representative" democracy

Cornell University

- [Wes Gurnee, DBS]
  **Fairmandering: A Column Generation Heuristic for Fairness-Optimized Political Districting.** Proceedings of the 1st SIAM Conference on Applied and Computational Discrete Algorithms (2021).
- [Nikhil Garg, Wes Gurnee, David Rothschild, DBS]
  **Combatting Gerrymandering with Social Choice: the Design of Multi-member Districts.** EC 2022: The 23rd ACM Conference on Economics and Computation, 560-561

Cornell University

## Fairness as the Objective
## Congressional Districting

- Use historical data to learn distribution for partisan split in any proposed district
- Choose a fairness measure such as "efficiency gap" or "proportionality gap"
- Optimize choice of statewide districting plan to optimize fairness

Cornell University

## Fairness as a Constraint

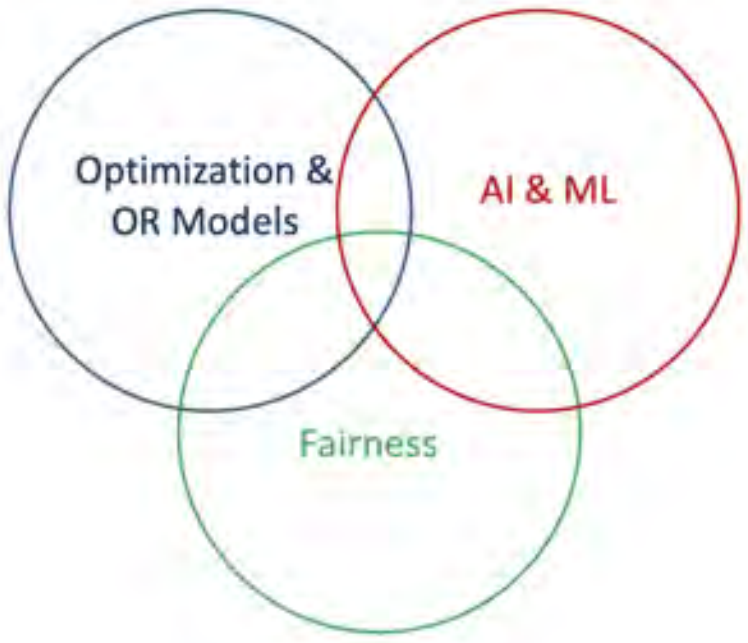

Cornell University

## Fairness as a Constraint
### Managing Ride-Share

- Use historical data to learn rider demand, driver supply, transit times
- Optimize pricing and dispatch policy to maximize long-term profit
- Ensure these policies are "fair"

- But what is "fair"?

- Between robustness & differential privacy, need a notion of optimizing indifference

# THANK YOU!

Figure B-6: Hamsa Bastani - Decision-Aware Reinforcement Learning

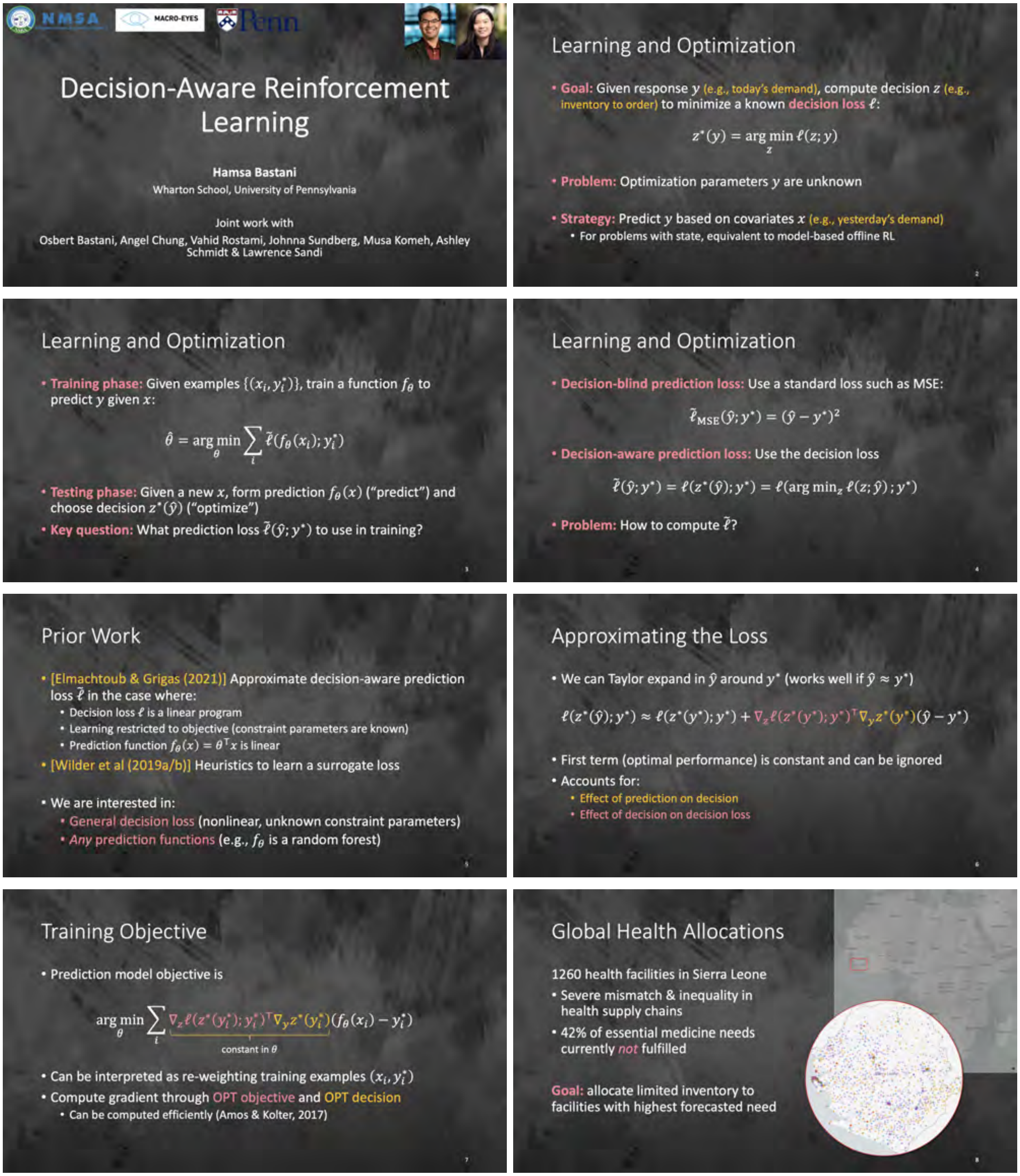

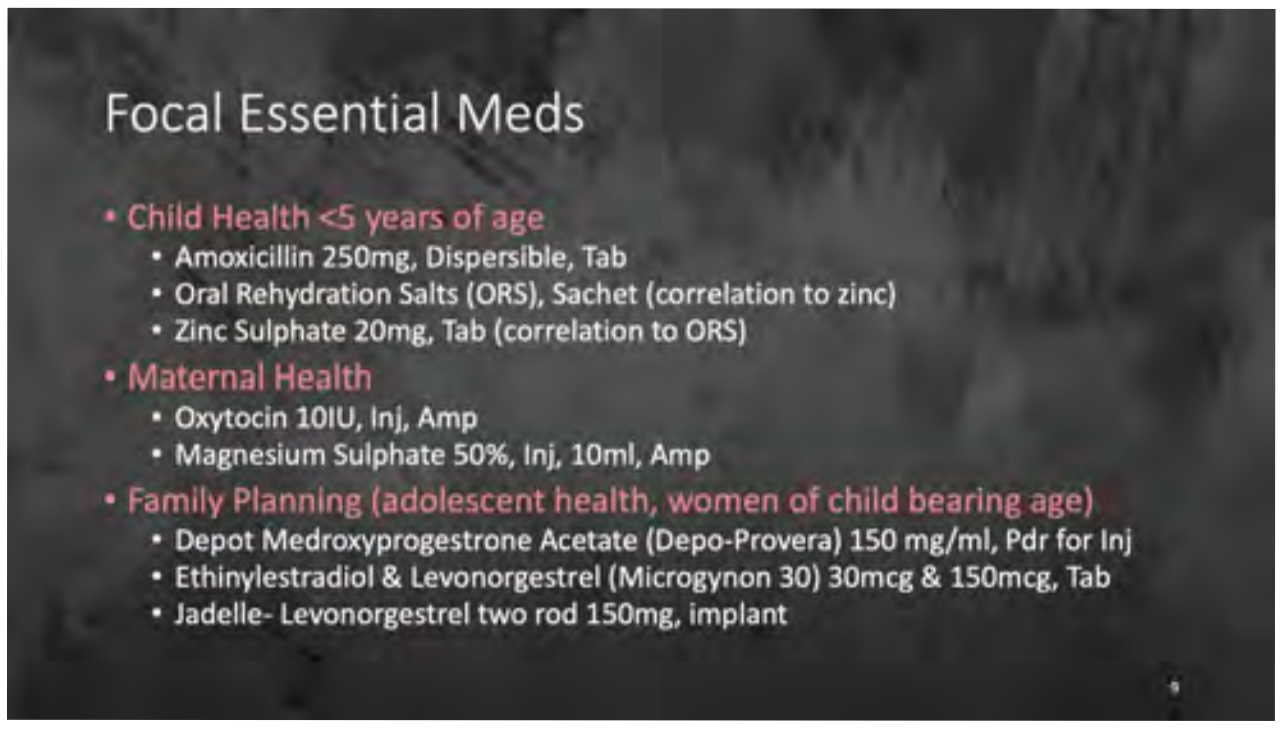
Focal Essential Meds
• Child Health <5 years of age
• Amoxicillin 250mg, Dispersible, Tab
• Oral Rehydration Salts (ORS), Sachet (correlation to zinc)
• Zinc Sulphate 20mg, Tab (correlation to ORS)
• Maternal Health
• Oxytocin 10IU, Inj, Amp
• Magnesium Sulphate 50%, Inj, 10ml, Amp
• Family Planning (adolescent health, women of child bearing age)
• Depot Medroxyprogestrone Acetate (Depo-Provera) 150 mg/ml, Pdr for Inj
• Ethinylestradiol & Levonorgestrel (Microgynon 30) 30mcg & 150mcg, Tab
• Jadelle- Levonorgestrel two rod 150mg, implant
9

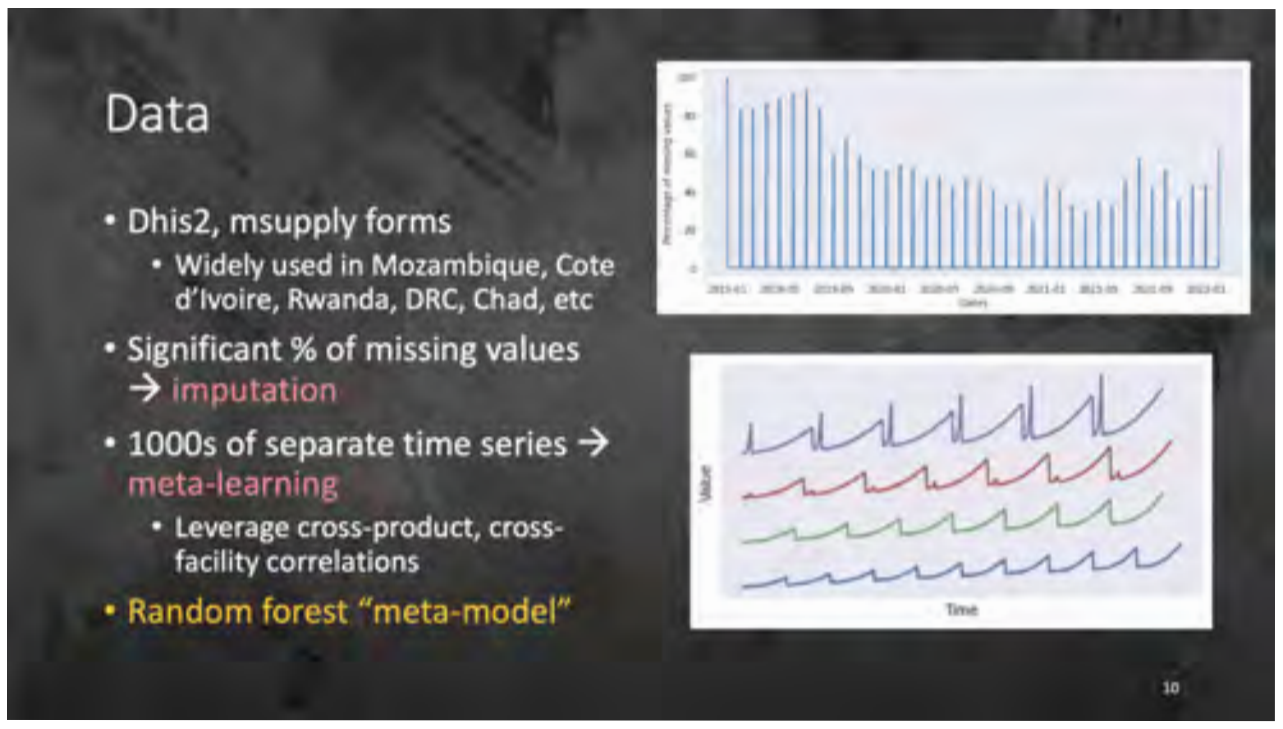
Data
• Dhis2, msupply forms
• Widely used in Mozambique, Cote d'Ivoire, Rwanda, DRC, Chad, etc
• Significant % of missing values → imputation
• 1000s of separate time series → meta-learning
• Leverage cross-product, cross-facility correlations
• Random forest "meta-model"
Value
Time
10

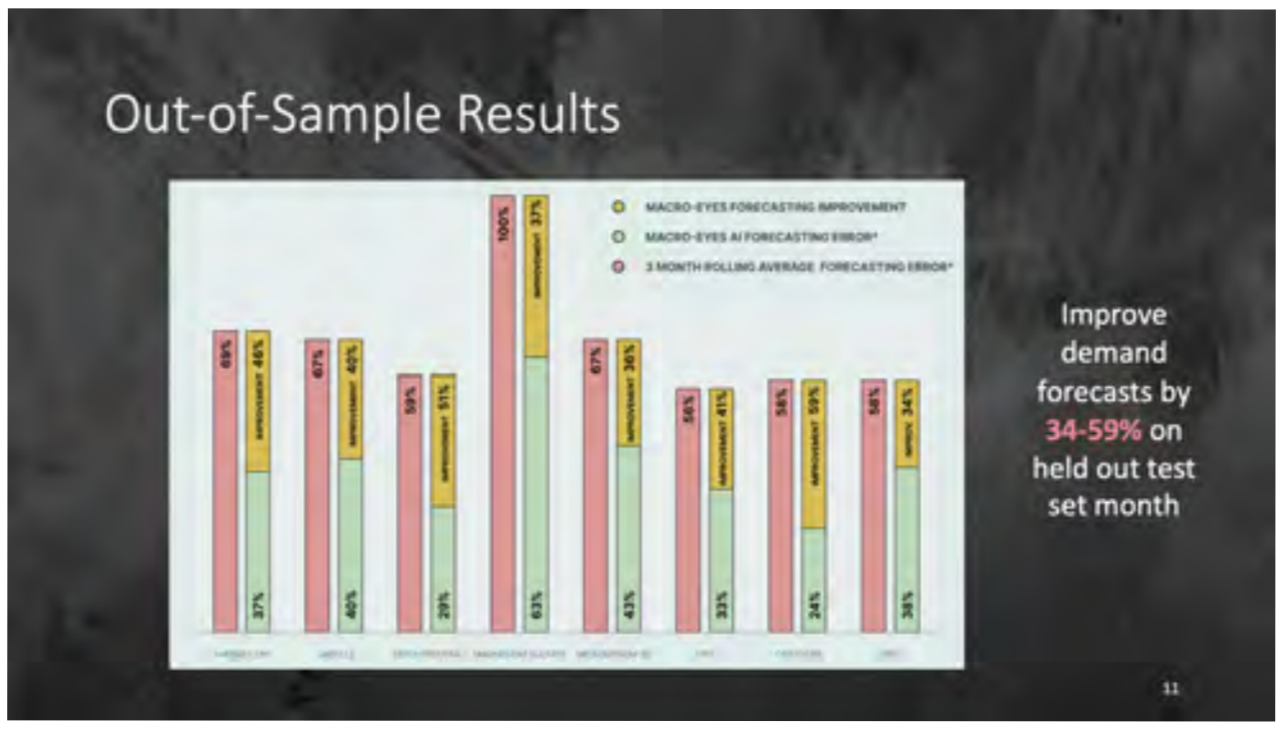
Out-of-Sample Results
MACRO-EYES FORECASTING IMPROVEMENT
MACRO-EYES AI FORECASTING ERROR*
3 MONTH ROLLING AVERAGE FORECASTING ERROR*
Improve demand forecasts by 34-59% on held out test set month
11

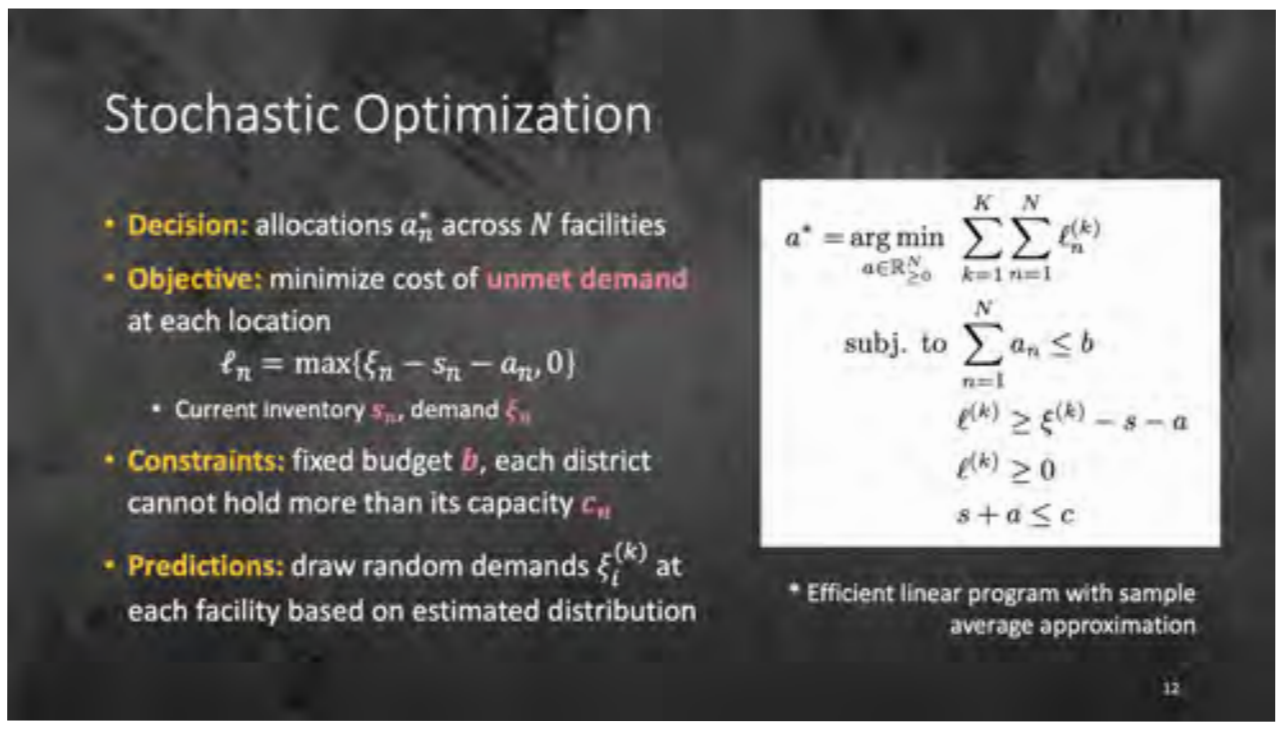
Stochastic Optimization
• Decision: allocations $a_n^*$ across $N$ facilities
• Objective: minimize cost of unmet demand at each location
$\ell_n = \max\{\xi_n - s_n - a_n, 0\}$
• Current inventory $s_n$, demand $\xi_n$
• Constraints: fixed budget $b$, each district cannot hold more than its capacity $c_n$
• Predictions: draw random demands $\xi_i^{(k)}$ at each facility based on estimated distribution
$a^* = \arg\min_{a \in \mathbb{R}^N_{\geq 0}} \sum_{k=1}^{K} \sum_{n=1}^{N} \ell_n^{(k)}$
subj. to $\sum_{n=1}^{N} a_n \leq b$
$\ell^{(k)} \geq \xi^{(k)} - s - a$
$\ell^{(k)} \geq 0$
$s + a \leq c$
* Efficient linear program with sample average approximation
12

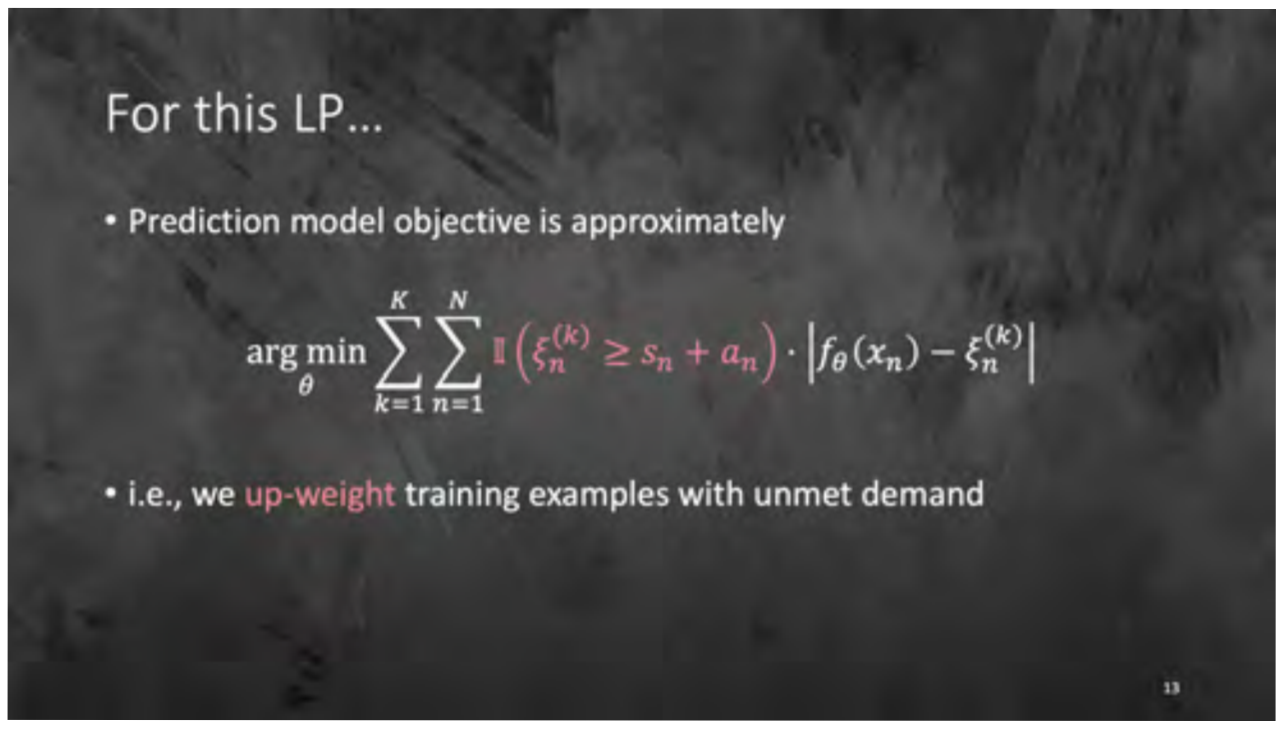
For this LP...
• Prediction model objective is approximately
$\arg\min_{\theta} \sum_{k=1}^{K} \sum_{n=1}^{N} \mathbb{I}\left(\xi_n^{(k)} \geq s_n + a_n\right) \cdot \left|f_\theta(x_n) - \xi_n^{(k)}\right|$
• i.e., we up-weight training examples with unmet demand
13

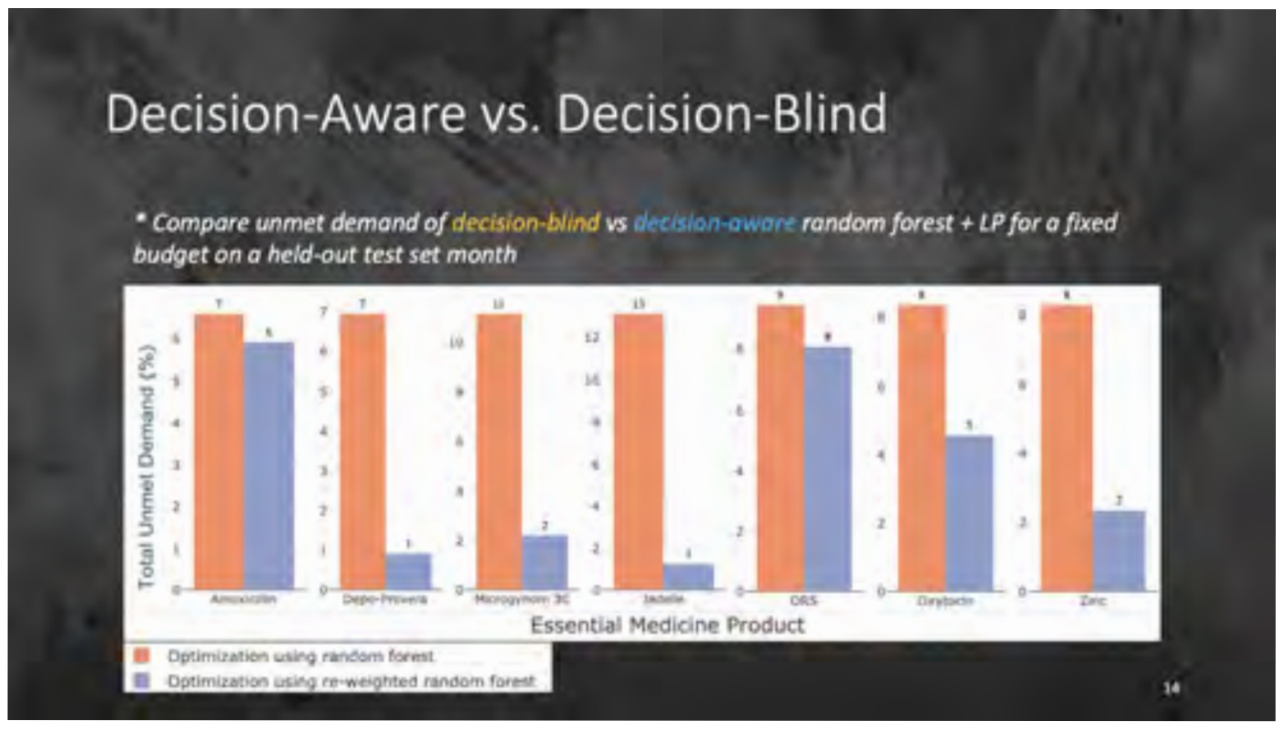
Decision-Aware vs. Decision-Blind
* Compare unmet demand of decision-blind vs decision-aware random forest + LP for a fixed budget on a held-out test set month
Total Unmet Demand (%)
Amoxicillin
Depo-Provera
Microgynon 30
Jadelle
ORS
Oxytocin
Zinc
Essential Medicine Product
Optimization using random forest
Optimization using re-weighted random forest
14

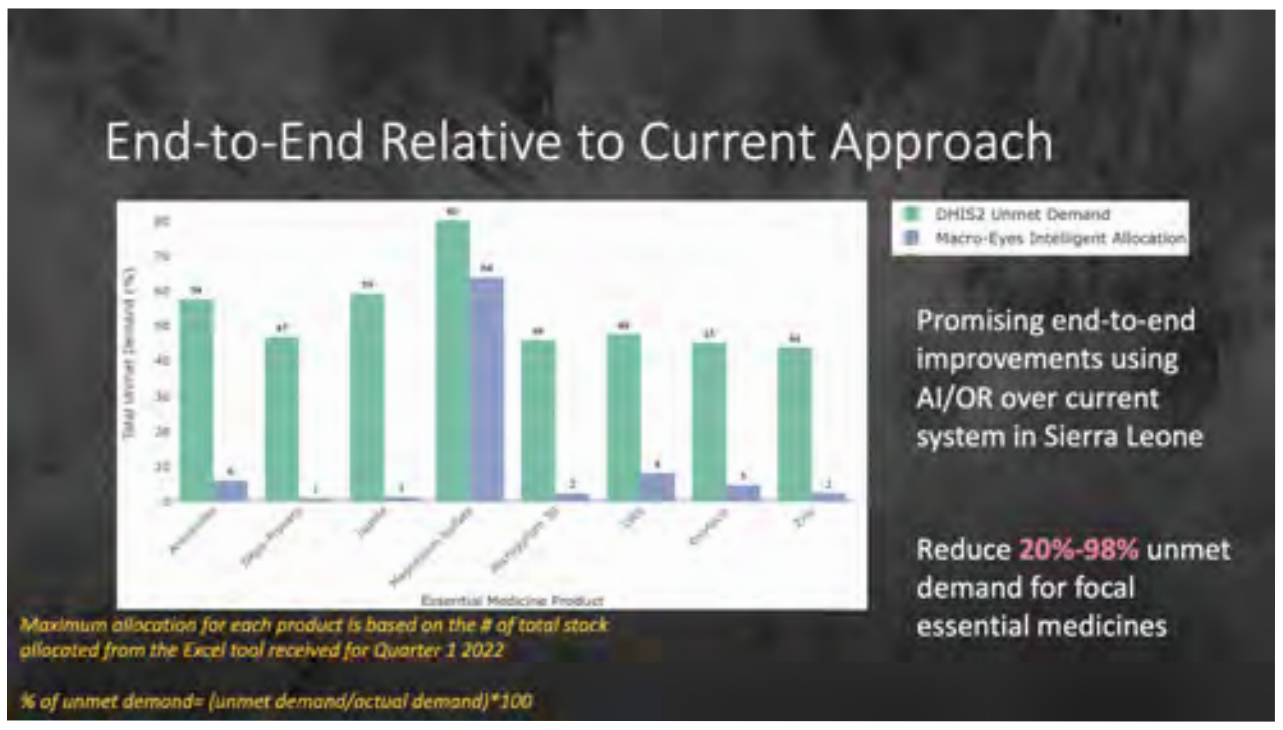
End-to-End Relative to Current Approach
DHIS2 Unmet Demand
Macro-Eyes Intelligent Allocation
Total Unmet Demand (%)
Essential Medicine Product
Promising end-to-end improvements using AI/OR over current system in Sierra Leone
Reduce 20%-98% unmet demand for focal essential medicines
Maximum allocation for each product is based on the # of total stock allocated from the Excel tool received for Quarter 1 2022
% of unmet demand= (unmet demand/actual demand)*100

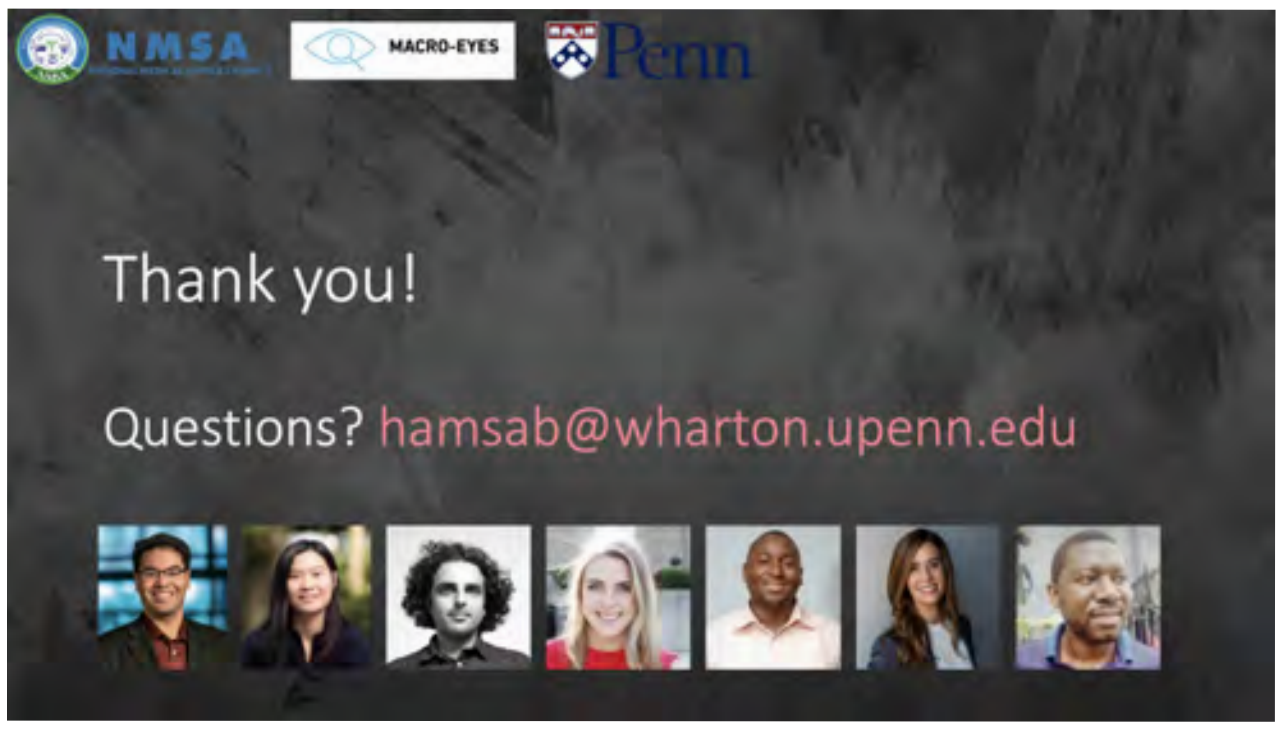
NMSA
MACRO-EYES
Penn
Thank you!
Questions? hamsab@wharton.upenn.edu

Figure B-7: Peter Frazier - Preference Learning for Stakeholder Management

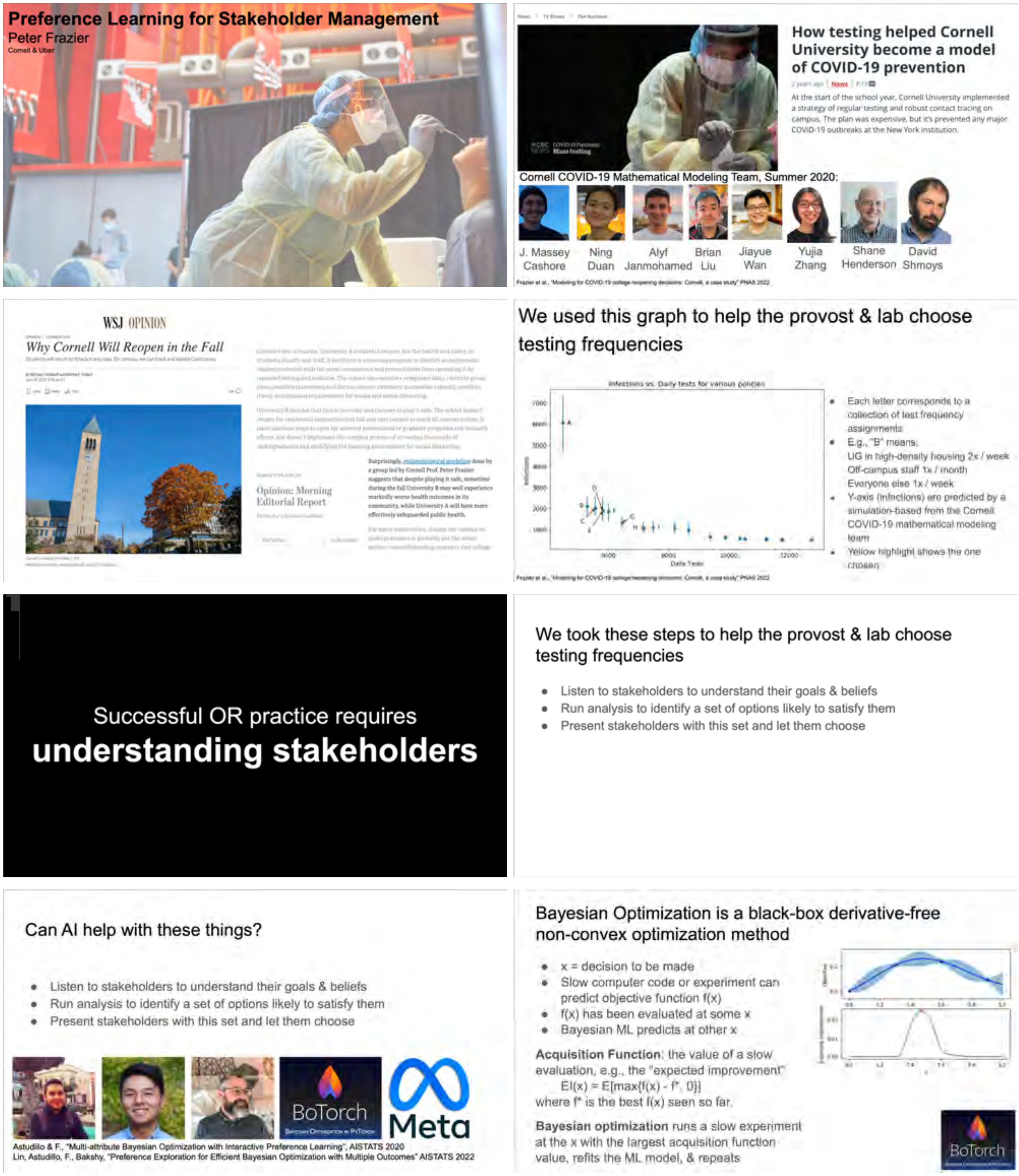

## Sometimes the objective is the utility of an outcome vector

- Slow experiment gives outcome vectors $[h_1(x), \ldots, h_k(x)]$
- u = utility function over outcomes
- Objective is $f(x) := u(\mathbf{h}(x))$
- Example:
  - $\mathbf{h}(x)$ = [quality(x), cost(x)]
  - $u(\mathbf{h}(x))$ = quality(x) - cost(x)

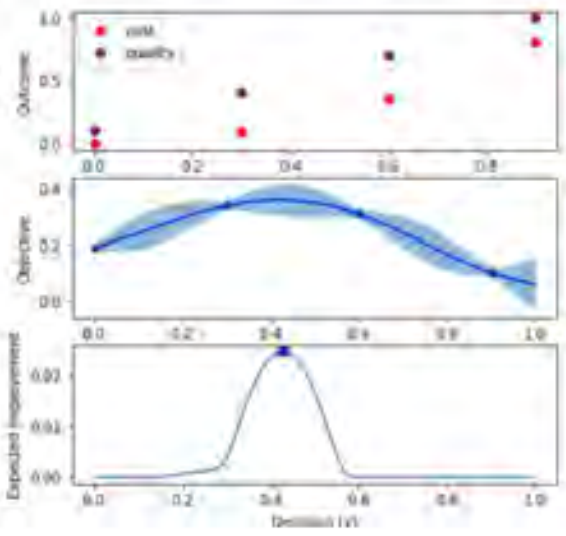

## Sometimes the objective is the utility of an outcome vector

You can also put a Bayesian ML model on $\mathbf{h}(x)$ and use
$EI(x) = E[\max\{u(\mathbf{h}(x)) - f^*, 0\}]$

Example:

- $\mathbf{h}(x)$ = [quality(x), cost(x)]
- $u(\mathbf{h}(x))$ = quality(x) - cost(x)

Astudillo & F. "Bayesian optimization of composite functions" ICML 2019

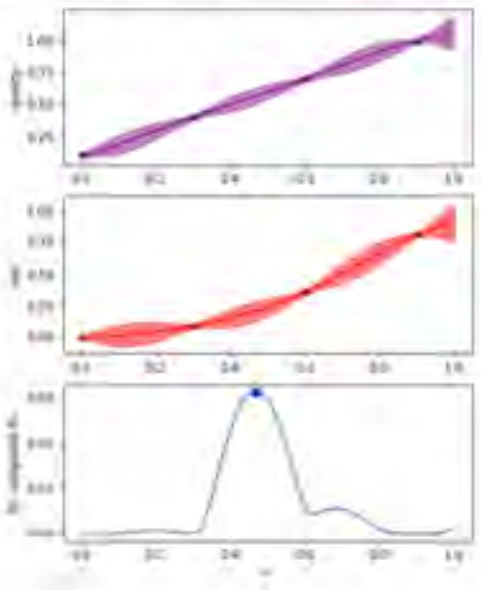

## If the utility function is unknown, we learn it with Bayesian preference learning

- Suppose we don't know the utility function
- Preference learning (for learning the utility fn):
  - Present the stakeholder pairs of outcome vectors. Ask, "Which do you prefer?"
  - Learn a Bayesian ML model on the utility function
  - Active learning decides which pairs to present
- Bayesian optimization (for choosing x to eval):
  - Prediction: Fit a GP to outcomes $\mathbf{h}(x)$
  - Acquisition function: $EI(x) = E[\max(u(\mathbf{h}(x)) - f^*, 0)]$ where the expectation E is over the random utility function $u(\cdot)$ and the outcome vector $\mathbf{h}(x)$
  - Run the slow computer code where EI(x) is largest
- Iterate between asking the stakeholder questions & running the slow computer code

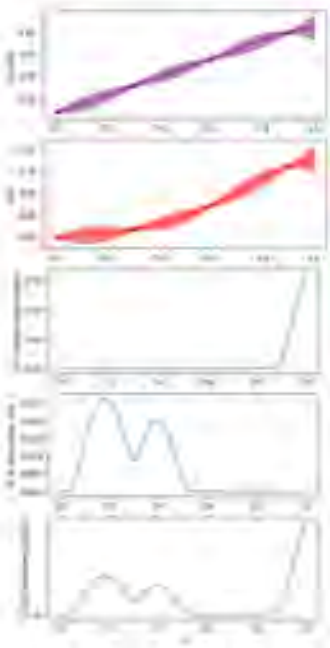

## This approach provides lots of value compared to multi-objective optimization

- In use at Meta for product improvement:
  - Stakeholder = product manager
  - Slow experiment = A/B experiment
  - Utility = Quality of Instagram, Facebook, etc.
- Can handle many outcomes; multi-objective optimization usually limited to 3 outcomes
- Can leverage preferences learned from active queries
- Can leverage passive stakeholder observations, e.g., choices in related decision problems

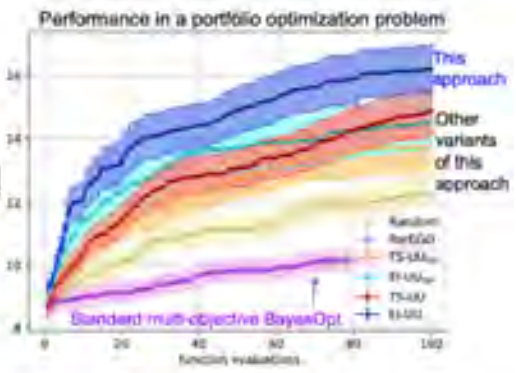

## There are many kinds of stakeholder engagement

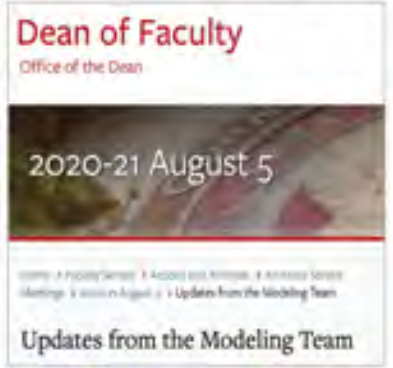

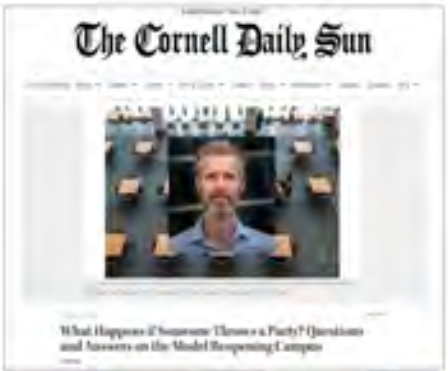

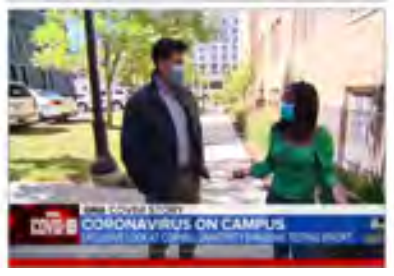

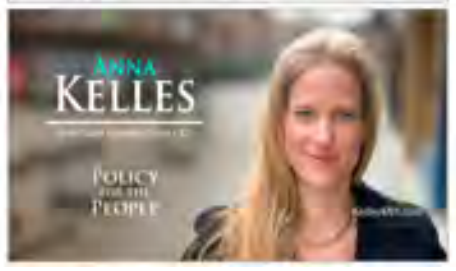

## Lots of room for innovation in AI-enabled stakeholder enagement for OR applications

- Understand stakeholder goals, beliefs & incentives
- Understand how groups of stakeholders influence each other
- Predict how stakeholders will react to communication
- Managing trust (in the OR analyst and her models)

Figure B-8: Kristian Lum - De-biasing “Bias” Measurement

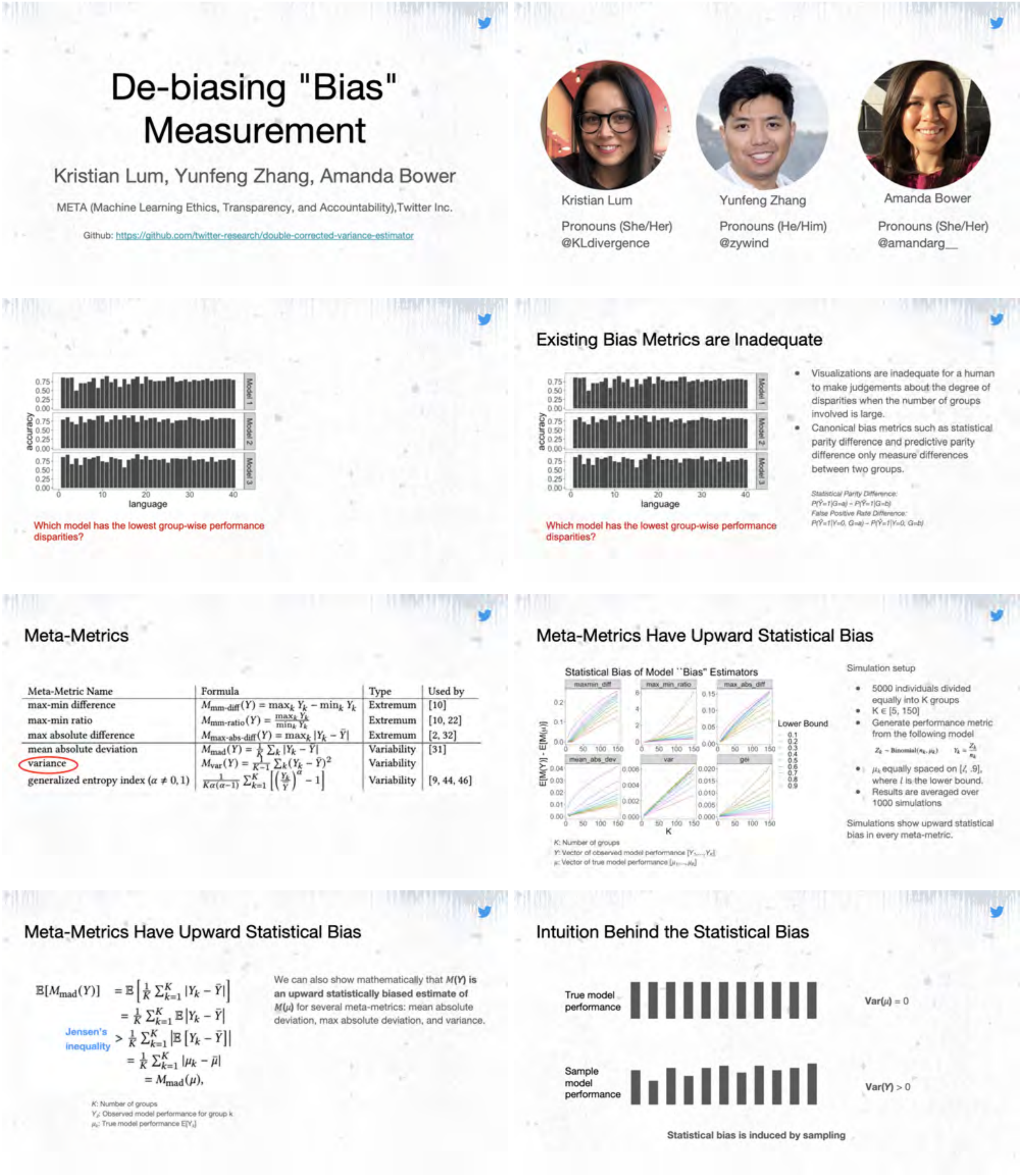

## Statistically Biased Meta-Metrics Are Problematic

1. The overestimation of the true group-wise performance disparity may cause unnecessary model adjustments. Because of fairness-accuracy trade-off, such adjustments could lead to less performant models for all individuals.
2. Statistically Biased meta-metrics cannot be compared fairly across different grouping methods. Different grouping could lead to different amount of bias.

## Correcting for Statistical Bias in The Variance Meta-Metric

$$\hat{M}_{\text{var}}(\mu) = M_{\text{var}}(Y) - \frac{1}{K}\sum_k \hat{\sigma}_k^2$$

Truncated:

$$\hat{M}_{\text{var}}(\mu) = \max(0, M_{\text{var}}(Y) - \tfrac{1}{K}\textstyle\sum_k \hat{\sigma}_k^2)$$

$\sigma_k$: Standard error of the observed model performance.
$M_{var}$: The between-group variance meta-metric.

- Introduced by Cochran (1954) and popularized by Hedge and Olkin (1985) for meta-analysis of between study variabi $\hat{\sigma}_k^2$.
- Larger I (e.g. caused by small sample size) leads to larger statistical bias.
- Untruncated version is statistically unbiased when statistically unbiased estimates of the standard errors are available. We follow the convention to use the truncated version to preclude negative variance estimates.

## Simulation shows that the correction works

n=5000, K=100

Equal group size, $n_k$=50
Unequal group size, $n_k$=10 to 90

Equal performance, $\mu_k$=0.8, $M_{var}(\mu)$=0
Unequal performance, $\mu_k$=0.1 to 0.9, $M_{var}(\mu)$=0.055

**Black vertical lines show the true variance.**

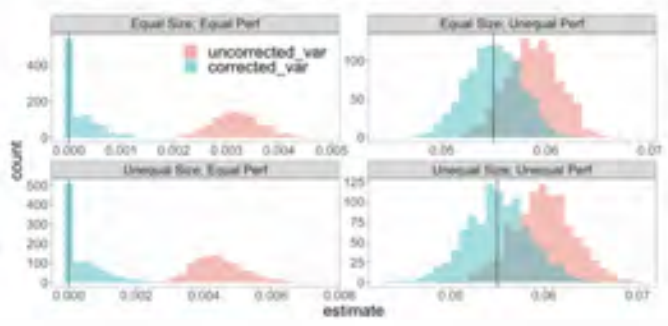

Simulation:

- Generate samples for each group using the parameters on the left.
- Calculate sampling variance of each group as: $\hat{\sigma}_k^2 = \frac{Y(1-Y)}{n_k}$
- Apply the $M_{var}(\mu)$ equation.
- Repeat 1000 times.

**The distributions of corrected variance estimates are centered at the truth.**

## Uncertainty Quantification for MetaMetrics

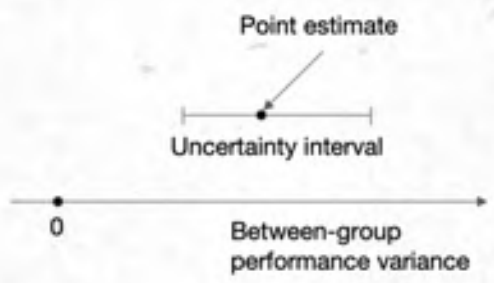

Uncertainty intervals inform decision makers about the confidence of the statistical estimate.

## Uncertainty Quantification By Bootstrapping

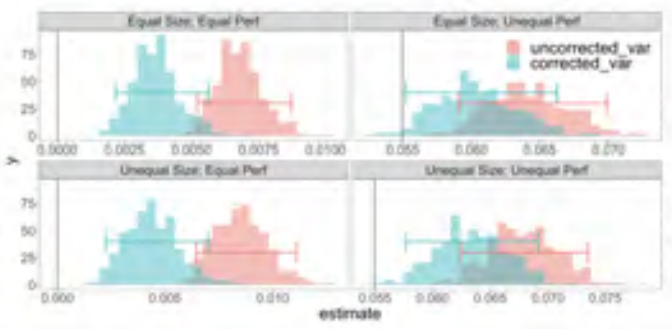

$X_k \sim \text{Binomial}(n_k, \mu_k)$ $\quad Y_k = \frac{X_k}{n_k}$

$X_k^* \sim \text{Binomial}(n_k, Y_k)$ $\quad Y_k^* = \frac{X_k^*}{n_k}$

Naive application of the corrected variance estimator to each bootstrap sample results in upwardly shifted uncertainty intervals that do not even cover the true value.

## Correcting for Statistical Bias in Uncertainty Intervals

Bootstrapping induces variance by itself:

$$\text{Var}(Y_k^*) \quad = \quad \mathbb{E}(\text{Var}(Y_k^* \mid Y_k)) + \text{Var}(\mathbb{E}(Y_K^* \mid Y_K))$$

Bootstrap induced var $\qquad$ Sampling var $\hat{\sigma}_k^2 = \frac{Y(1-Y)}{n_k}$

Sampling variance of bootstrapped $Y_k^*$ is:

$$\hat{\sigma}_k^{2*} = \frac{2Y_k^*(1-Y_k^*)}{n_k} - \frac{Y_k^*(1-Y_k^*)}{n_k^2}$$

## Corrected Uncertainty Quantification

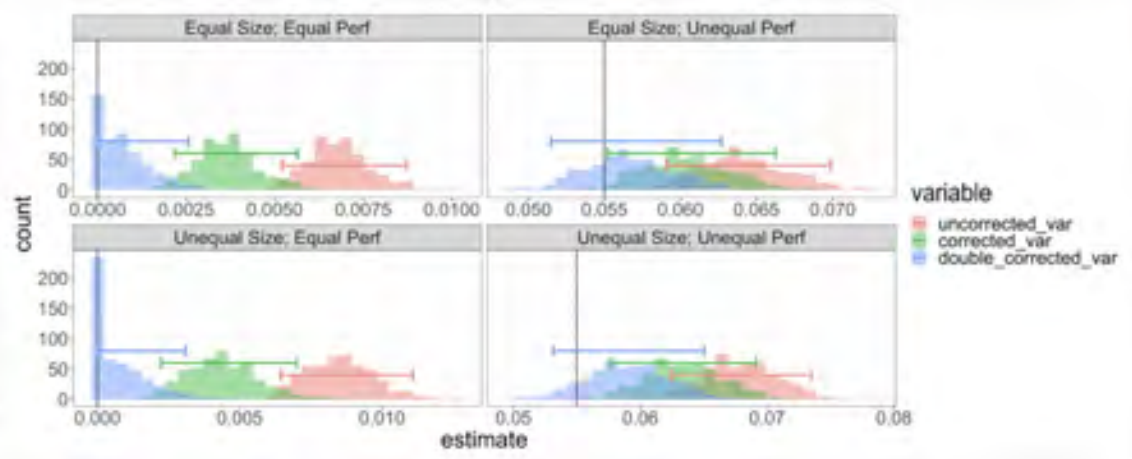

## Corrected Uncertainty Quantification

| | uncorrected_var | corrected_var | double_corrected_var |
|---|---|---|---|
| Equal Size; Equal Perf | 0.0 | 0.0 | 99.7 |
| Unequal Size; Equal Perf | 0.0 | 0.0 | 99.3 |
| Equal Size; Unequal Perf | 15.4 | 67.6 | 94.9 |
| Unequal Size; Unequal Perf | 10.4 | 60.4 | 93.0 |

**Table 3: Empirical coverage of the 95% bootstrap intervals over 1000 replicates.**

## Application on The Adult Income Dataset

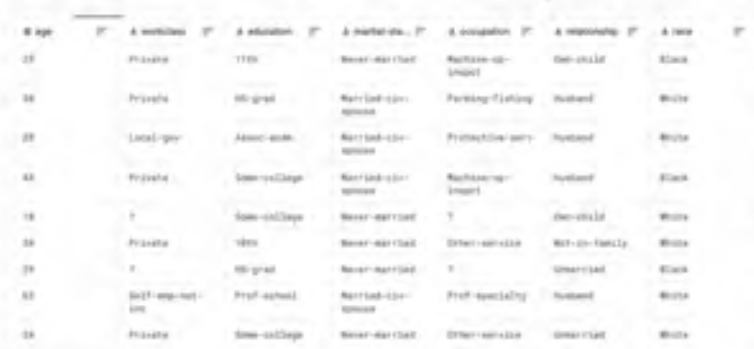

- ~48K individuals from the 1994 census database.
- 14 features
- Label: whether an individual's annual income was above $50K
- Split the data into 70% train and 30% test
- Trained a gradient-boosted trees classifier.
- 87% accuracy.

## Application on The Adult Income Dataset

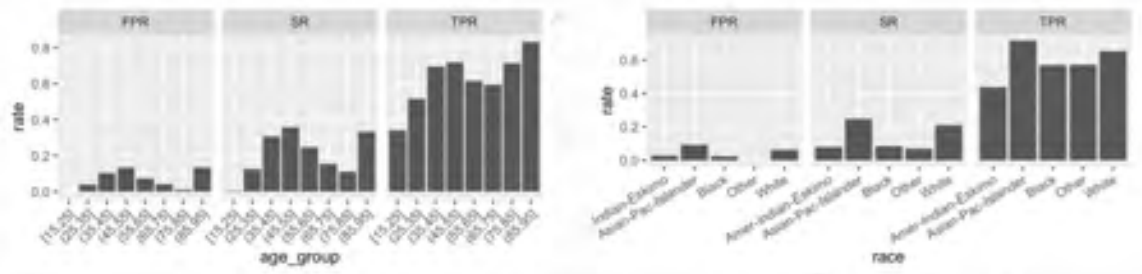

Age: 8 groups. 15-95 at 10 year interval.
Race: 5 groups. White, Black, American Indian and Eskimo, Asian and Pacific Islander, and other.
Performance metrics: False positive rate, selection rate, and true positive rate.

## Application on The Adult Income Dataset

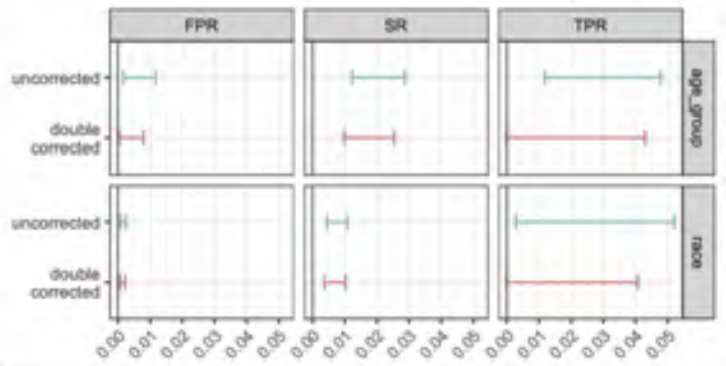

The double corrected uncertainty intervals show that in some cases, had we used standard methods, we could have **erroneously** concluded that there are large disparities with statistical confidence.

## Contributions and Conclusion

1. We identified meta-metrics for measuring group-wise model performance disparities, particularly in consideration of large numbers of groups.
2. We showed that these meta-metrics are statistically biased measurements.
3. We developed an unbiased estimator for between-group variance based on prior work.
4. We also developed a double-corrected estimator for quantifying the uncertainty of between-group variance.

**Future work**

- Examine other methods for measuring between-group variability, particularly those from the meta-analysis literature.
- Investigate corrections other metametrics such as max-min difference.

**Caveat:** Meta-metrics cannot capture the entirety of the impact of ML systems. Small measured disparities should not be taken as a guarantee that the system is fair or free from adverse impacts.

Figure B-9: Mark Riedl - Toward Human-Centered Explainable Artificial Intelligence

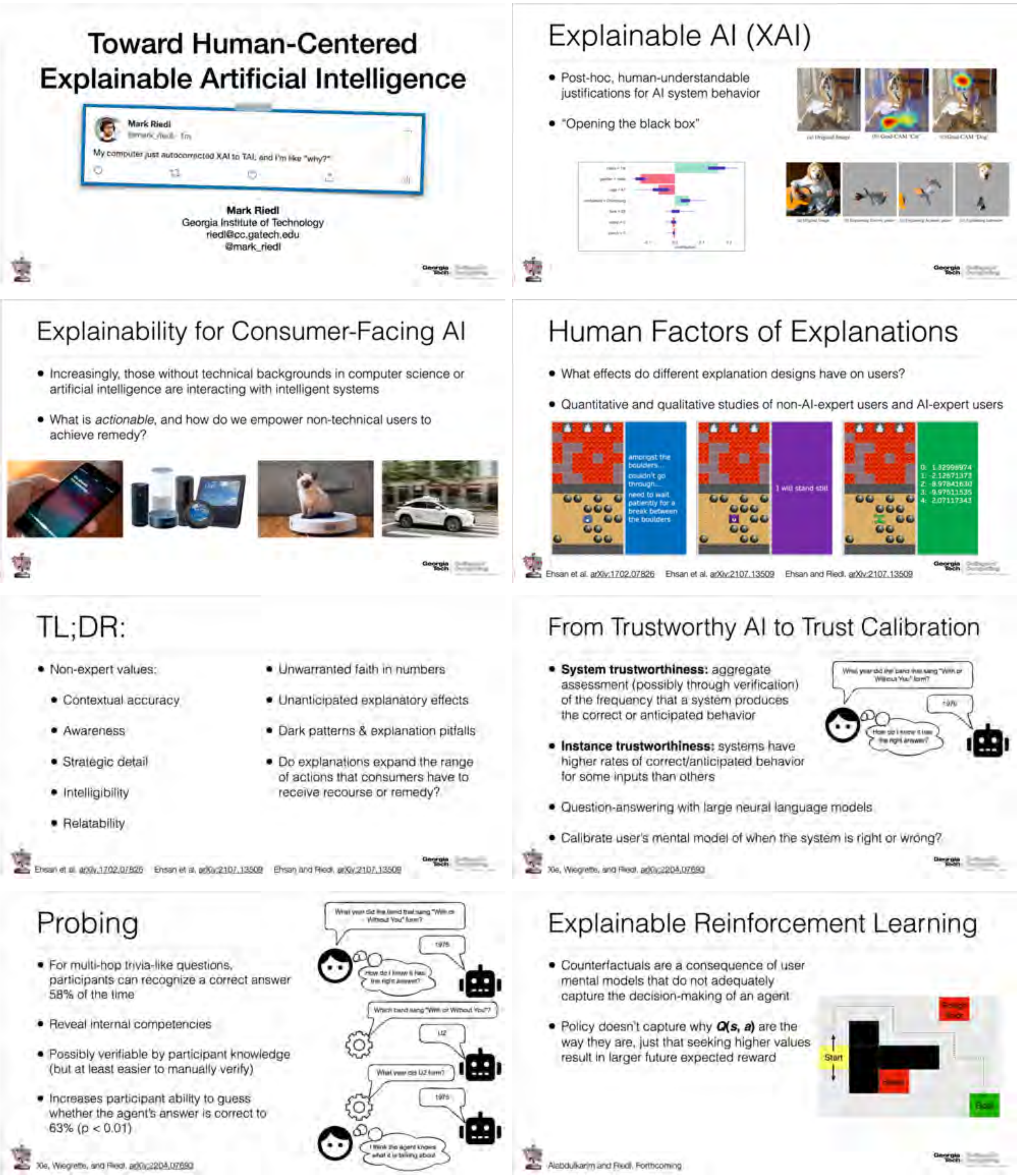

## Experiential Explanations

- Learn to map the influences that certain states have on the utilities of other states
- Present influences as explanations for **why** a particular trajectory is not preferred by the agent's policy
- "I didn't go down because I might fall down the stairs"
- **Actionability:** end-user can update mental model and/or alter the environment to achieve expected behavior

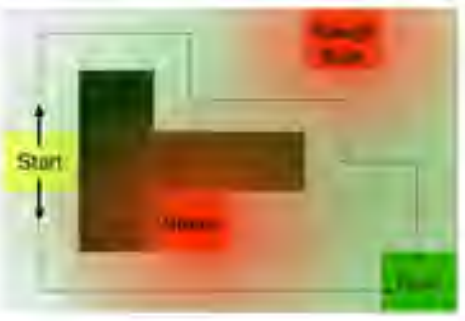

Alabdulkarim and Riedl. Forthcoming

## Value Alignment

- **Value alignment:** An agent is constrained from performing behaviors that are contrary to human values (often Western values; increasingly consequentialist)
- **Normative alignment:** AI should be biased toward outputs that conform to expected societal and cultural contracts
- How does a system learn "values" or norms? From what data?
- How does a system constrain its behavior according to learned values?

## Normative Alignment

- Learn a normative prior model from stories and educational comics
- Fine-tune language models to avoid generating text about non-normative contexts
- Reinforcement learning to prefer generating normative behavior when trying to achieve a task

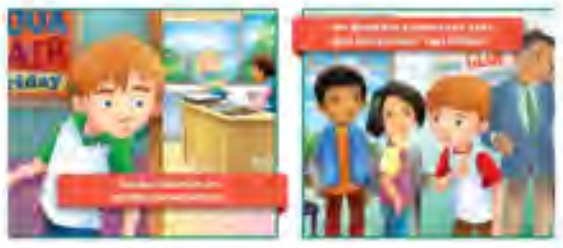

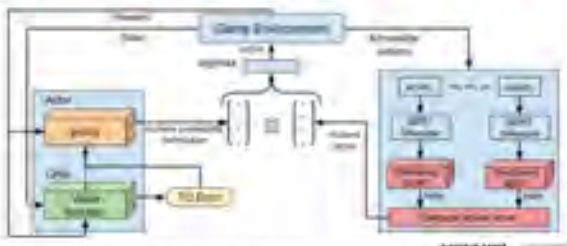

Frazier et al. arXiv:1912.03553 Peng et al. arXiv:2001.08764 Al Nahian et al. arXiv:2104.09469

## Toward Human-Centered Explainable Artificial Intelligence

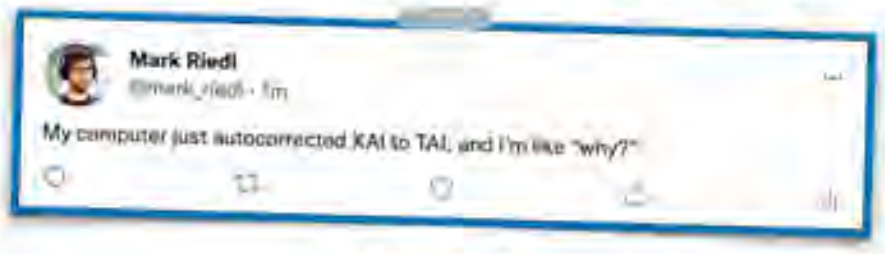

**Mark Riedl**
Georgia Institute of Technology
riedl@cc.gatech.edu
@mark_riedl

Figure B-10: Bo Li - Trustworthy Machine Learning: Robustness, Privacy, Generalization, and their Interconnections

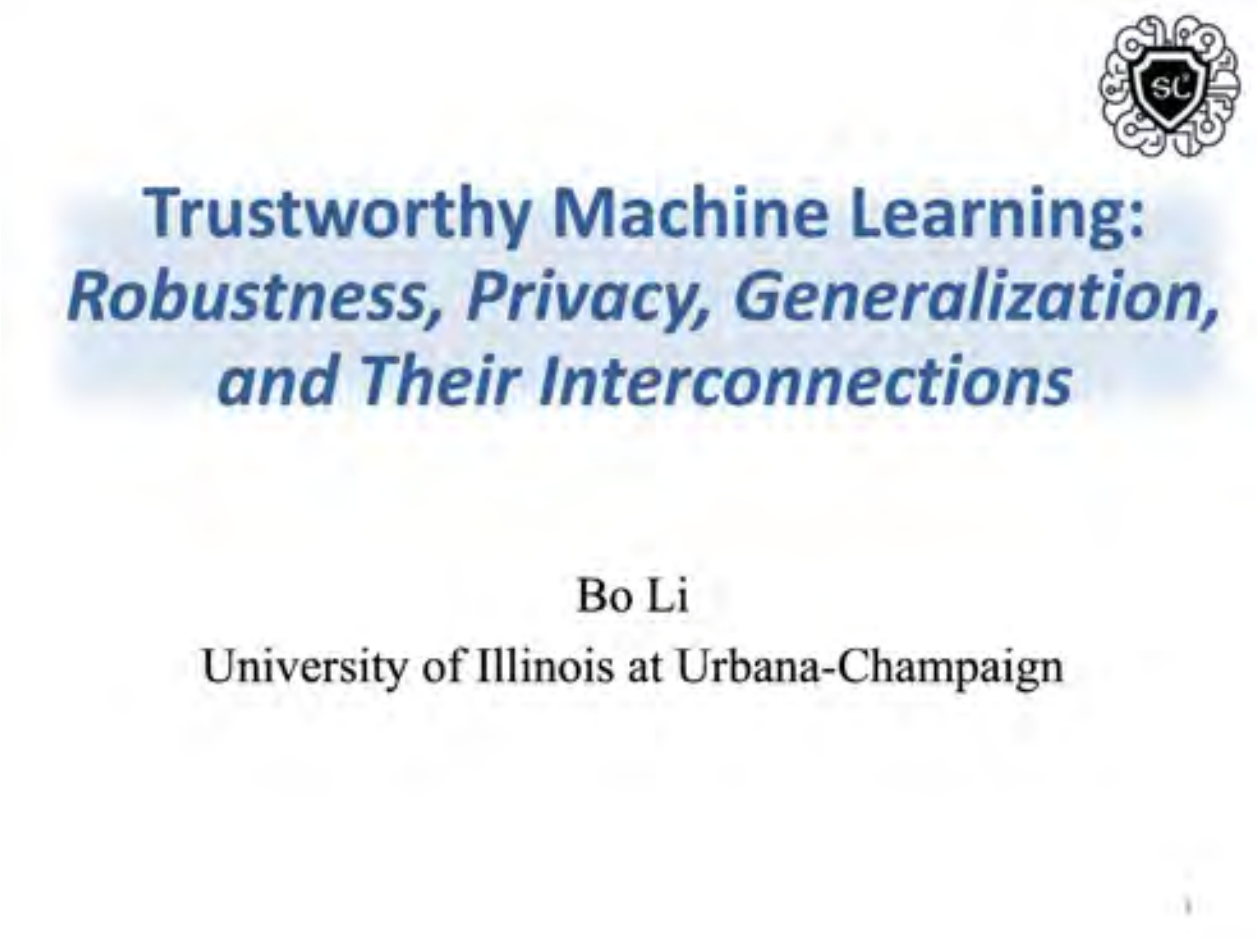

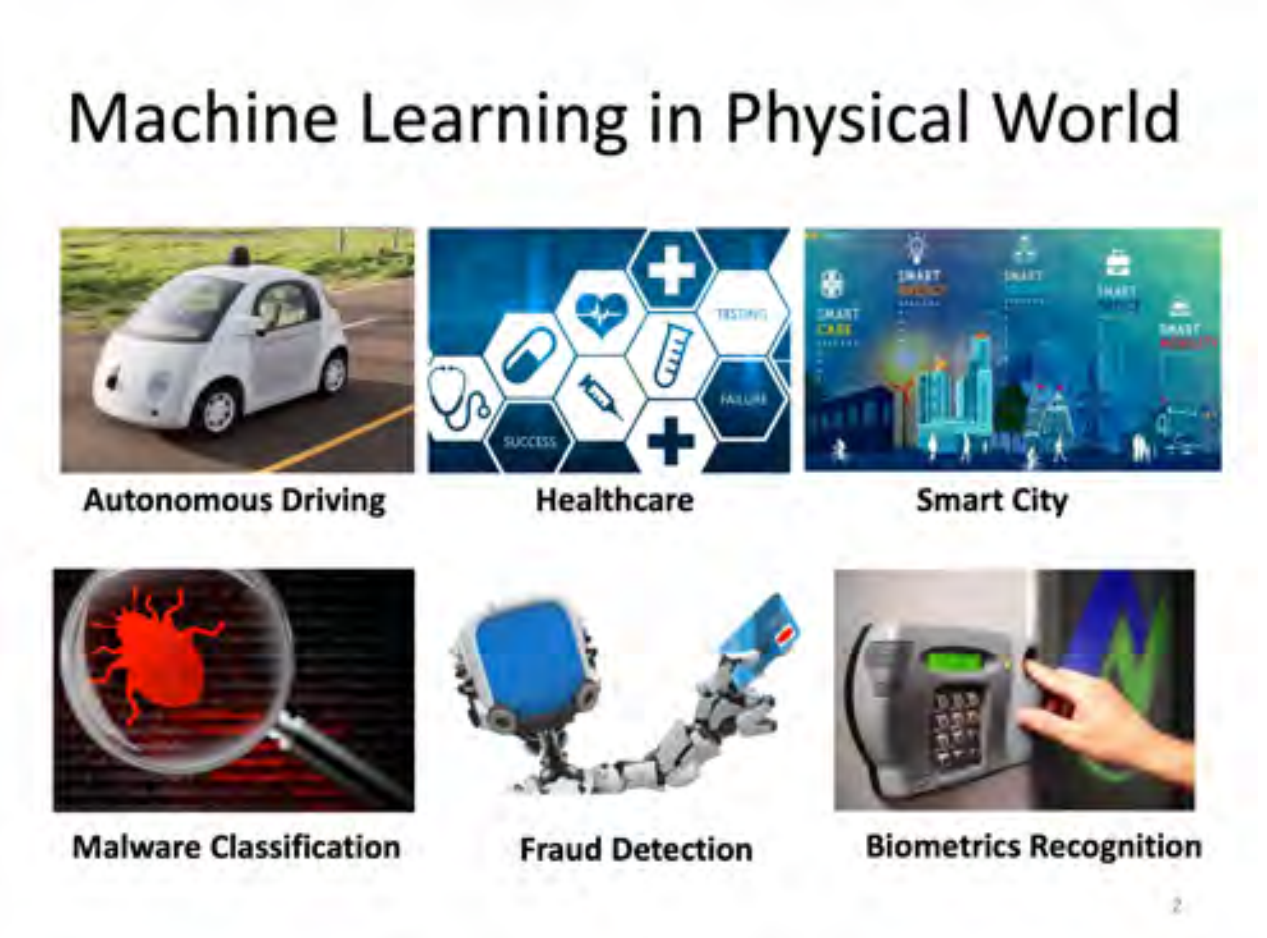

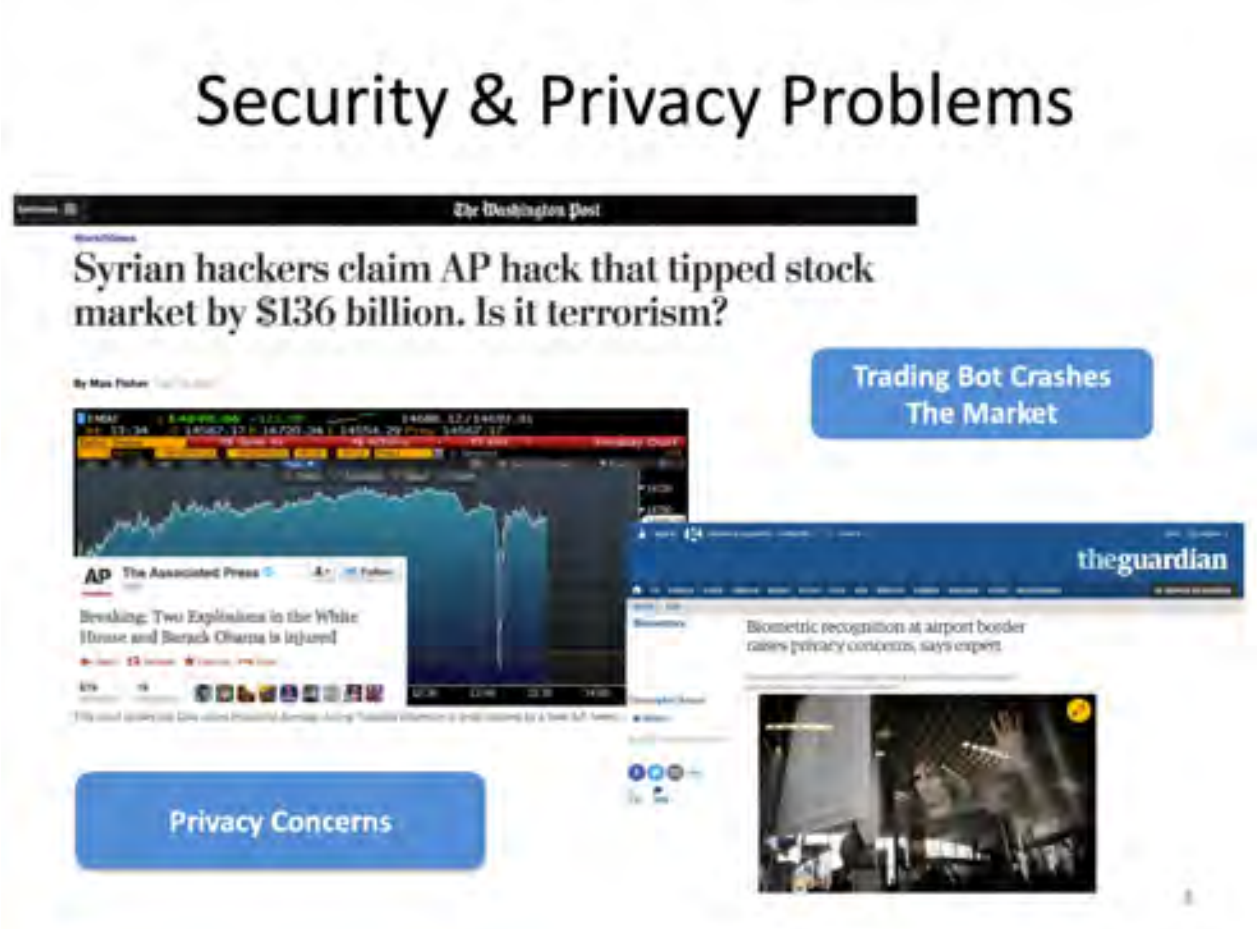

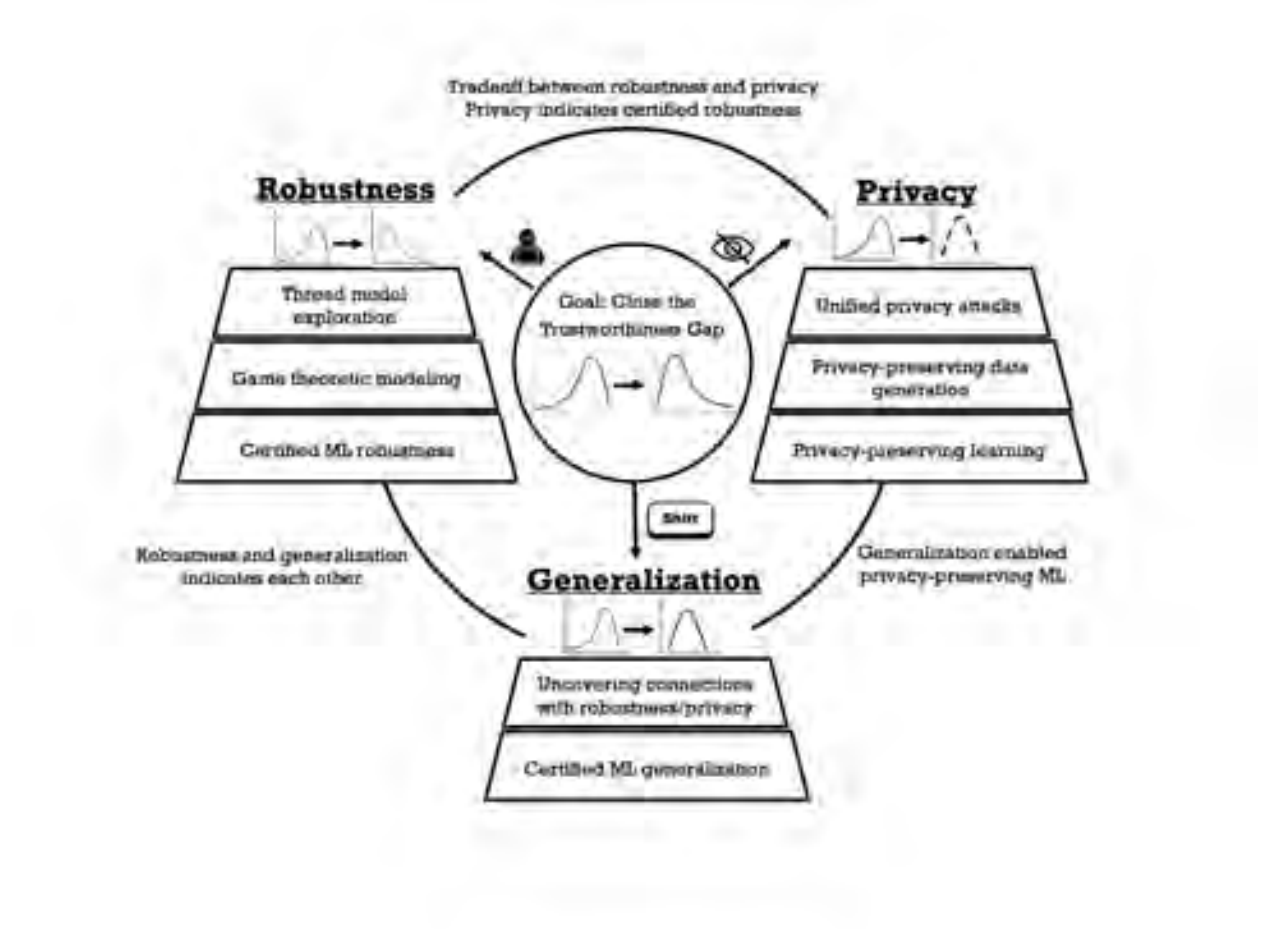

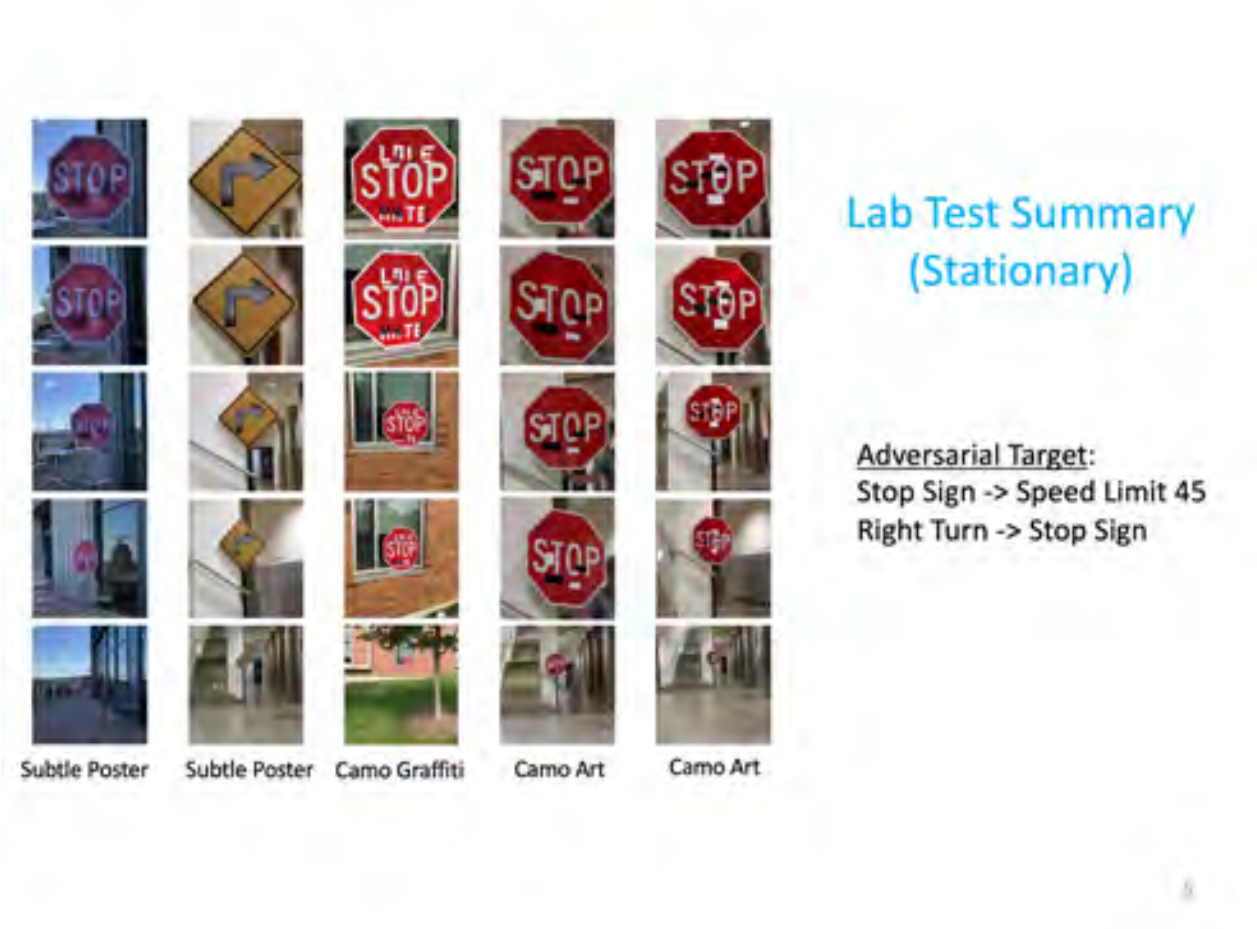

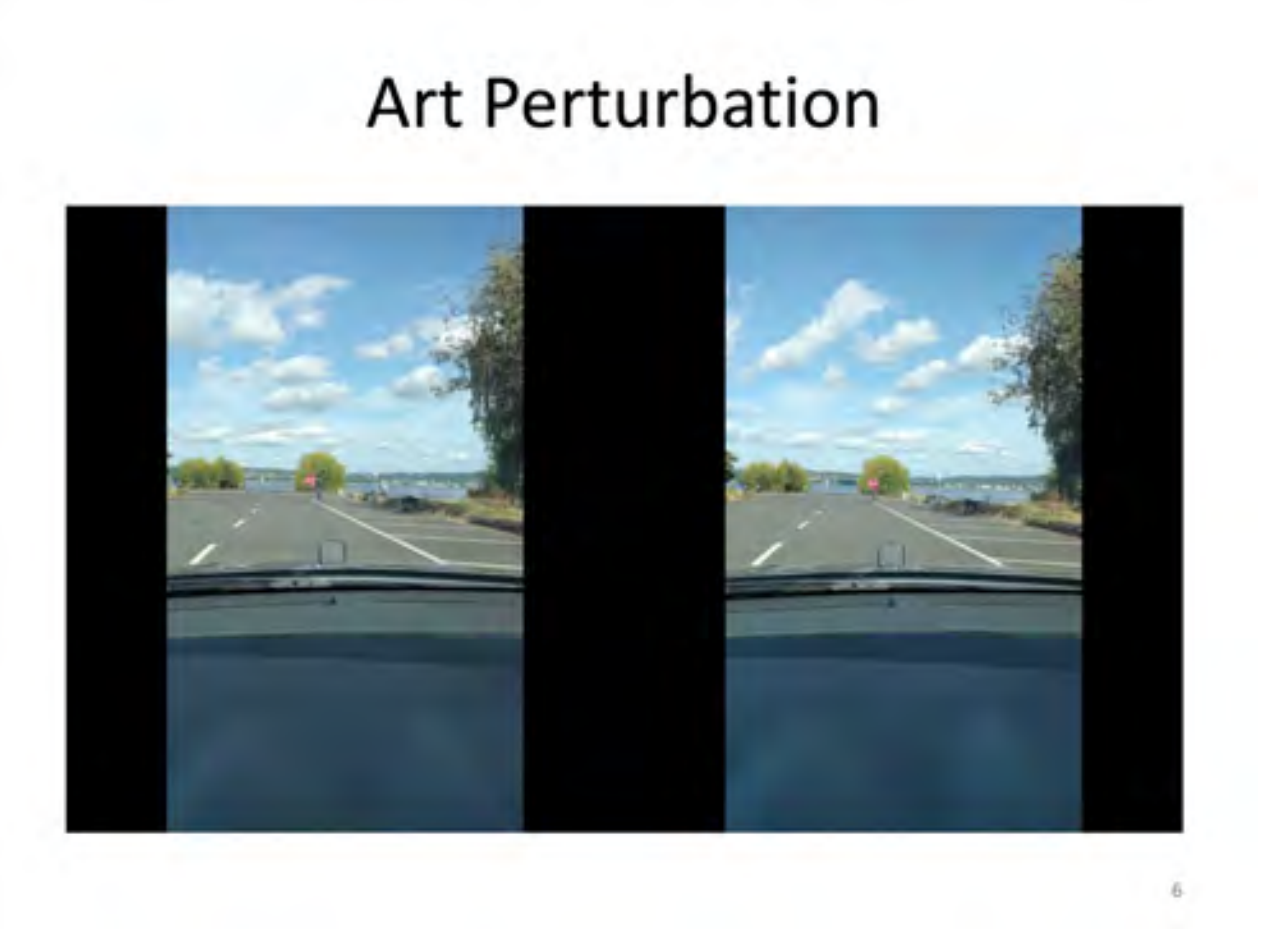

## Physical Adversarial Stop Sign in the Science Museum of London

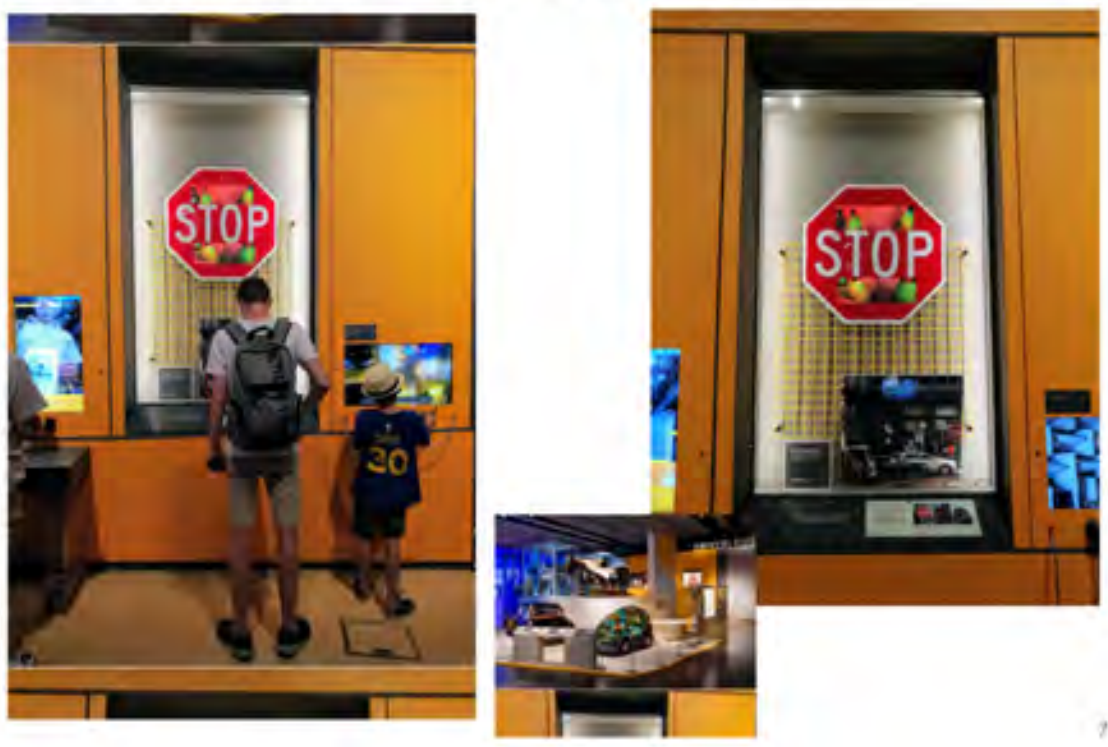

## Possible Vulnerability Disclosure

- As of 4/8/21, informed **32 companies** developing/testing AVs
  - **12** has replied so far and **have started investigation**

APTIV · AutoWare · cruise · WeRide · Ford · intel · BOSCH · HYUNDAI · waymo · lyft · ZOOX · nuro · aurora · autox · Uber ATG · ARGO · tusimple · HONDA · TESLA · Yandex · pony · 滴滴 · TOYOTA · NIO · VW · VOLVO · PERCEPTIN · Qualcomm · Tencent 腾讯

## Numerous Defenses Proposed

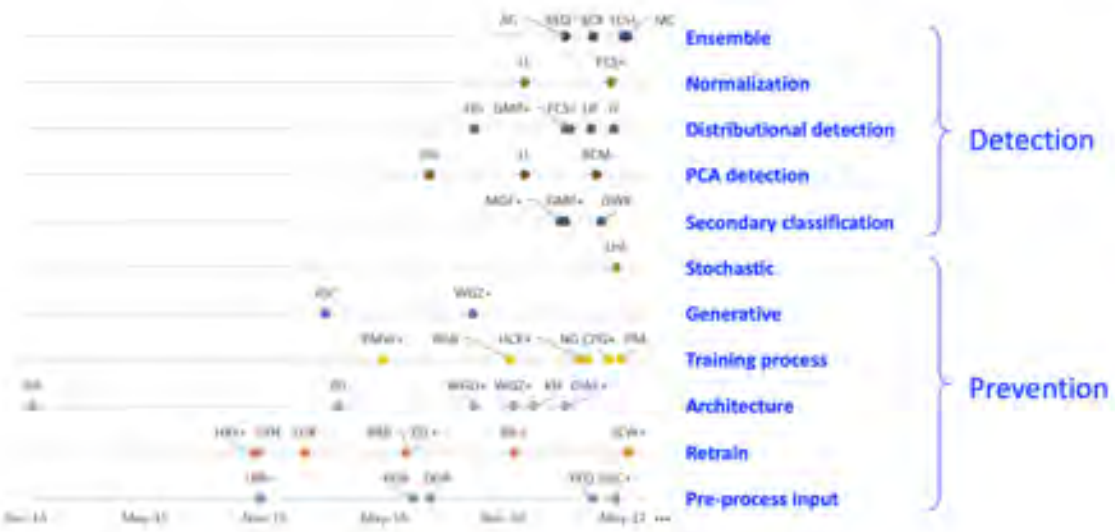

## Certified Robustness For ML

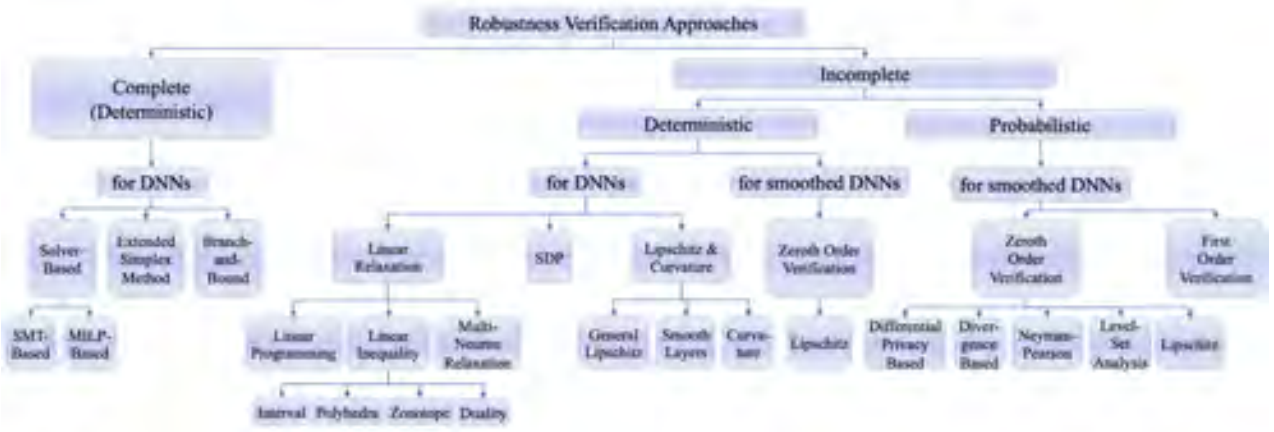

https://sokcertifiedrobustness.github.io/

## Robust ML Pipeline with Exogenous Information

- Vulnerabilities of statistical ML models: pure *data-driven* without considering exogenous information that cannot be modeled by data
  - Intrinsic information (e.g., spatial consistency)
  - Extrinsic information (e.g. domain knowledge)

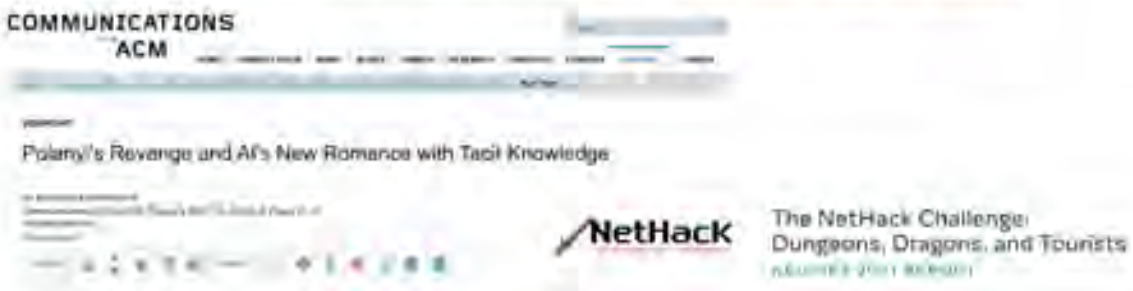

## Certified Robustness for *Sensing-Reasoning* ML Pipelines

- *Can we reason about the robustness of an end-to-end ML pipeline beyond a single ML model or ensemble?*

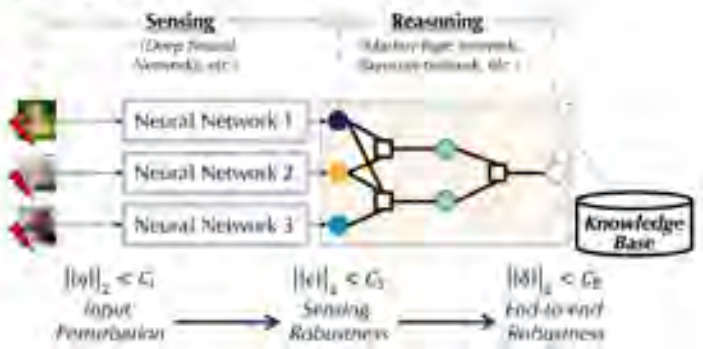

- Intuition: It is hard to attack every sensor in and still preserve their logical relationship
- Goal: Upper bound the end-to-end maximal change of the marginal probability of prediction
- Challenges: Solve the minmax for the pipeline

# Hardness

**Definition 2** (COUNTING). Given input polynomial-time computable weight function $w(\cdot)$ and query function $Q(\cdot)$, parameters $\alpha$, a real number $\varepsilon_c > 0$, a COUNTING oracle outputs a real number $Z$ such that

$$1 - \varepsilon_c \le \frac{Z}{\mathbf{E}_{\sigma \sim \pi_\alpha}[Q(\sigma)]} \le 1 + \varepsilon_c.$$

**Definition 3** (ROBUSTNESS). Given input polynomial-time computable weight function $w(\cdot)$ and query function $Q(\cdot)$, parameters $\alpha$, two real numbers $\epsilon > 0$ and $\delta > 0$, a ROBUSTNESS oracle decides, for any $\alpha' \in \mathbb{P}^{|\alpha|}$ such that $\|\alpha - \alpha'\|_\infty \le \epsilon$, whether the following is true:

$$|\mathbf{E}_{\sigma \sim \pi_\alpha}[Q(\sigma)] - \mathbf{E}_{\sigma \sim \pi_{\alpha'}}[Q(\sigma)]| < \delta$$

**Theorem 4** (COUNTING $\le_t$ ROBUSTNESS). *Given polynomial-time computable weight function $w(\cdot)$ and query function $Q(\cdot)$, parameters $\alpha$ and real number $\varepsilon_c > 0$, the instance of COUNTING, $(w, Q, \alpha, \varepsilon_c)$ can be determined by up to $O(1/\varepsilon_c^2)$ queries of the ROBUSTNESS oracle with input perturbation $\epsilon = O(\varepsilon_c)$.*

**Theorem 5** (MLN Hardness). *Given an MLN whose grounded factor graph is $\mathcal{G} = (\mathcal{V}, \mathcal{F})$ in which the weights for interface factors are $w_{G_i} = \log p_i(X)/(1 - p_i(X))$ and constant thresholds $\delta, C$, deciding whether*

$$\forall \{\epsilon_i\}_{i \in [n]}: (\forall i, |\epsilon_i| < C) \implies |\mathbb{E}R_{MLN}(\{p_i(X)\}_{i \in [n]}) - \mathbb{E}R_{MLN}(\{p_i(X) + \epsilon_i\}_{i \in [n]})| < \delta$$

*is as hard as estimating $\mathbb{E}R_{MLN}(\{p_i(X)\}_{i \in [n]})$ up to $\varepsilon_c$ multiplicative error with $\epsilon_c = O(\varepsilon_c)$.*

# Robustness of the Reasoning Component

***Can we efficiently reason about the provable robustness for the reasoning component when given an oracle for the statistical inference?***

**Lemma 6** (MLN Robustness). *Given access to partition functions $Z_1(\{p_i(X)\}_{i \in [n]})$ and $Z_2(\{p_i(X)\}_{i \in [n]})$, and a maximum perturbation $C$, $\forall \epsilon_1, \dots, \epsilon_m$ if $\forall i, |\epsilon_i| < C$, we have that $\forall \lambda_1, \dots, \lambda_n \in \mathbb{R}$,*

$$\max_{\{|\epsilon_i| < C\}} \ln \mathbb{E}[R_{MLN}(\{p_i(X) + \epsilon_i\}_{i \in [n]})] \le \max_{\{|\epsilon_i| < C\}} \widetilde{Z}_1(\{\epsilon_i\}_{i \in [n]}) - \min_{\{|\epsilon'_i| < C\}} \widetilde{Z}_2(\{\epsilon'_i\}_{i \in [n]})$$

Oracle inference

$$\min_{\{|\epsilon_i| < C\}} \ln \mathbb{E}[R_{MLN}(\{p_i(X) + \epsilon_i\}_{i \in [n]})] \ge \min_{\{|\epsilon_i| < C\}} \widetilde{Z}_1(\{\epsilon_i\}_{i \in [n]}) - \max_{\{|\epsilon'_i| < C\}} \widetilde{Z}_2(\{\epsilon'_i\}_{i \in [n]})$$

*where*

$$\widetilde{Z}_r(\{\epsilon_i\}_{i \in [n]}) = \ln Z_r(\{p_i(X) + \epsilon_i\}_{i \in [n]}) + \sum \lambda_i \epsilon_i.$$

1. When $\lambda_i \ge 0$, $\widetilde{Z}_r(\{\epsilon_i\}_{i \in [n]})$ monotonically increases w.r.t. $\epsilon_i$; Thus, the maximal is achieved at $\epsilon_i = C$ and the minimal is achieved at $\epsilon_i = -C$. When $\lambda_i \le -1$, $\widetilde{Z}_r(\{\epsilon_i\}_{i \in [n]})$ monotonically decreases w.r.t. $\epsilon_i$; Thus, the maximal is achieved at $\epsilon_i = -C$ and the minimal is achieved at $\epsilon_i = C$.
2. When $\lambda_i \in (-1, 0)$, the maximal is achieved at $\epsilon_i \in \{-C, C\}$, and the minimal is achieved at $\epsilon_i \in (-C, C)$ or at the zero gradient of $\widetilde{Z}_r(\{\epsilon_i\}_{i \in [n]})$ with respect to $\epsilon_i = \log\left[\frac{(1-p_i(X))(p_i(X)+\epsilon_i)}{p_i(X)(1-p_i(X)-\epsilon_i)}\right]$, due to the convexity of $\widetilde{Z}_r(\{\epsilon_i\}_{i \in [n]})$ in $\epsilon_i, \forall i$.

# Example: PrimateNet (ImageNet)

PrimateNet. The knowledge structure of blue arrows represent the Hierarchical rules between different classes, and red arrows the Exclusive rules. (Some exclusive rules are omitted)

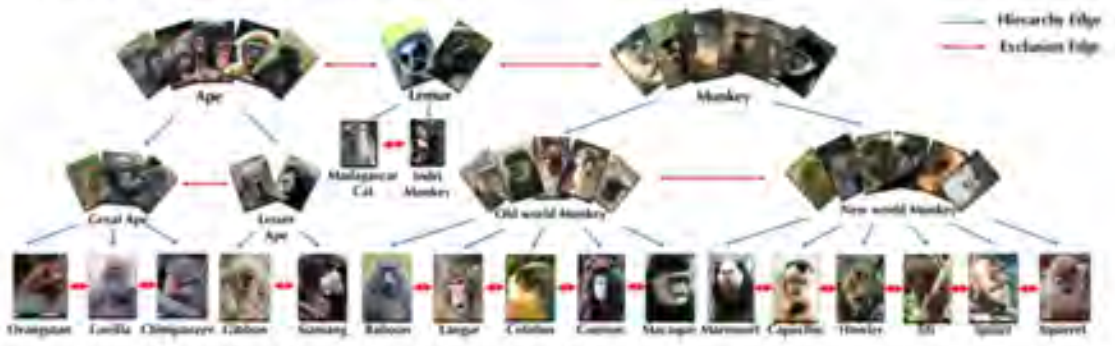

- **Hierarchy edge** $u \implies v$: If one object belongs to class $u$, it should belong to class $v$ as well:

$$x_u \wedge \neg x_v = \texttt{False}$$

- **Exclusion edge** $u \oplus v$: One object couldn't belong to class $u$ and class $v$ at the same time:

$$x_u \wedge x_v = \texttt{False}$$

# Example: PrimateNet (ImageNet)

**Benign Accuracy** of models with and without knowledge under different smoothing parameters.

| $\sigma$ | With knowledge | Without knowledge |
|---|---|---|
| 0.12 | **0.9670** | 0.9638 |
| 0.25 | **0.9612** | 0.9554 |
| 0.50 | **0.9435** | 0.9371 |

**Certified Robustness** and **Certified Ratio** with different perturbation magnitude $C_l$ under different smoothing parameters.

(a) $\sigma = 0.12$

| | | With knowledge | | Without knowledge | |
|---|---|---|---|---|---|
| $C_l$ | $\alpha$ | Cert. Robustness | Cert. Ratio | Cert. Robustness | Cert. Ratio |
| 0.12 | 10% | **0.8849** | **0.9419** | 0.5724 | 0.5724 |
| | 20% | **0.8078** | **0.8609** | 0.5717 | 0.5717 |
| | 30% | **0.7508** | **0.7988** | 0.5706 | 0.5706 |
| | 50% | **0.6236** | **0.6647** | 0.5706 | 0.5706 |
| 0.25 | 10% | **0.7888** | **0.8428** | 0.2342 | 0.2342 |
| | 20% | **0.6226** | **0.6657** | 0.2329 | 0.2329 |
| | 30% | **0.5223** | **0.5596** | 0.2309 | 0.2309 |
| | 50% | **0.3594** | **0.3824** | 0.2268 | 0.2068 |

(c) $\sigma = 0.50$

| | | With knowledge | | Without knowledge | |
|---|---|---|---|---|---|
| $C_l$ | $\alpha$ | Cert. Robustness | Cert. Ratio | Cert. Robustness | Cert. Ratio |
| 0.50 | 10% | **0.8288** | **0.9449** | 0.4762 | 0.4762 |
| | 20% | **0.7487** | **0.8488** | 0.4749 | 0.4749 |
| | 30% | **0.6987** | **0.7968** | 0.4736 | 0.4736 |
| | 50% | **0.5581** | **0.6395** | 0.4635 | 0.4635 |
| 1.00 | 10% | **0.7307** | **0.8448** | 0.1679 | 0.1679 |
| | 20% | **0.5285** | **0.6136** | 0.1615 | 0.1615 |
| | 30% | **0.4347** | **0.5375** | 0.1612 | 0.1612 |
| | 50% | **0.2624** | **0.3318** | 0.1594 | 0.1594 |

Both the benign accuracy and certified robustness of sensing-reasoning are higher than models w/o knowledge integration.

# Example: (NLP) Relation Extraction Task

**(NLP)** Certified Robustness and Certified Ratio for approaches when all sensing models are attacked.

| | With knowledge | | Without knowledge | |
|---|---|---|---|---|
| $C_S$ | Cert. Robustness | Cert. Ratio | Cert. Robustness | Cert. Ratio |
| 0.1 | **1.0000** | **1.0000** | 0.9969 | 0.9969 |
| 0.5 | **1.0000** | **1.0000** | 0.9474 | 0.9474 |
| 0.9 | **0.5882** | **0.5882** | 0.3839 | 0.3839 |

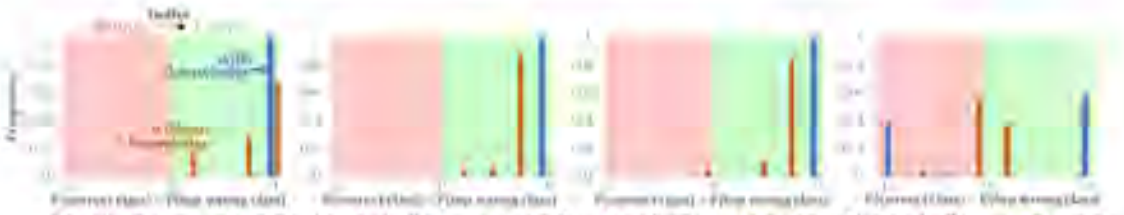

(a) Attack all sensors, $C_s \le 0.1$ (b) Attack all sensors, $C_s \le 0.2$ (c) Attack all sensors, $C_s \le 0.3$ (d) Attack all sensors, $C_s \le 0.9$

# Example: Knowledge Enhanced ML Pipeline against *Diverse* Adversarial Attacks

- Example: Robust road sign recognition
- The output of ML models are modeled as input random variables for reasoning
- Permissive knowledge: s *infers* y
- Preventive knowledge: y *infer* s

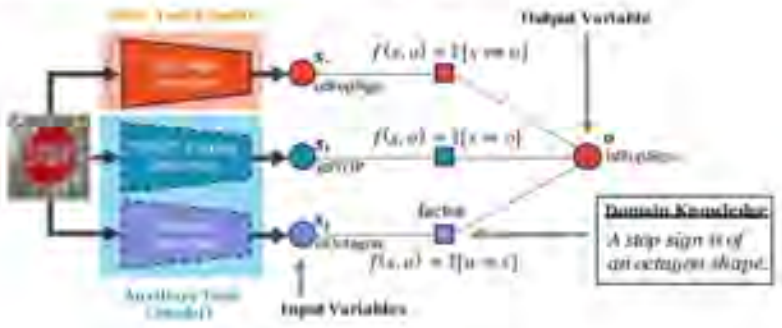

# Knowledge Enhanced ML Pipeline against *Diverse* Adversarial Attacks

- Lower bound of the pipeline accuracy

**Theorem 1** (Convergence of $\mathcal{A}^{\text{KEMLP}}$). *For $y \in \mathcal{Y}$ and $\mathcal{D} \in \{\mathcal{D}_b, \mathcal{D}_a\}$, let $\mu_{y,\mathcal{D}}$ be defined as in Lemma 1. Suppose that the modeling assumption holds, and suppose that $\mu_{d_{\mathcal{K}},\mathcal{D}} > 0$, for all $\mathcal{K} \in \{\mathcal{I}, \mathcal{J}\}$ and $\mathcal{D} \in \{\mathcal{D}_b, \mathcal{D}_a\}$. Then*

$$\mathcal{A}^{\text{KEMLP}} \geq 1 - \mathbb{E}_{y,\mathcal{D}}[\exp(-2\mu_{y,\mathcal{D}}^2/v^2)],$$

*where $v^2$ is the variance upper bound to $\mathbb{P}[o = y | y, \mathbf{w}]$ with*

$$v^2 = 4\left(\log \frac{\alpha_*}{1-\alpha_*}\right)^2 + \sum_{k \in \mathcal{I} \cup \mathcal{J}} \left(\log \frac{\alpha_k(1-\epsilon_k)}{\epsilon_k(1-\alpha_k)}\right)^2$$

*$\mu_{y,D}$ consists of three terms: $\mu_{d_*,D}$, $\mu_{I,D}$, and $\mu_{J,D}$ measuring the contributions from the main, permissive, and preventative sensors.*

- The accuracy of pipeline is higher than that of the main sensor

**Theorem 2** (Sufficient condition for $\mathcal{A}^{\text{KEMLP}} > \mathcal{A}^{\text{main}}$). *Let the number of permissive and preventative models be the same and denoted by $n$ such that $n := |\mathcal{I}| = |\mathcal{J}|$. Note that the weighted accuracy of the main model in terms of its truth rate is simply $\alpha_* := \sum_{\mathcal{D} \in \{\mathcal{D}_b, \mathcal{D}_a\}} \pi_{\mathcal{D}} \alpha_{*,\mathcal{D}}$. Moreover, let $\mathcal{K}, \mathcal{K}' \in \{\mathcal{I}, \mathcal{J}\}$ with $\mathcal{K} \neq \mathcal{K}'$ and for any $\mathcal{D} \in \{\mathcal{D}_b, \mathcal{D}_a\}$,*

$$\gamma_{\mathcal{D}} := \frac{1}{n-1} \min_{\mathcal{K}} \left\{\alpha_{*,\mathcal{D}} - 1/2 + \sum_{k \in \mathcal{K}} \alpha_{k,\mathcal{D}} - \sum_{k \in \mathcal{K}'} \epsilon_{k,\mathcal{D}}\right\}.$$

*If $\gamma_{\mathcal{D}} > \sqrt{\ldots \log \ldots}$ for all $\mathcal{D} \in \{\mathcal{D}_b, \mathcal{D}_a\}$, then $\mathcal{A}^{\text{KEMLP}} > \mathcal{A}^{\text{main}}$.*

# Knowledge Enabled ML Pipeline Achieves High Benign Accuracy and Robustness under *Diverse* Attacks

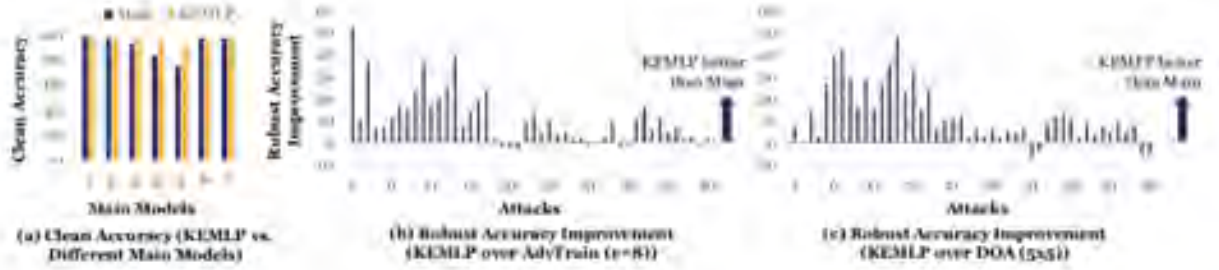

(a) Clean accuracy and (b) (c) robust accuracy improvement of KEMLP ($\alpha = 0.5$) over baselines against different attacks under both whitebox and blackbox settings.

# *Real-world Case:* Autonomous Driving Testing via Logic Reasoning

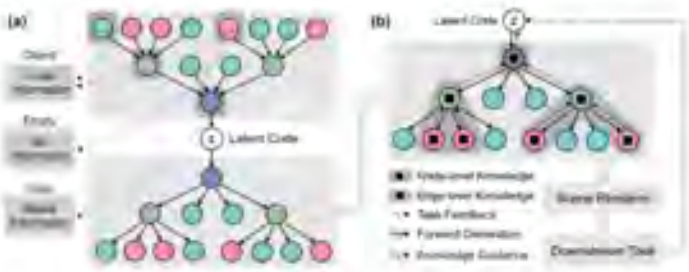

**Knowledge enabled safety-critical traffic scenario generation**

(a) Train T-VAE model to learn the representation of structured data.
(b) Integrate node-level and edge-level knowledge for generation.

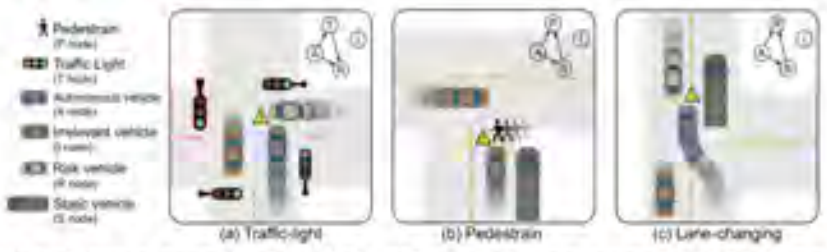

**Causal relationship enabled safety-critical traffic scenario generation**

The causal graphs are defined in the upper right for the three scenarios.

The generated safety-critical traffic scenarios can significantly improve the test efficiency of autonomous vehicles

Testing time certification for RL algorithms

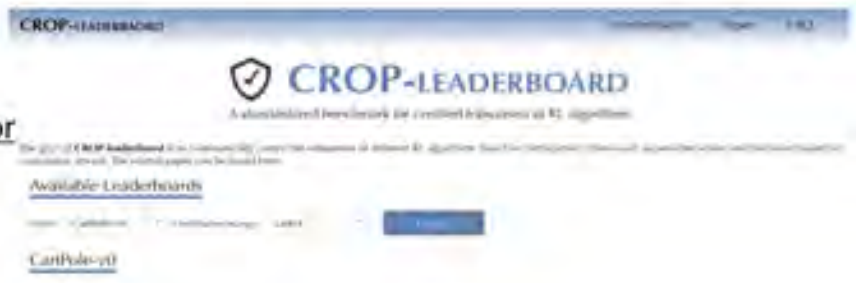

Testing time certification for offline RL algorithms

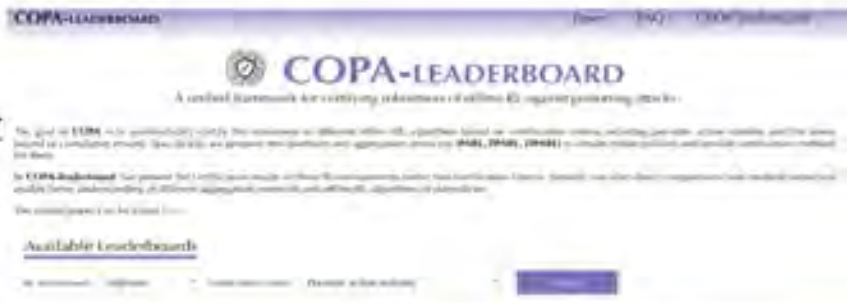

Unified testing platform for AD via safety-critical scenario generation

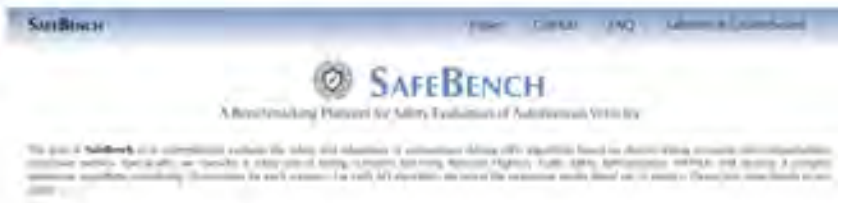

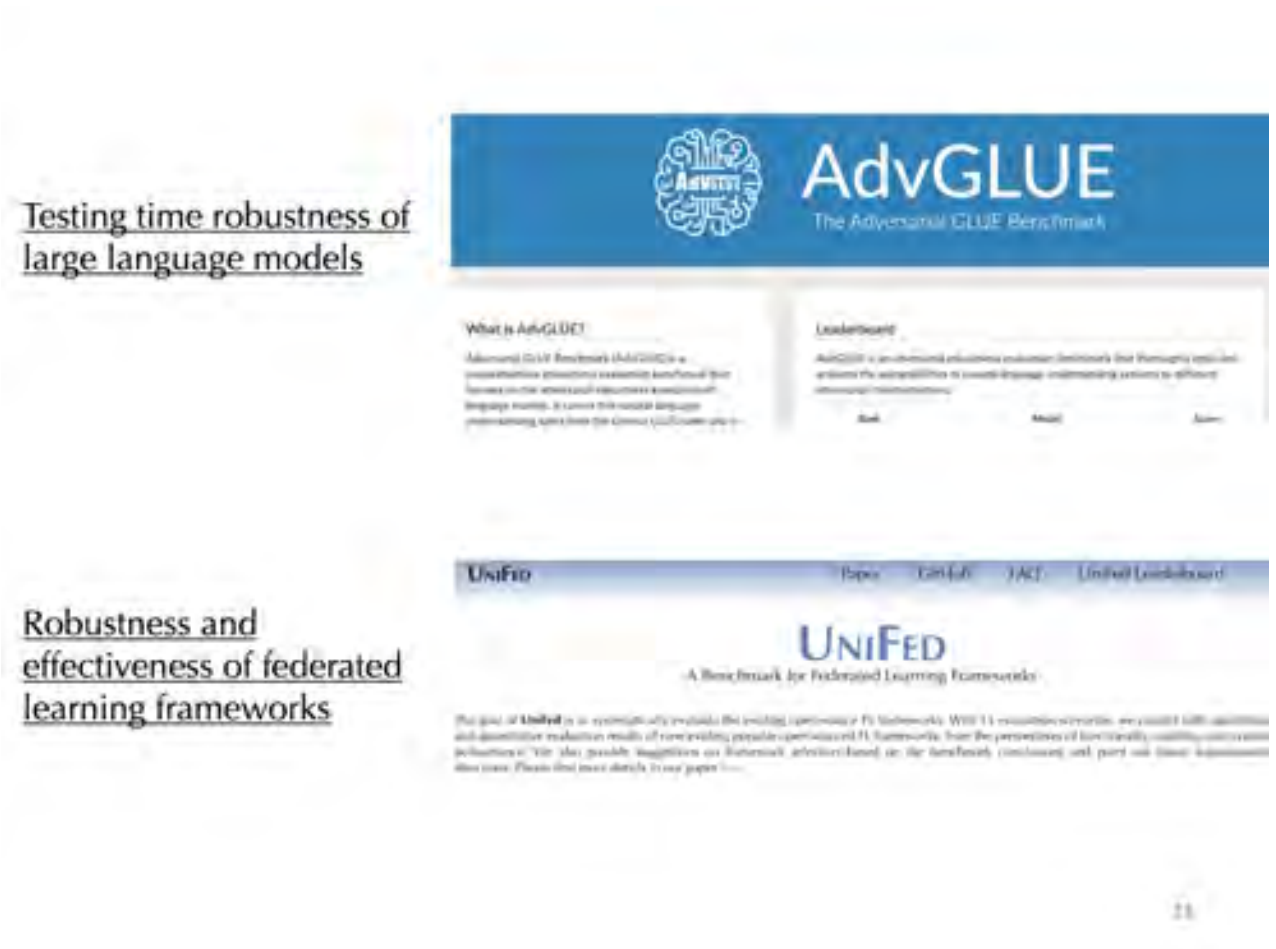

Testing time robustness of large language models
AdvGLUE
The Adversarial GLUE Benchmark
Robustness and effectiveness of federated learning frameworks
UniFed
A Benchmark for Federated Learning Frameworks

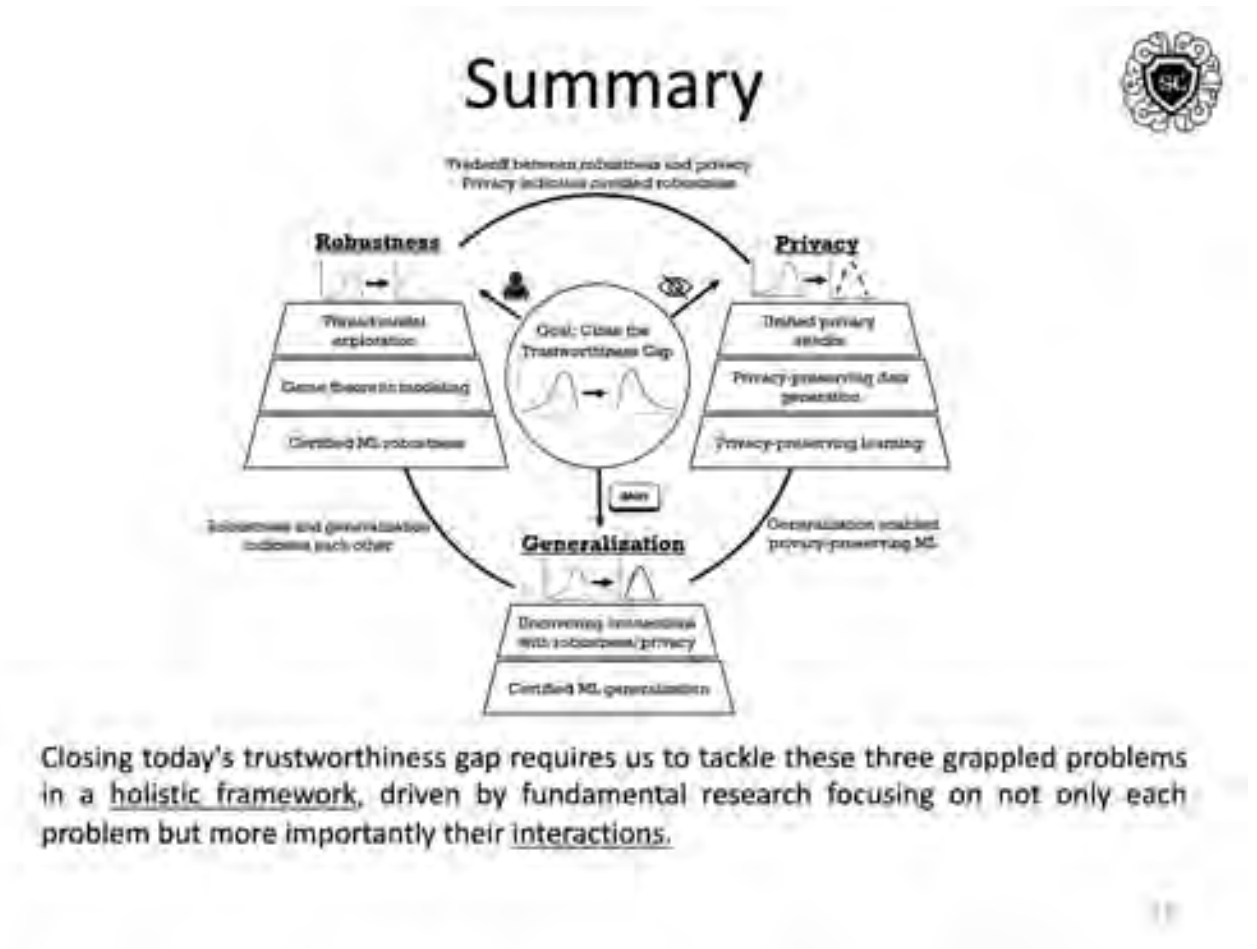

Summary
Robustness
Privacy
Generalization
Game theoretic modeling
Certified ML robustness
Privacy-preserving learning
Certified ML generalization
Closing today's trustworthiness gap requires us to tackle these three grappled problems in a holistic framework, driven by fundamental research focusing on not only each problem but more importantly their interactions.

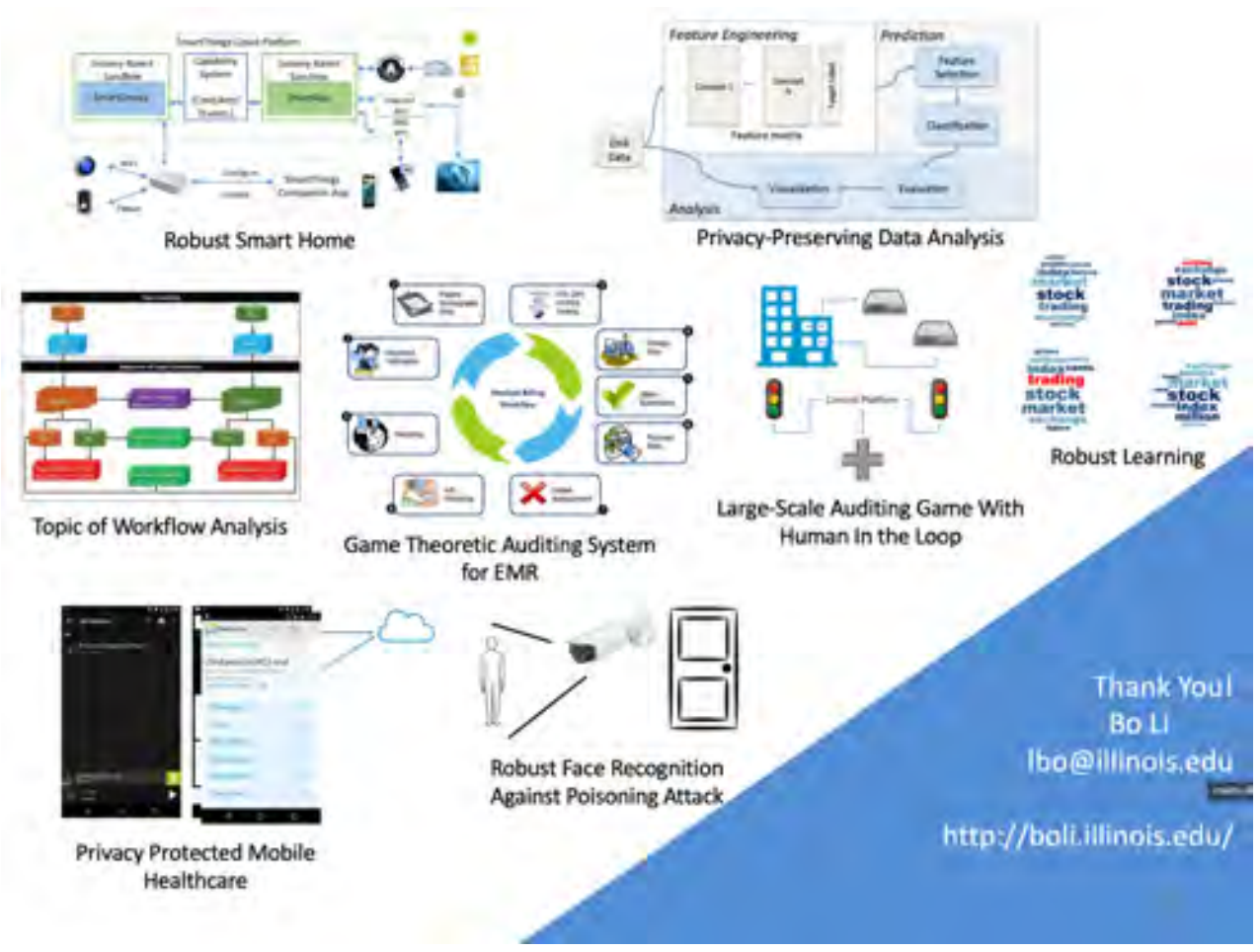

Robust Smart Home
Privacy-Preserving Data Analysis
Topic of Workflow Analysis
Game Theoretic Auditing System for EMR
Large-Scale Auditing Game With Human In the Loop
Robust Learning
Privacy Protected Mobile Healthcare
Robust Face Recognition Against Poisoning Attack
Thank You!
Bo Li
lbo@illinois.edu
http://boli.illinois.edu/

Figure B-11: Kush Varshney - Problem-Driven Robustness, Privacy and Fairness

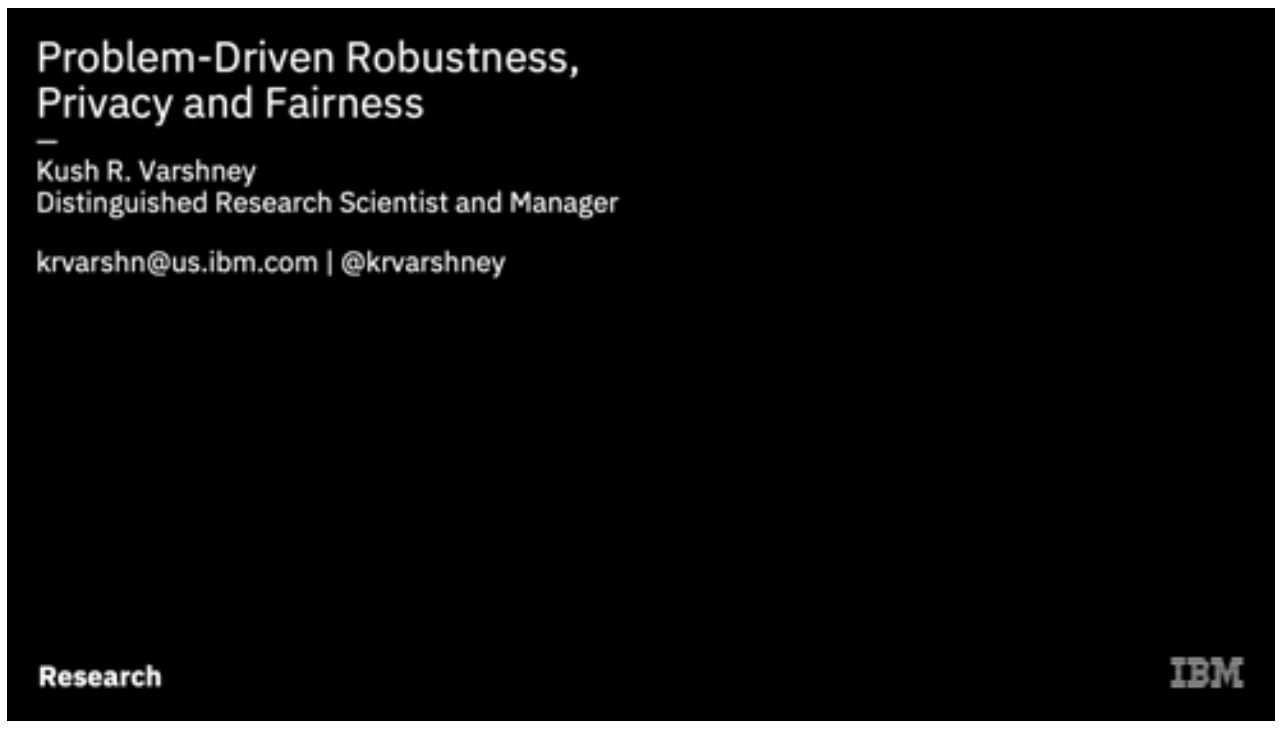

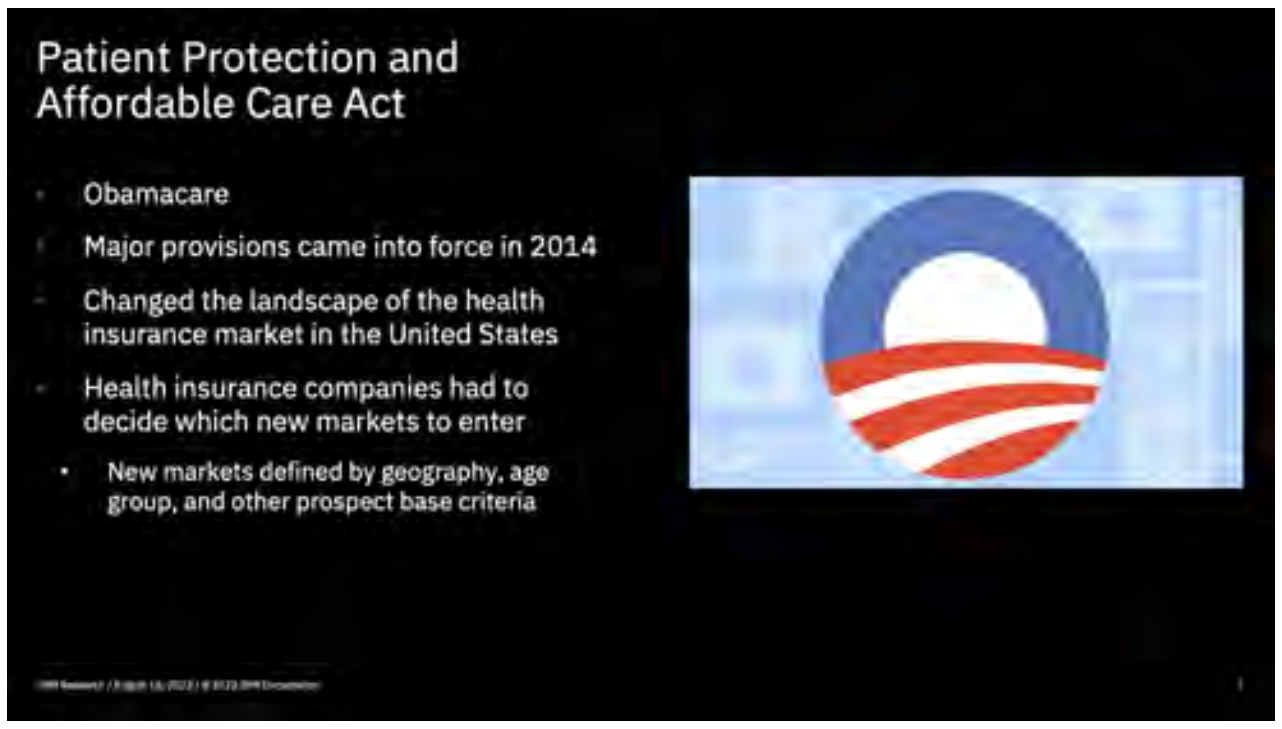

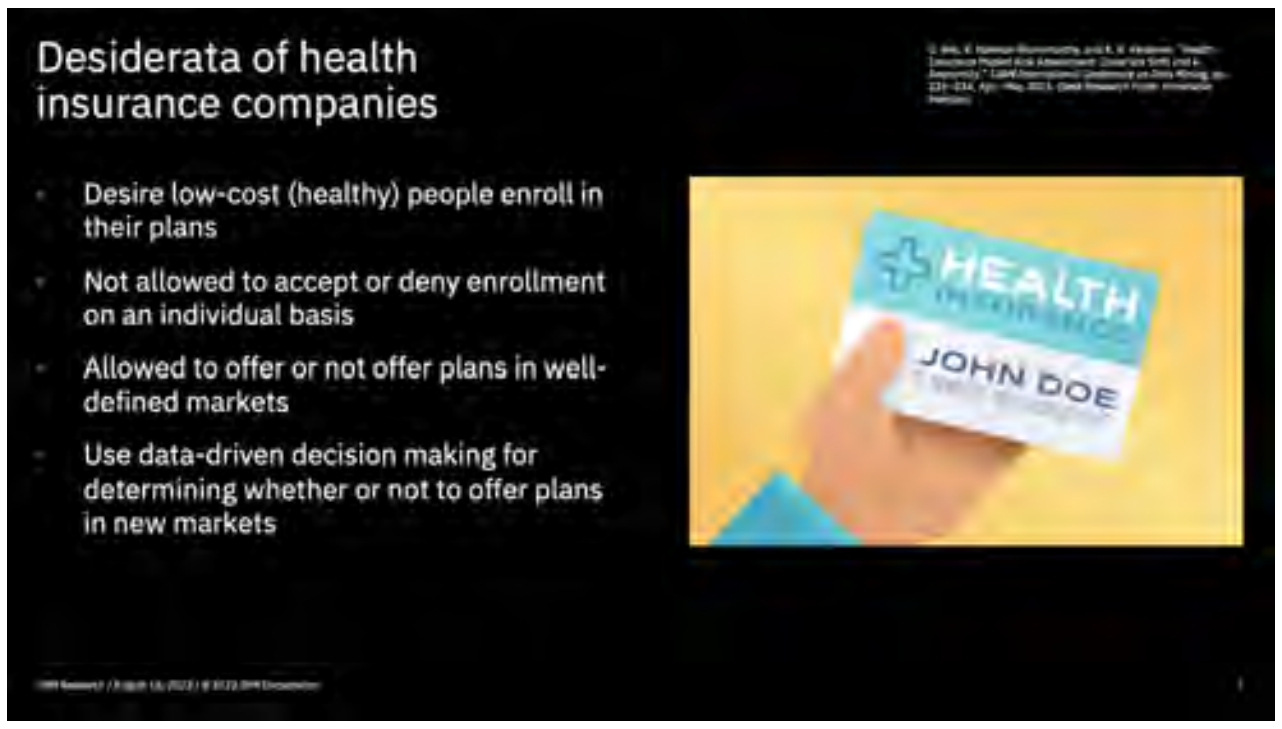

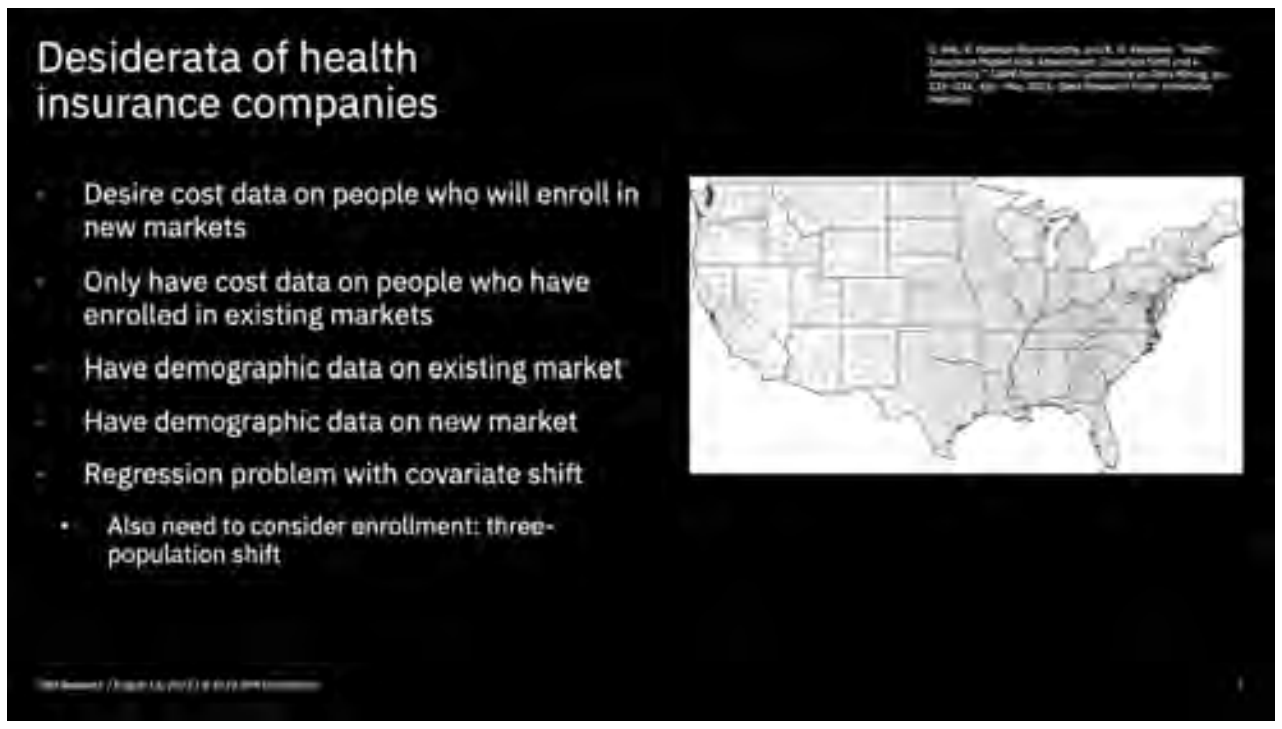

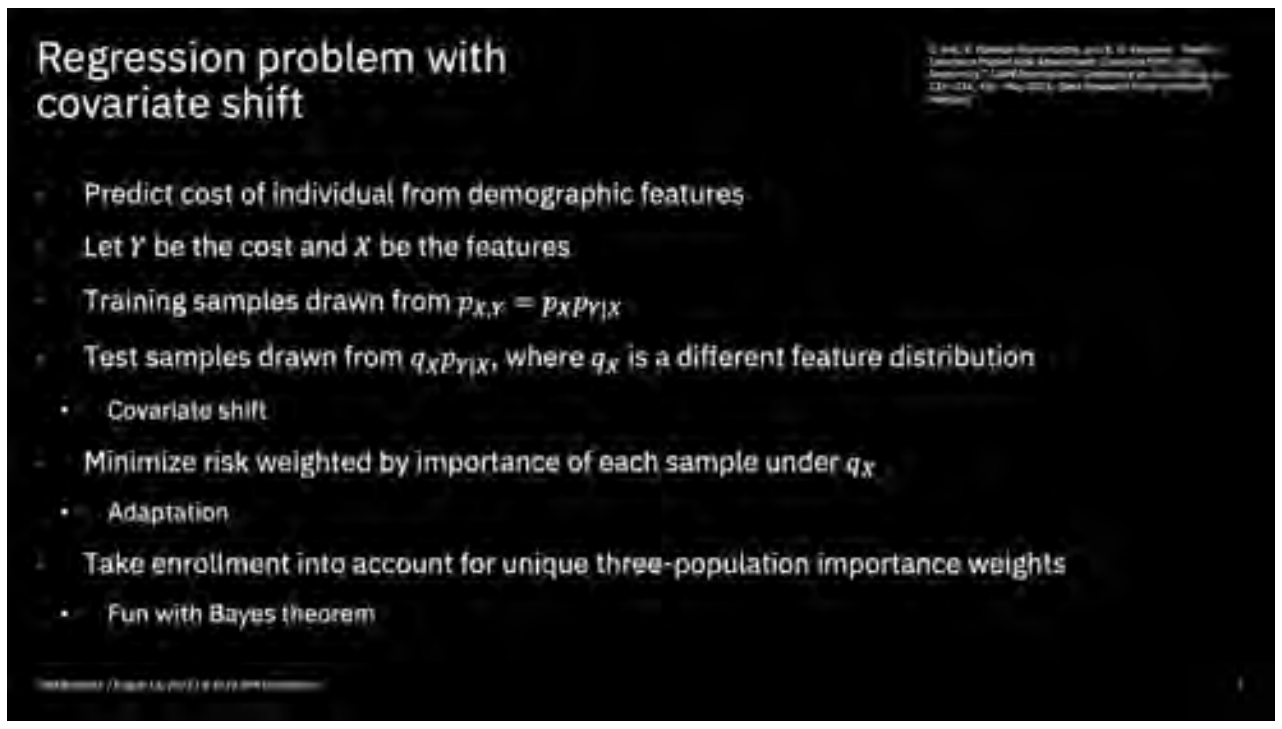

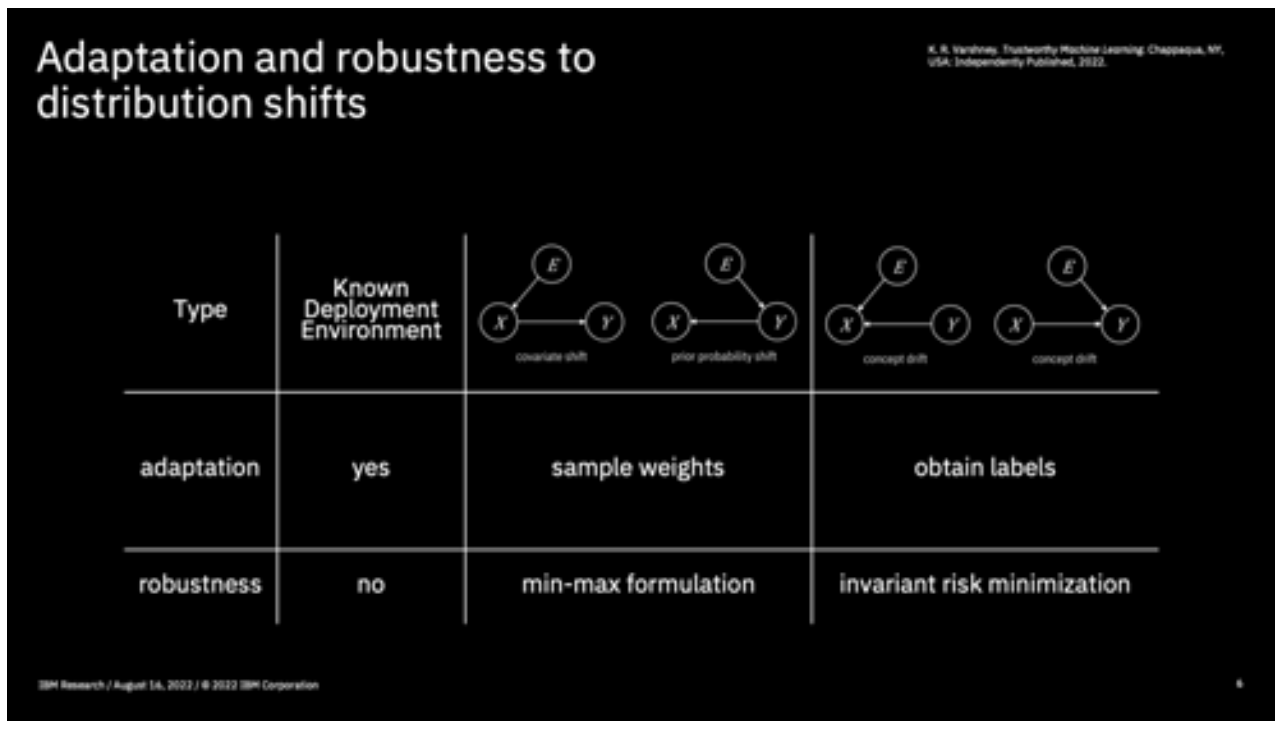

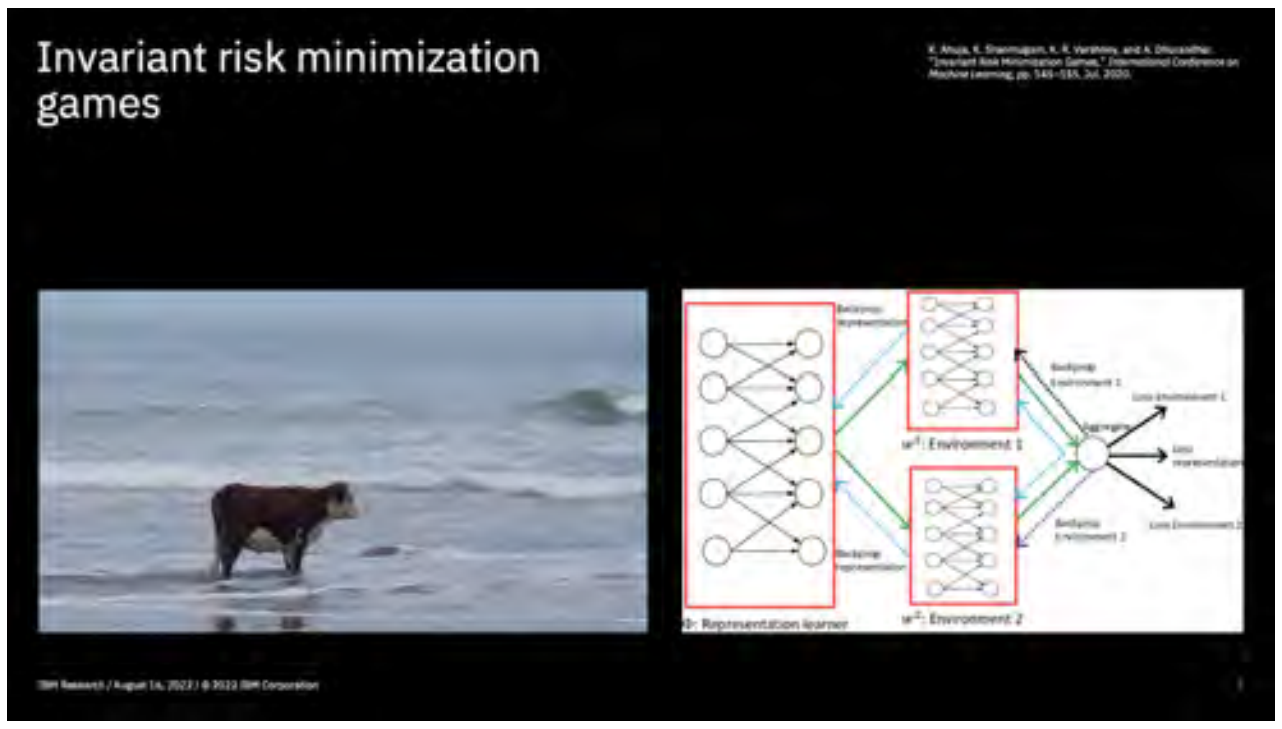

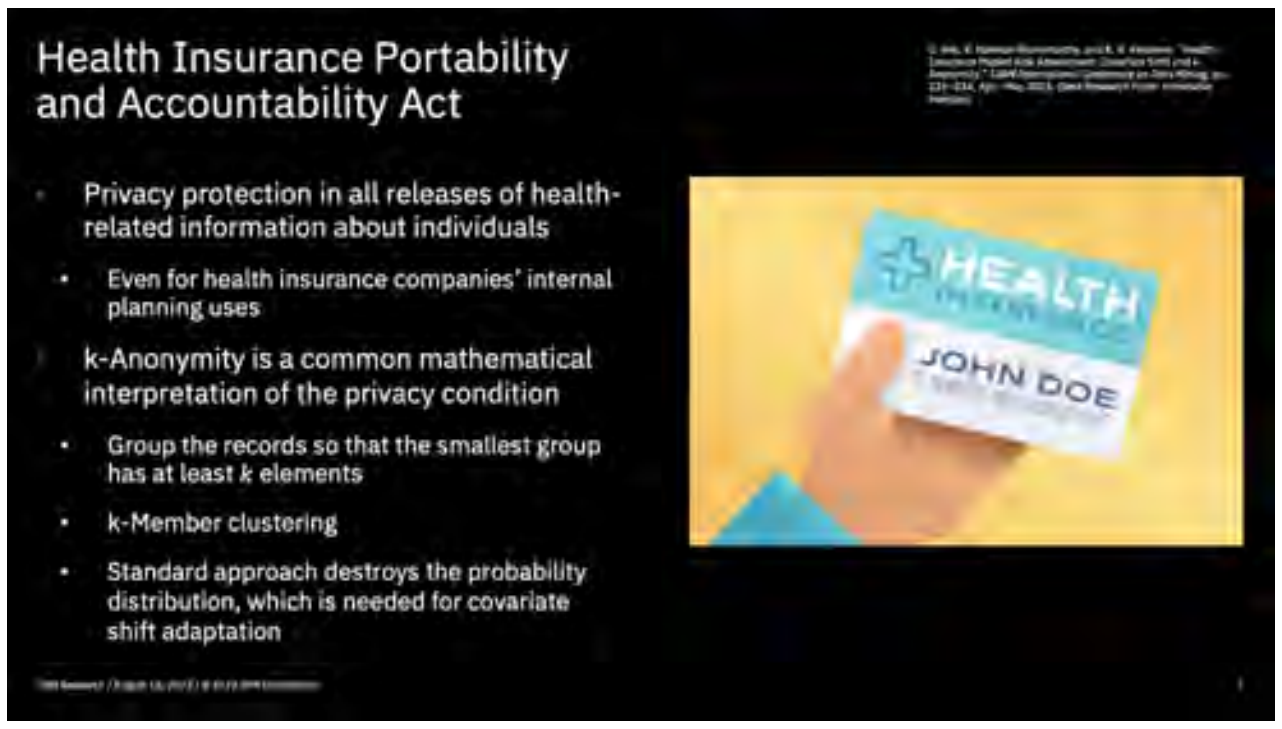

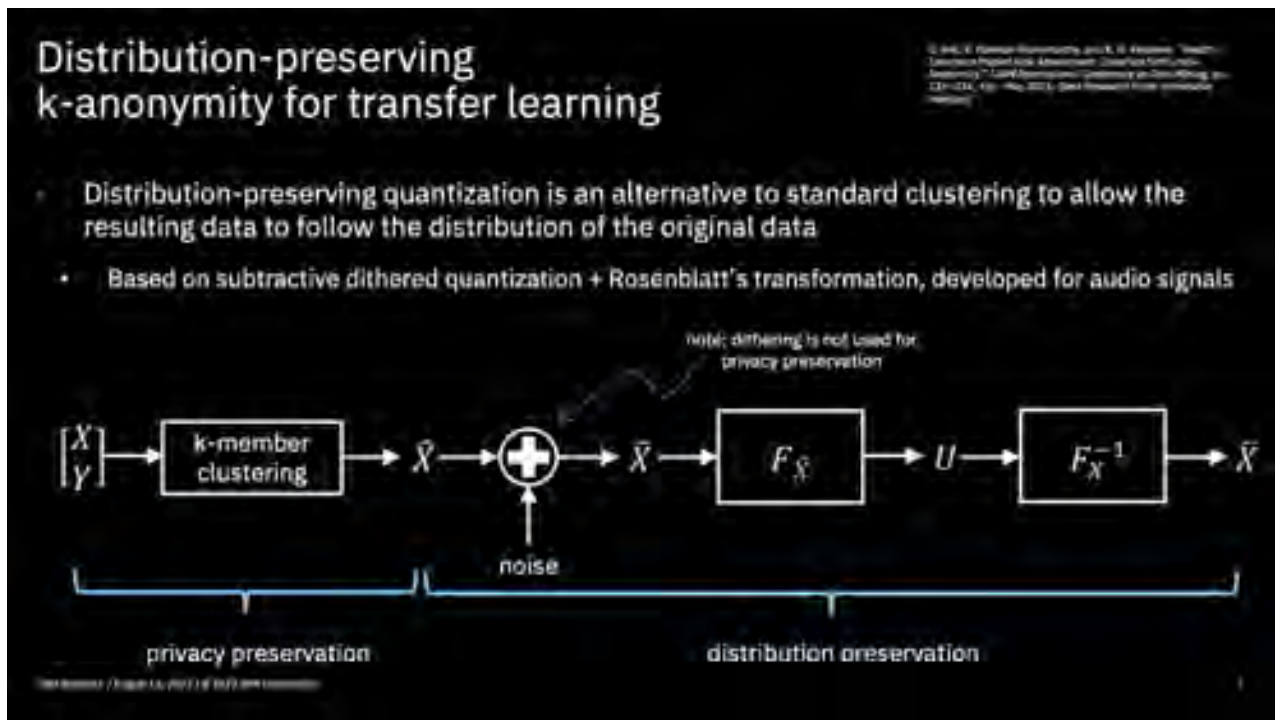
Distribution-preserving
k-anonymity for transfer learning
Distribution-preserving quantization is an alternative to standard clustering to allow the resulting data to follow the distribution of the original data
Based on subtractive dithered quantization + Rosenblatt's transformation, developed for audio signals
note: dithering is not used for privacy preservation
k-member clustering
noise
privacy preservation
distribution preservation

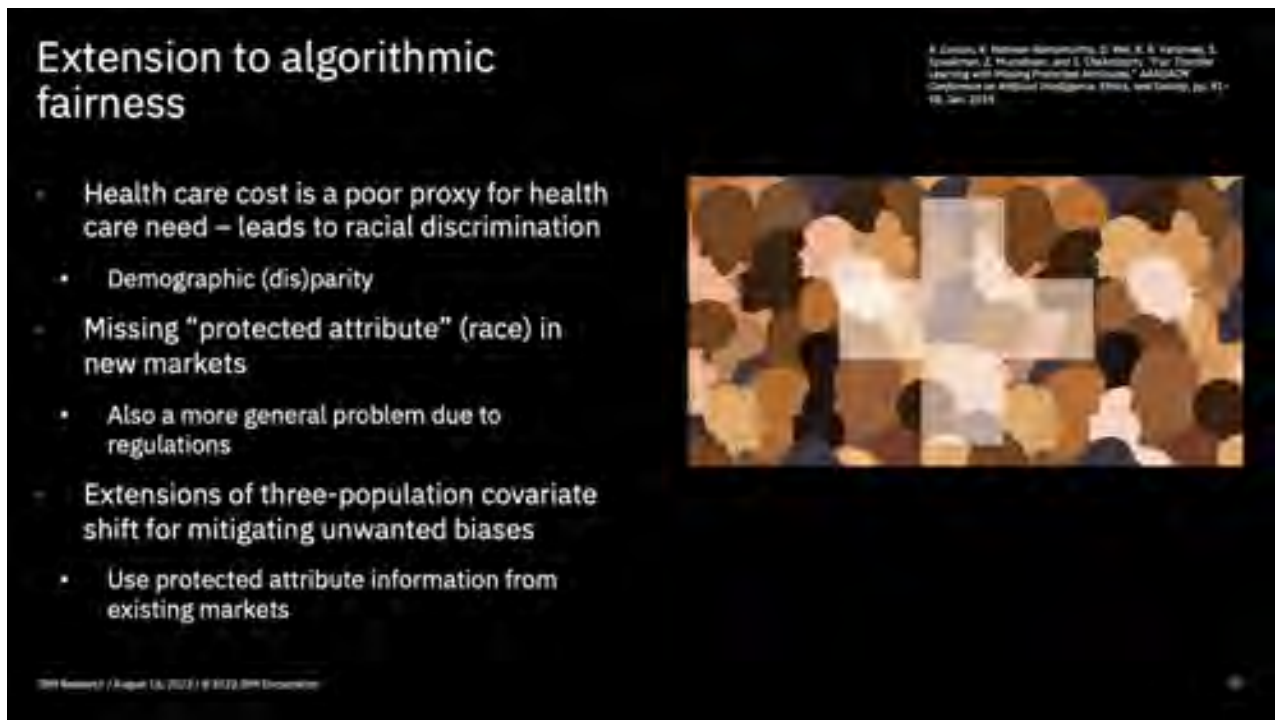
Extension to algorithmic fairness
Health care cost is a poor proxy for health care need – leads to racial discrimination
Demographic (dis)parity
Missing "protected attribute" (race) in new markets
Also a more general problem due to regulations
Extensions of three-population covariate shift for mitigating unwanted biases
Use protected attribute information from existing markets

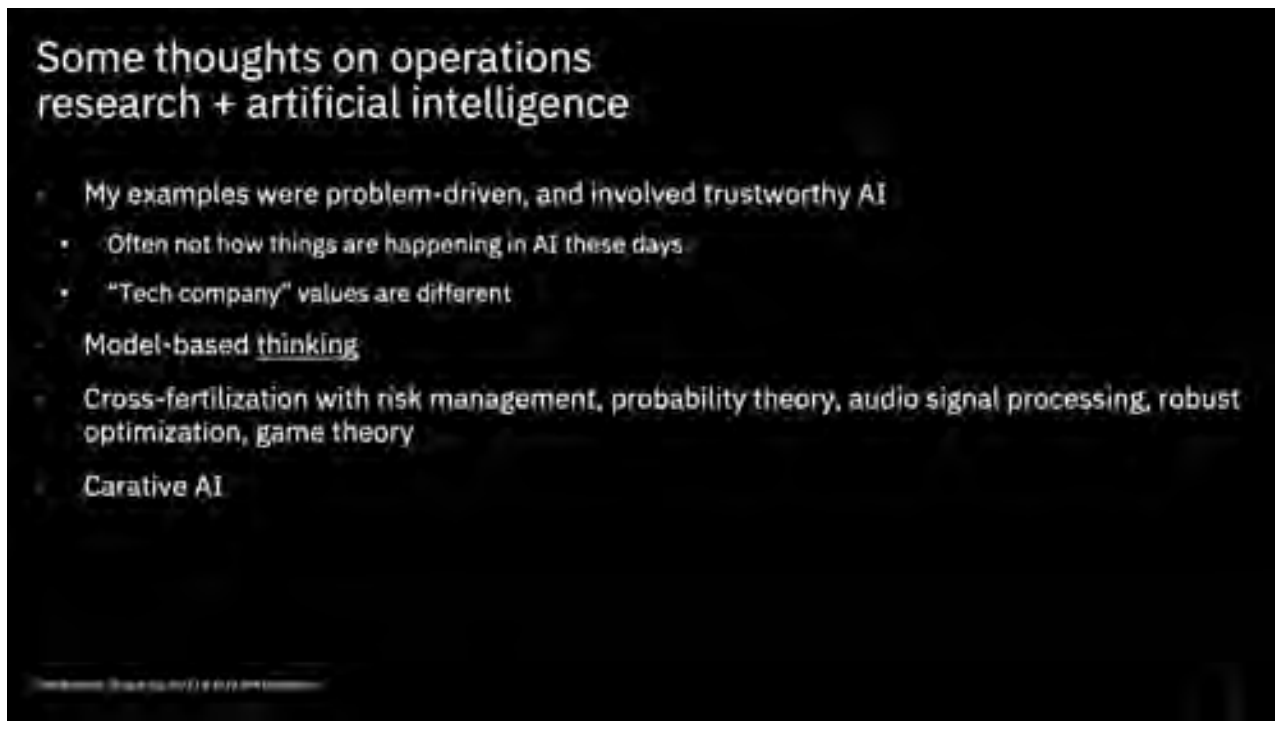
Some thoughts on operations research + artificial intelligence
My examples were problem-driven, and involved trustworthy AI
Often not how things are happening in AI these days
"Tech company" values are different
Model-based thinking
Cross-fertilization with risk management, probability theory, audio signal processing, robust optimization, game theory
Carative AI

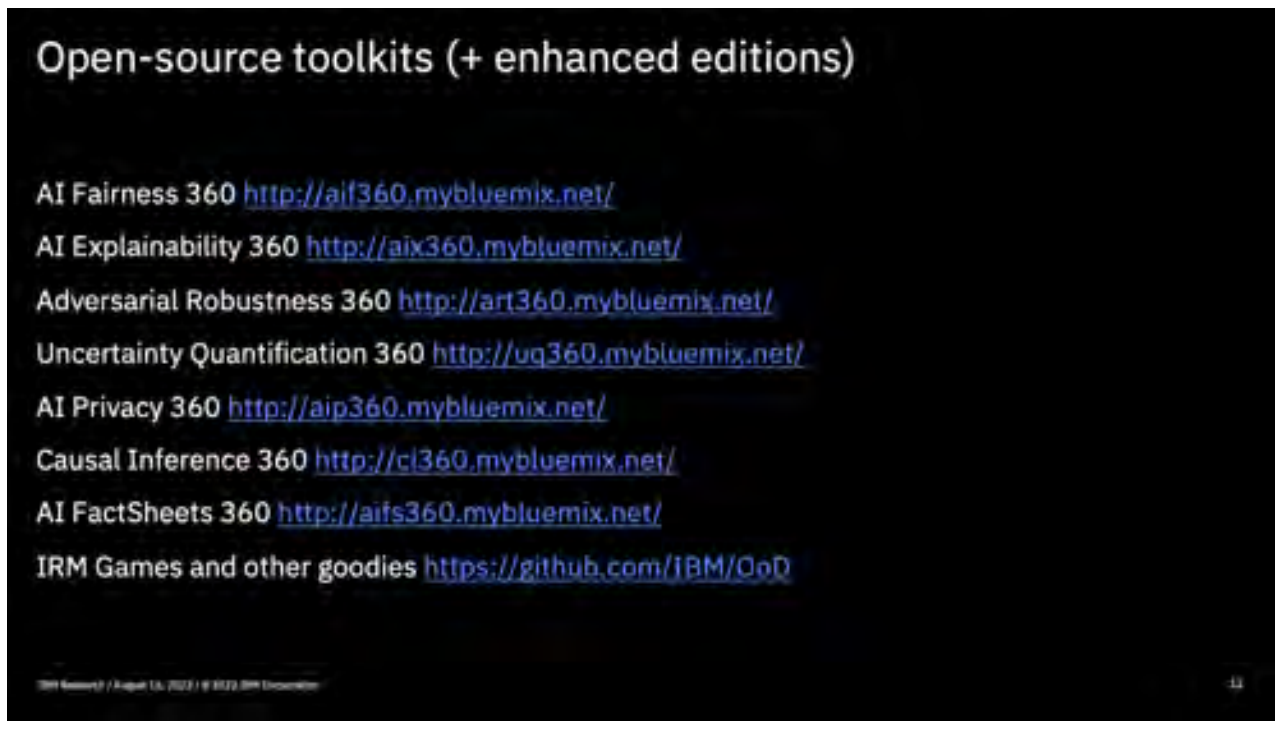
Open-source toolkits (+ enhanced editions)
AI Fairness 360 http://aif360.mybluemix.net/
AI Explainability 360 http://aix360.mybluemix.net/
Adversarial Robustness 360 http://art360.mybluemix.net/
Uncertainty Quantification 360 http://uq360.mybluemix.net/
AI Privacy 360 http://aip360.mybluemix.net/
Causal Inference 360 http://ci360.mybluemix.net/
AI FactSheets 360 http://aifs360.mybluemix.net/
IRM Games and other goodies https://github.com/IBM/OoD

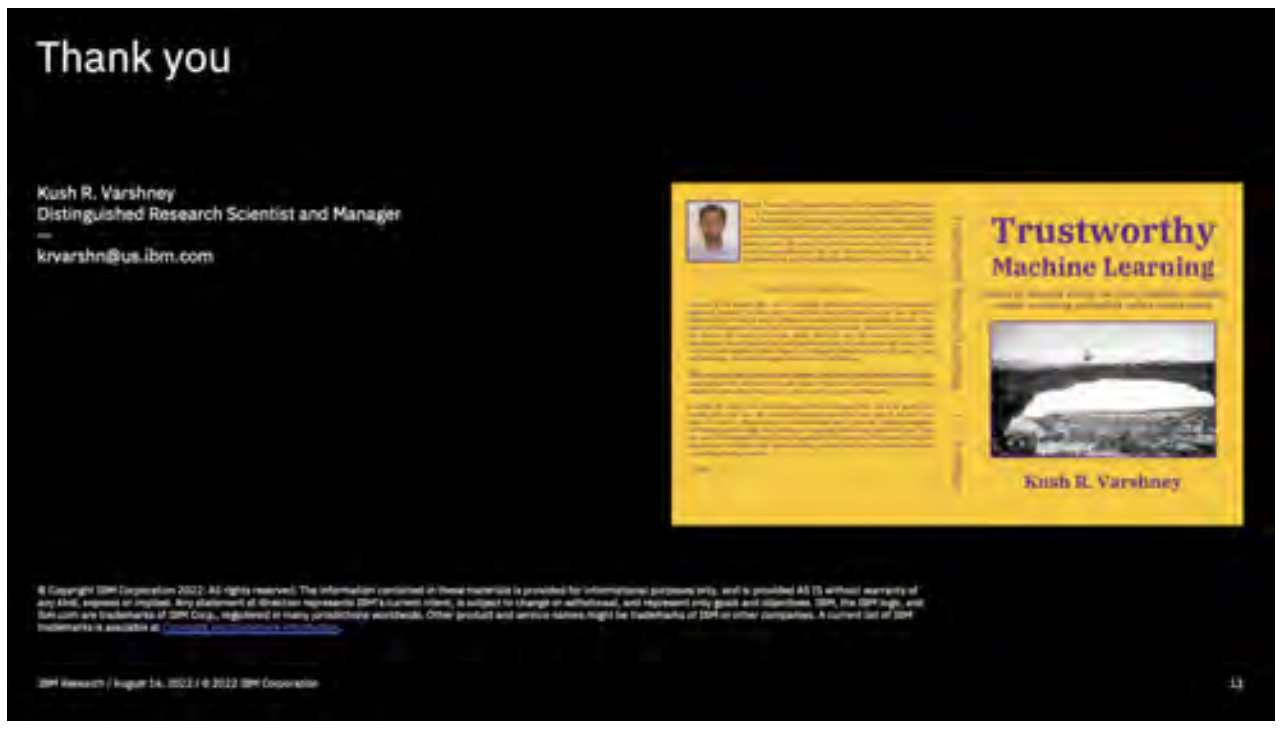

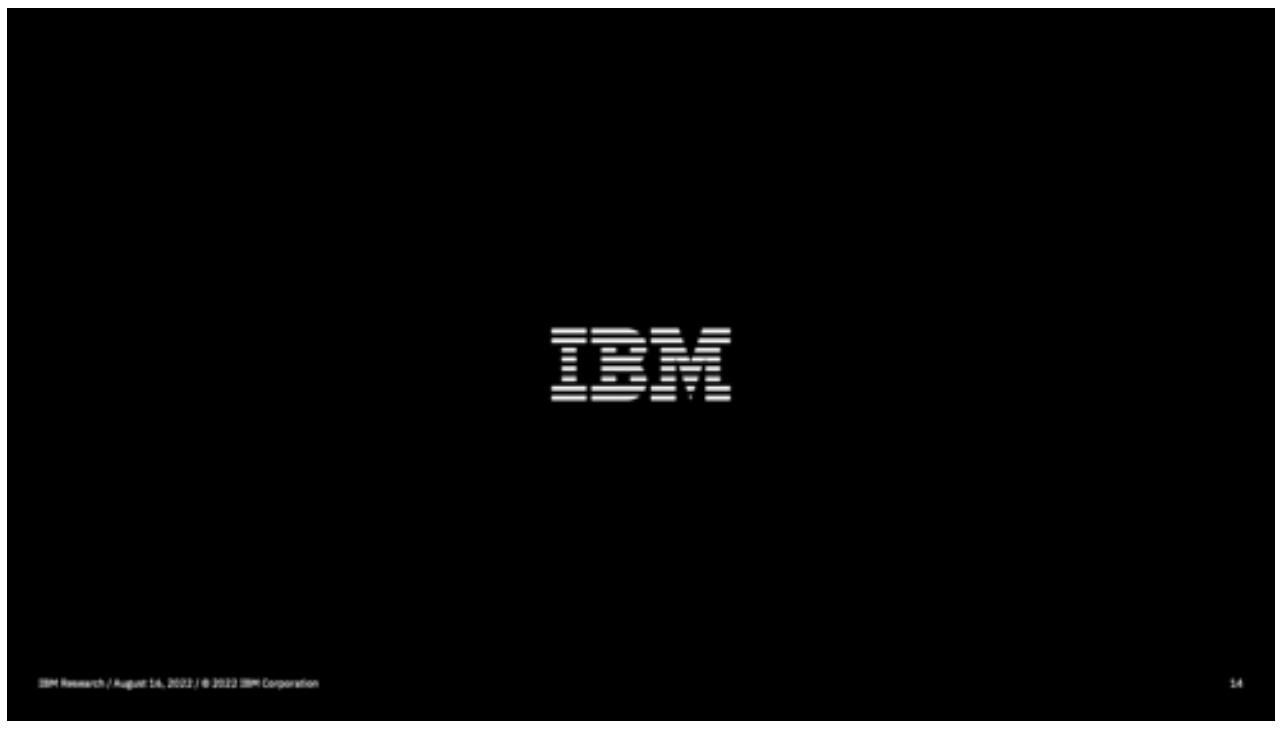
IBM

## Figure B-12: John Abowd - Some Lessons from the 2020 U.S. Census Disclosure Avoidance System

### Some Lessons from the 2020 U.S. Census Disclosure Avoidance System

John M. Abowd
Chief Scientist and Associate Director for Research and Methodology
U.S. Census Bureau

Computing Community Consortium, INFORMS, ACM SIGAI
Artificial Intelligence/Operations Research Workshop II
Panel C: Robustness/Privacy, Tuesday, August 16, 2022, 3:30pm

United States Census Bureau

*The views expressed in this talk are my own and not those of the U.S. Census Bureau. DMS Project ID: P-7502798. DRB Clearance numbers: CBDRB-FY20-DSEP-001, CBDRB-FY22-DSEP-003, CBDRB-FY22-DSEP-004.*

### Acknowledgements

Co-authors on the HDSR paper: Robert Ashmead, Ryan Cumings-Menon, Simson Garfinkel, Christine Heiss, Robert Johns, Daniel Kifer, Philip Leclerc, Ashwin Machanavajjhala, Brett Moran, William Sexton, Matthew Spence, Pavel Zhuravlev
The 2020 Census Disclosure Avoidance System TopDown Algorithm · Special Issue 2: Differential Privacy for the 2020 U.S. Census (mit.edu)

Annual Review of Statistics paper (with Michael Hawes): [2206.03524] Confidentiality Protection in the 2020 US Census of Population and Housing (arxiv.org)

This presentation also includes work by the Census Bureau's 2020 Disclosure Avoidance System development team, Census Bureau colleagues, and our collaborators from the following Census Bureau divisions and outside organizations: ADCOM, ADDC, ADRM, CED, CEDDA, CEDSCI, CES, CSRM, DCMD, DITD, ESMD, GEO, POP, TAB, CDF, Econometrica Inc., Galois, Knexus Research Corp, MITRE, NLT, TI, and Tumult Labs.

I also acknowledge and greatly appreciate the ongoing feedback we have received from external stakeholder groups that has contributed to the design and improvement of the Disclosure Avoidance System.

United States Census Bureau

### Bottom Line Up Front:

*Going from suppression to differential privacy is much easier than going from publishing all the microdata to differential privacy.*

United States Census Bureau

### Translation:

*2020 Census data clients had accuracy expectations that modern privacy protection can't support (2010 Census basically released all the microdata, although not intentionally).*

United States Census Bureau

### Forecast:

*AI applications, particularly in industry, are going to face the same conundrum. Advertising executives are not going to like the privacy-protected models. (Conventional AI applications are inherently disclosive.)*

United States Census Bureau

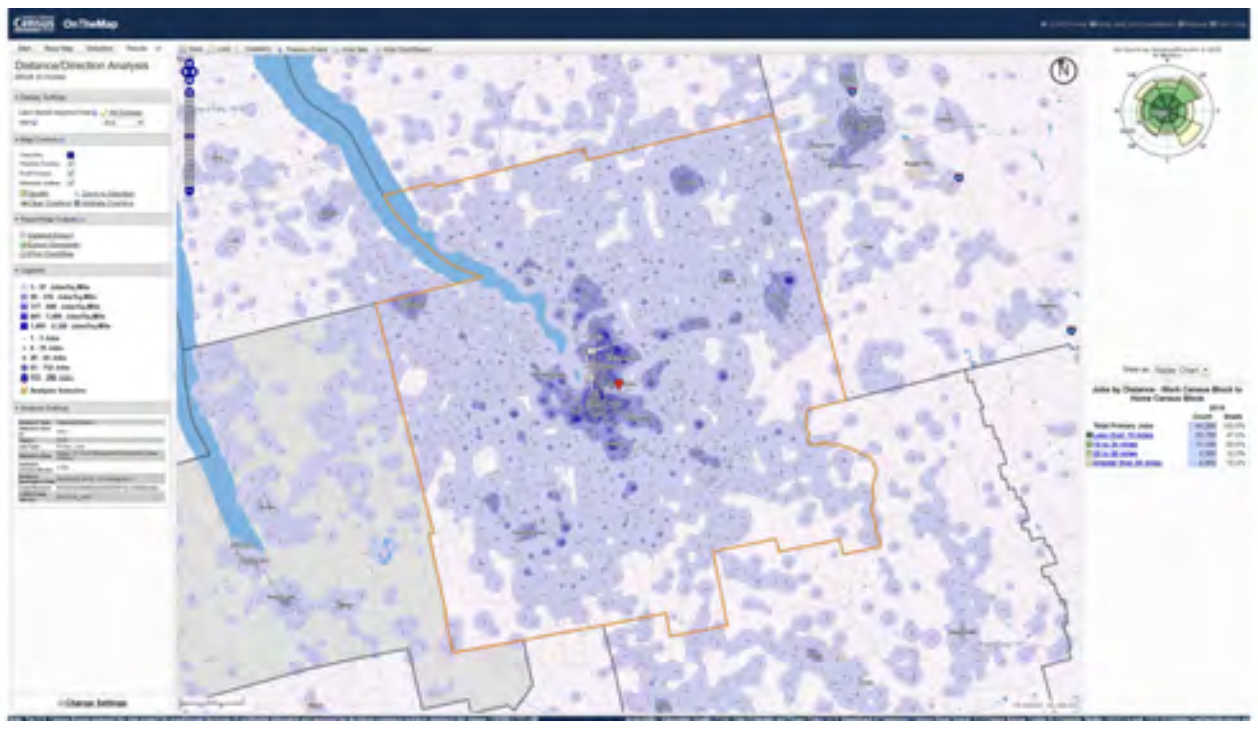

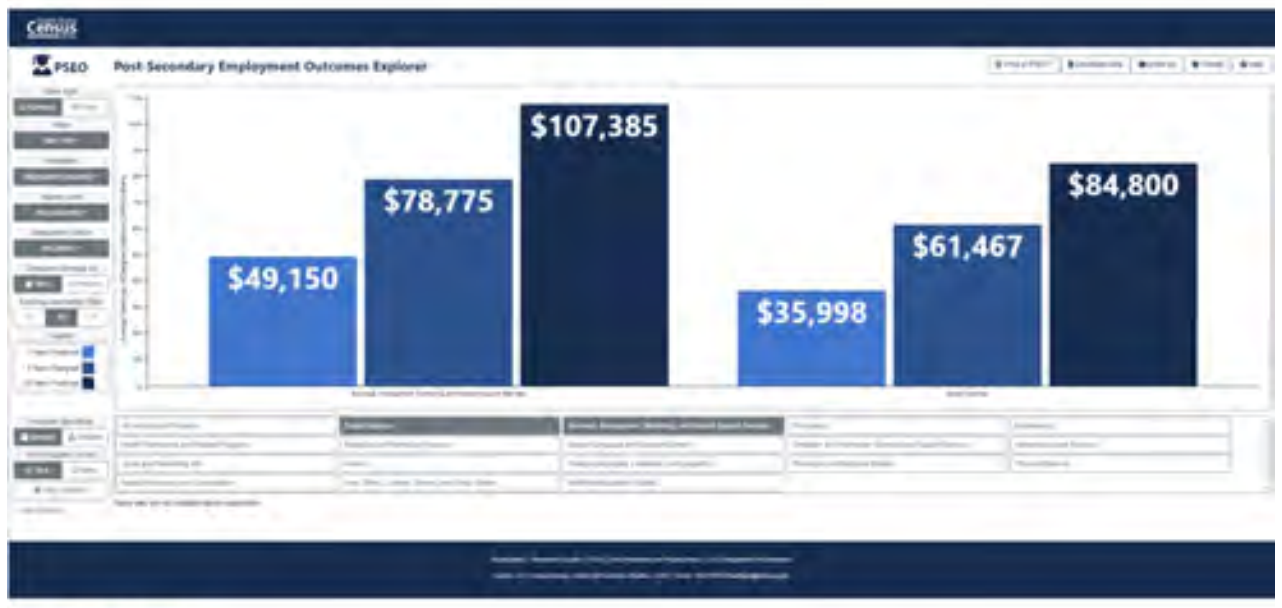

United States Census Bureau

Major data products from the 2020 Census:

- Apportion the House of Representatives (April 26, 2021)
- Supply data to all state redistricting offices (August 12, 2021)
- Demographic and housing characteristics (May 2023)
- Detailed demographic and housing characteristics (Part A August 2023; Part B TBD; Supplemental DHC TBD)
- American Indian, Alaska Native, Native Hawaiian data (Included in Part A Detailed DHC; August 2023)

For the 2010 Census, this was *more than 150 billion* statistics from 15GB total data.

United States Census Bureau

## Reconstructing the 2010 Census-I

- The 2010 Census collected information on the age, sex, race, ethnicity, and relationship (to householder) status for 308,745,538 million individuals. (about 1.5 billion confidential data points; Garfinkel et al. 2019)
- The 2010 Census data products released over 150 billion statistics
- Internal Census Bureau research confirms that the confidential 2010 Census microdata can be accurately reconstructed from the publicly released tabulations
- This means that all the tabulation variables for 100% of the person records on the confidential data file can be accurately reproduced from the published tabulations
- Based on Dinur and Nissim (2003) and Dwork and Yekhanin (2008)
- *A violation of the 2010 Census contemporaneous disclosure avoidance standards for 2010 Census microdata files*

## Reconstructing the 2010 Census-II

- A violation of the 2010 Census contemporaneous disclosure avoidance standards for 2010 Census microdata
  - The reconstructed microdata are not a sample; there is one record for every person enumerated in the 2010 Census, and the geographic identifier on that record is always correct (matches the geographic identifier on the confidential input file—the Hundred-percent Detail File, which was swapped)
  - The reconstructed microdata have geography identifiers with an average population of 29 (50, if only occupied blocks are counted)
  - The reconstructed microdata have U.S.-level demographic cells (race, ethnicity) with fewer than 10,000 persons
- The standards for releasing microdata from the 2010 Census required (McKenna, 2019)
  - Sample (10% rate was used)
  - Restrict geographic identifiers to areas with at least 100,000 persons (Public-use Microdata Areas)
  - Collapse demographic categories until the national population in 1-way marginals contains at least 10,000 persons
  - The standards for tabular data permitted universe files, block geography, and low-U.S. population demographic groups (McKenna, 2018) on the assumption that microdata reconstruction was infeasible
- These are the reason the Data Stewardship Executive Policy Committee instructed the 2020 Census not to use swapping as the main protection for 2020 Census products from the reconstruction evidence alone: swapping plus aggregation did not protect the 2010 Census confidential microdata properly

## Reconstructing the 2010 Census: What did we find?

Table 1 Agreement Rates (Reconstruction to CEF) by Block Size

| Block Size | Total | 1-9 | 10-49 | 50-99 | 100-249 | 250-499 | 500-999 | 1,000+ |
|---|---|---|---|---|---|---|---|---|
| Agreement | 91.8% | 74.0% | 93.0% | 93.1% | 92.1% | 91.3% | 90.6% | 91.5% |

DRB clearance number CBDRB-FY22-DSEP-004; Source: Hawes (2022).

- Block, sex, age (exact/binned in 38 categories), race (OMB 63 categories), and ethnicity were reconstructed:
  - Exactly for 91.8% of the population
  - Exactly 74.0% in the smallest population blocks, but 93.0% in blocks with 10-49 people and 93.1% in blocks with 50-99 people
  - An external user can confirm that these solutions correspond to the exact record in the confidential data for 65% of all blocks using only the published data because there is provably one and only one reconstruction possible in these blocks. That user can identify population uniques on any combination of reconstructed variables.

*This is one of the principal failures of the 2010 tabular disclosure avoidance methodology — swapping provided protection for households deemed "at risk," primarily those in blocks with small populations, whereas for the for the entire 2010 Census 57% of the persons are population uniques on the basis of block, sex, age (in years), race (OMB 63 categories), and ethnicity. Furthermore, 44% are population uniques on block, age and sex. Aggregation provided no additional protection for most blocks.*

Table 5
Distribution of Population and Population Uniques by Block Population Size

| Block Population Bin | Number of Blocks in Bin | 2010 Census Population in Bin | Cumulative Population | Percent of Population in Bin | Cumulative Percent of Population | Population Uniques (block, sex, age) in Bin | Percent of (block, sex, age) Uniques in Bin |
|---|---|---|---|---|---|---|---|
| TOTAL | 11,078,297 | 308,745,538 | | | | 135,432,888 | 43.87% |
| 0 | 4,871,270 | 0 | 0 | 0.00% | 0.00% | | |
| 1-9 | 1,823,665 | 8,069,681 | 8,069,681 | 2.61% | 2.61% | 7,670,927 | 95.06% |
| 10-49 | 2,671,753 | 67,597,683 | 75,667,364 | 21.89% | 24.51% | 53,435,603 | 79.05% |
| 50-99 | 994,513 | 69,073,496 | 144,740,860 | 22.37% | 46.88% | 40,561,372 | 58.72% |
| 100-249 | 540,455 | 80,020,916 | 224,761,776 | 25.92% | 72.80% | 27,258,556 | 34.06% |
| 250-499 | 126,344 | 42,911,477 | 267,673,253 | 13.90% | 86.70% | 5,297,867 | 12.35% |
| 500-999 | 40,492 | 27,028,992 | 294,702,245 | 8.75% | 95.45% | 1,051,924 | 3.89% |
| 1000+ | 9,805 | 14,043,293 | 308,745,538 | 4.55% | 100.00% | 156,639 | 1.12% |

DRB clearance number CBDRB-FY21-DSEP-003

Table 2 Reidentification rates for population uniques

| Match file | Universe | Putative rate[a] | Confirmed rate[b] | Precision[c] |
|---|---|---|---|---|
| Commercial | All data defined persons[d] | 60.2% | 24.8% | 41.2% |
| | Population uniques[e] | 23.1% | 21.8% | 94.6% |
| CEF | All data defined persons[d] | 97.0% | 75.5% | 77.8% |
| | Population uniques[e] | 93.1% | 87.2% | 93.6% |

[a]The number of records that agree on block, sex, and age (exact/binned), divided by the total number of records in the universe. [b]The number of records that agree on PIK (the Census Bureau's internal person identifier), block, sex, age (exact/binned), race, and ethnicity, divided by the total number of records in the universe. [c]The number of confirmed reidentifications [records that agree on PIK, block, sex, age (exact/binned), race, and ethnicity] divided by the number of putative reidentifications [records that agree on block, sex, and age (exact/binned)]. [d]All individuals with a unique PIK identifier within the block (276 million persons for the 2010 Census). [e]All data defined individuals who are unique in their block on sex and exact/binned age. DRB clearance number CBDRB-FY22-DSEP-004; Data are from Abowd et al. (under review) released in Hawes (2022). Abbreviations: CEF, Census Edited File; DRB, Disclosure Review Board; PIK, Protected Identification Key.

Table 3 Reidentification rates for population uniques of the block's modal and non-modal races

| Match file | Population uniques[a] | Putative rate | Confirmed rate | Precision |
|---|---|---|---|---|
| Commercial | All population uniques | 23.1% | 21.8% | 94.6% |
| | Of the modal race | 25.3% | 24.2% | 95.3% |
| | Of the nonmodal races | 13.7% | 12.2% | 89.2% |
| CEF | All population uniques | 93.1% | 87.2% | 93.6% |
| | Of the modal race | 94.8% | 91.3% | 96.3% |
| | Of the nonmodal races | 86.2% | 70.2% | 81.5% |

[a]Individuals who are unique in their block on sex and exact/binned age. DRB clearance number CBDRB-FY22-DSEP-004. Data are from Abowd et al. (under review) released in Hawes (2022). Abbreviations: CEF, Census Edited File; DRB, Disclosure Review Board.

- *This is not a statistical use, and both the Census Act (13 U.S. Code §§ 8(b) & 9) and CIPSEA (44 U.S. Code § 3561(11) 'Statistical Purpose') clearly prohibit releasing data that support not-statistical uses.*

## All 2020 Census Publications

- Will all be processed by a collection of differentially private algorithms (Dwork et al. 2006a, 2006b; Dwork 2006) using the zero-Concentrated DP privacy-loss accounting framework (Bun and Steinke 2016) implemented with the discrete Gaussian mechanism (Canonne et al. 2020, 2021)
- Using a total privacy-loss budget set as policy, not hard-wired, determined by the Data Stewardship Executive Policy Committee
- Production code base, technical documents, and extensive demonstration products based on the 2010 Census confidential data have all been released to the public
- More information: https://www.census.gov/newsroom/blogs/research-matters/2019/10/balancing_privacyan.html

## TopDown Algorithm System Requirements

- The 2020 Disclosure Avoidance System's TopDown Algorithm (TDA) implemented formal privacy protections for the P. L. 94-171 Redistricting Data Summary File
- Planned for use in the Demographic Profiles, Demographic and Housing Characteristics (DHC), and Special Tabulations of the 2020 Census
- TDA system requirements include:
  - Input/output specifications
  - Invariants
  - Edit constraints and structural zeros
  - Tunable utility/accuracy for pre-specified tabulations
  - Privacy-loss budget asymptotic consistency
  - Transparency

## What is a histogram?

| Record ID | Block | Race | ... | Sex |
|---|---|---|---|---|
| 1 | 1001 | Black | ... | Male |
| 2 | 1001 | Black | ... | Male |
| 3 | 1001 | Asian | ... | Female |
| 4 | 1001 | Asian | ... | Female |
| 5 | 1001 | Black | ... | Male |
| 6 | 1001 | AIAN | ... | Female |
| 7 | 1001 | AIAN | ... | Male |
| 8 | 1001 | Black | ... | Female |
| 9 | 1001 | Black | ... | Female |

Microdata: One record per respondent

⇆

| Attribute Combination (Block/Race/.../Sex) | # of Records |
|---|---|
| 1001/AIAN/.../Male | 1 |
| 1001/AIAN/.../Female | 1 |
| 1001/Asian/.../Male | 0 |
| 1001/Asian/.../Female | 2 |
| 1001/Black/.../Male | 3 |
| 1001/Black/.../Female | 2 |
| ... | ... |

Histogram: Record count for each unique combination of attributes (including location), equivalent to the fully saturated contingency table, vectorized, and with structural zeros removed or imposed by constraint

## Noisy Measurements

- TDA allocates shares of the total privacy-loss budget by geographic level and by query
- Each query of the confidential data will have noise added to its answer
- The noise is taken from a probability distribution with mean=0, and variance determined by the share of the privacy-loss budget allocated to that query at that geographic level
- These noisy measurements are independent of each other, and can include negative values, hence the need for post-processing

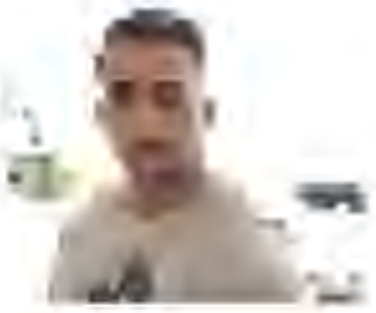

## Zero-Concentrated Differential Privacy (zCDP)

- Privacy-loss parameter: $\rho$ (Bun and Steinke 2016)
- $\rho$-based privacy-loss budgets can be converted to any single point along a continuum of $(\varepsilon, \delta)$ pairs. Analysis of the privacy protection afforded by a $\rho$ budget should use the entire continuum, not a single $(\varepsilon, \delta)$ point. Some formulas provide tighter bounds on the $(\varepsilon, \delta)$ curve implied by a particular value of $\rho$. TDA uses this one:

$$\varepsilon = \rho + 2\sqrt{-\rho \log_e \delta}$$

- Noise distribution: discrete Gaussian (Canonne et al. 2020, 2021)
- The expected variance of any noisy measurement can be estimated by knowing the total privacy-loss budget and the share of $\rho$ allocated to that query at that geographic level [see Appendix B of Abowd et al. (2022) technical paper]

## The TopDown Approach

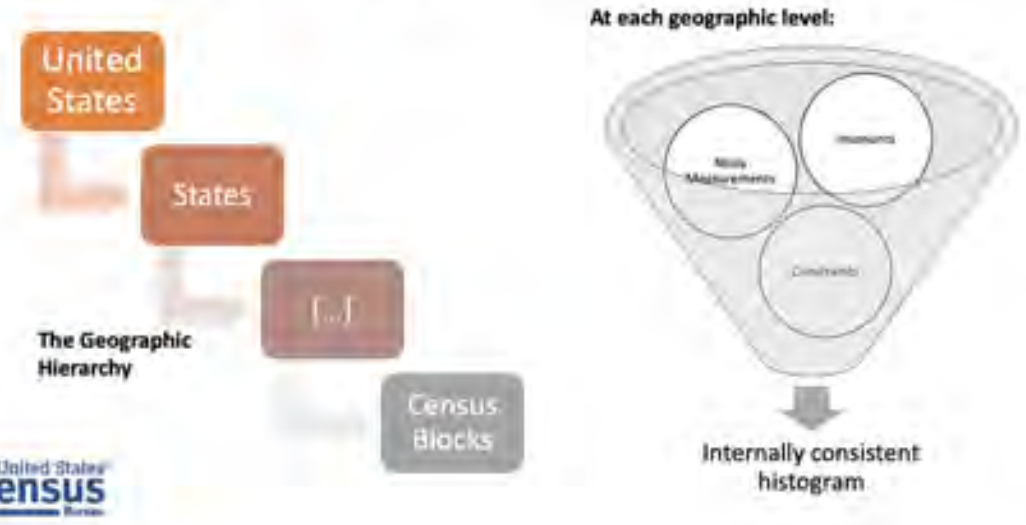

## Naïve Method: BottomUp or Block-by-Block

- Apply differential privacy algorithms to the most detailed level of geography
- Build all geographic aggregates from those components as a post-processing

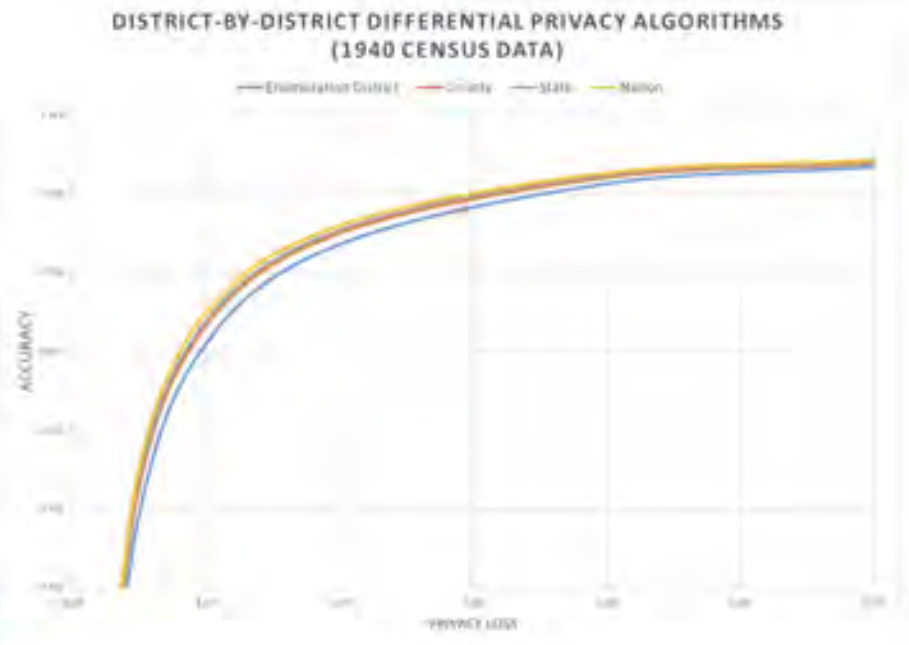

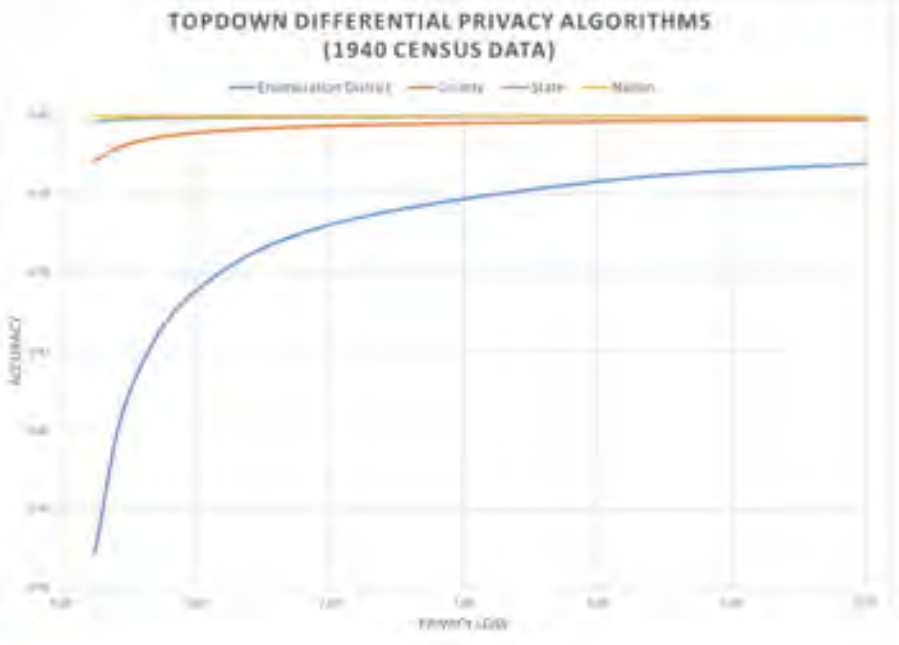

## Benefits of TDA Compared to Block-by-block

- TDA is in stark contrast with naïve alternatives (e.g., block-by-block or bottom-up)
- TDA disclosure-limitation error does not increase with number of contained Census blocks in the geographic entity (on spine)
- TDA yields increasing relative accuracy as the population being measured increases (in general), and increased count accuracy compared to block-by-block
- TDA "borrows strength" from upper geographic levels to improve count accuracy at lower geographic levels (e.g., for sparsity)

*If you feed TDA 16.6 billion differentially private measurements (23 trillion for DHC), it will do a good job that completely satisfies no one.*

*(This was predicted in Abowd and Schmutte 2019.)*

## Accurate, but to whom?

- DAS operates under interpretable formal privacy guarantees, given privacy-loss budgets
- Accuracy properties depend upon the output metric (use case)
- Distinct groups of data users will have a particular analyses they wish to be accurate
- Tuning accuracy for a given analysis can reduce accuracy for other analyses
- Policy makers must consider reasonable overall accuracy metrics for privacy tradeoffs

## Deep Dive: Redistricting Data

- Legislative districts for politically defined entities of arbitrary size
- Must be (approximately) equal populations in each district
- Districts must be consistent with Section 2 scrutiny under the 1965 Voting Rights Act
  - Large minority populations cannot be clustered into a few districts
  - Majority-minority districts (approximately 50%+ minority population) must be drawn when feasible
- Focus statistics: total population, ratio largest race/ethnic population to total population

## Multi-pass Post-processing

- The sparsity of many queries (i.e., prevalence of zeros and small counts) has the potential to introduce bias in TDA's post-processing
- To address the sparsity issue, TDA processing is performed in a series of passes
- At certain geographic levels, the algorithm constructs histograms for a subset of queries in a series of passes for that level, constraining the histogram for each pass to be consistent with the histogram produced in the prior pass
- Example for the P.L. 94-171 Redistricting Data Summary File:
  - Pass 1: Total Population
  - Pass 2: Remaining tabulations supporting P.L. 94-171 Redistricting Data

## Tabulation Geographic Hierarchy

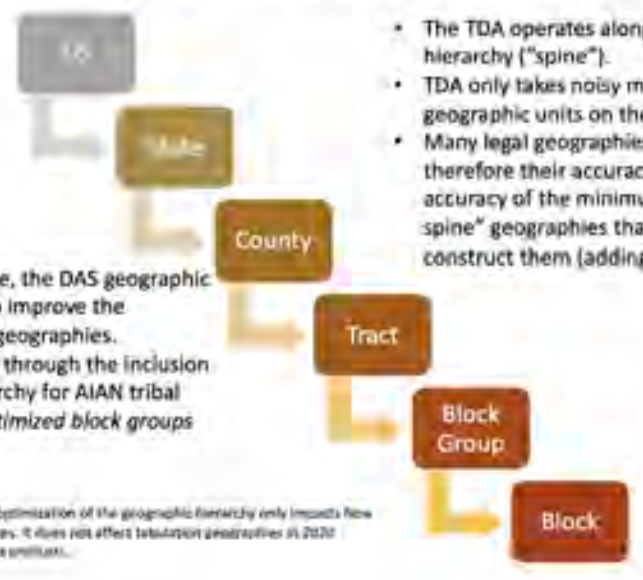

- The TDA operates along a geographic hierarchy ("spine").
- TDA only takes noisy measurements for geographic units on the hierarchy.
- Many legal geographies are "off-spine," therefore their accuracy is impacted by the accuracy of the minimum number of "on-spine" geographies that can be used to construct them (adding or subtracting).

- To address this challenge, the DAS geographic hierarchy was altered to improve the accuracy of "off-spine" geographies.
- This was done primarily through the inclusion of a branch of the hierarchy for AIAN tribal areas and the use of *optimized block groups (not shown)*.

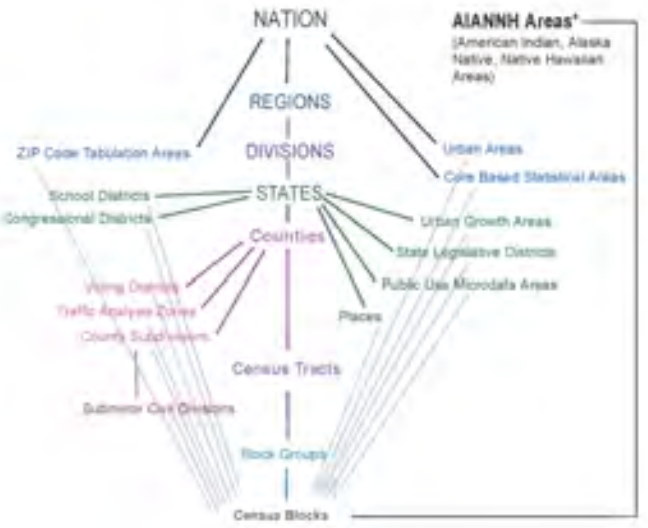

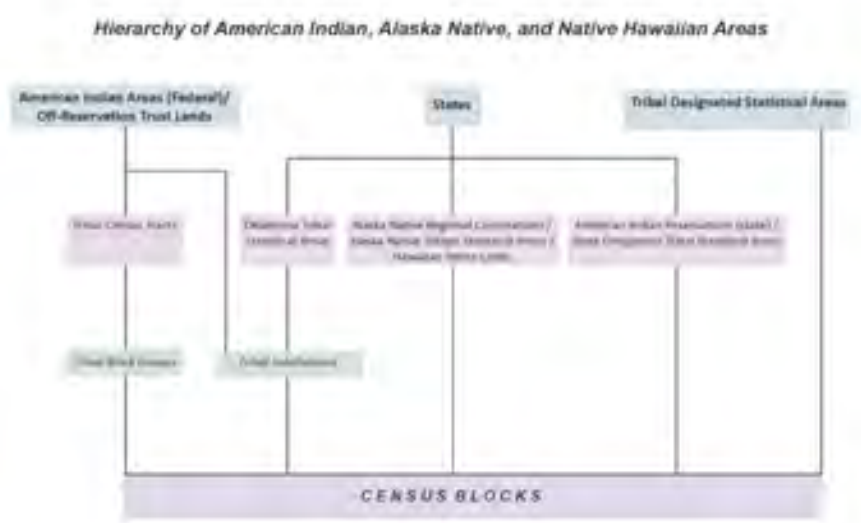

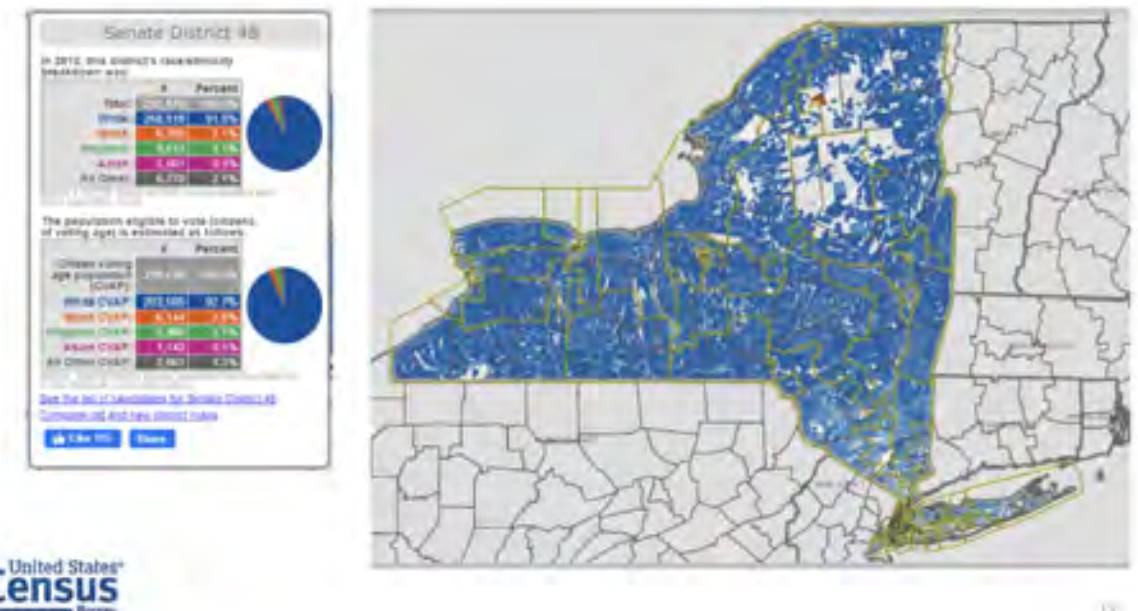

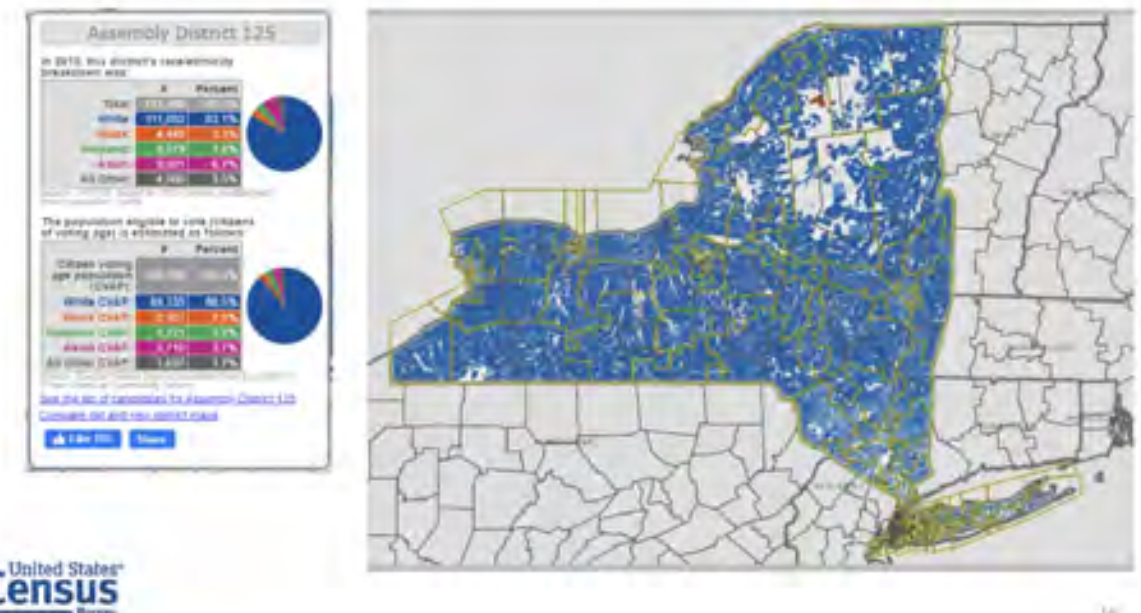

## How to reconcile these statistics

- Construct error metrics of the form

$$Pr[|TDA - CEF| \le \alpha] \ge 1 - \beta$$

- Less than $\alpha$ error with probability at least 1-$\beta$ for a target minimum population
- Statistical interpretation: absolute differences (≈RMSE differences) greater than $\alpha$ are outside the 1-$\beta$ confidence interval
- A single statistic can be used to tune the redistricting application

$$\frac{Population\ of\ Largest\ Race\ or\ Ethnic\ Group}{Total\ Population}$$

- Calculated for the TopDown Algorithm (TDA) output and the 2020 Census (CEF)
- Implemented successfully for the production code release
- *In the production data: minimum population of 200 to 249 for political areas and 450 to 499 for block groups to achieve 95% accuracy ($\alpha$ = 0.05) at least 95% of the time ($\beta$ = 0.05) See Wright and Irimata (2021)*

## What do the redistricting data do?

- Total differentially private measurements (queries): 16.6 billion
- Global ρ = 2.63 [($\varepsilon$, $\delta$) = (18.19, $10^{-10}$) and infinitely many other pairs] U.S. persons and housing units
- Total block-level tables 29.4 million
- Total block-level statistics 3.4 billion
- Total independent block-level statistics 1.5 billion
- Accuracy of populations and largest race/ethnic group fit for redistricting and Voting Rights Act scrutiny for populations of at least 200-249, which is much smaller than legal entity subject to VRA

**Figure 2.** Mean Absolute Error of the County Total Population among the Least Populous Counties (Population Under 1,000) by Demonstration Data Product Vintage

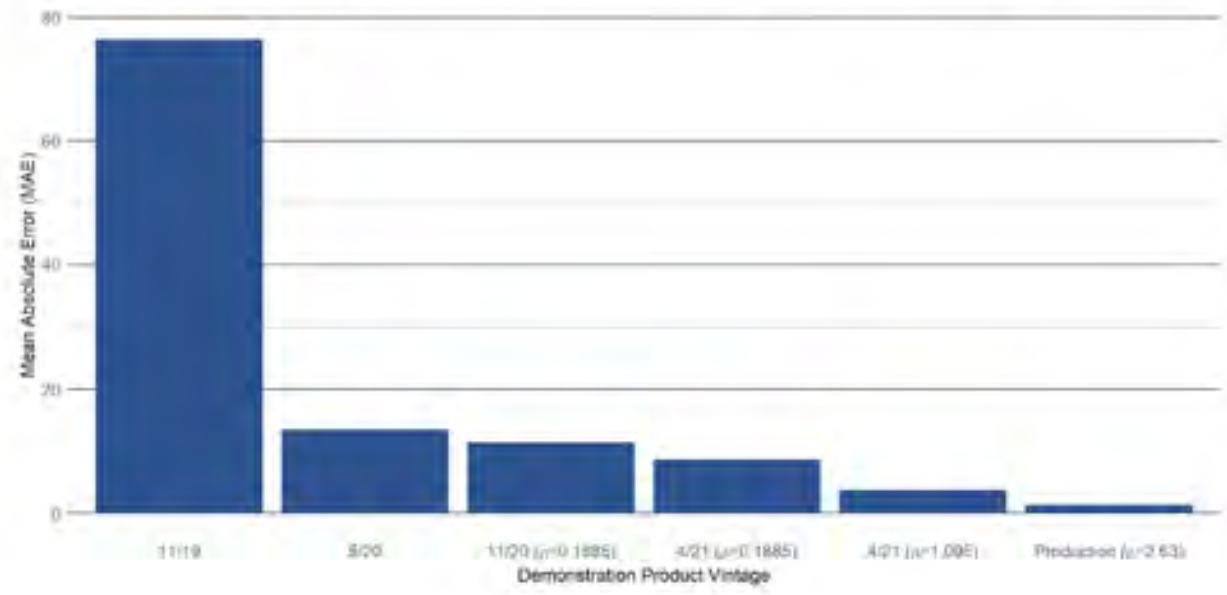

**Figure 3.** Mean Absolute Error of the Total Population for Federal American Indian Reservation/Off-Reservation Trust Lands by Demonstration Data Product Vintage

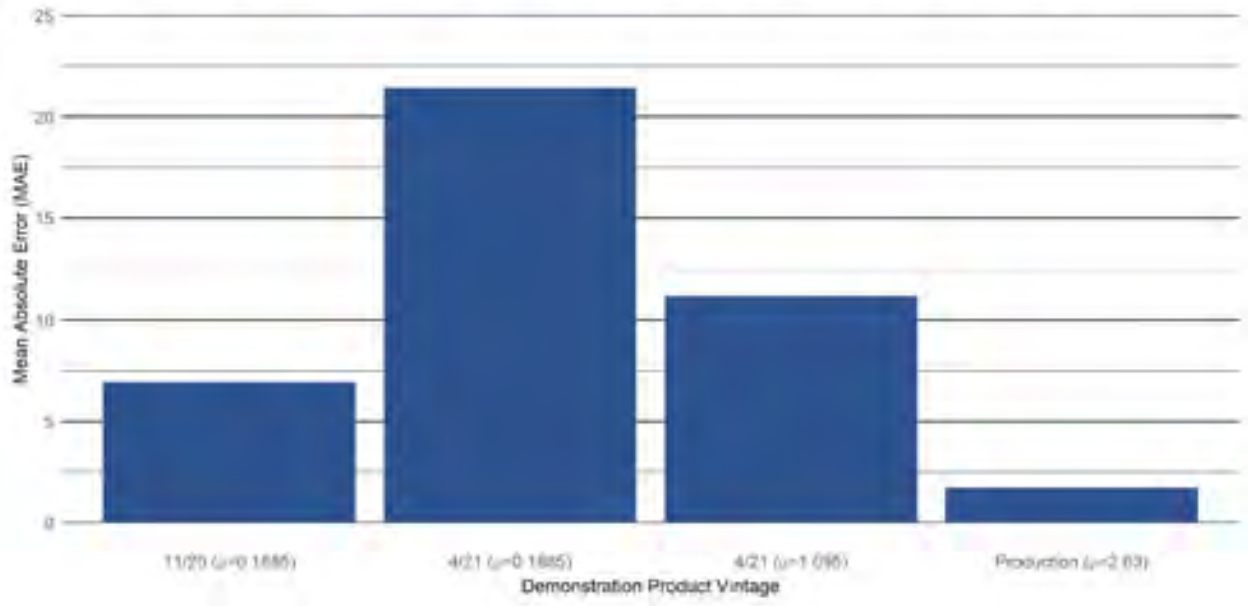

**Figure 4.** Mean Absolute Error of the Total Population among All Incorporated Places by Demonstration Data Product Vintage

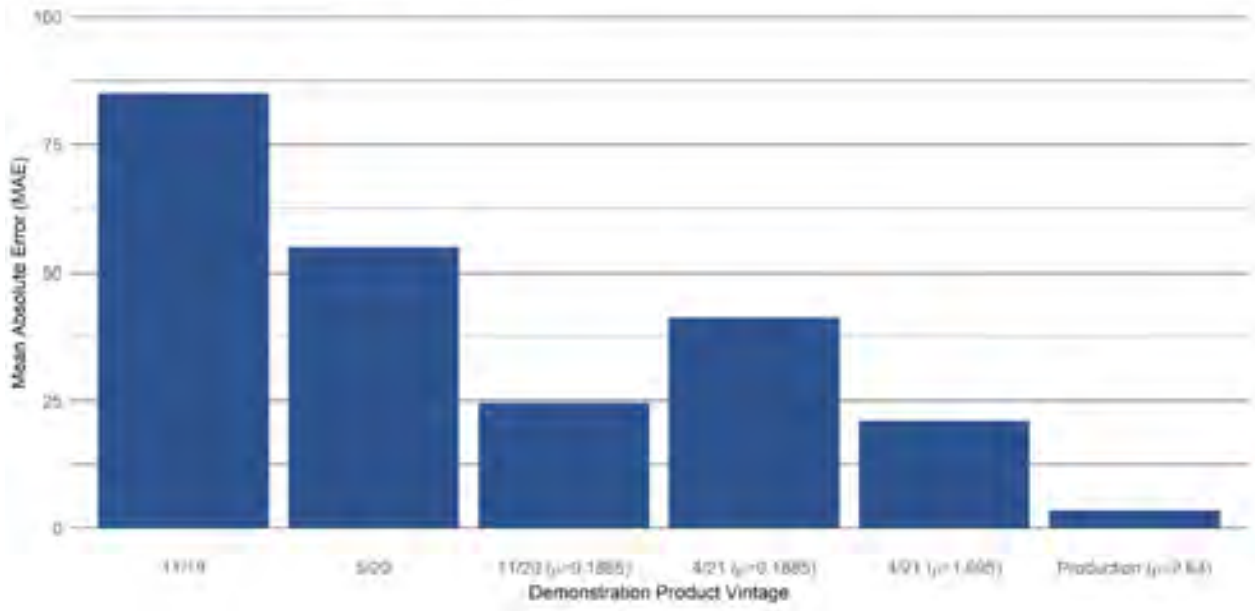

**Figure 5.** Mean Absolute Error of the Total Population among Tracts for Hispanic x Race Alone Populations by Demonstration Data Product Vintage

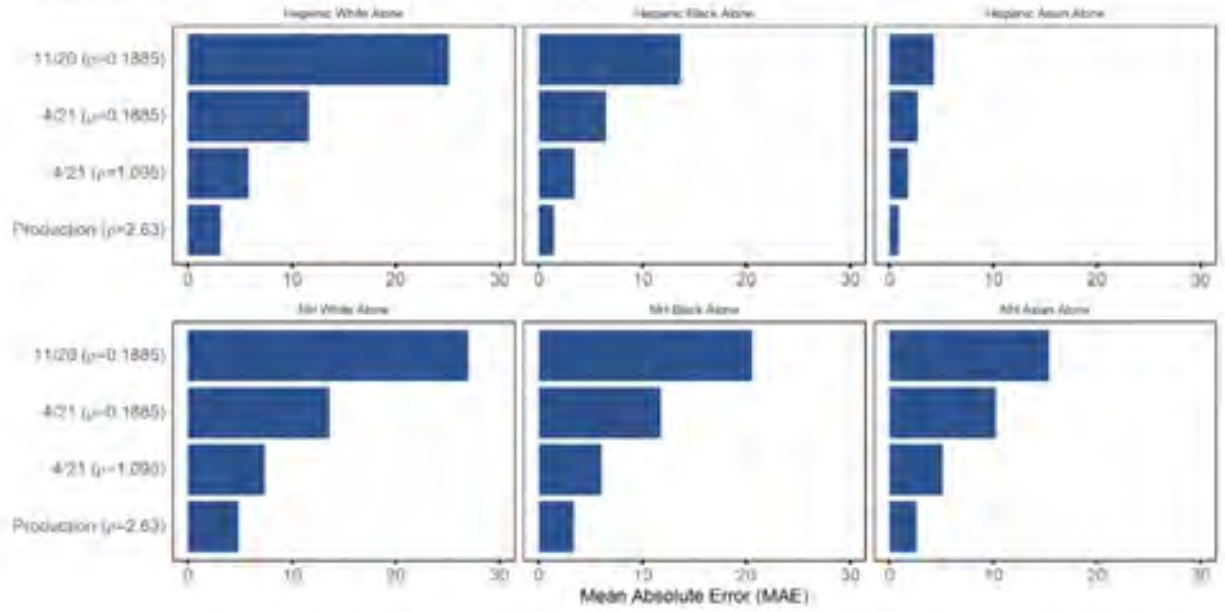

## Addressing Other Biases

**Block Groups**

| April 2021 PPMF: Diversity Quintile | Mean Difference in Total Population | Production Settings: Diversity Quintile | Mean Difference in Total Population |
|---|---|---|---|
| 0 – Least Diverse | 5.04 | 0 – Least Diverse | -0.375 |
| 1 | 4.24 | 1 | 1.009 |
| 2 | 0.99 | 2 | 0.997 |
| 3 | -2.21 | 3 | -0.303 |
| 4 – Most Diverse | -8.07 | 4 – Most Diverse | -1.352 |

**Tracts**

| April 2021 PPMF: Diversity Quintile | Mean Difference in Total Population | Production Settings: Diversity Quintile | Mean Difference in Total Population |
|---|---|---|---|
| 0 – Least Diverse | 15.95 | 0 – Least Diverse | 0.029 |
| 1 | 11.15 | 1 | 0.045 |
| 2 | 3.01 | 2 | 0.000 |
| 3 | -6.17 | 3 | -0.020 |
| 4 – Most Diverse | -23.94 | 4 – Most Diverse | -0.053 |

United States Census Bureau

## Block-Level Inconsistencies Due to DAS-induced Uncertainty

| Inconsistency | April 2021 $\rho$=1.095 Count of Blocks | Production Settings $\rho$=2.63 Count of Blocks |
|---|---|---|
| Occupied Housing Units > Household Population | 203,519 | 303,984 |
| Zero Occupied Housing Units; > 0 Household Population | 674,598 | 505,840 |
| Zero Household Population; > 0 Occupied Housing Units | 77,947 | 148,836 |
| Everyone in Block Under 18 | 90,534 | 163,884 |
| > 10 Persons Per Household | 87,342 | 121,376 |

United States Census Bureau

## Privacy-loss Budget Allocation (by geographic level)

Privacy-loss Budget Allocation 2021-06-08
Person Tables (Production Settings)
United States

| | |
|---|---|
| Global $\rho$ | 2.56 |
| Global $\varepsilon$ (incl. units) | 18.19 |
| delta | $10^{-10}$ |

| | $\rho$ Allocation by Geographic Level |
|---|---|
| US | 104/4099 |
| State | 1440/4099 |
| County | 447/4099 |
| Tract | 687/4099 |
| Optimized Block Group* | 1256/4099 |
| Block | 165/4099 |

Privacy-loss Budget Allocation 2021-06-08
Units Tables (Production Settings)
United States

| | |
|---|---|
| Global $\rho$ | 0.07 |

| | $\rho$ Allocation by Geographic Level |
|---|---|
| US | 1/205 |
| State | 1/205 |
| County | 7/82 |
| Tract | 364/1025 |
| Optimized Block Group* | 1759/4100 |
| Block | 99/820 |

United States Census Bureau

## Privacy-loss Budget Allocation (by query)

| Query | Per Query $\rho$ Allocation by Geographic Level: US | State | County | Tract | Optimized Block Group* | Block |
|---|---|---|---|---|---|---|
| TOTAL (1 cell) | | 3773/4097 | 3126/4097 | 1567/4102 | 1705/4099 | 5/4097 |
| CENRACE (63 cells) | 52/4097 | 6/4097 | 10/4097 | 4/2051 | 3/4099 | 9/4097 |
| HISPANIC (2 cells) | 26/4097 | 6/4097 | 10/4097 | 5/4102 | 3/4099 | 5/4097 |
| VOTINGAGE (2 cells) | 26/4097 | 5/4097 | 10/4097 | 5/4102 | 3/4099 | 5/4097 |
| HHINSTLEVELS (3 cells) | 26/4097 | 6/4097 | 10/4097 | 5/4102 | 3/4099 | 5/4097 |
| HHGQ (8 cells) | 26/4097 | 6/4097 | 10/4097 | 5/4102 | 3/4099 | 5/4097 |
| HISPANIC*CENRACE (126 cells) | 130/4097 | 12/4097 | 28/4097 | 1933/4102 | 1055/4099 | 21/4097 |
| VOTINGAGE*CENRACE (126 cells) | 130/4097 | 12/4097 | 28/4097 | 10/2051 | 9/4099 | 21/4097 |
| VOTINGAGE*HISPANIC (4 cells) | 26/4097 | 6/4097 | 10/4097 | 5/4102 | 3/4099 | 5/4097 |
| VOTINGAGE*HISPANIC*CENRACE (252 cells) | 26/241 | 2/241 | 101/4097 | 67/4102 | 24/4099 | 71/4097 |
| HHGQ*VOTINGAGE* HISPANIC*CENRACE (2,016 cells) | 189/241 | 230/4097 | 754/4097 | 241/2051 | 1288/4099 | 3945/4097 |

United States Census Bureau

**Table 4** Accuracy of 2010 Census, enhanced Swap, and DP: mean absolute error (in persons) for age group population counts at the county level

| Age group | 2010 Census | Enhanced swap | DP |
|---|---|---|---|
| 0-17 years | 0 | 256.41 | 9.84 |
| 18-64 years | NA[a] | 494.16 | 12.83 |
| 65 years and over | NA[a] | 431.37 | 12.66 |

[a]Error statistics for the impact of swapping as applied to the published 2010 Census are confidential. The 2010 Census swapping algorithm kept the number of non-voting age individuals (0-17 years) invariant but did inject noise into the age groups within the voting age population. DRB clearance number CBDRB-FY22-DSEP-003. Data are from Devine & Spence (2022). Abbreviations: DP, differential privacy; DRB, Disclosure Review Board; NA, not available.

United States Census Bureau

**Table 5** Reidentification statistics for 2010 Census, enhanced swap, and DP

| Reidentification Statistic | 2010 Census | Enhanced swap | DP |
|---|---|---|---|
| Putative reidentification rate | 97.0% | 75.4% | 44.4% |
| Confirmed reidentification rate | 75.5% | 46.6% | 27.4% |
| Precision rate | 77.8% | 61.8% | 61.7% |
| Precision for population uniques (nonmodal race) | 81.4% | 33.4% | 24.0% |

DRB clearance number CBDRB-FY22-DSEP-004. Data are from Abowd et al. (under review) released in Hawes (2022). External Matching File: Census Edited File. Abbreviations: DP, differential privacy; DRB, Disclosure Review Board.

- Tables 4 and 5 illustrate that TDA is a much more efficient disclosure avoidance mechanism for controlling accuracy and confidentiality than swapping with aggregation, as also shown in Abowd and Schmutte 2019.

United States Census Bureau

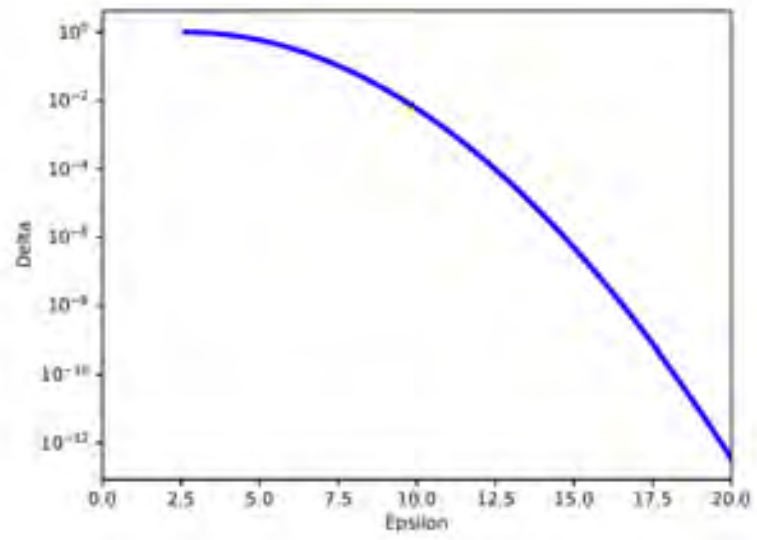

FIGURE 5. Bayesian $(\varepsilon, \delta)$ curve under the semantics of Section 7.4.2 for $\rho = 2.63$

United States Census Bureau

Source: Kifer et al. In preparation.

| Significance Level | Power (Gaussian) | Power (DGM) | zCDP Upper Bound |
|---|---|---|---|
| 0.01 | 0.032 | 0.032 | 0.037 |
| 0.05 | 0.12 | 0.12 | 0.14 |
| 0.10 | 0.21 | 0.21 | 0.24 |

TABLE 2. **Block within Custom Block Group:** Likelihood ratio test significance level/power tradeoff for block-level queries (1) if Gaussian noise is used, (2) if discrete Gaussian noise is used, (3) guaranteed upper bound if an arbitrary $\rho$-zCDP mechanism with $\rho = 0.1115$ is used.

$$\sup_{x>1} \frac{\alpha^x(1-\beta)^{1-x} + (1-\alpha)^x\beta^{1-x}}{e^{\rho x(1-x)}} \leq 1$$

where $\alpha$ is the level (probability of a Type I error), $\beta$ is the probability of a Type II error, and $(1-\beta)$ is the power of the likelihood ratio test for correctly attaching a block-id to a record when block group, age, sex, race and ethnicity are known for zCDP.
$H_0$: $N(0,1/(2\rho))$; $H_1$: $N(1,1/(2\rho))$

United States Census Bureau

Source: Kifer et al. In preparation.

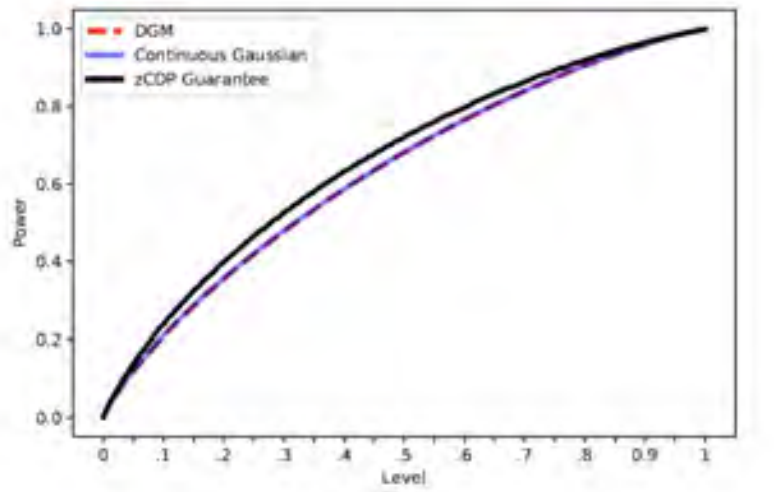

FIGURE 6. **Block within Custom Block Group**: Level (x-axis) vs. power (y-axis) curves for (1) the Gaussian mechanism over block-level queries at production settings for redistricting data ($\rho = 0.1115$), (2) the likelihood ratio test of the discrete Gaussian block-level noisy queries at production settings for redistricting data.

Source: Kifer et al. In preparation.

## Privacy protection out of the shadows

- Certain privacy practices for previous censuses depended upon obfuscation
- 2020 DAS demonstration data are the most transparent view into Census Bureau privacy practices ever
- We appreciate and are excited to assess feedback from our external partners

Stay Informed:
Subscribe to the 2020 Census Data Products Newsletters

*Search "Disclosure Avoidance" at www.census.gov or click the graphic

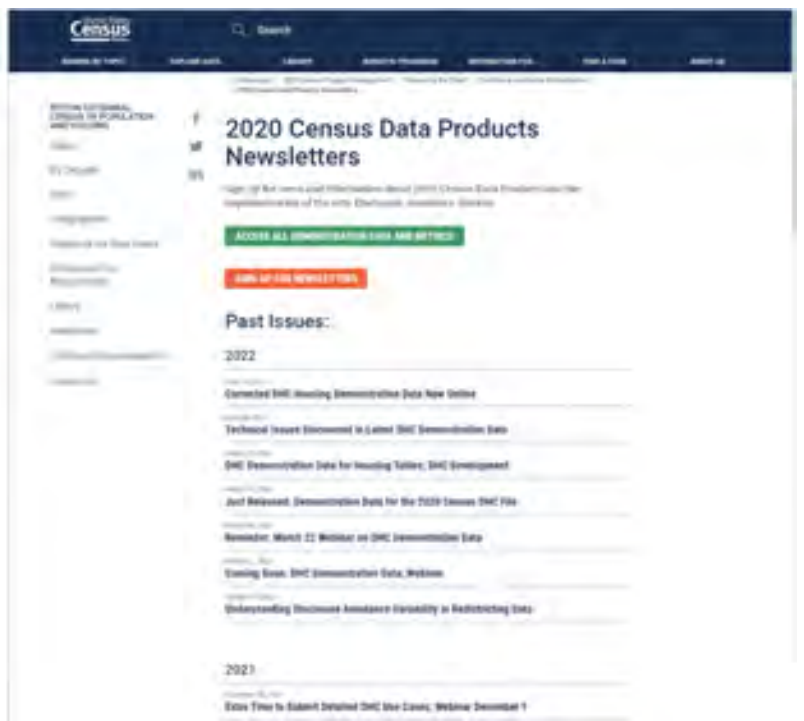

Stay Informed:
Visit Our Website

*Search "Disclosure Avoidance" at www.census.gov or click the graphic

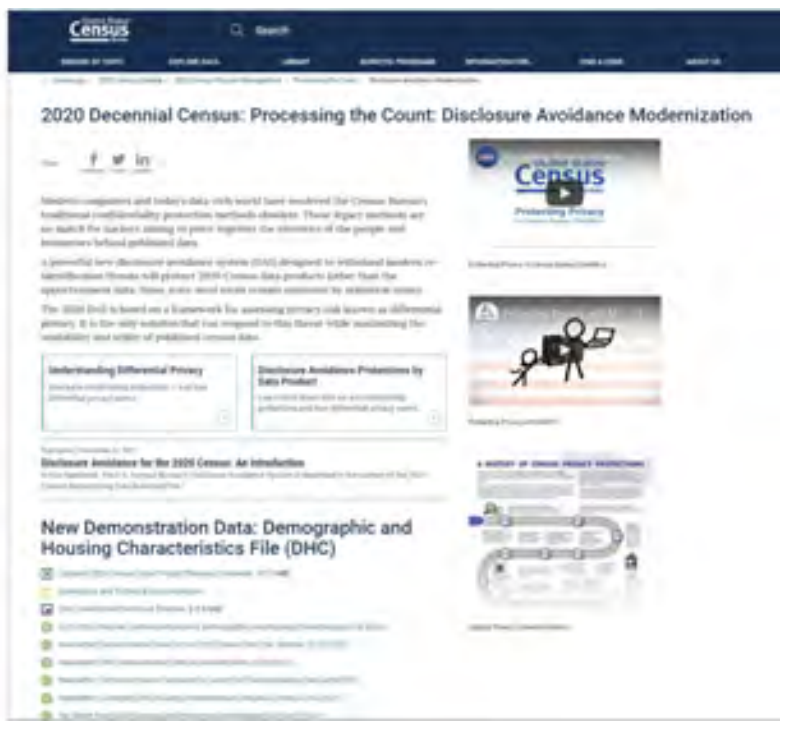

** Video **

[Protecting Privacy in Census Bureau Statistics]

*Find it on our website and YouTube Page

Search "Disclosure Avoidance" at www.census.gov or click the graphic

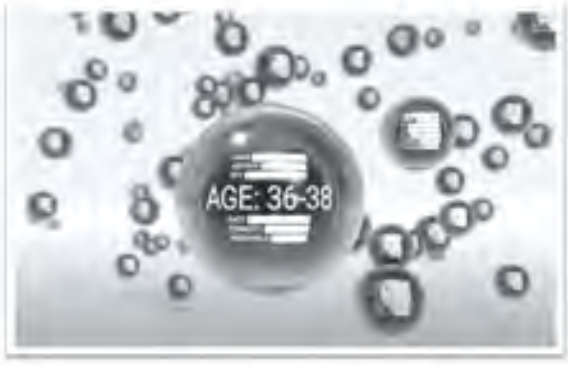

## Selected Additional Resources

- Code: uscensusbureau/DAS_2020_Redistricting_Production_Code: Official release of source code for the Disclosure Avoidance System (DAS) used to protect against the disclosure of individual information based on published statistical summaries. (github.com)
- Technical: *HDSR* The 2020 Census Disclosure Avoidance System TopDown Algorithm
- Updates: Developing the DAS: Demonstration Data and Progress Metrics (census.gov)

## Selected References

Thank you.

John.Maron.Abowd@census.gov

Figure B-13: Zachary Lipton - Adapting Predictors under Causally Structured Distribution Shift

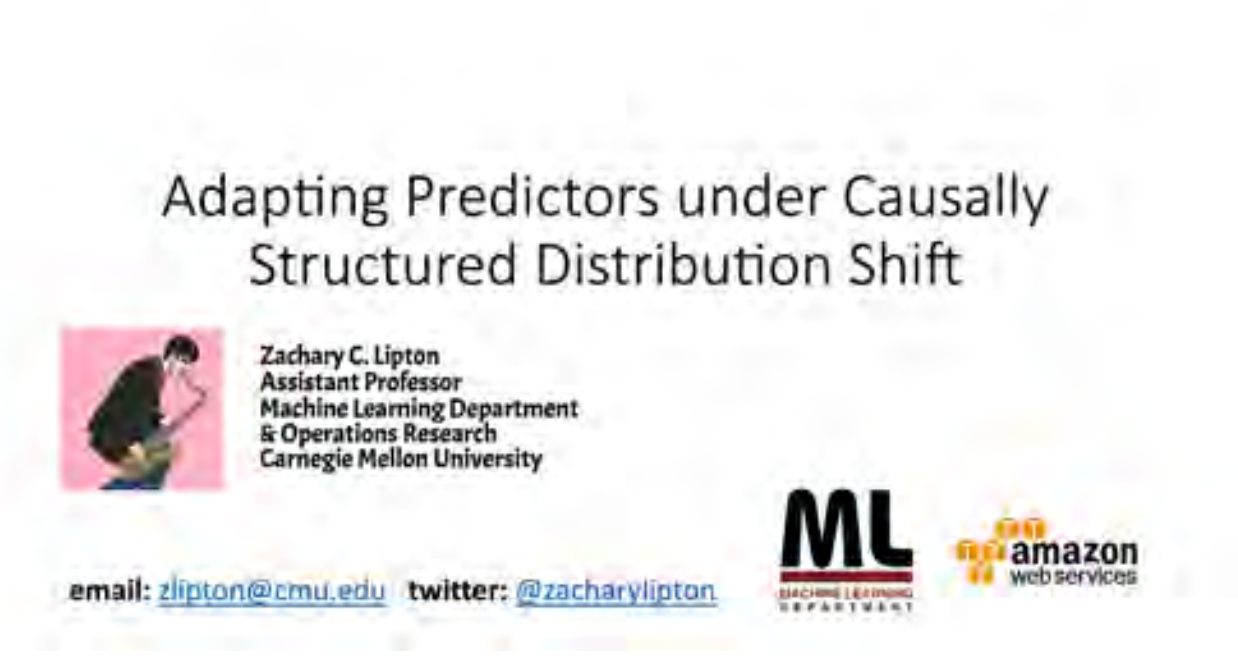

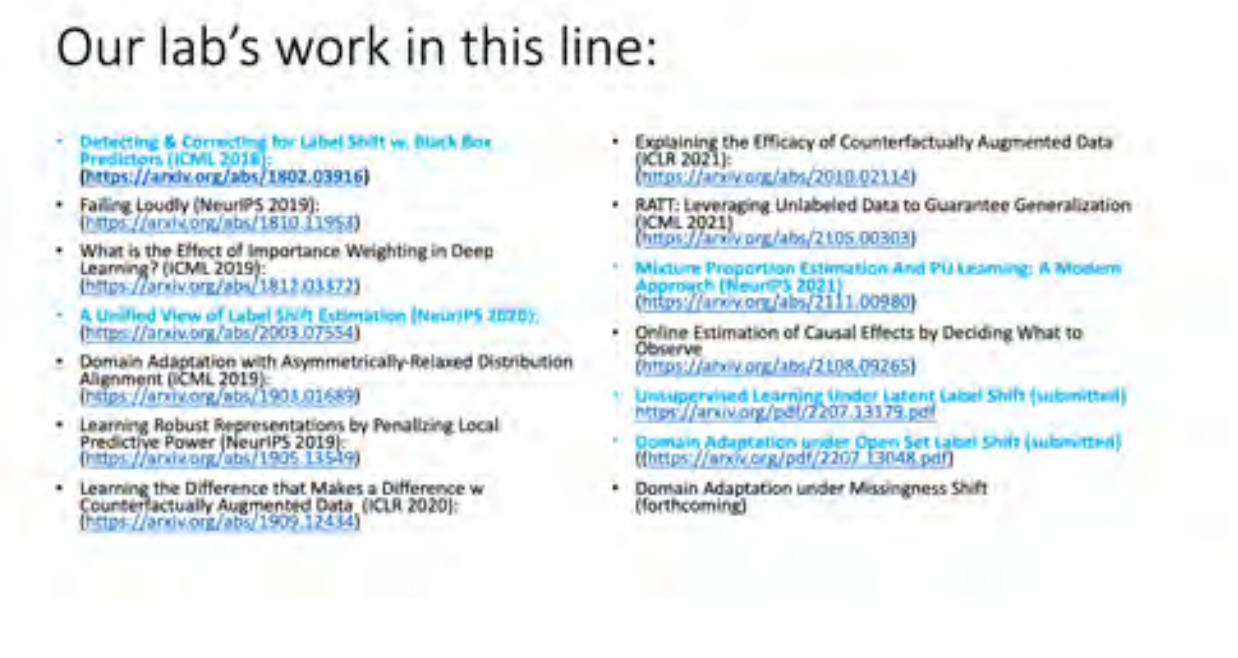

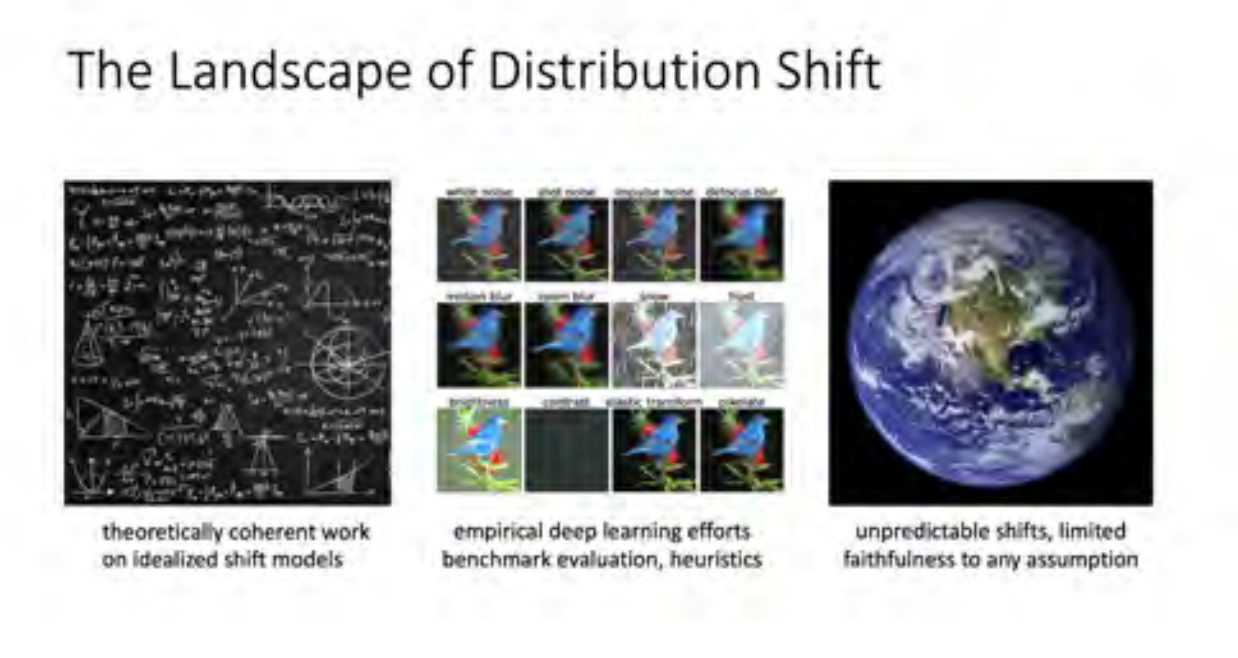

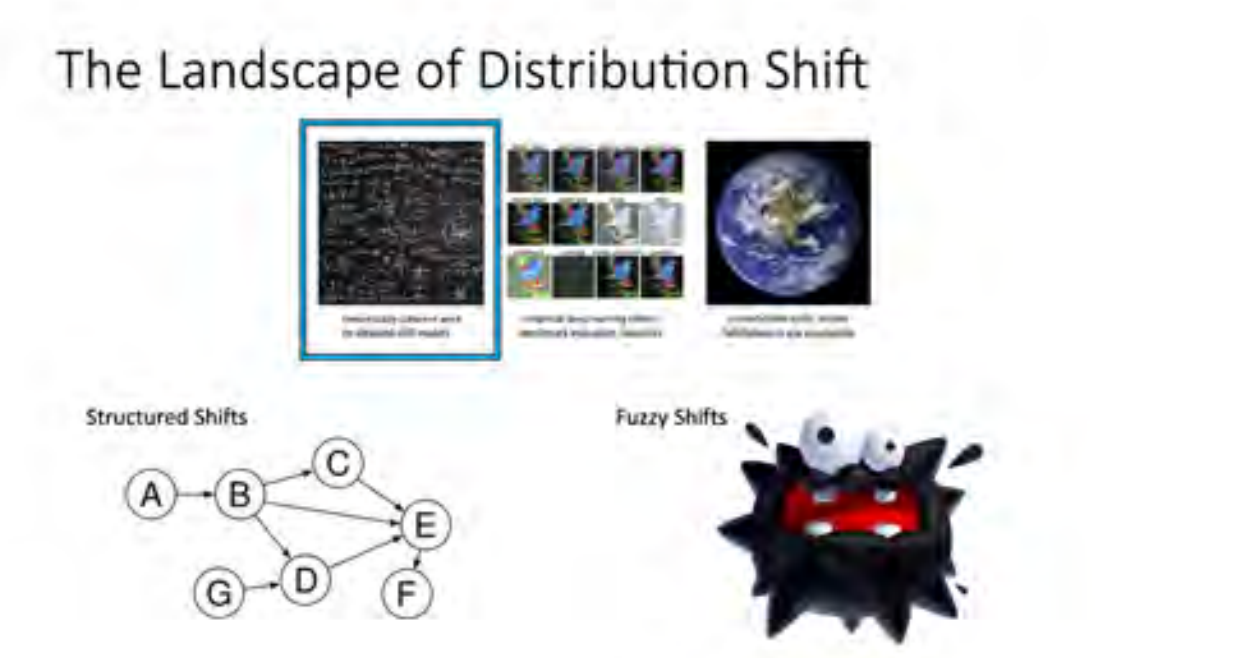

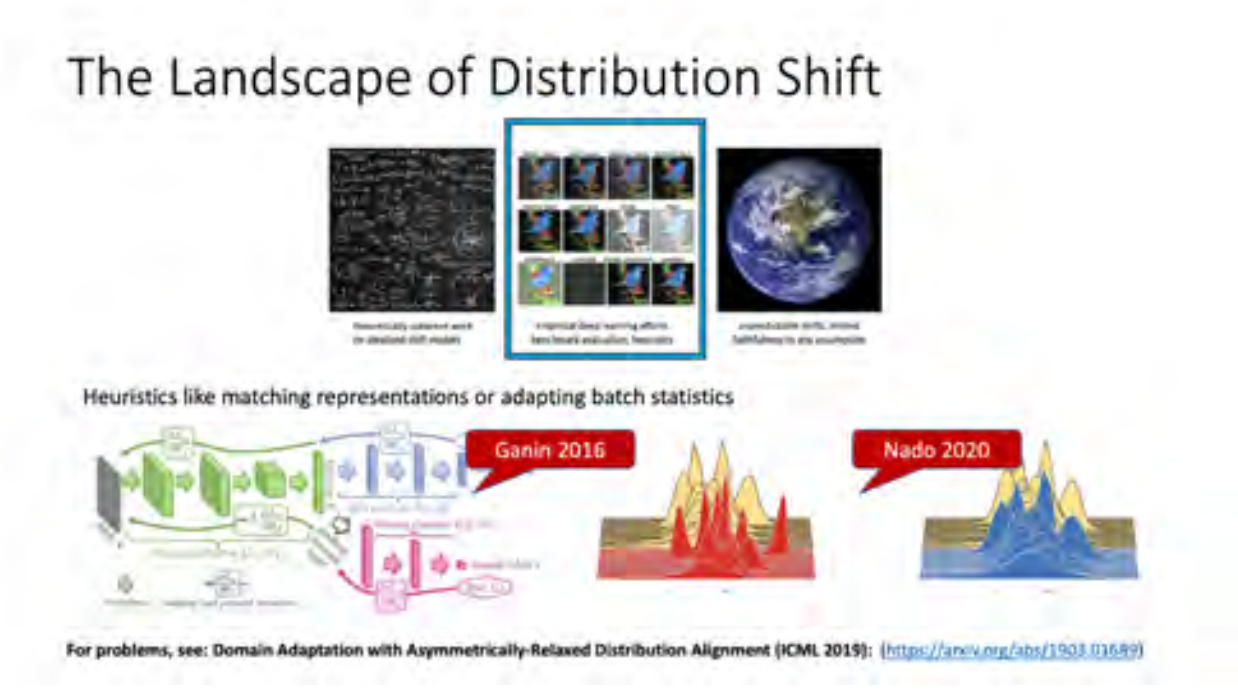

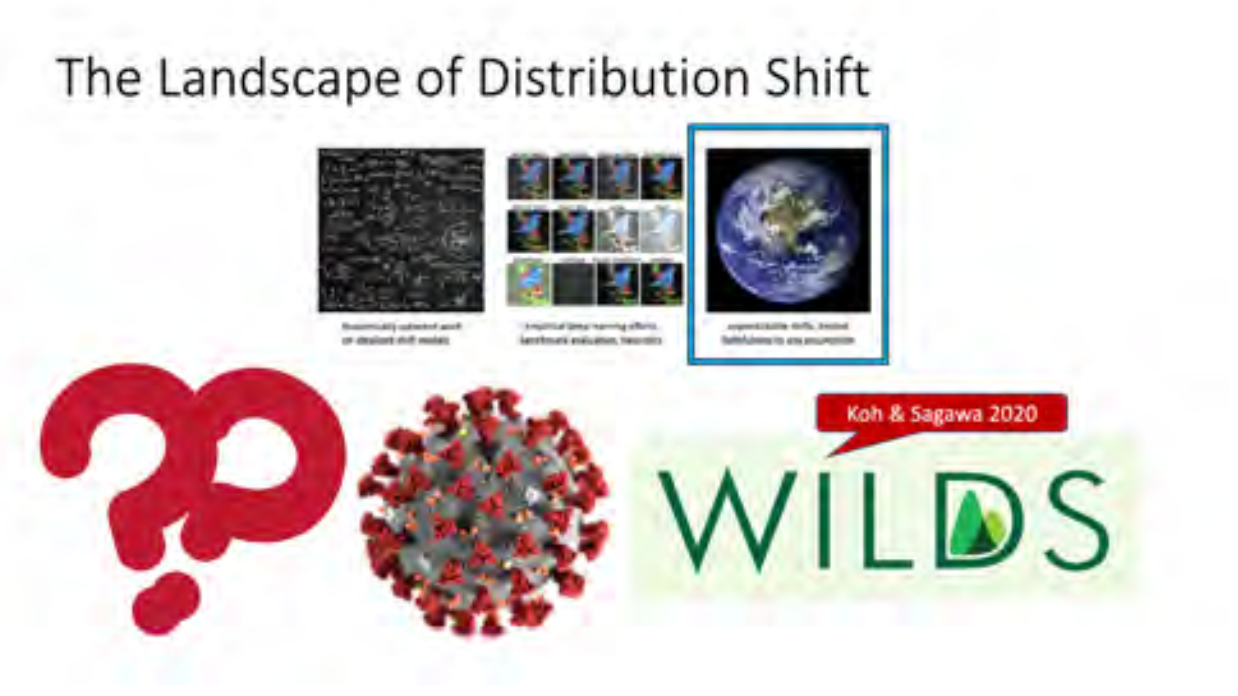

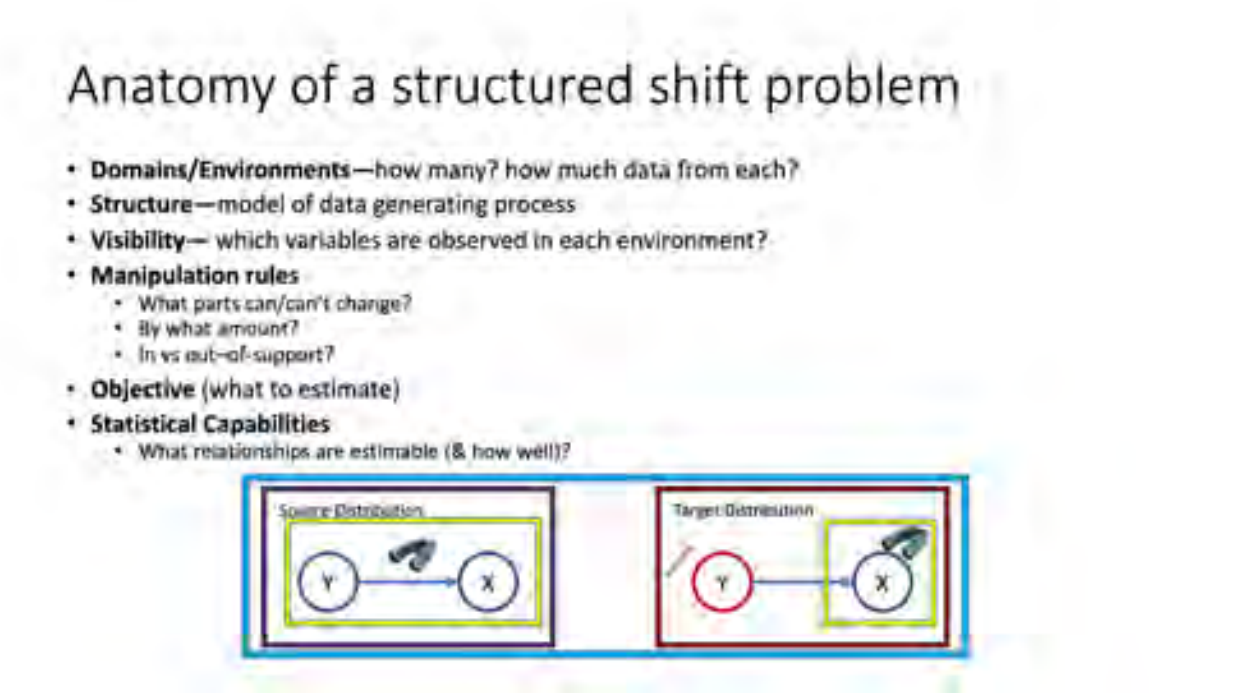

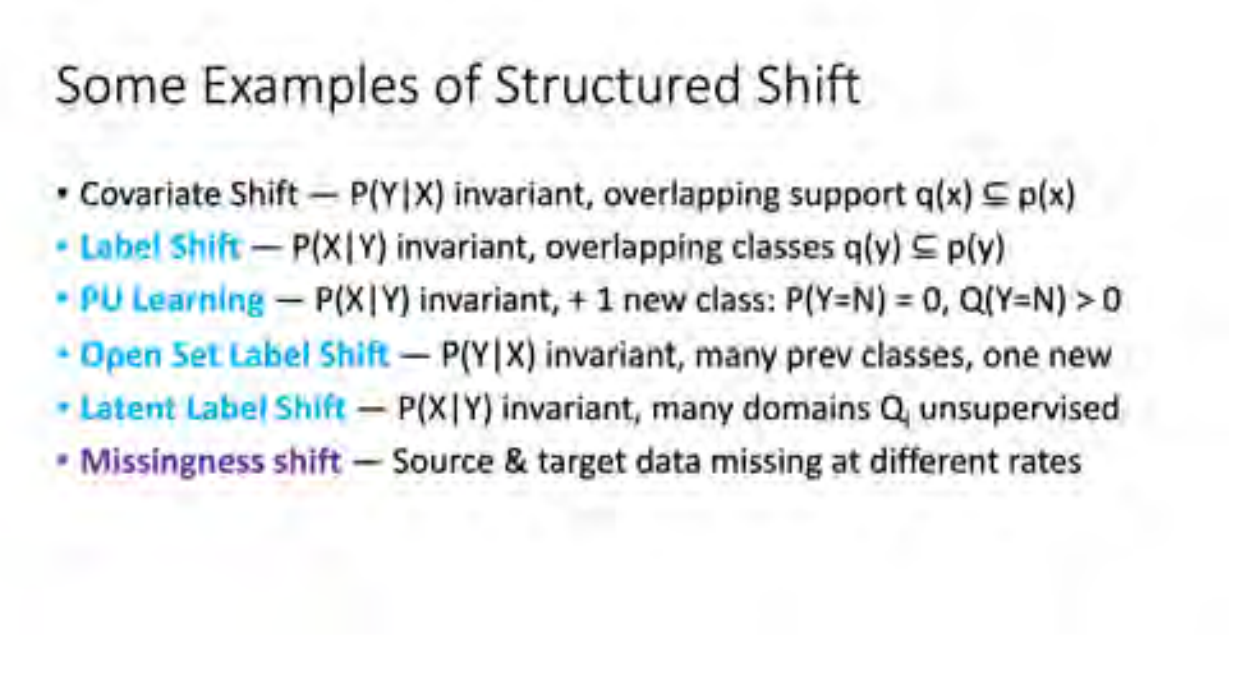

## Two Obstacles to Practicality

- Identification is nice but we need practical estimators for high-dim data
- Graph-structure is not the only identification problem (eg, overlap violations)

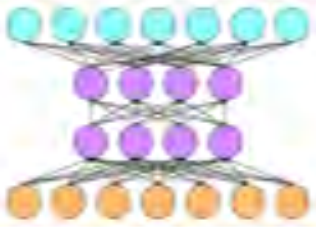

- Assumptions too rigid, performance under fuzzy violations unknown

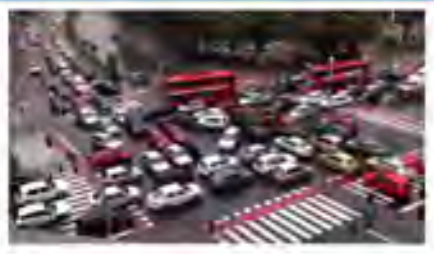

## The Move: Leveraging Black Box Predictors

- No theory says we should be able to predict well (even on iid data) w high-dimensional, arbitrarily non-linear data (e.g. images, speech)
- However, we want to show that when it's possible to learn good iid classifiers, we can leverage these black boxes to get target classifiers

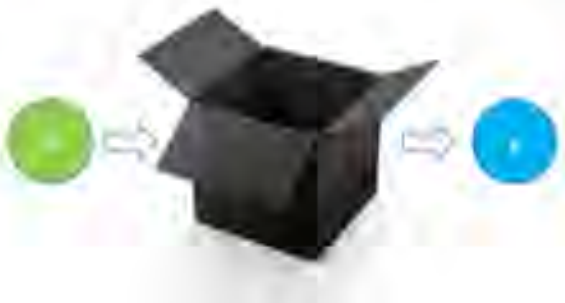

## Label Shift (aka Target Shift)

- Assume *p*(**x**,y) changes, but the conditional p(x|y) is fixed

$$q(y, \mathbf{x}) = q(y)p(\mathbf{x}|\mathbf{y})$$

- Corresponds to anticausal assumption, (y causes x)
- Assumptions: for all y such that q(y) > 0, p(y) > 0

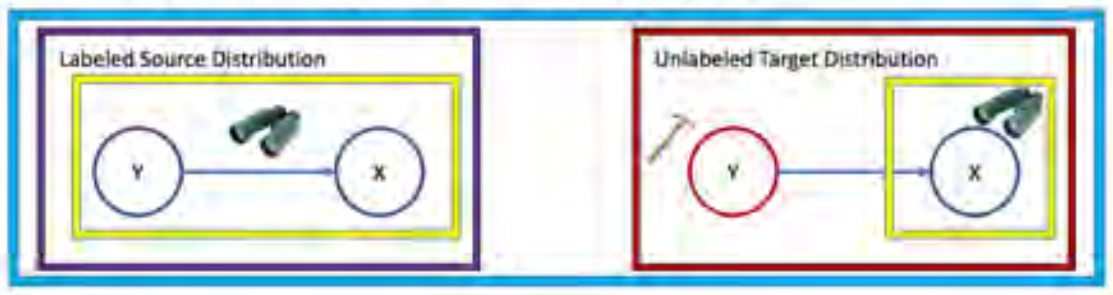

## Label Shift Identification

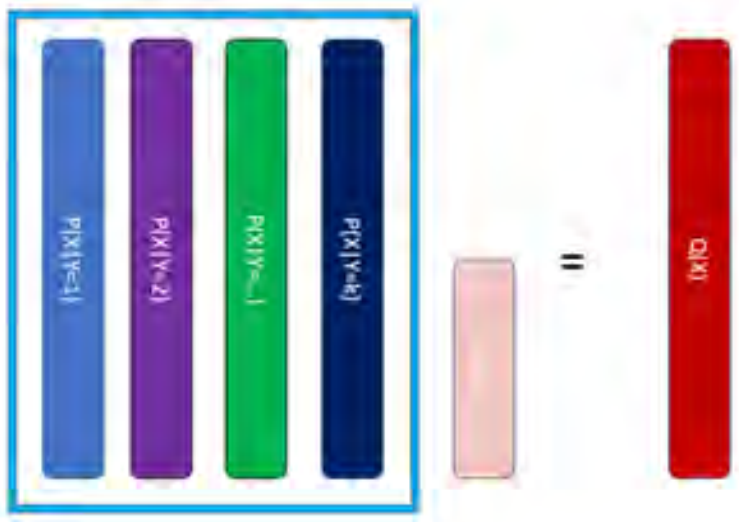

## Approach 1: Black Box Shift Estimation (BBSE)

- Consistent estimator with interpretable error bounds
- Accuracy doesn't depend (directly) on ambient dimension
- Use black box for dim-reduction (d → 1)
- Stronger *f* → tighter error bounds
- Inaccurate, uncalibrated *f* → BBSE still consistent

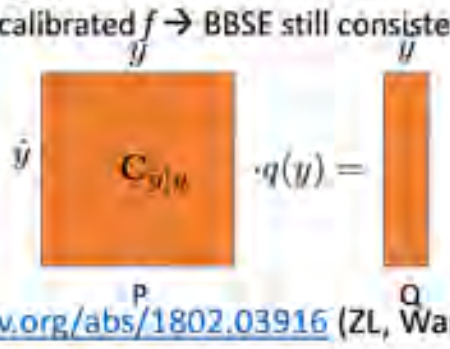

https://arxiv.org/abs/1802.03916 (ZL, Wang, Smola—ICML 2018)

## Estimation error in theory & practice

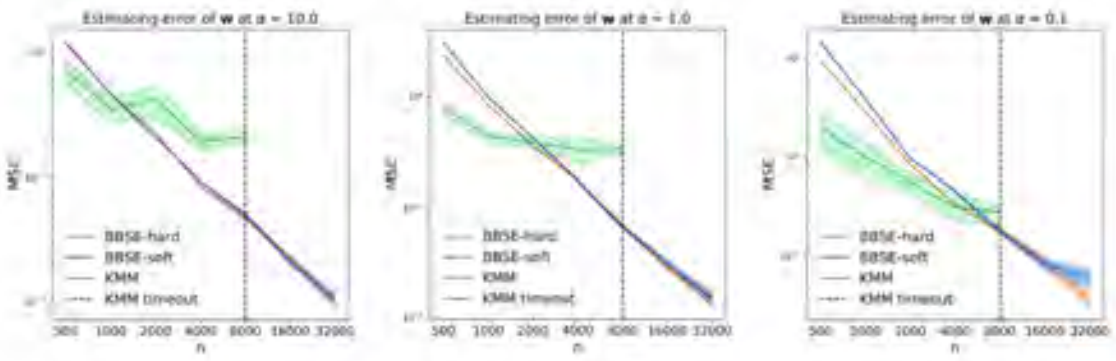

$$\|\hat{\boldsymbol{w}} - \boldsymbol{w}\|_2^2 \le \frac{C}{\sigma^2_{\min}}\left(\frac{\|\boldsymbol{w}\|^2 \log n}{n} + \frac{k \log m}{m}\right)$$

## Black Box Shift Correction (CIFAR10 w IW-ERM)

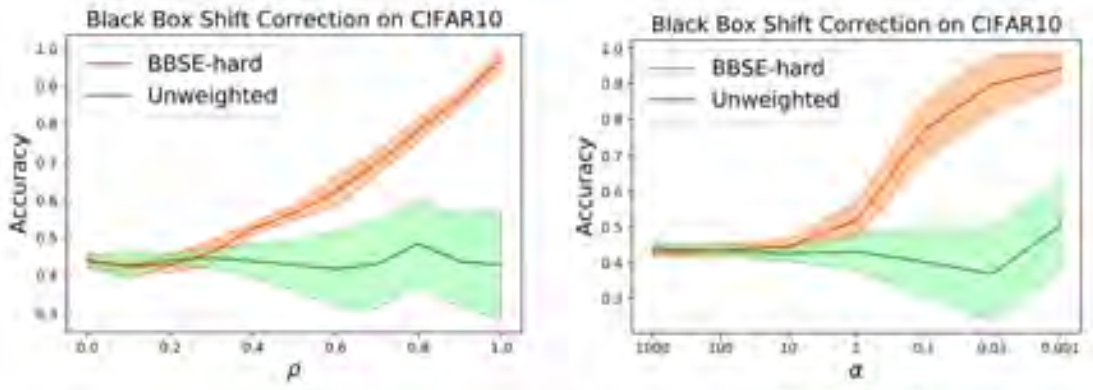

tweak-one shift

Dirichlet shift

## A General Pipeline for Detecting Shift

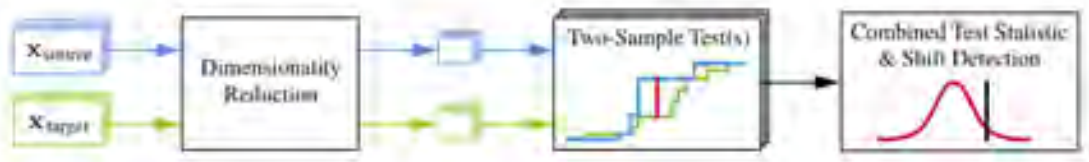

Failing Loudly: An Empirical Study of Methods for Detecting Dataset Shift
(Rabanser, Gunneman, Z. — Neurips 2019)
(https://arxiv.org/abs/1810.11953)

## What is the effect of importance weighting in deep learning?

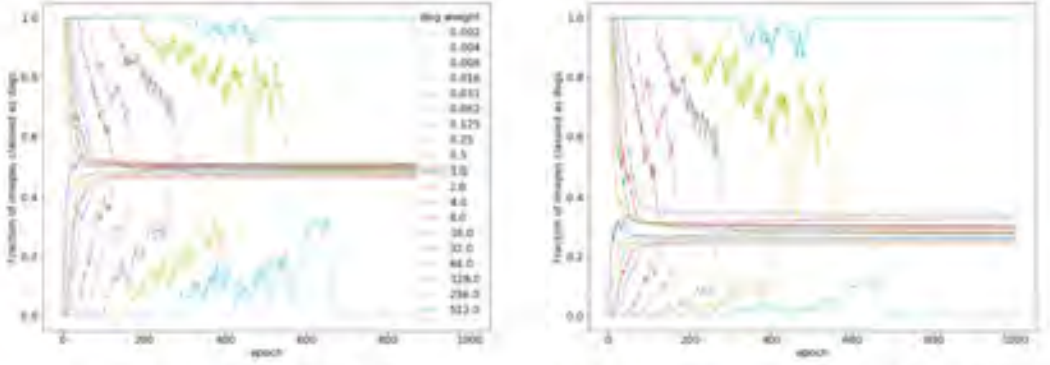

(a) CIFAR10 cat and dog test images (b) CIFAR10 test images from the other eight classes

(Byrd, Z. ICML 2019) — www.arxiv.org/abs/1812.03372

## Label Shift with a New Class

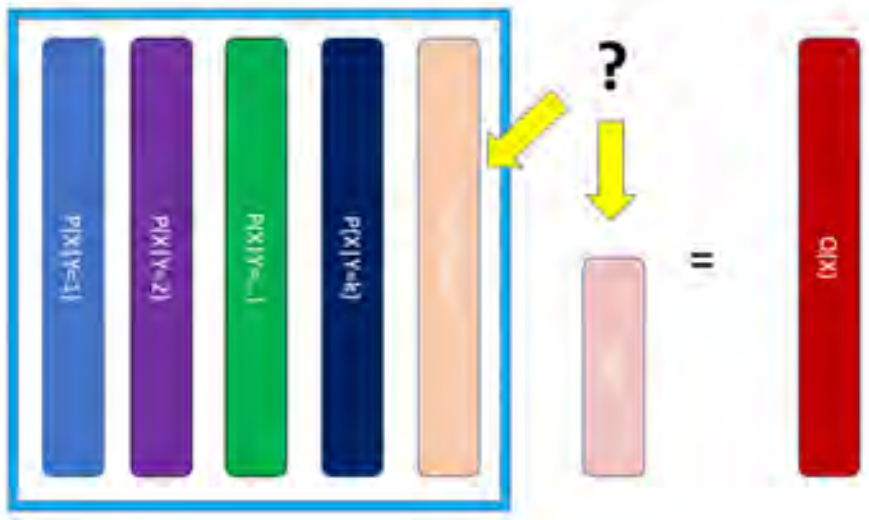

## PU Mixture Proportion Identification (irreducibility / positive subdomain)

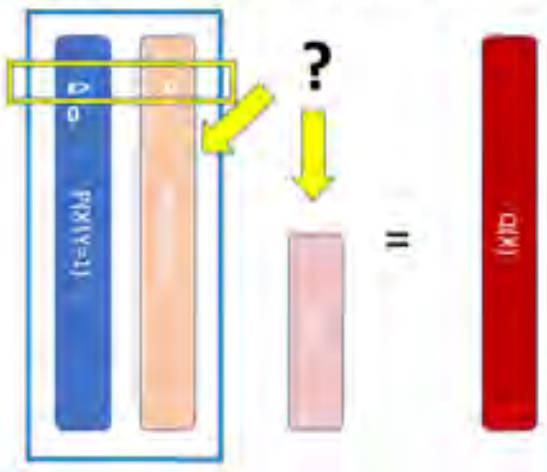

Garg, Balakrishnan, Smola, ZL (https://arxiv.org/abs/2111.00980)

## MPE: Traditional Approaches

- Elkan and Notto '08 discuss several approaches to estimate mixture proportion.
  - No guarantees and empirically perform bad
- Density estimation in input space [Ramaswamy+2013]. Curse of dimensionality in high dimensional datasets
- Recent methods that use classifier to reduce dimensionality
  - Need Bayes predictor for guarantees

## Estimation Strategy: Domain Discrimination

- Pool data
  - assign positive points label +,
  - assign unlabeled points label u
- Train a + vs. *u* classifier
- Assumption: when positive subdomain exists **most confidently predicted positive examples** are mostly positive

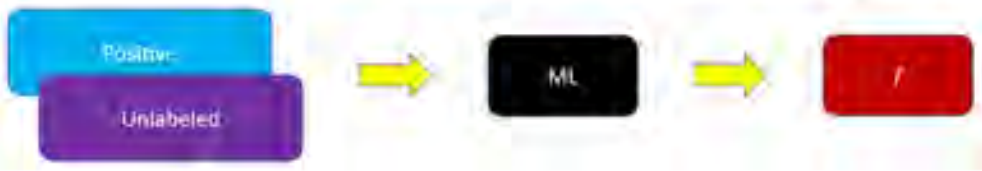

## Our Estimator: Best Bin Estimation (BBE)

- Our approach: dimensionality reduction $f(x)$.
  - Use a classifier to transform $z = f(x)$
  - $q_u(z) = \alpha q_p(z) + (1-\alpha) q_n(z)$
  - Estimate $\alpha^* = \min_z q_u(z)/q_p(z) \geq \alpha$
  - However, point estimates suffer from high variance
- Learn $f$ to classify training v.s. test!
  - Pure top bin property: For $f(x) \geq t$, $P_n(f(x)) \approx 0$.
- **Our estimator:** $\hat{\alpha}(t) \approx P_u(f(x) \geq t)/P_p(f(x) \geq t)$

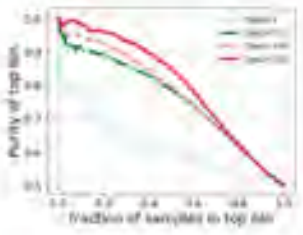

## Identifying cutoff for the top bin

- How to choose $t$?
  - A Naive approach: $t = 0.5$.
  - A better approach: $t = \arg\min_t \hat{\alpha}(t) + \mathrm{conf}\,(\alpha(t))$
- Understanding Bias vs Variance tradeoff in the top bin

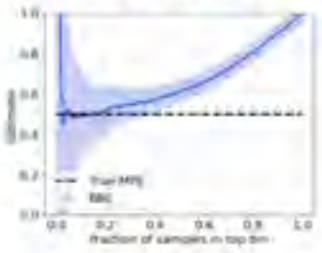

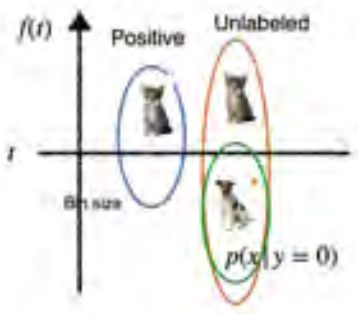

## Theoretical Result

**Theorem (Error rate of BBE)**

Define $c^* = \mathrm{argmin}_{c \in [0,1]} q_u(c)/q_p(c)$. For $\min(n_p, n_u) \geq 2\log(4/\delta)/q_p^2(c^*)$ and for every $\delta > 0$, the the mixture proportion estimator $\hat{\alpha}$ (in Algo 1) satisfies the following with probability $1-\delta$

$$|\hat{\alpha} - \alpha^*| \leq \frac{c_1}{q_p(c^*)}\left(\sqrt{\log(4/\delta)/n_u} + \sqrt{\log(4/\delta)/n_p}\right)$$

## Obtaining the Classifier: Conditional Value Ignoring Risk (CVIR)

- We propose a simple objective

**Algorithm 2** PU learning with Conditional Value Ignoring Risk (CVIR) objective

1: Rank samples $x_u \in X_u$ according to their loss values $\ell(f_\theta(x_u), -1)$.
2: $X_n := X_{u,1-\alpha}$ where $X_{u,1-\alpha}$ denote the lowest ranked $1-\alpha$ fraction of samples.
3: Train model $f_\theta$ for one epoch on $(X_p^1, X_n^1)$

- We can theoretically show that our loss will correctly discard positives from unlabeled when data is separable

## (TED)$^n$: Combining CVIR and BBE

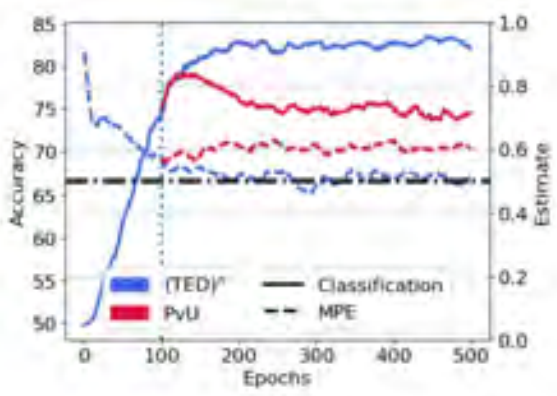

## MPE Results

| Dataset | Model | (TED)$^n$ | BBE* | Dedpul* | AlphaMax* | EN | KM2 | TiCE |
|---|---|---|---|---|---|---|---|---|
| Binarized CIFAR | ResNet | **0.018** | 0.072 | 0.075 | 0.125 | 0.175 | | |
| | All Conv | 0.041 | **0.038** | 0.046 | 0.09 | 0.23 | 0.181 | 0.251 |
| | FCN | 0.184 | 0.175 | **0.151** | 0.3 | 0.355 | | |
| CIFAR Dog vs Cat | ResNet | **0.074** | 0.120 | 0.113 | 0.17 | 0.205 | 0.11 | 0.203 |
| | All Conv | **0.073** | 0.093 | 0.098 | 0.19 | 0.274 | | |
| Binarized MNIST | FCN | **0.021** | 0.028 | 0.027 | 0.09 | 0.067 | 0.102 | 0.247 |
| MNIST17 | FCN | **0.003** | 0.008 | 0.006 | 0.075 | 0.065 | 0.03 | 0.117 |
| IMDb | BERT | **0.008** | 0.011 | 0.016 | 0.07 | 0.12 | – | - |

## Classification Results

| Dataset | Model | (TED)$^n$ (unknown $\alpha$) | CVIR (known $\alpha$) | PvU* (known $\alpha$) | Dedpul* (unknown $\alpha$) | nnPU (known $\alpha$) | uPU* (known $\alpha$) | PvN |
|---|---|---|---|---|---|---|---|---|
| Binarized CIFAR | ResNet | **82.7** | 82.6 | 78.3 | 78.4 | 76.8 | 75.8 | 86.9 |
| | All Conv | 76.8 | **77.1** | 74.1 | 76.9 | 72.1 | 71.3 | 76.7 |
| | FCN | 63.2 | **65.9** | 61.4 | 62.5 | 63.9 | 64.8 | 65.1 |
| CIFAR Dog vs Cat | ResNet | **76.1** | 74.0 | 71.6 | 70.9 | 72.6 | 69.5 | 80.4 |
| | All Conv | **72.2** | 71.0 | 70.1 | 70.5 | 68.4 | 65.2 | 77.9 |
| Binarized MNIST | FCN | 95.9 | **96.4** | 94.5 | 95.2 | 95.9 | 95.0 | 96.7 |
| MNIST17 | FCN | **98.6** | **98.6** | 98.7 | 98.1 | 98.2 | 98.4 | 99.0 |
| IMDb | BERT | **87.6** | 87.4 | 86.1 | 87.3 | 86.2 | 85.9 | 89.1 |

## Domain Adaptation under Open Set Label Shift (OSLS)

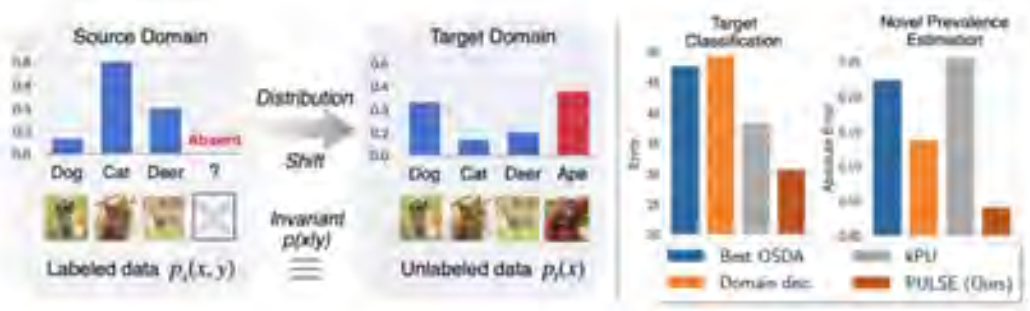

Garg, Balakrishnan, ZL (https://arxiv.org/pdf/2207.13048.pdf)

## Unsupervised Learning w Latent Label Shift

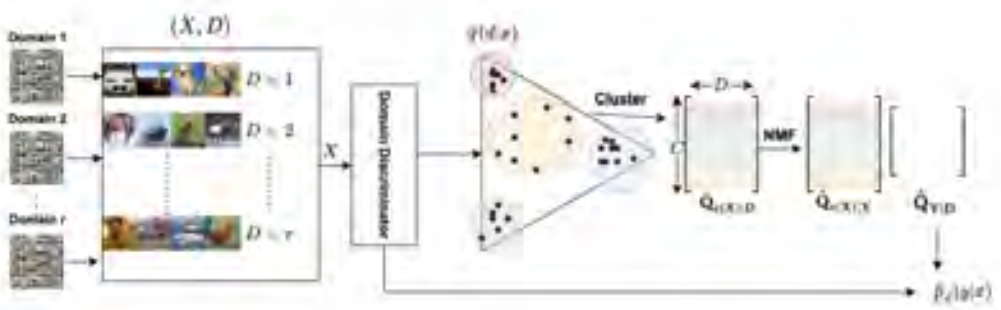

Roberts, Maini, Garg, ZL (https://arxiv.org/pdf/2207.13179.pdf)

## Thanks!

Figure B-14: Ruoxuan Xiong - Design and Analysis of Panel Data Experiments

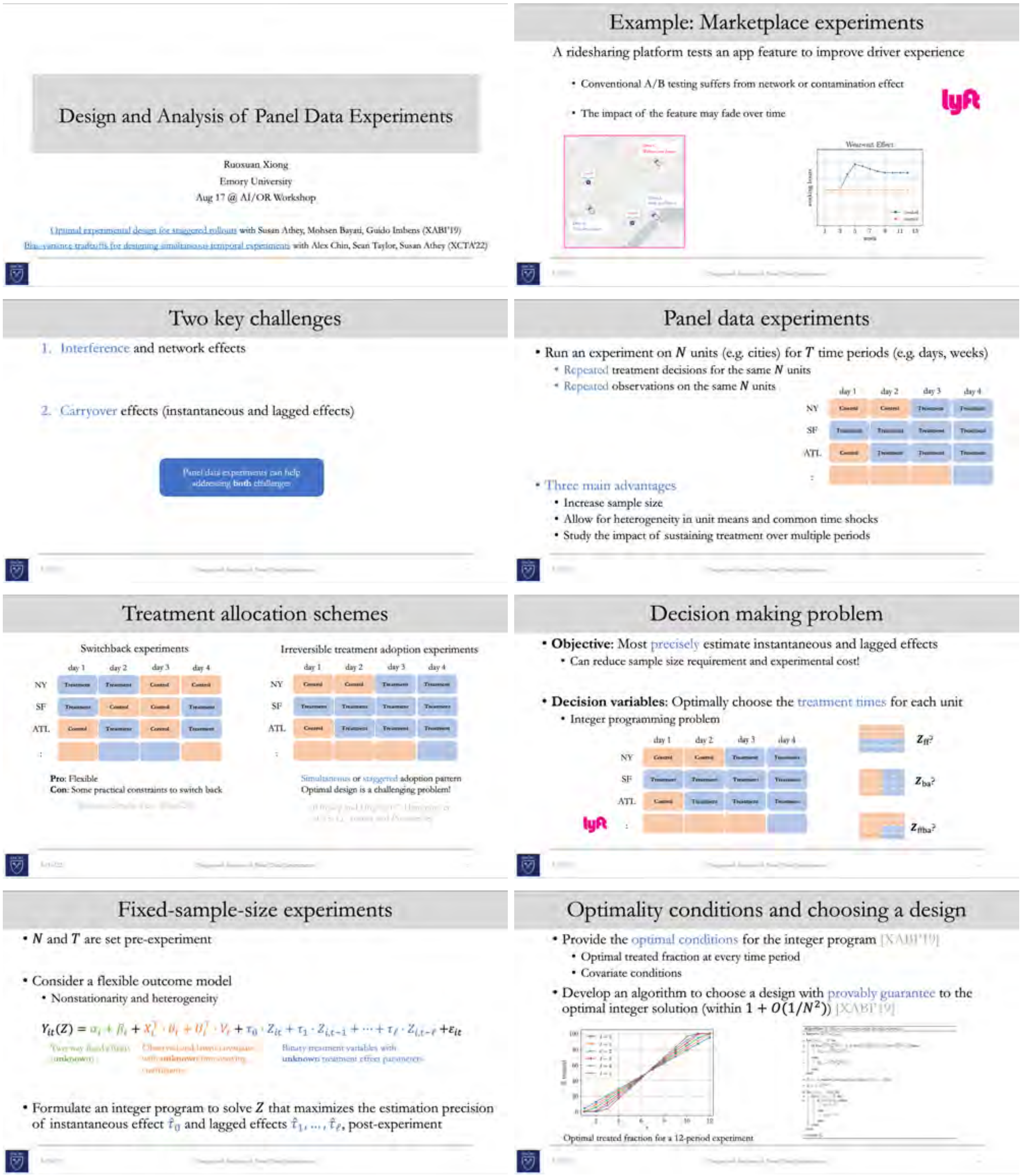

## Sequential experiments

- $N$ is fixed and $T$ varies (we can early stop the experiment)
- More flexible and cost-effective!
- Key challenges
  - Experiment termination rule
    - What is an appropriate rule and how to implement the rule
  - Peeking challenge
    - Treatment effect estimation based by experiment termination rule
  - Infeasibility to optimize treatment times pre-experiment
    - Optimal solution depends on $T$

## An initial stab at the problem

- We propose the Precision-Guided Adaptive Experimentation (P-GAE) algorithm
  - Leverage ideas from dynamic programming, empirical Bayes, and sample splitting
  - Key features
    - Adaptive treatment decisions
    - Precision-based experiment termination rule
    - Valid statistical inference post-experiment
  - We provide theoretical guarantees for P-GAE
    - Asymptotic: consistency, normality, and efficiency

## Micro-level perspective

- Micro-level data: raw data are **events**, like rider checking price, outcome is whether rider requested a ride
  - Large sample size, but analysis is more challenging
- Additional considerations when analyzing event data

Irregular event density

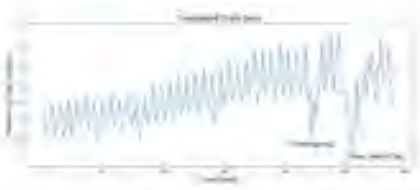

Correlation in event outcomes

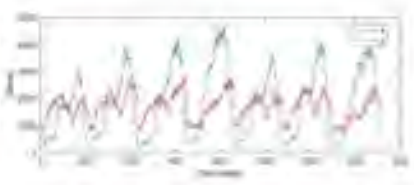

## Additional considerations

- Additional considerations when analyzing event data

Spillover effects

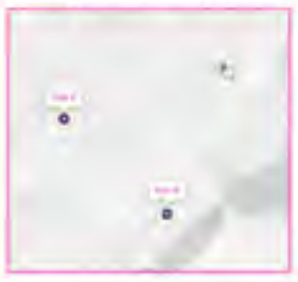

Simultaneous experiments

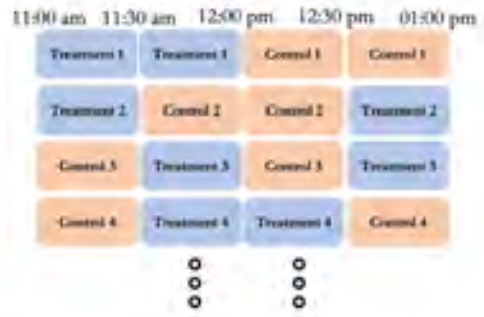

## Error analysis and design of experiments

- Analyze the mean-squared error (MSE) in treatment effect estimation
  - Bias affected by event density, spillover effects, simultaneous experiments
  - Variance affected by event density, correlation in event outcomes
- Study how partition time and space (irregularly) to minimize MSE

# Thank you!

## Figure B-15: Yu Ding - Causal Inference in Engineering Applications

### Causal Inference in Engineering Applications

Yu Ding, Ph.D.

Mike and Sugar Barnes Professor
Texas A&M University

Associate Director for Research Engagement
Texas A&M Institute of Data Science

Aug 17, 2022

### Natural experiments and causal inference

- Natural experiments "*are observational studies and are not controlled in the traditional sense of a randomized experiment.*" (source: Wikipedia)
- Causal inference aims at
  - determining which factors have a genuine cause-and-effect relationship with the response
  - quantifying the effect on the response due to the action taken.
- Huge impact in economics and social sciences.
- In some engineering applications, conducting controlled experiment is either too costly or infeasible. Then, causal inference becomes relevant.

### Wind energy application

- Wind energy (the same for solar) is known as *variable* renewable energy.
- Its fuel input is not controllable. Nor are other environmental conditions, which affect the wind power production, too.

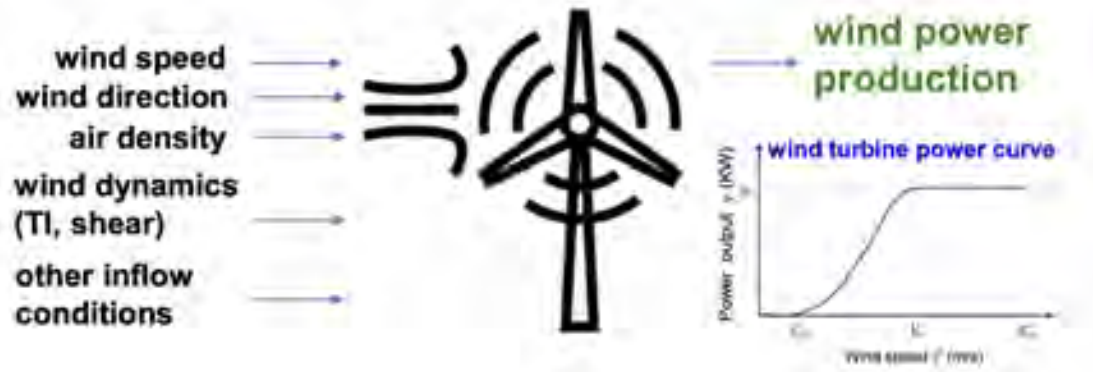

### Size matters

- Typical wind tunnels are an order of magnitude smaller than the size of commercial turbines.

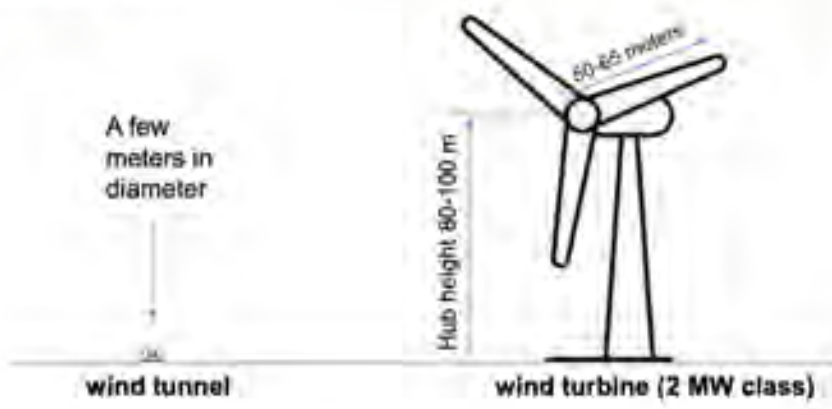

- Controlled experiments can be done on lab-scaled small turbines, but extrapolating to the commercial setting causes huge inaccuracy.

### Where does causal inference help?

- Both aims are relevant, but the second aim, i.e., the effect quantification, is particularly so.
- To counter a turbine's fast deterioration, a popular solution is to retrofit wind turbine, especially the blades. Such retrofit is known as **turbine upgrade**, meant to be **performance enhancement**.

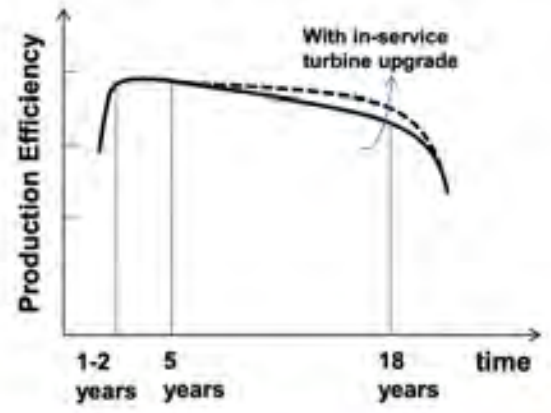

Service providers sell all kinds of performance enhancement options.

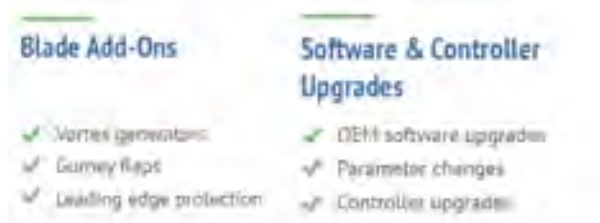

Staffell and Green (2014). "How does wind farm performance decline with age?" *Renewable Energy*, 66:775–786.

Tasker, Bechant, Post. "Are Power Curve Upgrades Worth It? Measuring the ROI of Turbine Upgrades" *American Wind Energy Association* Seminar, Sept 15, 2020.

### Vortex generator installation

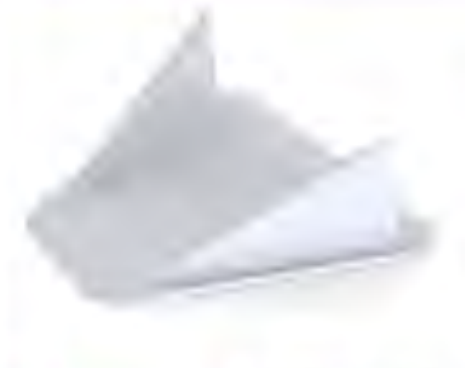

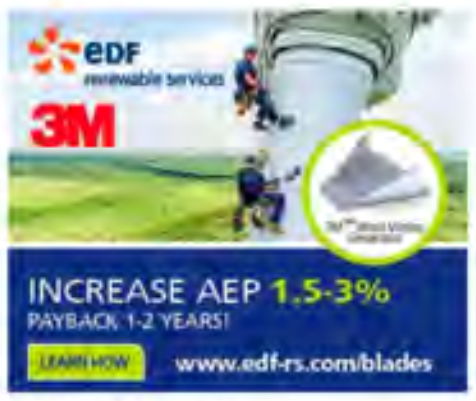

- Take vortex generators for example (picture courtesy of SMART BLADE® GmbH).
- Installing VGs on one turbine costs about $10K, so for a 200-turbine farm, the total cost is $2M.
- The key question is how soon can the wind farm operator pay back the expense?
- Depends on the increase in AEP after VG installation.

AEP = Annual Energy Production

## Causal inference for effect quantification

- If confirmed 1.5% – 3% improvement, as EDF advised, it makes a difference!

  3% improvement in AEP = $1.2 million revenue increase

  (for a wind farm with 200 2MW-class turbines at 6 cents/kWh price)

- The general consensus is that the VG benefit in commercial settings is moderate, producing likely 1% – 5% increase in AEP.

- Detecting this small improvement and attributing its effect to VG installation is where causal inference can help with.

## Covariate matching—A classical causal inference method

- Basic idea (Rubin 1973[†]): carefully select data points for making the environmental conditions probabilistically comparable BEFORE and AFTER.
- Specific techniques (Shin et al. 2018[††]):
  1. Hierarchical subgrouping
  2. Controlling for unmeasured factors
  3. One-to-one matching with replacement
  4. Robustness check on the order of matching

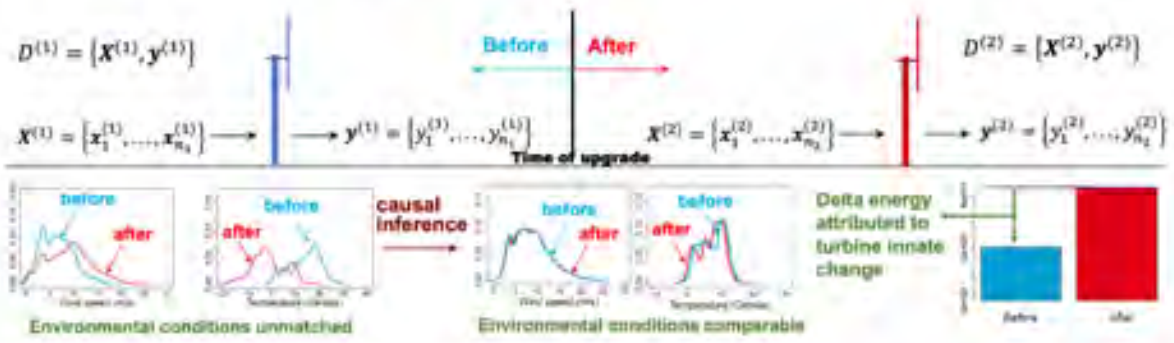

† Rubin (1973). "Matching to remove bias in observational studies." *Biometrics*, **29**: 159-183.
†† Shin, Ding, and Huang (2018). "Covariate matching methods for testing and quantifying wind turbine upgrades," *Annals of Applied Statistics*, **12**: 1271-1292.

## Matching results

- Two turbines are involved: a test turbine and a control turbine
- Before matching, the change in environmental conditions **confounds** the effect of a turbine upgrade

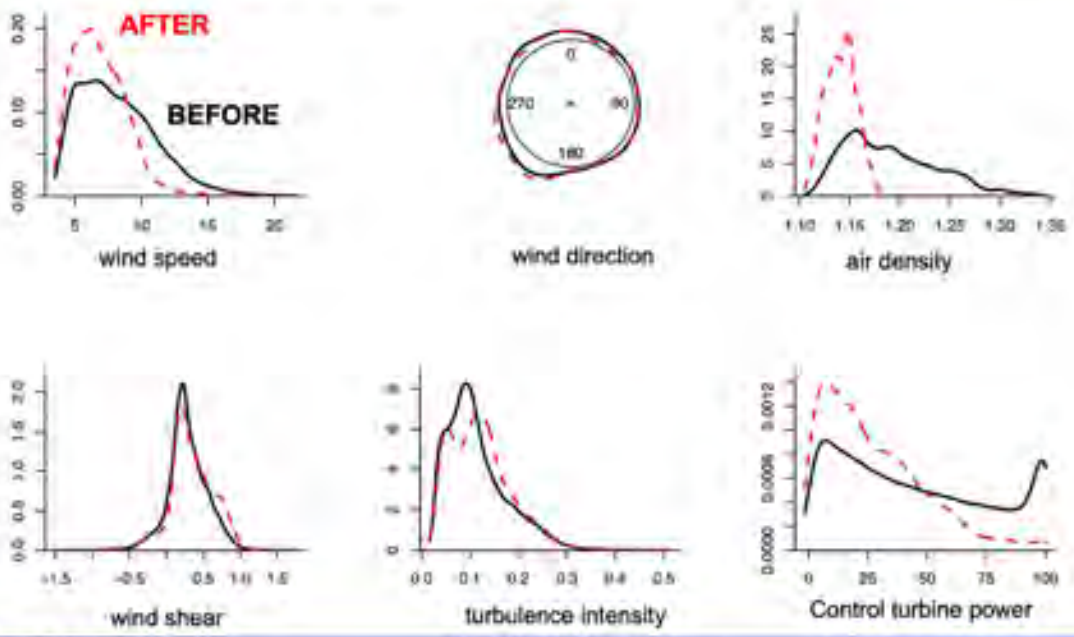

## Matching results

- Two turbines are involved: a test turbine and a control turbine
- After matching, the density curves are probabilistically comparable.

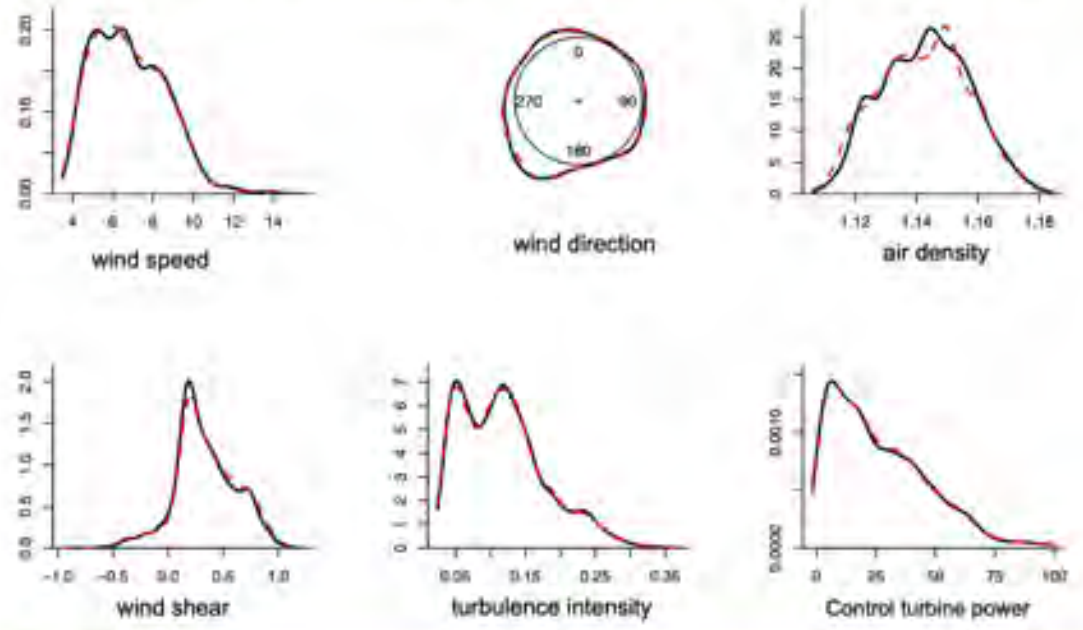

## Further need for bias-reduction

- Even though a visual inspection shows reasonable alignment, there may still be a few percentage difference between two functional curves, large enough if compared with the anticipated upgrade effect.
- We explored a number of advanced options for covariate matching but those do not help and some even makes the bias worse.
- Bias reduction is still a pressingly needed.

## Causal Inference Nested in a Three-Step Procedure

- Three methodological components.

| Which input variables to use? | Which data subsets to use? | How to compare nonparametric functions? |
|---|---|---|
| J. of Am. Stat. Assoc. (2015), Technometrics (2022, online) | Annals of Applied Stat. (2018) | Technometrics (2022) |
| Nonparametric regression<br>Temporal overfitting | Casual inference for natural experiments | Nonparametric comparison & uncertainty quantification |

- The solution procedure[†] is implemented in both `R` and `Python` packages, named "DSWE".

† Ding, Kumar, Prakash, Kio, Liu, Liu, and Li (2021) "A case study of space-time performance comparison of wind turbines on a wind farm," *Renewable Energy*, **171**: 735-746.

## Step 1. Build a model that avoids temporal overfitting

- Cause-and-effect of certain variables is unmistakeable, like wind speed, while conditional cause-and-effect of other variables are not so clear.
- When data are autocorrelated, deciding the conditional causal effect becomes tricky. Regular cross-validation tends to select an overfitting model.
- Our method, tempGP, is meant to extract the genuine effect despite the autocorrelation in data.

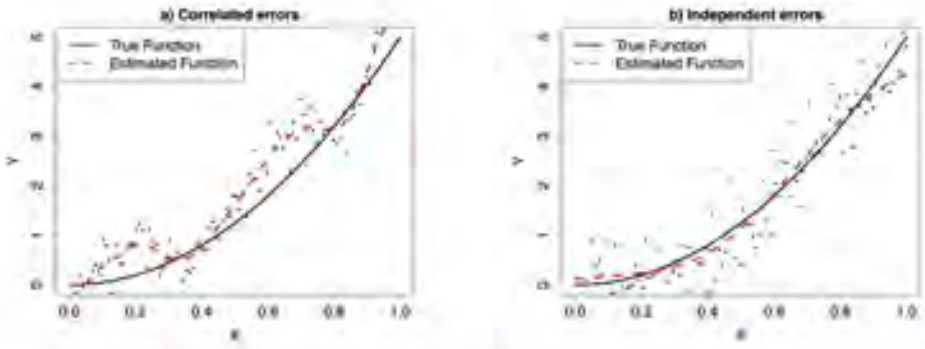

Prakash, Tuo, and Ding (2022) "The temporal overfitting problem with applications in wind power curve modeling," *Technometrics*, online published.

## Step 3. Inference and UQ of difference in nonparametric functions

Let $f_1(\cdot)$ and $f_2(\cdot)$ be two nonparametric functions, representing the performances of two turbines of the same period, or two periods of the same turbine.

$$H_0 : f_1(\boldsymbol{x}) = f_2(\boldsymbol{x}) \quad \forall \quad \boldsymbol{x} \in \mathcal{X}$$
$$H_1 : f_1(\boldsymbol{x}) \neq f_2(\boldsymbol{x}) \quad \exists \quad \boldsymbol{x} \in \mathcal{X}.$$

1. The functional estimates, $\hat{f}_1(\cdot)$ and $\hat{f}_2(\cdot)$, are used, as $f_1(\cdot)$ and $f_2(\cdot)$ are unknown.
2. Gaussian Process modeling provides foundation for uncertainty quantification.
3. Karhunen Loève (KL) expansion makes computation feasible.

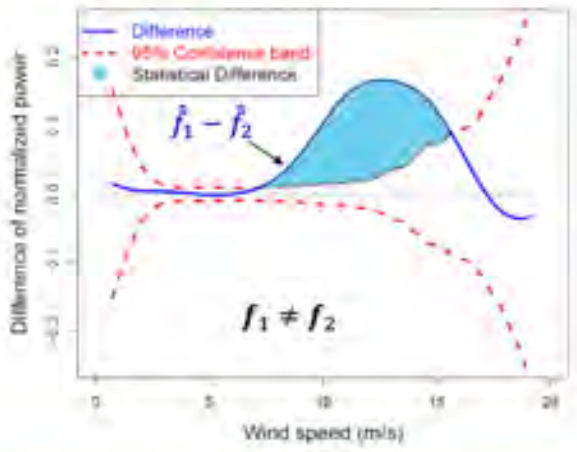

Prakash, Tuo, and Ding (2022) "Gaussian process aided function comparison using noisy scattered data," *Technometrics*, **64**: 92-102.

## A simulated study

The power in the test period is multiplied by a factor of $1 + r$ for those corresponding to wind speed greater than 9 m/s.

| | | | | | | | | |
|---|---|---|---|---|---|---|---|---|
| $r$ | 2% | 3% | 4% | 5% | 6% | 7% | 8% | 9% |
| $r'$ | 1.25% | 1.87% | 2.49% | 3.11% | 3.74% | 4.36% | 4.98% | 5.60% |
| $\Delta$ in Ding et al. (2021) | 1.12% | 1.77% | 2.73% | 3.43% | 3.69% | 4.69% | 5.16% | 5.59% |
| $\Delta/r'$ | 0.90 | 0.95 | 1.10 | 1.10 | 0.99 | 1.08 | 1.04 | 1.00 |
| UPG in Shin et al. (2018) | 1.74% | 2.21% | 2.68% | 3.16% | 3.63% | 4.11% | 4.58% | 5.05% |
| UPG/$r'$ | 1.39 | 1.18 | 1.08 | 1.02 | 0.97 | 0.94 | 0.92 | 0.90 |
| DIFF in Lee et al. (2015) | 1.97% | 2.56% | 3.15% | 3.73% | 4.30% | 4.86% | 5.42% | 5.97% |
| DIFF/$r'$ | 1.58 | 1.37 | 1.27 | 1.20 | 1.15 | 1.11 | 1.09 | 1.07 |

- Ding et al. (2021): functional fitting + covariate matching + UQ.
- Shin et al. (2018): covariate matching only.
- Lee et al. (2015): functional fitting only.

## Industrial case studies

- 2014, a blind study, with EDPR North America. [ASME TurboExpo Conference Proceedings, GT 2015, Montreal, Canada, June 15-19.]
- 2015-2016, with SMART BLADE® GmbH (Germany). [*Renewable Energy*, 2017, volume 113, pp. 1589–1597.]
- 2017-2018, with EDP Renewables (Spain/US). [A specialized `R` package named `gainML`].

  The above studies focused on VG installation.
- 2019-2020, with Goldwind (China). Evaluation of software and control logic options. [*Renewable Energy*, 2021, volume 171, pp. 735–746].
- 2020-2021, with Wind Energy Institute of Canada (WEICan). Evaluation of leading-edge protection options on blades. [*Wind Energy*, 2022, volume 25, pp. 1203–1221].

## Takeaways

- Causal inference to wind engineering: simple idea, real impact.
- For the aim of effect quantification, both bias reduction and uncertainty quantification are still challenging problems.
- Data characteristics deviating from standard iid assumptions raises new needs in causal inference.
- **Other applications**. Brian Denton recently published a perspective paper in *IISE Transactions*, entitled "*Frontiers of medical decision-making in the modern age of data analytics*". The whole Section 2 is dedicated to the discussion of leveraging observational data to build IE/OR models.